\documentclass[journal]{IEEEtran}
\usepackage{tikz}
\usepackage{tikz}

\usepackage{amssymb}
\usepackage{supertabular}

\usepackage{bbm}
\usepackage{rotating}
\usepackage{dsfont}

\usepackage{bm}
\usepackage{makecell}
\usepackage{array}
\usepackage{eucal}
\usepackage{graphicx}
\usepackage{epstopdf}
\usepackage{amsfonts,amsthm}
\usepackage{setspace}
\usepackage{cite}
\usepackage[colorlinks,linkcolor=blue,anchorcolor=blue,citecolor=blue]{hyperref}
\usepackage{subfig}
\usepackage{multirow,booktabs}

\usepackage{amsmath,mathrsfs}
\usepackage[misc]{ifsym}

\usepackage{amsmath,graphicx}
\usepackage{epstopdf}
\usepackage{amsfonts,amsthm}
\usepackage{algorithmic}
\usepackage[ruled,linesnumbered,boxed]{algorithm2e}
\usepackage{setspace}
\usepackage{cite}
\usepackage[colorlinks,linkcolor=blue,anchorcolor=blue,citecolor=blue]{hyperref}
\usepackage{subfig}
\usepackage{subfloat}
\usepackage{multirow,booktabs}

\usepackage{amsmath}
\usepackage{amssymb}
\usepackage[misc]{ifsym}
\usepackage{url}

\usepackage{booktabs}
\usepackage{multirow}

\usepackage{color,soul}
\definecolor{myred}{RGB}{255,0,0}
\definecolor{myblue}{RGB}{0,0,255}
\usepackage{threeparttable}

\usepackage{hyperref}
\hypersetup{hypertex=true,
            colorlinks=true,
            linkcolor=red,
            anchorcolor=green,
            citecolor=blue}

\newtheorem{Definition}{Definition}[section]
\newtheorem{Theorem}{Theorem}[section]
\newtheorem{Remark}{Remark}[section]
\newtheorem{Lemma}{Lemma}[section]
\newtheorem{Assumption}{Assumption}[section]

\newcommand{\tabincell}[2]{\begin{tabular}{@{}#1@{}}#2\end{tabular}}

\ifCLASSINFOpdf
\else
\fi

\begin{document}

\title{Hyperspectral Anomaly Detection Fused 
Unified Nonconvex Tensor Ring Factors Regularization}

\author{
Wenjin~Qin,
~Hailin~Wang,~\IEEEmembership{Student Member,~IEEE,}
~Hao~Shu,
~Feng~Zhang, 
~Jianjun~Wang,~\IEEEmembership{Member,~IEEE,}
~Xiangyong~Cao,~\IEEEmembership{Member,~IEEE,}
~Xi-Le~Zhao, and
~Gemine~Vivone,~\IEEEmembership{Senior Member,~IEEE}
%
\thanks{
This work was supported in part by the National Key Research and Development Program of China under Grant 2023YFA1008502; in part by the National Natural Science Foundation of China's Regional Innovation Development Joint Fund under Grant U24A2001; in part by the Natural Science Foundation of Chongqing, China, under Grant
CSTB2023NSCQ-LZX0044; in part by the National Natural Science Foundation of China under Grant 12301594, Grant 12201505, 
Grant 12101512;  in part by the Chongqing Talent Project, China, under Grant cstc2021ycjh-bgzxm0015;
in part by the Fundamental Research Funds for the Central Universities under Grant SWU-KR25013;
 and in part by
 the Initiative Projects for Ph.D. in China West Normal University under Grant 22kE030.
 (Corresponding author: Jianjun Wang.)
%
}
\thanks{Wenjin Qin,  Feng Zhang,  and Jianjun Wang are
with the School of Mathematics and Statistics, Southwest University, Chongqing 400715, China (e-mail:
qinwenjin2021@163.com,  zfmath@swu.edu.cn, wjj@swu.edu.cn).} %
%
\thanks{Hailin Wang and Hao Shu are  with the School of Mathematics and Statistics, Xi'an
Jiaotong University, Xi'an 710049, China (e-mail: wanghailin97@163.com, haoshu812@gmail.com).}
%
\thanks{
Xiangyong Cao is with School of Computer Science and Technology and Ministry of Education Key Lab For Intelligent Networks and
Network Security, Xi'an Jiaotong University, Xi'an 710049, China (e-mail: caoxiangyong@mail.xjtu.edu.cn).
}
\thanks{
Xi-Le Zhao is  with the School of Mathematical Sciences/Research Center for Image and Vision Computing,
University of Electronic Science and Technology of China, Chengdu 611731, China (e-mail: xlzhao122003@163.com).
}
%
\thanks{
Gemine Vivone is with the Institute of Methodologies for Environmental Analysis, CNR-IMAA, 85050 Tito Scalo, Italy, and also with the National Biodiversity Future Center (NBFC), 90133 Palermo, Italy (e-mail: gemine.vivone@imaa.cnr.it).
}
\vspace{-1.0cm}
}
\maketitle
\begin{abstract}
In recent years, tensor decomposition-based approaches for \textit{hyperspectral anomaly detection} 
(HAD) have gained significant attention in the field of remote sensing.
 %
 \textcolor[rgb]{0.00,0.00,0.00}{However, existing methods often fail to flexibly %
 and effectively extract
  both the global correlations and local smoothness of the background components in 
 \textit{hyperspectral images} (HSIs).}
To mitigate this critical issue,
%
we put forward a novel HAD method named HAD-EUNTRFR, which incorporates an enhanced unified nonconvex tensor ring (TR) factors regularization.
 %
 \textcolor[rgb]{0.00,0.00,0.00}{In the HAD-EUNTRFR framework, the raw HSIs are first decomposed into background and anomaly components
  using
the idea of tensor robust principal component analysis.}
 %
 The TR decomposition is then employed to capture the spatial-spectral correlations within the background component.
Additionally, we introduce a unified and efficient nonconvex regularizer, induced by  
 \textit{tensor singular value decomposition} (T-SVD), to simultaneously encode 
 the low-rankness and sparsity of the $3$-D gradient TR factors into  a unique
concise form.
The above 
characterization scheme
enables the interpretable gradient TR factors to inherit the low-rankness and smoothness of the original background.
 To further enhance anomaly detection, we design a generalized nonconvex regularization term to exploit the group sparsity of the anomaly component.
%
\textcolor[rgb]{0.00,0.00,0.00}{Based upon the above,  we ultimately propose a scalable and reliable nonconvex HAD model.}
To solve the resulting doubly nonconvex model, we develop a highly efficient optimization algorithm based on the \textit{alternating direction method of multipliers} (ADMM) framework.
\textcolor[rgb]{0.00,0.00,0.00}{Theoretical results on
convergence analysis for the proposed  algorithm
are derived.}
%
Experimental results on several benchmark datasets demonstrate that our proposed method outperforms existing 
\textit{state-of-the-art} (SOTA)
approaches 
 in terms of detection accuracy.
\end{abstract}

\IEEEpeerreviewmaketitle

\vspace{-0.4cm}
\section{\textbf{Introduction}}
 \textit{Hyperspectral images} (HSIs) capture a wealth of spatial and spectral information, with numerous spectral bands that provide a continuous and detailed representation of a scene.
  These rich spectral-spatial  information
   serves as a strong foundation for analyzing and identifying various ground objects.
\textcolor[rgb]{0.00,0.00,0.00}{As a result, HSIs have found widespread applications in fields such as
fusion \cite{Vivone2025, vivone2023multispectral 
},
restoration \cite{he2020non, pang2024hir
},
 denoising \cite{zhang2023hyperspectral55,  wang2023hyperspectral},
super-resolution \cite{he2022hyperspectral,  
xu2025nonlinear},
unmixing \cite{YangBin2018,   GuJiafeng2022,   
 Minglei2024},
classification \cite{cao2018hyperspectral, 
 cao2020hyperspectral}, and anomaly detection \cite{xu2022hyperspectral,su2021hyperspectral, hu2022hyperspectral}.}
%
%
Among these, \textit{hyperspectral anomaly detection} (HAD) has gained significant attention due to its importance in both civilian and military applications \cite{stein2002anomaly, nasrabadi2013hyperspectral}.
The primary goal of HAD is to detect anomalies by distinguishing them from the surrounding natural background.
 This process involves classifying a pixel as either an anomaly or part of the background based on the criterion that the spectral signature of an anomaly significantly deviates from that of the surrounding background. However, HAD task is inherently challenging because it often lacks prior knowledge of the spectral signatures for both the target anomaly and the background.
%

\vspace{-0.35cm}
\subsection{\textbf{Related Works}}

In recent years, HAD literature  
has seen significant growth, with methods evolving across various
advanced technical frameworks.
These methods can be grouped into the folowing four  main categories:
%
\textcolor[rgb]{0.00,0.00,0.00}{(i) approaches  using   statistical theory    \cite{reed1990adaptive, liu2021multipixel,  molero2013analysis, kwon2005kernel, guo2014weighted},
  (ii) approaches using deep learning  \cite{li2017transferred, jiang2020discriminative,  fu2021hyperspectral, xiang2022hyperspectral,
 jiang2021lren,  li2023lrr ,   wang2023pdbsnet, gao2023bs,wnagdegang, cheng2024deep, 
 GTHAD2025},
(iii)  approaches using matrix decomposition/representation  \cite{
  sun2014low55, zhang2015low,   li2020low,zhang2020improved,  chang2021orthogonal, 
   li2024feedback,
 li2014collaborative, tan2019anomaly, 
 wu2022hyperspectral,
 li2015hyperspectral, zhao2017hyperspectral ,
 xu2015anomaly, cheng2019graph, cheng2020total55, wang2023learning, guo2023learnable, ren2024hadgsm},
  and (iv) approaches using    tensor  decomposition/representation
 \cite{
 li2020prior, feng2022hyperspectral, shang2023hyperspectral,
 wang2020anomaly, 
 zhao2023hyperspectral, wang2023guaranteed,qin2024tensor, xiao2024hyperspectral33,
  wang2022learning1,  qin2023generalized1, 
 sun2022hyperspectral ,   yu2024generalized
 }.}

The typical  model of the first category is the \textit{Reed-Xiaoli} (RX) algorithm \cite{reed1990adaptive},
 which adheres to the fundamental premise that background pixels obey the
 Gaussian distribution, whereas anomalous pixels do not.
Additionally, Liu et al. \cite{liu2021multipixel} introduced two adaptive anomaly detectors induced by
 statistical knowledge   to fulfill  the HAD  task  in the presence of Gaussian noise.
To address the challenge of detecting anomalies in HSIs with complex backgrounds,
some  strengthened versions  of RX methods have been proposed,   such as
the  locally adaptable  RX  detector \cite{molero2013analysis},
the kernel-RX  \cite{kwon2005kernel}, and the weighted-RXD \cite{guo2014weighted}.
These improvements have significantly boosted the effectiveness of RX-based approaches in identifying complex anomalies.


The second type of  anomaly detectors
can be further divided into supervised, unsupervised, and self-supervised categories.
Supervised methods utilize \textit{convolutional neural networks} (CNN) to extract deep features and identify anomalies using labeled data \cite{li2017transferred}.
 However, the efficacy of this approach depends heavily on the availability of ground-truth data for training.
%
Consequently, several innovative unsupervised  architectures have been proposed for the HAD tasks,
including \textit{autoencoder} (AE) networks and \textit{generative adversarial networks} (GANs).
For example,
Jiang et al. \cite{jiang2020discriminative} suggested   an unsupervised discriminative reconstruction constrained GAN for HAD.
Xiang et al. \cite{xiang2022hyperspectral}  developed   a novel guided-AE-based
HAD method  to reduce the feature representation for the anomaly targets.
Li et al.  \cite{li2023lrr} put forward  an  interpretable deep unfolding network, called LRR-Net,  for the  HAD task.
\textcolor[rgb]{0.00,0.00,0.00}{At last, self-supervised methods \cite{wang2023pdbsnet, gao2023bs, wnagdegang} generate pseudolabels by predicting certain tasks without labeled data.}
Typically,  novel blind-spot self-supervised reconstruction networks were devised for the HAD task
in PDBSNet \cite{wang2023pdbsnet} and  BS$^{3}$LNet \cite{gao2023bs}.
Generally speaking, this category of HAD methods has significant advantages in extracting features from HSIs,
attributable to the strong  learning capabilities of neural networks.

%
The third type of HAD methods
include
\textit{collaborative representation} (CR)-based approaches \cite{li2014collaborative, tan2019anomaly, 
wu2022hyperspectral},
\textit{sparse representation} (SR)-based approaches \cite{ li2015hyperspectral, zhao2017hyperspectral},
and
\textit{low-rank representation} (LRR)-based approaches \cite{zhang2020improved,
 sun2014low55, zhang2015low,   li2020low,  chang2021orthogonal, 
 li2024feedback,
xu2015anomaly, cheng2019graph, cheng2020total55, wang2023learning, guo2023learnable, ren2024hadgsm}.
%
%
Depending on whether the strategy of  background dictionary construction  is employed,
 LRR-based approaches  can be further subdivided into two branches:
\textit{low-rank and SR} (LRaSR)  methods \cite{xu2015anomaly, cheng2019graph, cheng2020total55, wang2023learning, guo2023learnable, ren2024hadgsm}
and
\textit{low-rank and sparse matrix decomposition} (LRaSMD) methods
\cite{
sun2014low55, zhang2015low,   li2020low, zhang2020improved, chang2021orthogonal, 
li2024feedback}.
The core idea behind LRaSR methods is modeling the background as the product of a background dictionary and a coefficient matrix, while applying regularization constraints to the coefficient matrix. In contrast, LRaSMD-based methods decompose HSI data directly into a background matrix, an anomaly matrix, and a noise matrix, applying various prior constraints to each of these components to achieve the HAD task. However, these algorithms often involve transforming a $3$-D HSI cube into a $2$-D matrix, resulting in the loss of crucial structural information embedded in the HSI cube.



 \textcolor[rgb]{0.00,0.00,0.00}{Compared to the representation in vector/matrix structure,  tensor tends to more faithfully and accurately  uncover
  the intrinsic multidimensional structural information within HSI data
\cite{qin2022low,qin2023nonconvex, hou2021robust, wang2022tensor55555,  
wang2021generalized, liu2023tensor,
qin2024tensor}.}
Motivated by the advantages of tensor representation,
 the fourth kind of
HAD methods  has emerged as a captivating research focus.
The key to solving this kind of HAD problem   is how to excavate the prior structures of HSI data finely,
and encode them as certain regularization items  for guiding a sound separation of the background and anomaly components.
Among all tensor priors, the \textit{global low-rankness} (denoted as ``$\textbf{L}$")  property is particularly important.
%
Nevertheless,
different tensor decompositions define various notions of tensor rank.
\textcolor[rgb]{0.00,0.00,0.00}{
The most mainstream ones are related to the 
 \textit{CANDECOMP/PARAFAC} (CP) decomposition
 \cite{wang2020anomaly 
 },
the  Tucker decomposition
 \cite{li2020prior, 
 shang2023hyperspectral, ZhaoXiaobin2025}, the
 \textit{Tensor Singular Value Decomposition} (T-SVD)
 \cite{ 
  wang2022learning1, sun2022hyperspectral ,  qin2023generalized1, 
  zhao2023hyperspectral, wang2023guaranteed,yu2024generalized}, and the
 \textit{Tensor Ring} (TR) decomposition
 \cite{feng2023hyperspectral, qin2024tensor}.}
As a result,   many researchers have explored different strategies to characterize the $\textbf{L}$ 
prior in the context of HAD.
For instance,
Wang et al. \cite{wang2022learning1}  originally introduced the concept of tensor LRR into HAD task,
which  exploits the  $\textbf{L}$ prior of background tensor via the weighted \textit{tensor nuclear norm} (TNN).
\textcolor[rgb]{0.00,0.00,0.00}{To improve 
detection capacity, 
 the weighted tensor Schatten-$p$ ($0<p<1$) norm was utilized to
estimate the low-rank background  in the HAD   method named S$^{2}$ELR \cite{zhao2023hyperspectral}.}
\textcolor[rgb]{0.00,0.00,0.00}{To make full use of temporal continuity and spatial correlation, a 
novel HAD method based on nonconvex  tensor Gamma-norm
was proposed in \cite{zhao2023hyperspectral2222}.}
In addition,  Qin et al. \cite{qin2023generalized1} replaced the convex  TNN with a generalized nonconvex surrogate
and developed an effective  low-rank tensor representation  model for the HAD task.


 %
 In addition to  investigating the low-rank characteristics, 
 several studies have also focused on exploring
the  local smoothness (denoted as ``$\textbf{S}$") property  of the background to 
jointly boost
detection performance  \cite{li2020prior, sun2022hyperspectral ,shang2023hyperspectral,  wang2023guaranteed,  yu2024generalized,
 feng2023hyperspectral, qin2024tensor, bjZhaoXiaobin2024, ZhaoXiaobin2025}.
 For instance, Li et al. \cite{li2020prior} designed a new HAD algorithm, called PTA, which leverages the spectral low-rank property and spatial smoothness of the background.
 \textcolor[rgb]{0.00,0.00,0.00}{Zhao et al. \cite{ZhaoXiaobin2025} proposed a novel HAD
approach based on tensor adaptive reconstruction cascaded with
global and local feature fusion.}
 %
 %
%
By  utilizing  TNN and \textit{total variation} (TV) regularization constraints,
Sun et al.  \cite{sun2022hyperspectral} suggested  a novel HAD  algorithm, named LARTVAD.
Furthermore,  TR factorization and TV regularization constraints were introduced to explore the low-rankness and piecewise-smoothness  of the background  in all   dimensions \cite{feng2023hyperspectral}.
%
%
Nevertheless, the 
 performance of these 
 methods is   highly
 affected by    the trade-off  parameters imposed between $\textbf{L}$ and $\textbf{S}$ regularizers.
  %
 To alleviate the above issue,
under the high-order T-SVD framework \cite{qin2022low, qin2023nonconvex},
Wang et al.  \cite{wang2023guaranteed} developed a new regularizer, named T-CTV, to simultaneously encode
the $\textbf{L}$+$\textbf{S}$ priors of the background  with a unique concise term.
 Similarly, another method in  \cite{shang2023hyperspectral} discovered the sparsity of the core tensor from the Tucker decomposition of the gradient tensor and developed a regularization term that simultaneously captures the 
 $\textbf{L}$+$\textbf{S}$ priors of the background.
 %
   Despite these advances, the mentioned HAD methods still have limitations and cannot fully and flexibly extract prior information from complex backgrounds.

 %
\begin{figure*} 
\renewcommand{\arraystretch}{0.0}
\setlength\tabcolsep{0.3pt}
\centering
\begin{tabular}{c}
\centering
\includegraphics[width=6.8in, height=3.8in]{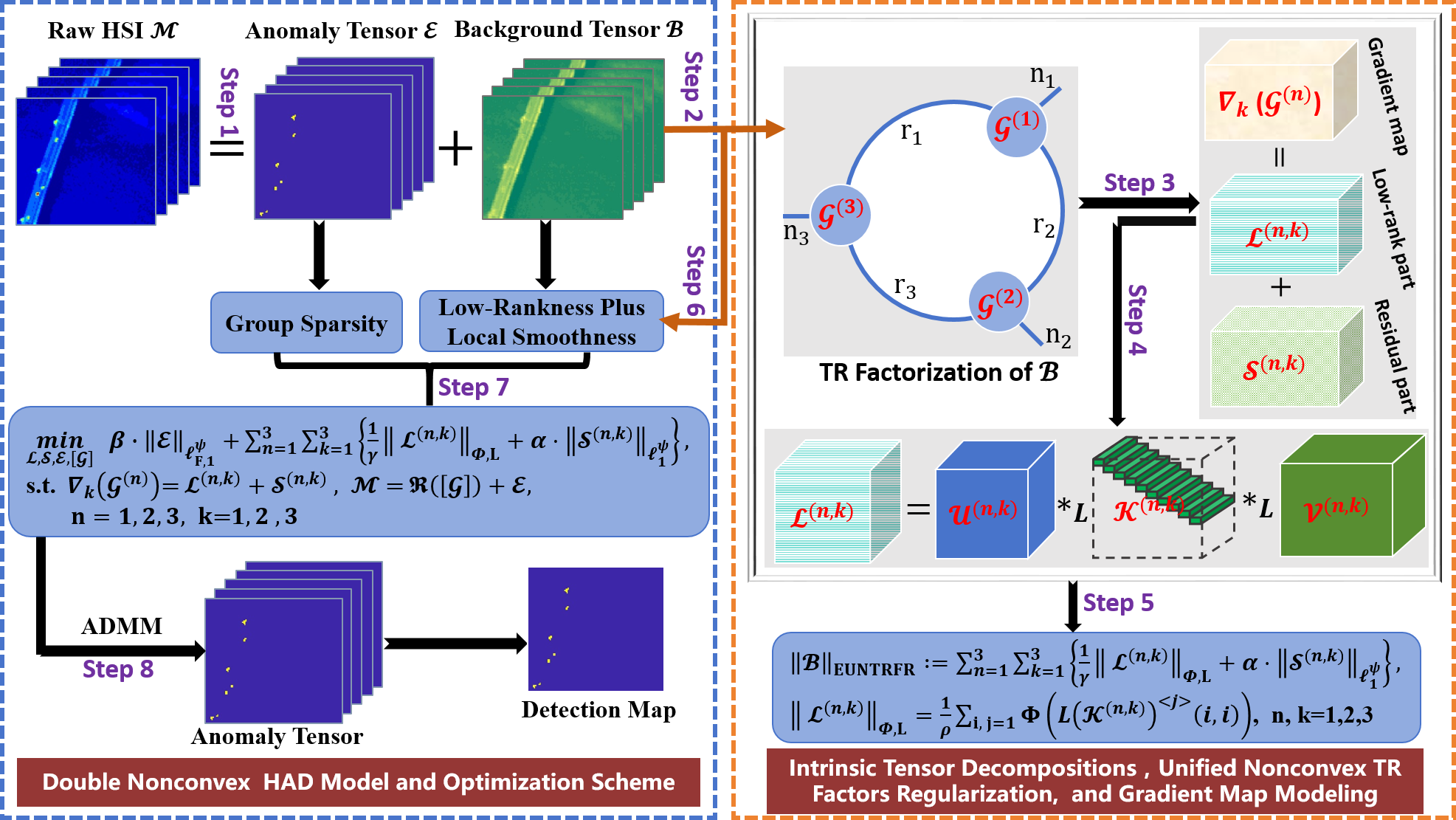} 
\end{tabular}
\caption{\textcolor[rgb]{0.00,0.00,0.00}{The detailed flowchart of the proposed
HAD-EUNTRFR method.
\textbf{Module 1 (Right side):}
Under a new prior representation paradigm, this module focuses on investigating  novel unified nonconvex tensor ring  factor regularization
strategies;
%
\textbf{Module 2 (Left side):} 
Based on the new regularization 
 methods tailored for the background 
  and 
  anomaly tensors,
the second module primarily aims to design  effective, scalable and reliable  HAD 
 model,
as well as 
optimization  algorithm with convergence guarantees.}}
\label{fig_chart}
\vspace{-0.655cm}
\end{figure*}

\vspace{-0.4cm}
\subsection{\textbf{Research Motivations}}

\subsubsection{\textbf{Why consider utilizing TR framework?}}
 The TR factorization  \cite{zhao2016tensor} approximates 
 a high-dimensional tensor as  a  multilinear products over a sequence of cyclically contracted  
 low-dimensional cores.
 %
%
In practice, TR factorization can enhance the compression capability and also improve the interpretability of latent factors.
The TR rank remains consistently invariant regardless of the cyclic permutation of the factors,
thereby flexibly and effectively 
capturing the underlying intermodal redundancy
within the tensor data.
Compared with other tensor decompositions,
the TR  decomposition  possesses more flexible, 
powerful and generalized modeling 
capacities 
in a wide range of applications 
\cite{yuan2019tensor, xu2021hyperspectral,
zhang2024hyperspectral, zhao2016tensor,
he2022hyperspectral, chen2022hyperspectral,
zhang2023hyperspectral, feng2023hyperspectral,
qin2024tensor,
wu2023tensor}.
Motivated by these merits,
in this paper,   we consider applying the advanced  TR factorization to the HAD problem.



\subsubsection{\textbf{Why consider 
devising 
 enhanced  unified
nonconvex TR 
factors regularization?}}

The TR rank minimization and low-rank 
 factorization   schemes
\cite{yuan2019tensor, xu2021hyperspectral,
zhang2024hyperspectral, zhao2016tensor,
he2022hyperspectral, chen2022hyperspectral,
zhang2023hyperspectral, feng2023hyperspectral,
qin2024tensor,
wu2023tensor}
 have  become a hot research topic
on the purpose of accurately uncovering the latent space of TR factors.
\textcolor[rgb]{0.00,0.00,0.00}{For example,
the nuclear norm regularization of the third TR cores  with
mode-$2$ unfolding was introduced to further exploit the global spectral low-rank property of 
high-resolution HSIs
\cite{he2022hyperspectral}.}
Moreover,
to boost the recovery performance,
Yuan et al. \cite{yuan2019tensor} imposed the the 
nuclear norm  regularization on
the unfolding matrices  of each  TR factor. 
Recently,
%
Wu et al. \cite{wu2023tensor} empirically and theoretically    investigated
the  physical interpretation of gradient TR factors, 
and found that
the mode-$2$ unfoldings of gradient TR factors inherit the    $\textbf{L}$+$\textbf{S}$ properties of the original tensor. 
Inspired by this exploration, they consider the low-rankness and sparsity priors of
gradient factors  to boost the performance and robustness of TR-based model.
\textcolor[rgb]{0.00,0.00,0.00}{However, although acceptable performance can be achieved from the above TR-factors-based modeling 
methods, 
they generally 
 unfolded the TR factors 
 to mode-$n$ matrix which may result in loss of optimality in the representation.
 Therefore, this paper considers investigating  a more essential regularization   method to reveal  
 the structural information of TR factors,
 and then applies it to \textit{hyperspectral anomaly detection} 
  task.}  



\textcolor[rgb]{0.00,0.00,0.00}{Existing HAD 
methods (e.g., \cite{feng2023hyperspectral, shang2023hyperspectral}) based on low-rank tensor 
modeling  suffer
from some issues, such as insufficient 
priors representation,
loose convex approximation, 
 and the absence of unified regularization form.
 The method we propose is intended to mitigate the aforementioned problems.}
%
\textcolor[rgb]{0.00,0.00,0.00}{Driven by  the proven  effectiveness and advantages of prior characterization through 
gradient maps-based modeling  strategy \cite{
 peng2020enhanced, 
peng2022exact, wang2023guaranteed 
}, 
 %
 we consider imposing an 
 effective   regularizer  that resembles 
  the T-CTV   
 constraint  \cite{wang2023guaranteed} 
 on the $3$-D TR 
 factors.}
This regularizer is intended to deeply capture the internal structure of TR factors, 
 serving as a more advantageous option compared to the matrix rank minimization or low-rank matrix factorization techniques employed in
 \cite{yuan2019tensor,he2022hyperspectral, wu2023tensor}.
 \textcolor[rgb]{0.00,0.00,0.00}{In other words,
 the key techniques underlying
 our  regularization  item will  draw on the idea of the TCTV-based HAD  method \cite{wang2023guaranteed},
  thus enabling it to simultaneously characterize the $\textbf{L}$+$\textbf{S}$   priors  of the background tensor.} 
 However, the convex T-CTV regularization scheme 
 still has some room for further
 improvement.
 Firstly,
the  T-CTV does not coordinate the low-rankness and sparsity of gradient maps well,
and its robustness requires further enhancement.
\textcolor[rgb]{0.00,0.00,0.00}{Secondly,  the T-CTV equally treats each singular component of the background  tensor in the gradient domain
 and neglects the 
 physical significance of the different singular values,
 which may lead to  a biased approximation 
 and 
 fail  to retain 
 some 
 major
 information.}
To resolve 
 the above issues, \textcolor[rgb]{0.00,0.00,0.00}{with the help of 
 a novel nonconvex regularization paradigm,
  we ultimately consider establishing  a generalized  and efficient
 method
  to encode intrinsic prior structures underlying gradient TR factors.} 
Within our
 prior characterization scheme, the interpretable gradient TR factors will adeptly 
 inherit
the $\textbf{L}$+$\textbf{S}$  
priors of the original background.
More importantly, how to enhance the robustness of the proposed regularization  term is also taken into account.

\vspace{-0.25cm}
\subsection{\textbf{Proposed HAD  Method}}

 In this study,    we put forward a novel 
 unified nonconvex   HAD method termed HAD-EUNTRFR.   
 The detailed flowchart of the proposed  framework is provided in Figure \ref{fig_chart}. Specifically, in \textbf{Step 1},
 the original HSI ${\boldsymbol{\mathcal{M}}}$ are separated into the background and anomaly components
 (i.e., ${\boldsymbol{\mathcal{B}}}$ and
${\boldsymbol{\mathcal{E}}}$)
using the concept of \textit{tensor robust principal component analysis} (TRPCA).
 %
 \textcolor[rgb]{0.00,0.00,0.00}{The TR decomposition  is then performed on the background tensor
 ${\boldsymbol{\mathcal{B}}}$
 in \textbf{Step 2}, i.e.,
${\boldsymbol{\mathcal{B}}}=
\Re
(\textbf{[}{\boldsymbol{\mathcal{G}}}\textbf{]})$, $\textbf{[}{\boldsymbol{\mathcal{G}}}\textbf{]}:=
\{{\boldsymbol{\mathcal{G}}}^{(1)}, {\boldsymbol{\mathcal{G}}}^{(2)}, {\boldsymbol{\mathcal{G}}}^{(3)}\}$,}
\textcolor[rgb]{0.00,0.00,0.00}{where
$
\Re
(\textbf{[}{\boldsymbol{\mathcal{G}}}\textbf{]})$
is the TR decomposition of ${\boldsymbol{\mathcal{B}}}$,
$\{{\boldsymbol{\mathcal{G}}}^{(k)}\}_{k=1}^{3}$ are the TR factors.}
In \textbf{Step 3},
a
joint low-rank plus sparse decomposition is performed on each gradient TR factor
(namely, TR factors in the gradient domain),
 i.e.,  
$\nabla_{k}  ({\boldsymbol{\mathcal{G}}}^{(n)} )= {\boldsymbol{\mathcal{L}}}^{(n,k)}+ {\boldsymbol{\mathcal{S}}}^{(n,k)}, n,k=1,2,3$.
\textcolor[rgb]{0.00,0.00,0.00}{The low-rank part is further decomposed by T-SVD in \textbf{Step 4}, i.e.,
${\boldsymbol{\mathcal{L}}}^{(n,k)}  = {\boldsymbol{\mathcal{U}}}^{(n,k)}  {{*}_{\mathfrak{L}}} {\boldsymbol{\mathcal{K}}}^{(n,k)} {*}_{\mathfrak{L}}
{ ({\boldsymbol{\mathcal{V}}}^ {(n,k)})}^{\mit{T}}$,}
\textcolor[rgb]{0.00,0.00,0.00}{where
${\boldsymbol{\mathcal{U}}}^{(n,k)},   { {\boldsymbol{\mathcal{V}}}^ {(n,k)}} $ are  orthogonal tensors,
 ${\boldsymbol{\mathcal{K}}}  ^{(n,k)}$ is  a f-diagonal tensor, ${{*}_{\mathfrak{L}}}$ denotes the \textit{tensor-tensor product}}.
In \textbf{Step 5},
we devise an 
enhanced  
unified
nonconvex TR  factors regularization   
(\textit{please see \ref{nonconvex-regulari} for more details}).
 %
 \textcolor[rgb]{0.00,0.00,0.00}{The regularization scheme discussed in this 
 text
  is designed to be robust against the  TR rank selection, while leveraging the prior information of gradient TR factors effectively.
This 
approach
 enables the interpretable gradient
TR factors to skillfully inherit the low-rankness and smoothness 
of the
original background.}
 The procedure outlined in \textbf{step  2-5} constitutes 
 %
a new joint $\textbf{L}$+$\textbf{S}$ prior characterization paradigm for the background tensor.
In \textbf{Steps 6-7},
 the method combines the proposed EUNTRFR regularizer with another  anomaly-sparsity regularizer, leading to the formulation of a new generalized nonconvex HAD model 
(\textit{please see \ref{non-model} for more details}).
%
The formulated  model is then solved in \textbf{Step 8} using 
Algorithm  \ref{algorithm1had},
which is derived from the ADMM framework, as detailed in Section 
\ref{non-algorithms}.

\vspace{-0.25cm}
\subsection{\textbf{Main Contributions}}
  The main contributions of this article are as follows:

  1)
  We introduce an innovative unified nonconvex HAD method, termed   HAD-EUNTRFR, by integrating several  crucial 
  technologies such as 
   TRPCA, tensor decompositions,
  tensor-correlated TV regularization, and low-rank plus  sparse gradient map modeling. This method
   concisely and  effectively
     captures both global correlations and local smoothness in the background within the spectral-spatial domains, while also addressing the structured sparsity of anomalies.


  2)
%
Building on the powerful representation capabilities of TR factorization, we propose a novel generalized nonconvex HAD model. This model incorporates a unified nonconvex regularization term known as UNTRFR, and its enhanced version, EUNTRFR
(\textit{please see \ref{nonconvex-regulari} for more details}). These components efficiently encode both low-rankness and sparsity in the gradient TR factors, providing a concise and effective approach for anomaly detection.

  3)
 In the algorithmic development, we provide  
 new solution paradigm
 tailored for 
 key subproblems involving a family of generalized nonconvex functions.
 We then
 derive an optimization algorithm based on the ADMM framework, enabling the efficient solution of the formulated HAD model.
   \textcolor[rgb]{0.00,0.00,0.00}{Theoretical results on convergence analysis for the proposed nonconvex algorithm are provided.}
  Experimental results on extensive HSI datasets show that our approach significantly outperforms current state-of-the-art
  methods, effectively suppressing the background and enhancing the detection of anomalous targets.


\vspace{-0.3cm}
\section{\textbf{notations and preliminaries}}\label{nota}
%
Some
frequently-employed symbols and preliminaries
are summarized in this section.

\textcolor[rgb]{0.00,0.00,0.00}{We use $a$, $\textbf{a}$, $\textbf{A}$, and $\bm{\mathcal{A}}$ to denote scalars, vectors, matrices, and tensors, respectively.
For a matrix $\bm{{A}} \in \mathbb{R}^{n_1 \times  n_2}$,
$\bm{I}_{n} \in \mathbb{R} ^{n\times n}$, $\operatorname{Tr}(\bm{A})$,
 $ {{\bm{A}}}^{\mit{H}} ({{\bm{A}}}^{\mit{T}})$,
  $ \langle {{\bm{A}}},{{\bm{B}}} \rangle=\operatorname{Tr} ({{\bm{A}}}^{\mit{H}} \cdot {{\bm{B}}})$ and
   ${\|{{\bm{A}}}\|}_{\star}= {\big(\sum_{i}  \; \big|\sigma_{i}({{\bm{A}}})\big| \big)}$
 denote its
 identity matrix,  trace, conjugate transpose (transpose), inner product and  nuclear   norm, respectively.
For an order-$N$ tensor $\boldsymbol{\mathcal{A}} \in \mathbb{R}^{n_1 \times  \cdots \times n_N}$,
$\boldsymbol{\mathcal{A}}_ {i_1,\cdots,i_N}$  denotes
its $(i_1,\cdots,i_N)$-th element,
${{\bm{A}}_{(k)}} \in \mathbb{R}^{n_k\times
   \prod_{j\neq k } {n_j}
    } $
     denotes its
      \textit{classical mode-$k$ unfolding},
      $ {{\bm{A}}} _{<k>} \in \mathbb{R}^{   n_k \times (n_{k+1}  \cdots n_N n_1 \cdots n_{k-1})  } $
denotes its  \textit{reversed mode-$k$ unfolding},
%
and
     $
    {\boldsymbol{\mathcal{A}}}^{<j>}:={\boldsymbol{\mathcal{A}}}{(:,:,i_3,\cdots, i_{N})},
    j={\sum_{a=4}^{N} }   {(i_a-1){\Pi}_{b=3}^{a-1}n_b}+i_3$
is called as  its 
\textit{face slice}.
The \textit{inner product} of two tensors $\boldsymbol{\mathcal{A}}$ and $\boldsymbol{\mathcal{B}}$ with the same size is defined as
$\langle {\boldsymbol{\mathcal{A}}},{\boldsymbol{\mathcal{B}}} \rangle  = {\sum}_{j=1}^{n_3 n_4 \cdots n_N} \langle {\bm{\mathcal{A}}}^{<j>},{\bm{\mathcal{B}}}^{<j>} \rangle$.
The \textit{$\ell_1$-norm}, \textit{Frobenius norm},
  \textit{infinity 
  norm}
  and \textit{$\ell_{{\mathnormal{F}},1}$-norm 
  }
of $\boldsymbol{\mathcal{A}}$ are defined as
${\|{\boldsymbol{\mathcal{A}}}\|}_{1}= ({\sum _ {i_{1}\cdots i_{N}}
    }
    |{\boldsymbol{\mathcal{A}}_{i_{1} \cdots i_N}}|)$,
 ${\|{\boldsymbol{\mathcal{A}}}\|}_{\mathnormal{F}}= {(\sum_{i_{1} \cdots i_N} |{\boldsymbol{\mathcal{A}}_{i_{1} \cdots i_N}}|^{2})^\frac{1}{2}}$,
 $\| {\boldsymbol{\mathcal{A}}}\|_{\infty}  = 
\max_{i_1, \cdots, i_N}
| {\boldsymbol{\mathcal{A}}}_{i_1, \cdots, i_N }
|$,
$\| {\boldsymbol{\mathcal{A}}}\|_{\mathnormal{F},1} = 
\sum_{i_1, i_2}
\|{\boldsymbol{\mathcal{A}}}_{i_1 i_2 : \cdots : }
\|_{\mathnormal{F}}$,
respectively.
    \textit{The mode-$k$ product of tensor $\boldsymbol{\mathcal{A}}
    \in \mathbb{R}^{n_1 \times  \cdots \times n_k \times \cdots \times n_N}
    $ with matrix ${\bm{M}}
    \in \mathbb{R}^{J \times  n_k}
    $} is  $\boldsymbol{\mathcal{B}} =\boldsymbol{\mathcal{A}} \;{\times}_{k}\; \bm{M}
    \in \mathbb{R}^{n_1 \times  \cdots \times J \times \cdots \times n_N}
    $,
    where
  $ \boldsymbol{\mathcal{B}} =\boldsymbol{\mathcal{A}}{\times}_{n} {\bm{M} } $
   $\Longleftrightarrow$
   ${\bm{B}}_{(n)} = {\bm{M}} \cdot {\bm{A}}_{(n)}$.}

\textit{\textbf{
T-SVD 
framework}}:
\textcolor[rgb]{0.00,0.00,0.00}{Let  ${\mathfrak{L}}(\boldsymbol{\mathcal{A}})$ or $\boldsymbol{\mathcal{A}}_{{\mathfrak{L}}}$ represent   the result of invertible linear transforms $\mathfrak{L}$ on $\boldsymbol{\mathcal{A}} \in \mathbb{R}^{n_1\times \cdots \times  n_N}$, i.e.,}
 \begin{align}\label{trans}
 \textcolor[rgb]{0.00,0.00,0.00}{\mathfrak{L}(\boldsymbol{\mathcal{A}}) =\boldsymbol{\mathcal{A}} \;{\times}_{3} \;\bm{U}_{n_3} \;{\times}_{4} \;\bm{U}_{n_4} \cdots
    {\times}_{N} \;\bm{U}_{n_N},}
\end{align}
 \textcolor[rgb]{0.00,0.00,0.00}{where   the transform matrices $\bm{U}_{n_i} \in \mathbb{C}^{n_i \times n_i}$ of $\mathfrak{L}$ satisfies:}
   \begin{align}\label{orth}
  \textcolor[rgb]{0.00,0.00,0.00}{{{\bm{U}}}_{n_i} \cdot {{\bm{U}}}^{\mit{H}}_{n_i}={{\bm{U}}}^{\mit{H}}_{n_i} \cdot {{\bm{U}}}_{n_i}=\alpha_i {{\bm{I}}}_{n_i},  \forall i\in \{3,\cdots,N\},}
   \end{align}
   \textcolor[rgb]{0.00,0.00,0.00}{in which $\alpha_i>0$  is a  constant.
  The inverse operator of $\mathfrak{L}(\boldsymbol{\mathcal{A}})$  is defined as
   $ \mathfrak{L}^{-1} (\boldsymbol{\mathcal{A}}) =\boldsymbol{\mathcal{A}} \;{\times}_{N} \;\bm{U}^{-1}_{n_N} \;{\times}_{{N-1}} \;
   {   {\bm{U}} _  { {n} _{{N-1}}}  ^{-1}    }   \cdots
    {\times}_{3} \;\bm{U}^{-1}_{n_3}$,
    %
    and  $\mathfrak{L}^{-1} (\mathfrak{L}(\boldsymbol{\mathcal{A}}))=\boldsymbol{\mathcal{A}}$.}

    \textcolor[rgb]{0.00,0.00,0.00}{The \textit{tensor-tensor product} 
    of
     $\boldsymbol{\mathcal{A}}$ and $\boldsymbol{\mathcal{B}}$
    under transform $\mathfrak{L}$ 
     is defined as
    $
     \boldsymbol{\mathcal{C}}={\boldsymbol{\mathcal{A}}}{*}_{\mathfrak{L}} {\boldsymbol{\mathcal{B}}}
     =\mathfrak{L}^{-1}
\big({\mathfrak{L}}(\boldsymbol{\mathcal{X}}) \bigtriangleup
{\mathfrak{L}}(\boldsymbol{\mathcal{Y}})\big)
    $,
    where
$ \bigtriangleup  $  denotes the face-wise product
($\boldsymbol{\mathcal{P}}=\boldsymbol{\mathcal{M}} \bigtriangleup
     \boldsymbol{\mathcal{N}} $ implies 
    $ {\boldsymbol{\mathcal{P}}}^{<j>}={\boldsymbol{\mathcal{M}}}^{<j>}\cdot{\boldsymbol{\mathcal{N}}}^{<j>}, \;j=1,\cdots,n_3\cdots n_N$)
    \cite{qin2022low,qin2023nonconvex}.
     \textcolor[rgb]{0.00,0.00,0.00}{Transform $\mathfrak{L}$-based \textit{T-SVD} of $\boldsymbol{\mathcal{A}}$  is denoted as
    ${\boldsymbol{\mathcal{A}}}={\boldsymbol{\mathcal{U}}} {*}_{\mathfrak{L}}   {\boldsymbol{\mathcal{K}}} {*}_{\mathfrak{L}}
{\boldsymbol{\mathcal{V}}}^{\mit{T}}$,
where ${\boldsymbol{\mathcal{U}}} $ and
${\boldsymbol{\mathcal{V}}} $ are  orthogonal,
 ${\boldsymbol{\mathcal{K}}} $ is  f-diagonal,
  $ {{\boldsymbol{\mathcal{V}}}}^{\mit{H}} ({{\boldsymbol{\mathcal{V}}}}^{\mit{T}} )$
denotes conjugate  transpose (transpose) \cite{qin2022low,qin2023nonconvex}.}}

\textit{\textbf{
TR algebraic framework}}:
%
\textcolor[rgb]{0.00,0.00,0.00}{For an order-$N$ tensor ${\boldsymbol{\mathcal{A}}} \in \mathbb{R}^{ n_1\times \cdots \times n_N} $ with TR rank $[r_1, \cdots, r_N]$, 
its  TR decomposition is  represented as a sequence of circularly contracted   core tensor
 ${\boldsymbol{\mathcal{G}}}^{(k)}  \in \mathbb{R}^{ r_k\times n_k \times r_{k+1}} $,
  $k=1,2, \cdots, N$, with $r_{N+1}=r_1$.
  Specifically, the element-wise relation of tensor ${\boldsymbol{\mathcal{A}}} $ and its TR core tensors
$\{{\boldsymbol{\mathcal{G}}}^{(k)}\}_{k=1}^{N}$ is defined as}
\begin{align*}
\textcolor[rgb]{0.00,0.00,0.00}{{\boldsymbol{\mathcal{A}}}_{i_1 \cdots i_N }= \operatorname{Tr}\Big ( \prod_{k=1} ^ {N}
{\boldsymbol{\mathcal{G}}}^{(k)} (:,i_k,:) \Big ),}
\end{align*}
\textcolor[rgb]{0.00,0.00,0.00}{where $ {\boldsymbol{\mathcal{G}}}^{(k)} (:,i_k,:) \in  \mathbb{R}^{ r_k\times  r_{k+1}} $
is the $i_k$-th slice matrix of ${\boldsymbol{\mathcal{G}}}^{(k)}$ along mode-$2$.
In this article, the TR decomposition  is abbreviated as
${\boldsymbol{\mathcal{A}}}=
\Re
(\textbf{[}{\boldsymbol{\mathcal{G}}}\textbf{]})$, $\textbf{[}{\boldsymbol{\mathcal{G}}}\textbf{]}:=
\{{\boldsymbol{\mathcal{G}}}^{(1)}, \cdots, {\boldsymbol{\mathcal{G}}}^{(N)}\}$.}

\textcolor[rgb]{0.00,0.00,0.00}{Given a TR decomposition  ${\boldsymbol{\mathcal{A}}}= \Re (\textbf{[}{\boldsymbol{\mathcal{G}}}\textbf{]})$,
 its mode-$k$ unfolding matrix can be written as}
\begin{equation}\label{eq:theoremals}
\textcolor[rgb]{0.00,0.00,0.00}{\bm{ A}_{<k>} = \bm{G}  ^{(k)} _{(2)} \cdot (\bm{G}
^{(\neq k)}_{<2>})
^{\mit{T}},}
\end{equation}
\textcolor[rgb]{0.00,0.00,0.00}{where $
{\boldsymbol{\mathcal{G}}} ^{(\neq k)}$ is a subchain tensor obtained by merging all
 cores except the $k$-th core $ {\boldsymbol{\mathcal{G}}} ^{( k)}$
\cite{zhao2016tensor}.}

\vspace{-0.5cm}

\section{\textcolor[rgb]{0.00,0.00,0.00}{\textbf{Unified Nonconvex HAD 
Framework}}}  \label{model}


\subsection{\textbf{Unified Nonconvex Regularizers}} \label{nonconvex-regulari} 

\subsubsection{\textbf{Unified Anomaly-Sparsity  Regularizer}}
In order to effectively recognize the anomalous component  existed in HSI data,
we hence consider defining  the following unified  nonconvex  regularization penalty, i.e.,
\begin{equation} \label{unified-regulare11}
\Upsilon( {\boldsymbol{\mathcal{E}}})
:=
\psi
\big(
h(
{\boldsymbol{\mathcal{E}}}
)
\big),
\end{equation}
\textcolor[rgb]{0.00,0.00,0.00}{where $\psi (\cdot): \mathbb{R}  \rightarrow \mathbb{R}$
 stands for a generalized
nonconvex function.}
\textcolor[rgb]{0.00,0.00,0.00}{Here,  two types of sparse anomalies are taken into account.}
If  the anomaly tensor ${\boldsymbol{\mathcal{E}}}$  has structured  sparsity
  on the tubes, i.e.,  $h(\cdot)=\|\cdot\|_{{\mathnormal{F}},1}$,
then  we have
\begin{equation} \label{unified-regular22222}
\|{\boldsymbol{\mathcal{E}}}\|_  {\ell _ {\mathnormal{F},1}^{ \psi}}
:= \psi (\| {\boldsymbol{\mathcal{E}}}\|_{\mathnormal{F},1} ) =
 \sum_{{i_1}=1}^{n_1}  \sum_{{i_2}=1}^{n_2}
\psi
\big(
\|{\boldsymbol{\mathcal{E}}}_{i_1 i_2 :  }
\|_{\mathnormal{F}} \big
).
\end{equation}
If  the tensor ${\boldsymbol{\mathcal{E}}}$ is an entry-wise anomaly tensor, i.e.,
$h(\cdot)=\|\cdot\|_{1}$,  then we have
\begin{equation} \label{unified-regular22}
\|{\boldsymbol{\mathcal{E}}}\|_  {\ell _ {1}^{ \psi}}
:= \psi (\| {\boldsymbol{\mathcal{E}}}\|_{1} ) =
 \sum_{{i_1}=1}^{n_1}  \sum_{{i_2}=1}^{n_2}  \sum_{{i_3}=1}^{n_3}
\psi
\big(
|{\boldsymbol{\mathcal{E}}}_{i_1 i_2  i_3} 
|\big).
\end{equation}

\begin{Assumption} \label{assumpt}
\textcolor[rgb]{0.00,0.00,0.00}{The generalized nonconvex function 
\textcolor[rgb]{0.00,0.00,0.00}{$\psi (\cdot): \mathbb{R} \rightarrow \mathbb{R}$}
 satisfies the following assumptions:}
\begin{itemize}
     \item
     \textbf{(I)}:
   \textcolor[rgb]{0.00,0.00,0.00}{\textcolor[rgb]{0.00,0.00,0.00}{$\psi (\cdot): \mathbb{R} \rightarrow \mathbb{R}$}
     is proper, lower semi-continuous
     and symmetric with respect to y-axis;}

     \item
      \textbf{(II)}:
      $\psi (\cdot)$  is concave and monotonically increasing  on $[0,\infty)$
 with   $\psi(0)=0
$.
   \end{itemize}
\end{Assumption}


   Some
   popular nonconvex penalty functions satisfying   the above Assumption \ref{assumpt}
   are summarized as follows, where we only consider the parameters of the nonconvex function on $[0,\infty)$
   since they are
    symmetric with respect to y-axis.

\begin{itemize}

\item  \textbf{Lp \cite{marjanovic2012l_q}:}  the $\ell_{p}$  penalty  function is defined by $\psi^{\text{Lp}}(x)= |x|^{p}$,
 where $0<p<1$.
\item  \textbf{Log
\cite{gong2013general}:}  the logarithmic (Log) function
is defined by $\psi^{\text{Log}}(x)= \log(  1+ \frac{|x|}{\theta} ) $,
 where $\theta >0$.
 \item  \textbf{MCP \cite{zhang2010nearly}:}  the minimax concave penalty  (MCP)
 function is defined by
  $$
  \psi^{\text{MCP}}(x)=
  \begin{cases}
  x -  \frac{x^2}{2\eta}, & 0\leq  x \leq  \eta,\\
\frac{ \eta} {2}, & x >  \eta > 0.
\end{cases}$$

  \item  \textbf{Capped Lp:}  the capped 
 $\ell_{p}$  penalty  function is defined by $\psi^{\text{CapLp}}(x)=
 \min\{  1, \frac{x^p}{v^p} \}
 $,
 where $v>0$, $0<p<1$.

 \item  \textbf{Capped Log:}  the capped logarithm function  %
is defined by $\psi^{\text{CapLog}}(x)= \min\{  1, \frac{1}{\psi^{\text{Log}}(v)} \psi^{\text{Log}}(x)  \} $,
 where $v >0$.

 \item  \textbf{Capped MCP:}  the capped   MCP 
 function is defined by
 $$
  \psi^{\text{CapMCP}}(x)= \min\Big\{1,   \frac{2\eta}{v(2\eta -v)}   \psi^{\text{MCP}}(x)\Big\}, 0< v<\eta
  .
 $$
\end{itemize}
\noindent
It is worth noting that the above-mentioned capped folded concave functions originate from the  reference \cite{pan2021group}.
Besides, the proximal operator of $v$ with respect to $\psi(\cdot)$ is defined as
\begin{equation}\label{equ_gst1}
\operatorname{\textit{Prox}}_ {\psi, \mu} (v)=
\arg \min_{x} \Big \{ \mu \psi(x)
 + \frac{1}{2} {(x-v)}^2 \Big \},
\end{equation}
where $\mu>0$ is a penalty parameter. For the special forms of the nonconvex  function $\psi(\cdot)$,
their  proximal mappings often have analytical expressions, which can be found in  the relevant   references
\cite{marjanovic2012l_q, gong2013general, zhang2010nearly, pan2021group}.

\subsubsection{\textbf{Novel  Regularizer Encoding the
Prior Structures
of  TR Factors}}  \label{nonconvex-regulari-sss}
\textcolor[rgb]{0.00,0.00,0.00}{Based on the characteristics of HSI datasets,
we put forward a novel 
\textit{unified nonconvex TR factors regularization}  
(UNTRFR).} \textcolor[rgb]{0.00,0.00,0.00}{The proposed regularization 
scheme, along with its enhanced variant,
is specifically
tailored
to encode the crucial  prior structures (i.e., $\textbf{L}$+$\textbf{S}$)   of  the background tensor ${\boldsymbol{\mathcal{B}}}
\in\mathbb{R}^{n_1\times n_2 \times n_3}$.
 Before formally introducing this novel regularizer,
 we first define a \textit{generalized nonconvex   
 tensor correlated
total variation} (GNTCTV)  norm 
 and its enhanced version under the T-SVD framework.}

\vspace{-0.13255cm}
\begin{Definition}\label{def10diff}
\textbf{(Gradient tensor \cite{wang2023guaranteed})}
\textcolor[rgb]{0.00,0.00,0.00}{For   ${\boldsymbol{\mathcal{B}}} \in \mathbb{R}^{ n_1\times \cdots \times n_d} $,
its gradient tensor along the $k$-th mode is defined as}
\begin{align}\label{gradient-tensor}
\textcolor[rgb]{0.00,0.00,0.00}{\nabla_{k} ({\boldsymbol{\mathcal{B}}})}
&\textcolor[rgb]{0.00,0.00,0.00}{: =
\boldsymbol{\mathcal{B}} \;{\times}_{k} \;\bm{D}_{n_k}, \;\; k=1,2,\cdots, d,}
\end{align}
\textcolor[rgb]{0.00,0.00,0.00}{where $\bm{D}_{n_k}$ is a row circulant matrix
of $(-1,1, 0, \cdots,0)$.}
\end{Definition}

\vspace{-0.13255cm}

\begin{Definition}\label{gtctv}
\textcolor[rgb]{0.00,0.00,0.00}{For any  
${\boldsymbol{\mathcal{B}}} \in \mathbb{R}^{ n_1\times n_2 \times n_3} $,
denote $\Gamma \in \{1,2,3\}$ as a priori set consisting of directions along which  ${\boldsymbol{\mathcal{B}}} $
equips $\textbf{L}$+$\textbf{S}$ priors,
and 
 $\nabla_{k} ({\boldsymbol{\mathcal{B}}})$, $k\in \Gamma$
as its correlated gradient tensors.
Then,  its GNTCTV  norm is defined as} 
\begin{align}
\| {\boldsymbol{\mathcal{B}}}\|_{ \textit{GNTCTV}}
&:= \frac{1}{\gamma} \sum_{k=1} ^{ 3}
 \big\|
 \nabla_{k} ({\boldsymbol{\mathcal{B}}})
 \big\|_
{
\Phi,
\mathfrak{L} },
\notag \\ \label{GTCTV}
&
=\frac{1}{\gamma} \sum_{k=1} ^{ 3}
 \Big(
\frac{1}{\rho} \sum_{i=1}^{\min(n_1,n_2)} \sum_{j=1}^{n_3}
\Phi
\big(
 \sigma_{ij}^{(k)}
\big)
\Big),
\end{align}
where
\textcolor[rgb]{0.00,0.00,0.00}{$\gamma := \sharp\{\Gamma\}$ equals to the cardinality of $\Gamma$,
 $\rho>0$ 
 is a constant determined by transform $\mathfrak{L}$
 satisfying Formulas (\ref{trans})-(\ref{orth}),
${\boldsymbol{\mathcal{K}}}^{(k)}$  is  the T-SVD  component of
$\nabla_{k}  ({\boldsymbol{\mathcal{B}}} )  = {\boldsymbol{\mathcal{U}}}^{(k)}  {{*}_{\mathfrak{L}}} {\boldsymbol{\mathcal{K}}}^{(k)} {*}_{\mathfrak{L}}
{ ({\boldsymbol{\mathcal{V}}}^ {(k)})}^{\mit{T}}$,
 $\sigma_{ij}^{(k)}=  { {\mathfrak{L}}({{{\boldsymbol{\mathcal{K}}}}^ {(k)} })} ^ {<j>}  (i,i)$,
  and \textcolor[rgb]{0.00,0.00,0.00}{$\Phi (\cdot):\mathbb{R} \rightarrow \mathbb{R}$
    is 
  a generalized
nonconvex function,  which  has the same properties as the 
function $\psi(\cdot)$ defined in (\ref{unified-regulare11}).}}
\end{Definition}
%
\begin{Remark}
\textcolor[rgb]{0.00,0.00,0.00}{Drawing inspiration from the principles  of \textit{correlated total variation} (CTV) \cite{peng2020enhanced,peng2022exact}
 and %
 \textit{tensor CTV} (T-CTV)
 \cite{wang2023guaranteed}   methods,
we came up with the aforementioned GNTCTV regularizer (\ref{GTCTV}),}
which can  simultaneously characterize both $\textbf{L}$ and $\textbf{S}$ priors of HSI's background
 with a unique  term.
 This single regularizer can promote the two priors concurrently, and overcome
 the difficulty of tuning  the trade-off  parameter imposed between these two regularizers.
In addition, compared with the convex  T-CTV constraint,
the GNTCTV regularizer involving  a class of nonconvex penalty functions
considers the importance of different singular components of each  gradient tensor,
thus providing better approximation to the background tensor.
\textcolor[rgb]{0.00,0.00,0.00}{Note that
when the  nonconvex function $\Phi (\cdot)$ is set to be an  $\ell_1$ 
penalty function,
the GNTCTV  can be degenerated to  the T-CTV regularizer  proposed in \cite{wang2023guaranteed}.} 
%
%
When the order of  tensor ${\boldsymbol{\mathcal{B}}}$ is $2$, the regularizer (\ref{GTCTV}) is equivalent to
%
\begin{align} \label{gnctv}
\big \|
  {{\bm{B}}}
 \big\|_
{\operatorname{GNCTV} }
:=
 \big\|
 \nabla_{1} ({{\bm{B}}})
 \big\|_
{\Phi }
=
\sum_{i=1}^{\min(n_1,n_2) }
\Phi
\big(
  {{\bm{S}}}  (i,i)
\big),
\end{align}
%
\begin{figure} 
\renewcommand{\arraystretch}{0.0}
\setlength\tabcolsep{0.3pt}
\centering
\begin{tabular}{c}
\centering

\includegraphics[width=2.8in, height=1.5in]{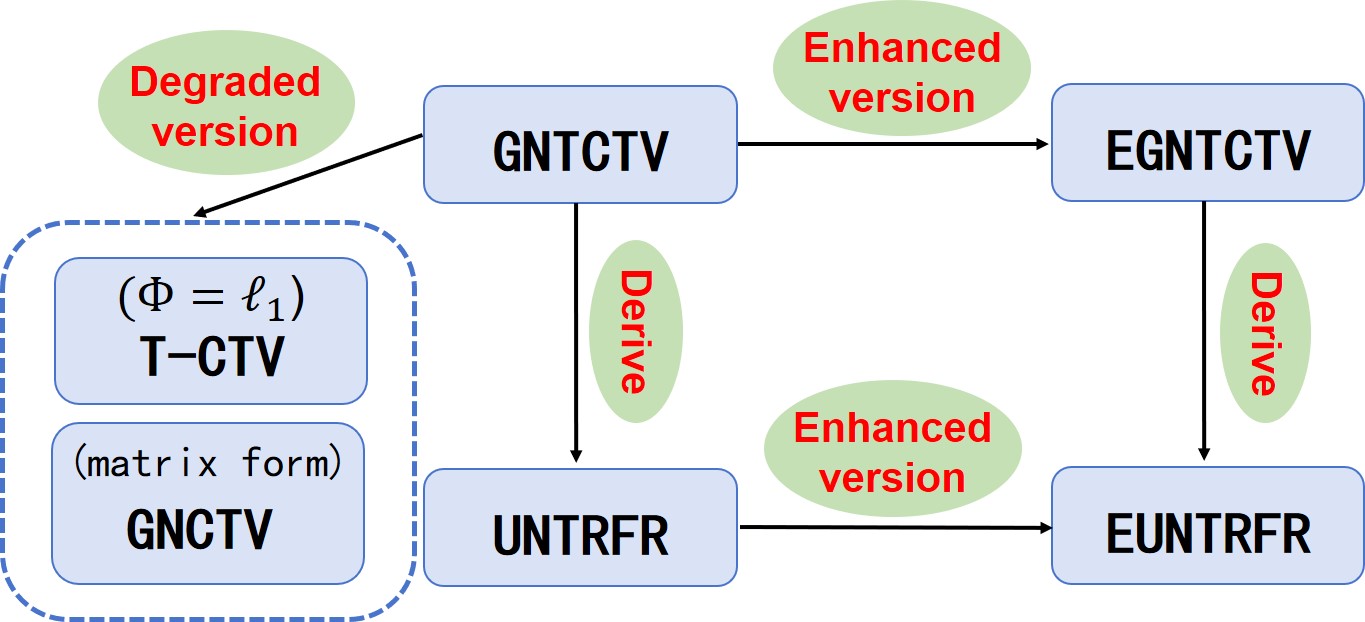} 
\end{tabular}
\caption{\textcolor[rgb]{0.00,0.00,0.00}{The relationship 
of
various 
regularization  methods.}
}
\label{fig_chart111}
\vspace{-0.70cm}
\end{figure}
where
${{\bm{S}}} $  is  the SVD  component of
$\nabla_{1}  ({{\bm{B}}}  )  = {{\bm{U}}}  {{\bm{S}}}
{{\bm{V}}}^{\mit{T}}$.
In this article, the regularizer  (\ref{gnctv}) is called  as the \textit{generalized nonconvex CTV} (GNCTV)
 of  matrix $\bm{B}$,
which  can be   viewed as the  nonconvex extensions of CTV norm  \cite{peng2022exact}.
\end{Remark}

\vspace{-0.1355cm}

To further  enhance the robustness of the formulated  regularizer (\ref{GTCTV}),
we first proposed decomposing  the gradient tensor $\nabla_{k}  ({\boldsymbol{\mathcal{B}}} ), k=1,2,3$,
into a low-rank term  $ {\boldsymbol{\mathcal{L}}}_{k}$ and a 
sparse  term   $ {\boldsymbol{\mathcal{S}}}_{k}$, i.e.,
$\nabla_{k}  ({\boldsymbol{\mathcal{B}}})= {\boldsymbol{\mathcal{L}}}_{k}+ {\boldsymbol{\mathcal{S}}}_{k}$.
 Then,  the  generalized nonconvex regularizers are   utilized to encode them. 
 In other words,  considering that the background tensor ${\boldsymbol{\mathcal{B}}} $ 
 may not have exact  low-rank structure 
in the gradient domain, 
 we introduced a residual term on the basis of the regularizer (\ref{GTCTV}), i.e.,  
 \begin{align}  \label{Egntctv}
\|{\boldsymbol{\mathcal{B}}}\|_{\text{EGNTCTV}}
&
:= \frac{1}{\gamma}  \sum_{k =1} ^{3}
 \|
 {\boldsymbol{\mathcal{L}}}_{k}
 \|_{
 \Phi,
\mathfrak{L}
 }
+ \alpha  \sum_{k =1} ^{ 3}
  \|{\boldsymbol{\mathcal{S}}}_{k}\|_ { \ell_{1} ^ { \psi} },
\end{align}
\textcolor[rgb]{0.00,0.00,0.00}{where $\alpha>0$ is a balancing parameter,
 $\|\cdot\|_ { \ell_{1} ^ { \psi} }$ and
$\|
 \cdot 
\|_
{
\Phi,
\mathfrak{L} }$
 are defined according to 
 Formulas (\ref{unified-regular22})  and (\ref{GTCTV}), respectively.}
 \textcolor[rgb]{0.00,0.00,0.00}{Note that 
the regularizer  (\ref{Egntctv}) is named  as the
\textit{enhanced  GNTCTV} (EGNTCTV) of
$ {\boldsymbol{\mathcal{B}}}$.}

Existing studies \cite{feng2023hyperspectral,wu2023tensor} have shown that  TR factors possess both global low-rankness and sparsity in gradient domain (i.e., the  $\textbf{L}$+$\textbf{S}$ properties of TR factors itself).
Differently from these methods, we investigate the low-rankness and sparsity of gradient TR factors
under the T-SVD framework,
rather than the matrix unfolding scheme that may result in loss of 
optimal  representation.
That is to say,
treating  the $3$-D TR factors  as the low-tubal-rank tensors and
borrowing the idea of the GNTCTV regularizer (\ref{GTCTV}),
 we propose the following 
 regularization item,   called UNTRFR,  to measure the 
  $\textbf{L}$+$\textbf{S}$
 properties of  TR factors of the background tensor:
\begin{align} \label{UNTRFR22}
\|{\boldsymbol{\mathcal{B}}}\|_{\text{UNTRFR}}
&
:= \frac{1}{\gamma} \sum_{n =1} ^{3} \sum_{k =1} ^{3}
 \|
  \nabla_{k}  ({\boldsymbol{\mathcal{G}}}^{(n)} )
 \|_{
 \Phi,
\mathfrak{L}
 },
\end{align}
\textcolor[rgb]{0.00,0.00,0.00}{where
$
\Re
(\textbf{[}{\boldsymbol{\mathcal{G}}}\textbf{]})$ (
$\textbf{[}{\boldsymbol{\mathcal{G}}}\textbf{]}:=
\{{\boldsymbol{\mathcal{G}}}^{(1)},  {\boldsymbol{\mathcal{G}}}^{(2)}, {\boldsymbol{\mathcal{G}}}^{(3)}\}$)
is the TR decomposition of ${\boldsymbol{\mathcal{B}}}$,
$\{{\boldsymbol{\mathcal{G}}}^{(k)}\}_{k=1}^{3}$ are the TR factors.
}


\textcolor[rgb]{0.00,0.00,0.00}{Similarly to the definition  of  the EGNTCTV regularizer  (\ref{Egntctv}),  we put forward 
 an  enhanced version   of UNTRFR regularizer
(\ref{UNTRFR22}), known as EUNTRFR.
Specifically,  we factorize the gradient TR factor $\nabla_{k}  ({\boldsymbol{\mathcal{G}}}^{(n)} ), n,k=1,2,3$
into a low-rank component ${\boldsymbol{\mathcal{L}}}^{(n,k)}$  and a sparse component ${\boldsymbol{\mathcal{S}}}^{(n,k)}$, i.e.,
$\nabla_{k}  ({\boldsymbol{\mathcal{G}}}^{(n)} )= {\boldsymbol{\mathcal{L}}}^{(n,k)}+ {\boldsymbol{\mathcal{S}}}^{(n,k)}$.
Then, the EUNTRFR 
regularizer is defined 
as follows}
\begin{align} \label{UNTRFR}
\|{\boldsymbol{\mathcal{B}}}\|_{\text{EUNTRFR}}
&
:= \frac{1}{\gamma}  \sum_{n =1} ^{3} \sum_{k =1} ^{3} 
 \|
 {\boldsymbol{\mathcal{L}}}^{(n,k)}
 \|_{
 \Phi,
\mathfrak{L}
 }
+ \alpha \sum_{n =1} ^{ 3} \sum_{k =1} ^{ 3} 
  \|{\boldsymbol{\mathcal{S}}}^{(n,k)}\|_ { \ell_{1} ^ { \psi} }.
\end{align}
\textcolor[rgb]{0.00,0.00,0.00}{Note that  the above 
regularizer (\ref{UNTRFR}) can be viewed as a 
robust extension %
of the regularization item (\ref{UNTRFR22}).}

\textcolor[rgb]{0.00,0.00,0.00}{Figure \ref{fig_chart111} clarifies the relationships among the various nonconvex
regularization methods (including GNCTV, 
GNTCTV, EGNTCTV, UNTRFR, EUNTRFR)  designed above.}

%
 %

\vspace{-0.4505cm}
\subsection{\textbf{Generalized Nonconvex Model}} 
\label{non-model}

In this subsection,  a novel   HAD   model using unified nonconvex factors regularization
is proposed on the basis of Subsection \ref{nonconvex-regulari},
i.e.,
\begin{align} \label{orin_nonconvex}
& \textcolor[rgb]{0.00,0.00,0.00}{ \min_{
{\boldsymbol{\mathcal{E}}},
{\boldsymbol{\mathcal{B}}}
}
 \|{\boldsymbol{\mathcal{B}}}\|_{\operatorname{\textit{EUNTRFR}}}
 +
 \beta \cdot \|{\boldsymbol{\mathcal{E}}}\|_  {\ell _ {\mathnormal{F},1}^{ \psi}} },
\textcolor[rgb]{0.00,0.00,0.00}{\text{s.t.}} \;\;
\textcolor[rgb]{0.00,0.00,0.00}{{\boldsymbol{\mathcal{M}}}={\boldsymbol{\mathcal{B}}}+{\boldsymbol{\mathcal{E}}},
}
\end{align}
%
%
\textcolor[rgb]{0.00,0.00,0.00}{where
${\boldsymbol{\mathcal{M}}}$, ${\boldsymbol{\mathcal{B}}}$ and  ${\boldsymbol{\mathcal{E}}}$
respectively denote 
observed  HSI, the background component and the abnormal component,
 the definitions  of
 $\|\cdot\|_  {\ell _ {\mathnormal{F},1} ^{ \psi} }$  and  $\|\cdot
 \|_{\text{EUNTRFR}}$ are consistent with
 Formulas  (\ref{unified-regular22222}) and  (\ref{UNTRFR}),
%
 %
 and   $\alpha, \beta> 0$  are the   trade-off parameters. 
  %
}

 \textcolor[rgb]{0.00,0.00,0.00}{The  proposed HAD   method  differs from the    previous researches fused $\textbf{L}$ and   $\textbf{S}$ properties
 (e.g.,  \cite{ li2020prior,feng2023hyperspectral, xiao2024hyperspectral33, 
 yu2024generalized, wang2023guaranteed, shang2023hyperspectral,qin2024tensor,
 he2023anomaly, qin2023generalized1, liu2023anomaly11, zhao2023hyperspectral }).
 For instance,
 in  TRDFTVAD method \cite{feng2023hyperspectral},
TR factorization and TV  regularization constraints were introduced to explore
the $\textbf{L}$+$\textbf{S}$ natures 
of the background tensor. 
However, this convex TRDFTVAD method did not take the prior characteristics of the TR factors into consideration,
 and 
 fail to  possess a more unified, efficient  and concise form.
 The proposed UNTRFR and EUNTRFR regularization strategies   alleviate these limitations.
Different from the nonconvex low-rank regularization methods that previously acted on the original
domain \cite{yu2024generalized, he2023anomaly, qin2023generalized1, liu2023anomaly11, zhao2023hyperspectral},
the method we propose is modeled from the perspective of gradient mapping,
enabling it to simultaneously mine the $\textbf{L}$+$\textbf{S}$  priors in a more concise yet effective form.}

 %

\textcolor[rgb]{0.00,0.00,0.00}{In previous work \cite{qin2024tensor}, a circular TR unfolding strategy was used alongside nonconvex constraints on the gradient maps of the background tensor to develop a generalized and effective regularizer. This approach effectively
extracts both the 
$\textbf{L}$ and   $\textbf{S}$
priors in a concise and unified manner.
  %
     \textcolor[rgb]{0.00,0.00,0.00}{Whereas, this paper proposes 
      novel unified nonconvex TR factors
     regularization methods, whose key techniques (i.e., GNTCTV, EGNTCTV) differ  
     from the convex TCTV-based HAD  approach  \cite{wang2023guaranteed}.  
     By  applying the new representation paradigm 
     to the $3$-D gradient TR factors derived from the background tensor,
     the proposed 
     method can encode 
     both 
     $\textbf{L}$ and   $\textbf{S}$  properties simultaneously.
    More importantly,
    to achieve an improved and more comprehensive solution for background and anomaly modeling, a family of generalized nonconvex functions and their
corresponding modified versions (i.e., capped folded concave functions) are included 
 in our  HAD model (\ref{orin_nonconvex}).
Therefore,
previous related methods \cite{qin2024tensor, wang2023guaranteed}   and the proposed one are 
entirely different regularization schemes for extracting
$\textbf{L}$+$\textbf{S}$ prior features.}}
 \begin{algorithm}[!htbp]
\setstretch{0.1150}
     \caption{{\textcolor[rgb]{0.00,0.00,0.00}{Generalized 
thresholding operator.
     }}
     }
     \label{random-wtst}
      \KwIn{
        $\bm{\mathcal{A}}\in\mathbb{R}^{n_1\times  n_2 \times n_3}$, $\lambda> 0$, $\psi(\cdot)$.
       }

\If{
$h(\cdot)=\|\cdot\|_{1}$
}
{
   $ {\hat{{\boldsymbol{\mathcal{E}}}}}_ {i_1 i_2 i_3}
=
\operatorname{\textit{Prox}}_ {\psi, \lambda} ({\boldsymbol{\mathcal{A}}}_ {i_1 i_2 i_3})
$\;
}

\ElseIf{
$h(\cdot)=\|\cdot\|_{\mathnormal{F},1}$
}
{

{
 \For{$i=1,2,\cdots, n_1$}
      {
      \For{$j=1,2,\cdots, n_2$}
 {
$$ {\hat{ {\boldsymbol{\mathcal{E}}} }_{ij :} }
  =
  \frac  {  {\boldsymbol{\mathcal{A}}}_{ij :  } }
 { { \|{\boldsymbol{\mathcal{A}}}_{ij :  } \|_{\mathnormal{F}}} } \cdot
  \operatorname{\textit{Prox}}_ {\psi, \lambda} ({ \|{\boldsymbol{\mathcal{A}}}_{ij :  } \|_{\mathnormal{F}}})
;
  $$
}
}}}
    		 {\color{black}\KwOut{
 $\hat{{\boldsymbol{\mathcal{E}}}}  \in\mathbb{R}^{n_1\times n_2 \times n_3} $.}}
    \end{algorithm}
%

 \begin{algorithm}[!htbp]
\setstretch{0.115}
     \caption{{\textcolor[rgb]{0.00,0.00,0.00}{GNTSVT operation, $ {\boldsymbol{\mathcal{D}}}_{\Phi, \tau}({\boldsymbol{\mathcal{A}}}, \mathfrak{L})$.}}}\label{random-wtsvt}
      \KwIn{ $\bm{\mathcal{A}}\in\mathbb{R}^{n_1\times n_2 \times n_3}$,
        transform:  $\mathfrak{L}$,  $ \Phi(\cdot)$, $\tau> 0$.}

      Compute the results of $\mathfrak{L}$ on $\boldsymbol{\mathcal{A} }$, i.e.,
       $\mathfrak{L}(\boldsymbol{\mathcal{A}})$\;

{\For{$v=1,2,\cdots, n_3$}{
{
$[{\mathfrak{L}}({\boldsymbol{\mathcal{U}}})^{<v>},{\mathfrak{L}}({\boldsymbol{\mathcal{S}}})^{<v>},{\mathfrak{L}}({\boldsymbol{\mathcal{V}}})^{<v>}]= \textrm{svd} \big(   {\mathfrak{L}} ({\boldsymbol{\mathcal{A}}})^{<v>}    \big)$ \;}

      $
\hat{\bm{S}} =  \operatorname{\textit{Prox}}_ {\Phi, \tau} \big [
\operatorname{diag} \big(  \mathfrak{L}(\boldsymbol{\mathcal{S}})^{<v>} \big) \big]
$;\\

${  \mathfrak{L}(\boldsymbol{\mathcal{C}})   }^{<v>}= {\mathfrak{L}(\boldsymbol{\mathcal{U}})}^{<v>}\cdot
\operatorname{diag} 
(\hat{\bm{S}})
\cdot
{   (\mathfrak{L}(\boldsymbol{\mathcal{V}})^{<v>})   }  ^{\mit{T}}$;\\
}
}

{

}

{\color{black}\KwOut{$ {\boldsymbol{\mathcal{D}}}_{\Phi, \tau}({\boldsymbol{\mathcal{A}}}, \mathfrak{L})
\leftarrow {\mathfrak{L}}^{-1}({\mathfrak{L}}({\boldsymbol{\mathcal{C}}}))$.}}
 \end{algorithm}
%


\vspace{-0.32cm}

\subsection{\textbf{Generalized solution paradigm 
 for Key Subproblems}}   \label {gnlstr-oper}
This subsection mainly presents the solution method of   two kinds of
 key subproblems  involved  in  our model (\ref{orin_nonconvex}), i.e.,
\begin{align}\label{E_prox1}
\textcolor[rgb]{0.00,0.00,0.00}{
\arg\min_{ \boldsymbol{\mathcal{E}}}
{\lambda} \cdot  \psi \big(h( {\boldsymbol{\mathcal{E}}} )\big)
 +
\frac {1}{2}
\|
{\boldsymbol{\mathcal{E}}}-  {\boldsymbol{\mathcal{A}}}
\|^2_{\mathnormal{F}},}
\\ \label{LS_prox1}
\arg\min_{ {\boldsymbol{\mathcal{L}}}}
\tau\cdot
\big\|
 {\boldsymbol{\mathcal{L}}}
 \big\|_
{
\Phi,
\mathfrak{L} }
+
\frac{1}{2}\|{\boldsymbol{\mathcal{L}}}
-{\boldsymbol{\mathcal{A}}}
\|^{2}_{\mathnormal{F}}.
\end{align}

\subsubsection{\textbf{Solve 
problem (\ref{E_prox1})  via Generalized Nonconvex Thresholding/Shrinkage Operator}}
 The  optimal solution to  the subproblem (\ref{E_prox1})
can be computed by the generalized nonconvex thresholding/shrinkage operator in an element-wise or  tube-wise
manner.
Please see Algorithm \ref{random-wtst} for more details.

\subsubsection{\textbf{Solve 
problem (\ref{LS_prox1})
via \textit{Generalized Nonconvex Tensor  Singular Value Thresholding}  (GNTSVT) Operator}}
The close-form  solution to the problem (\ref{LS_prox1}) can be obtained via
the following key   Theorem \ref{theorem_wtsvt33}. Note that
the computation process of GNTSVT  operation can be found in Algorithm \ref{random-wtsvt}.

\begin{Theorem}\label{theorem_wtsvt33}
 (\textbf{GNTSVT}) \cite{qiu2021nonlocal, wang2021generalized}
 Let the   T-SVD decomposition
  of ${\boldsymbol{\mathcal{A}}} \in\mathbb{R}^{n_1\times n_2 \times n_3}$ be
 ${\boldsymbol{\mathcal{A}}}={\boldsymbol{\mathcal{U}}} {*}_{\mathfrak{L}}   {\boldsymbol{\mathcal{S}}} {*}_{\mathfrak{L}}
{\boldsymbol{\mathcal{V}}}^{\mit{T}}$, $m=\min(n_1,n_2)$.
 For any $\tau \geq 0$, then  a global optimal solution to the
optimization problem (\ref{LS_prox1})
is given by
$$
\bm{\mathcal{D}}_{\Phi, \tau } ({\boldsymbol{\mathcal{A}}}, \mathfrak{L}) = {\boldsymbol{\mathcal{U}}}
{*}_{\mathfrak{L}}
\tilde {{\boldsymbol{\mathcal{Y}}}}
{*}_{\mathfrak{L}}
{\boldsymbol{\mathcal{V}}}
^{\mit{T}},
$$
where $\tilde {{\boldsymbol{\mathcal{Y}}}}$ is an f-diagonal tensor,
$\mathfrak{L} (\tilde {{\boldsymbol{\mathcal{Y}}}})^{<t>}
(i,i)
 =\operatorname{Prox}_{ \Phi,\tau}  \big(
 \mathfrak{L} ( {{\boldsymbol{\mathcal{S}}}})^{<t>} (i,i)
 \big)$,
$t \in \{ 1, \cdots, n_3\}$,  $i \in \{ 1,\cdots, m\}$,
and
$\operatorname{\textit{Prox}}_ {\Phi, \tau} (\cdot) $ denotes the  proximity operator of nonconvex penalty function
$\Phi(\cdot)$,
which   has the same properties as $\psi(\cdot)$.
\end{Theorem}

\vspace{-0.46cm}

\subsection{\textbf{Optimization Algorithm}}\label{non-algorithms}

In this subsection, the  well-known ADMM framework  \cite{boyd2011distributed} is adopted to solve the proposed  model (\ref{orin_nonconvex}).
The nonconvex model (\ref{orin_nonconvex})  can be equivalently reformulated as follows:
\begin{align}
& \min_{
{\boldsymbol{\mathcal{G}}} ^{(n)}, 
 {\boldsymbol{\mathcal{L}}}^{(n,k)},  {\boldsymbol{\mathcal{S}}}^{(n,k)},
{\boldsymbol{\mathcal{E}}}
}
 \sum_{n=1}^{3}
 \sum_{k=1}^{3}
 \Big\{ \frac{1}{\gamma}
 \|{\boldsymbol{\mathcal{L}}}^{(n,k)}\|_{ \Phi, {\mathfrak{L}}}
 +
 \alpha \cdot \|{\boldsymbol{\mathcal{S}}}^{(n,k)}\|_ { \ell_{1} ^ { \psi} }  \Big\}
 \notag \\
&
+
 \beta \cdot \|{\boldsymbol{\mathcal{E}}}\|_  {\ell _ {\mathnormal{F},1}^{ \psi}},
\notag \\
&\text{s.t.} \;\;
{\boldsymbol{\mathcal{M}}}= {\boldsymbol{\mathcal{B}}}+{\boldsymbol{\mathcal{E}}}=
\Re
(\textbf{[}{\boldsymbol{\mathcal{G}}}\textbf{]})
+{\boldsymbol{\mathcal{E}}},
\notag \\ \label{equ_nonconvex}
& \;\;\;\;\; \nabla_{k} ({\boldsymbol{\mathcal{G}}} ^{(n)})= {\boldsymbol{\mathcal{L}}}^{(n,k)}+ {\boldsymbol{\mathcal{S}}}^{(n,k)},
\;\; n, k \in \{1,2,3\}.
\end{align}
The  augmented Lagrangian function of (\ref{equ_nonconvex}) is
\begin{align}
&
\mathcal{F}(
{\boldsymbol{\mathcal{L}}}^{(n,k)}, {\boldsymbol{\mathcal{S}}}^{(n,k)},   {\boldsymbol{\mathcal{Q}}}^{(n,k)},
\textbf{[}{\boldsymbol{\mathcal{G}}}\textbf{]},
 {\boldsymbol{\mathcal{E}}},{\boldsymbol{\mathcal{Y}}})
 =
 \sum_{n=1}^{3}
 \sum_{k=1}^{3}
\Big\{ \frac{1}{\gamma}
 \|{\boldsymbol{\mathcal{L}}}^{(n,k)}\|_{ \Phi, {\mathfrak{L}}}
  \notag \\
 &
 +
 \alpha \cdot \|{\boldsymbol{\mathcal{S}}}^{(n,k)}\|_ { \ell_{1} ^ { \psi} }
 + \langle  {\boldsymbol{\mathcal{Q}}}^{(n,k)}    , \nabla_{k}({\boldsymbol{\mathcal{G}}}^{(n)}) - {\boldsymbol{\mathcal{L}}}^{(n,k)}   -{\boldsymbol{\mathcal{S}}}^{(n,k)}
 \rangle
 \notag \\
 & +
\frac {\mu}{2}
\|
\nabla_{k}({\boldsymbol{\mathcal{G}}}^{(n)}) - {\boldsymbol{\mathcal{L}}}^{(n,k)}   -{\boldsymbol{\mathcal{S}}}^{(n,k)}
  \|^{2}_{{{\mathnormal{F}}}} \Big\}
 +
 \beta \cdot \|{\boldsymbol{\mathcal{E}}}\|_  {\ell _ {\mathnormal{F},1}^{ \psi}}
 \notag \\ \label {lag-tttt}
&
%
+
\langle {\boldsymbol{\mathcal{Y}}},
{\boldsymbol{\mathcal{M}}} -
\Re
(\textbf{[}{\boldsymbol{\mathcal{G}}}\textbf{]})
-{\boldsymbol{\mathcal{E}}}
\rangle +
\frac {\mu}{2}
\|
{\boldsymbol{\mathcal{M}}} -
\Re
(\textbf{[}{\boldsymbol{\mathcal{G}}}\textbf{]})
-{\boldsymbol{\mathcal{E}}}
\|^{2}_{{{\mathnormal{F}}}},
\end{align}
where  $\mu$ is the regularization  parameter, and
${\boldsymbol{\mathcal{Q}}}^{(n,k)}$,  ${\boldsymbol{\mathcal{Y}}}$ are Lagrange multipliers.
It can be further expressed as
\begin{align}
&
\mathcal{F}(
{\boldsymbol{\mathcal{L}}}^{(n,k)}, {\boldsymbol{\mathcal{S}}}^{(n,k)},   {\boldsymbol{\mathcal{Q}}}^{(n,k)},
\textbf{[}{\boldsymbol{\mathcal{G}}}\textbf{]},
 {\boldsymbol{\mathcal{E}}},{\boldsymbol{\mathcal{Y}}})
 =
 \sum_{n=1}^{3}
 \sum_{k=1}^{3}
\Big\{ \frac{1}{\gamma}
 \|{\boldsymbol{\mathcal{L}}}^{(n,k)}\|_{ \Phi, {\mathfrak{L}}}  +  
  \notag \\
 &
 \alpha \cdot \|{\boldsymbol{\mathcal{S}}}^{(n,k)}\|_ { \ell_{1} ^ { \psi} }
 +
\frac {\mu}{2}
\|
\nabla_{k}({\boldsymbol{\mathcal{G}}}^{(n)}) - {\boldsymbol{\mathcal{L}}}^{(n,k)}   -{\boldsymbol{\mathcal{S}}}^{(n,k)}
+  \frac { {\boldsymbol{\mathcal{Q}}}^{(n,k)}} {  \mu}
  \|^{2}_{{{\mathnormal{F}}}} \Big\}
 \notag \\
 \label {lag-tttt11RRR}
&
+
 \beta \cdot \|{\boldsymbol{\mathcal{E}}}\|_  {\ell _ {\mathnormal{F},1}^{ \psi}}
+
\frac {\mu}{2}
\|  {\boldsymbol{\mathcal{M}}} -
\Re
(\textbf{[}{\boldsymbol{\mathcal{G}}}\textbf{]})
-{\boldsymbol{\mathcal{E}}} + {\boldsymbol{\mathcal{Y}}}/ \mu
\|^{2}_{{{\mathnormal{F}}}} + C,
\end{align}
where  $C$ is only the multipliers dependent squared items. Below, we show how to solve the subproblems for each involved variable.


\noindent \textcolor[rgb]{0.00,0.00,0.00}{{\textbf{Module 1: Update ${  ({{\boldsymbol{\mathcal{G}}}^{(n)}}) ^ {\{\nu+1\}}  },$ $n=1,2, 3$}}:}
Through  the matrix representation  of TR decomposition (\textit{please see Formula (\ref{eq:theoremals}) 
for details}) and let ${\bm{R}} _{1} =(
\bm{G}
^{(\neq n)}_{<2>})
^{\mit{T}}$.
Then,  the subproblem with respect to ${{\bm{{G}}}}^{(n)}_{(2)}$ can be constructed as 
\begin{align}
&
{ ({{\bm{{G}}}}^{(n)}_{(2)} )}
^ {\{\nu+1\}}
= \arg\min_{{{\bm{{G}}}}^{(n)}_{(2)}  }  
{\mu^ {\{\nu\}}  } / {2}
 \big \|
{\bm{{M}}}_{<n>} - {\bm{{G}}}^{(n)}_{(2)} {{\bm{R}} _{1}} ^ {\{\nu\}}
\notag \\ &
-{\bm{{E}}}_{<n>} ^{\{\nu\}}
+ 
 {  {{{\bm{Y}}}}_{<n>}^{\{\nu\}} }  /  {\mu^{\{\nu\}}}
\big  \|^{2}_{{{\mathnormal{F}}}}
 +
 {\mu^ {\{\nu\}}  } / {2}
 \big  \|
 \bm{D}_{n} {\bm{{G}}}^{(n)}_{(2)}  - {({\bm{{L}}}^{(n,2)}_{(2)}) }^{\{\nu\}}
  \notag \\  \label {coef-tensor} &
 -{({\bm{{S}}}^{(n,2)}_{(2)}) }^{\{\nu\}}
 +
 {({\bm{{Q}}}^{(n,2)}_{(2)}) }^{\{\nu\}}/  {\mu^{\{\nu\}}}
 \big  \|^{2}_{{{\mathnormal{F}}}}.
\end{align}
Taking the derivative in (\ref {coef-tensor}) with respect to ${\bm{{G}}}^{(n)}_{(2)}$, it obtains the following linear system:
\begin{align*}
& \bm{D}_{n}^{\mit{T}}
\bm{D}_{n} {\bm{{G}}}^{(n)}_{(2)} +
{\bm{{G}}}^{(n)}_{(2)} {{\bm{R}} _{1}} ^ {\{\nu\}} {({{\bm{R}} _{1}} ^ {\{\nu\}})}^{\mit{T}}
=
\big ( {\bm{{M}}}_{<n>}
-{\bm{{E}}}_{<n>} ^{\{\nu\}}
\notag \\
&
+ 
 {  {{{\bm{Y}}}}_{<n>}^{\{\nu\}} }  /  {\mu^{\{\nu\}}}
\big ) ( { { {\bm{R}} _{1} } ^ {\{\nu\}}} )^{\mit{T}}
+
 \bm{D}_{n} ^{\mit{T}}
\big(
 {({\bm{{L}}}^{(n,2)}_{(2)}) }^{\{\nu\}}+ {({\bm{{S}}}^{(n,2)}_{(2)}) }^{\{\nu\}}
 \notag \\  
 &
-
 {({\bm{{Q}}}^{(n,2)}_{(2)}) }^{\{\nu\}}/  {\mu^{\{\nu\}}} \big).
\end{align*}
The above formulation is  a classical Sylvester matrix equation,
which can be efficiently solved in the light of  the reference \cite{zheng2020double}.
Eventually, the updated TR factor tensor $  {{\boldsymbol{\mathcal{G}}}^{(n)}}  $ is achieved by the `fold' operation, i.e.,
\begin{align} \label{coreg}
 ( {{\boldsymbol{\mathcal{G}}}^{(n)}} )  ^{\{\nu+1\}}
=\operatorname{fold}_{2} \big ( ({\bm{{G}}}^{(n)}_{(2)}  )  ^{\{\nu+1\}}  \big),
\end{align}
\textcolor[rgb]{0.00,0.00,0.00}{where    $ \operatorname{fold}_{k} (  \cdot) $ 
denotes
the inverse operator  of mode-$k$ unfolding, which
  satisfies
$ {\boldsymbol{\mathcal{A}}} =\operatorname{fold}_{k} (  {\bm{{A}}} _{(k)} ) $, $  {\bm{{A}}} _{(k)} =   \operatorname{unfold}_{k} (   {\boldsymbol{\mathcal{A}}}) $.}

\noindent \textcolor[rgb]{0.00,0.00,0.00}{{\textbf{Module 2: Update $( { {\boldsymbol{\mathcal{S}}}^{(n, k)} ) }  ^{\{\nu+1\}}$}},
$n,k= 1,2,3$:}
Similarly,
extracting all items containing
$ \boldsymbol{\mathcal{S}}  ^{(n, k)} $
 from
(\ref {lag-tttt11RRR}), we have
\begin{align}
&  ( { {\boldsymbol{\mathcal{S}}}^{(n, k)} ) }  ^{\{\nu+1\}}=
\arg \min_{ \boldsymbol{\mathcal{S}}  ^{(n, k)} }
\alpha \cdot \|{\boldsymbol{\mathcal{S}}}^{(n,k)}\|_ { \ell_{1} ^ { \psi} }
+
\frac {{\mu}^{\{\nu\}}} {2}
\|  - {\boldsymbol{\mathcal{S}}}^{(n,k)}
\notag \\ \label {S_prox1} 
&+ \nabla_{k}(   {\boldsymbol{\mathcal{G}}}^{(n)}  ) ^{\{\nu+1\}}
-( {\boldsymbol{\mathcal{L}}}^{(n,k)} ) ^{\{\nu 
\}}
+  \frac {  ({\boldsymbol{\mathcal{Q}}}^{(n,k)} ) ^{\{\nu\}} } {  \mu ^{\{\nu\}}}
  \|^{2}_{{{\mathnormal{F}}}}.
\end{align}

\noindent \textcolor[rgb]{0.00,0.00,0.00}{{\textbf{Module 3: Update $( { {\boldsymbol{\mathcal{L}}}^{(n, k)} ) }  ^{\{\nu+1\}}$}},
$n,k=1,2,3$:}
For each $n =1,2,3$, $k=1, 2, 3$,  extracting all items containing
$ \boldsymbol{\mathcal{L}}  ^{(n, k)} $
 from
(\ref {lag-tttt11RRR}), we have
\begin{align}
&  ( { {\boldsymbol{\mathcal{L}}}^{(n, k)} ) }  ^{\{\nu+1\}}=
\arg \min_{ \boldsymbol{\mathcal{L}}  ^{(n, k)} }
\frac{1}{\gamma}
 \|{\boldsymbol{\mathcal{L}}}^{(n,k)}\|_{ \Phi, {\mathfrak{L}}}
 +
\frac {  {\mu}^{\{\nu\}}  }    {2}
\| - {\boldsymbol{\mathcal{L}}}^{(n,k)}
\notag \\ \label {L_prox1}
&
 + \nabla_{k}(   {\boldsymbol{\mathcal{G}}}^{(n)}  ) ^{\{\nu+1\}}
 - ({\boldsymbol{\mathcal{S}}}^{(n,k)}) ^{\{\nu +1 \}}
+  \frac {  ({\boldsymbol{\mathcal{Q}}}^{(n,k)} ) ^{\{\nu\}} } {  \mu ^{\{\nu\}}}
  \|^{2}_{{{\mathnormal{F}}}}.
\end{align}
%
This key subproblem is analogous to  the form of  the minimization problem  (\ref{LS_prox1}),
and its  close-form solution can be obtained by utilizing  the GNTSVT operation (see  \ref{gnlstr-oper} for details).

\noindent
\textcolor[rgb]{0.00,0.00,0.00}{
{\textbf{Module 4: Update ${\boldsymbol{\mathcal{E}}^{\{\nu+1\}}}$}}}:
\textcolor[rgb]{0.00,0.00,0.00}{
The optimization subproblem 
with respect to $\boldsymbol{\mathcal{E}}$ %
can  be written as}
\begin{align}
&
{\boldsymbol{\mathcal{E}}^{\{\nu+1\}}}=
\arg \min_{
\boldsymbol{\mathcal{E}}}
\beta  \|{\boldsymbol{\mathcal{E}}}\|_  {\ell _ {\mathnormal{F},1}^{ \psi}}
+
%
{\mu ^{\{\nu\}} } / {2}
\big \|  {\boldsymbol{\mathcal{M}}} -
\Re
(\textbf{[}{\boldsymbol{\mathcal{G}}}\textbf{]})
\notag \\ \label{e-pro}
&
-{\boldsymbol{\mathcal{E}}} +
{ {\boldsymbol{\mathcal{Y}}} ^{\{\nu\}}}  /  { \mu ^{\{\nu\}}  }
\big \|^{2}_{{{\mathnormal{F}}}}.
\end{align}
%
\noindent
The  subproblems (\ref{S_prox1}), (\ref{e-pro}) are equivalent to the form of the minimization problem (\ref{E_prox1}),
and their close-form solutions can be obtained by utilizing the 
 generalized nonconvex shrinkage operator.  Please see  \ref {gnlstr-oper} for details.

\noindent \textcolor[rgb]{0.00,0.00,0.00}{
{\textbf{Module 5: Update $  ({\boldsymbol{\mathcal{Q}}}^{(n,k)})  
^{\{\nu+1\}}$,   ${\boldsymbol{\mathcal{Y}}}^{\{\nu+1\}}$,  and
$\mu^{\{\nu+1\}}$  (Lagrange multipliers and   penalty parameter)}}}
Based on the rule of the ADMM framework, the lagrange multipliers are updated by the following equations:
\begin{align}\label{lagrange11}
{\boldsymbol{\mathcal{Y}}}^{\{\nu+1\}}
=
{\boldsymbol{\mathcal{Y}}}^{\{\nu\}}+
\mu^{\{\nu\}}
({\boldsymbol{\mathcal{M}}}-
\Re
(\textbf{[}{\boldsymbol{\mathcal{G}}}\textbf{]})
-{\boldsymbol{\mathcal{E}}}^{\{\nu+1\}}),
\end{align} 
\begin{align}
 & {  ({\boldsymbol{\mathcal{Q}}}^{(n,k)} ) }^{\{\nu+1\}}
   =
 { ({\boldsymbol{\mathcal{Q}}}^{(n,k)} ) }^{\{\nu\}}+
\mu^{\{\nu\}} \cdot {\boldsymbol{\mathcal{N}}} ^{\{\nu\}},
\end{align}
\begin{align}
\label{lagrange22}
& \mu^{\{\nu+1\}}
= \min \left( \mu^{\operatorname{max}},\vartheta \mu^{\{\nu\}} \right),
\end{align}
\noindent where
$
{\boldsymbol{\mathcal{N}}}   ^{\{\nu\}} =
\nabla_{k}(   {\boldsymbol{\mathcal{G}}}^{(n)}  ) ^{\{\nu+1\}}
-( {\boldsymbol{\mathcal{L}}}^{(n,k)} ) ^{\{\nu+1\}} - ( {\boldsymbol{\mathcal{S}}}^{(n,k)} ) ^{\{\nu+1\}}
$,
$\vartheta$ stands for the growth rate.

\noindent \textcolor[rgb]{0.00,0.00,0.00}{{\textbf{Module 6: Update the detection map  $\bm{T}  ^{\{\nu+1\} } $}}:
Finally, the anomaly detection map can be obtained by}
\begin{align}\label{mapmap}
\bm{T} ^{\{ \nu+1\}} (i_1,i_2)= \sqrt {\sum_{i_3=1} ^{n_3} |{\boldsymbol{\mathcal{E}}} ^{\{ \nu+1\}} (i_1,i_2, i_3)|^{2}}.
\end{align}

 The entire ADMM optimization framework 
 is summarized in Algorithm
\ref{algorithm1had}.

\begin{algorithm}[!htbp]
\setstretch{0.3} 
\caption{Solve the 
HAD model (\ref{orin_nonconvex}) via ADMM.
}\label{algorithm1had}

 \KwIn{
HSI data:
${\boldsymbol{\mathcal{M}}}$,
$ \Phi(\cdot),  \psi(\cdot)$, 
 $\alpha$,  $\beta$,
 $\mathfrak{L}$,
  TR  rank: $[r_1, r_2, r_3]$.
  }

 \textbf{Initialize:} For $n,k=1,2,3$,  $ {   ({\boldsymbol{\mathcal{L}}}^{(n, k) }) ^{\{0\}}   }=  
 {   ({\boldsymbol{\mathcal{S}}}^{(n, k) }) ^{\{0\}}   }={   ({\boldsymbol{\mathcal{Q}}}^{(n, k) }) ^{\{0\}}   }=
 {\boldsymbol{\mathcal{E}} }  ^{\{0\}}
 =
 {\boldsymbol{\mathcal{Y}}}^{\{0\}}
 =
 \boldsymbol{0}$,
 random sample
  ${\boldsymbol{\mathcal{G}}}^{(n) }$  by distribution $N \sim (0,1)$,
 $\vartheta$,
$\mu^{\{0\}}$,
$\mu^{\max}$,
$\varpi$,
$\nu_{\max}=500$\;

\While{\text{not converged}}
{

 \textcolor[rgb]{0.00,0.00,0.00}{Update $ { ({\boldsymbol{\mathcal{G}}} ^{(n)}) ^{\{\nu+1\}}}$ by computing
 (\ref{coreg});} \\ 

Update ${ ({{\boldsymbol{\mathcal{S}} }^{(n,k)}} ) ^{\{\nu+1\}}}$  by computing (\ref {S_prox1})
;\\

Update ${ ({{\boldsymbol{\mathcal{L}} }^{(n,k)}} ) ^{\{\nu+1\}}}$  by computing (\ref {L_prox1})
;\\

  \textcolor[rgb]{0.00,0.00,0.00}{Update $\boldsymbol{ {\boldsymbol{\mathcal{E}}}}^{\{\nu+1\}}$ by computing (\ref{e-pro});}\\


 \textcolor[rgb]{0.00,0.00,0.00}{Update   $(  {\boldsymbol{\mathcal{Q}}}^{(n,k)} 
 )^{\{\nu+1\}}, {\boldsymbol{\mathcal{Y}}}^{\{\nu+1\}}, \mu^{\{\nu+1\}}$
 by computing (\ref{lagrange11})-(\ref{lagrange22});}\\  

Update $ {\boldsymbol{\mathcal{B}}} ^{\{\nu+1\}} $ by computing
$ {\boldsymbol{\mathcal{B}}} ^{\{\nu+1\}} = \Re (\textbf{[}{\boldsymbol{\mathcal{G}}}\textbf{]})$; \\

Obtain the   detection map $\bm{T} ^{\{\nu+1\} } $ by (\ref{mapmap})\;

 Check the convergence conditions
\begin{align*}
\operatorname{Error}=
\| {{\boldsymbol{\mathcal{B}}} }  ^{\{\nu+1\}}-
 {{\boldsymbol{\mathcal{B}}} }   ^{\{\nu\}}
\|{_{\mathnormal{F}}} /
\|
 {{\boldsymbol{\mathcal{B}}} }   ^{\{\nu\}}
\|{_{\mathnormal{F}}}
 \leq \varpi;
\end{align*}
}
{\color{black}\KwOut{
background
  ${\boldsymbol{\mathcal{L}}}$, anomalies ${\boldsymbol{\mathcal{E}}}$,
  detection map $ \bm{T}$.
  }}
\end{algorithm}
\vspace{-0.3cm}

\subsection{\textbf{Complexity Analysis}}

Given an input HSI data with the size of  $ {n_1 \times n_2 \times n_3}$,
we analyze the  per-iteration complexity of Algorithm \ref{algorithm1had},
which  mainly needs  to update  $\{{{\boldsymbol{\mathcal{G}}}^{(n)}}\}_{n=1}^{3}$,
$\{{{\boldsymbol{\mathcal{L}}}^{(n,k)}}\}_{n,k =1}^{3}$,
$\{{{\boldsymbol{\mathcal{S}}}^{(n,k)}}\}_{n,k =1}^{3}$,
${\boldsymbol{\mathcal{E}}}$, 
$\{{{\boldsymbol{\mathcal{Q}}}^{(n,k)}}\}_{n,k =1}^{3}$,
and
${\boldsymbol{\mathcal{Y}}}$, 
respectively.
For simplicity, we set $N=n_1=n_2=n_3$,  $R=r_1=r_2=r_3$, and $R<N$.
Specifically, \textbf{1)}
the time complexity of calculating the first term  that mainly involves SVD,  FFT, and matrix multiplication
 is $\boldsymbol{\mathcal{O}}\big( N^3 R^2 +R^6 \big)$;
 \textbf{2)}
 The update of  the second term, which   mainly involves the
 GNTSVT operation and matrix multiplication,
  equals to
 $\boldsymbol{\mathcal{O}} \big( N^2 R^2 \big) $; 
\textbf{3)}
 the time complexity of computing the third  term is $\boldsymbol{\mathcal{O}}\big( R^2 N^2  \big)$;
\textbf{4)} Upadating ${\boldsymbol{\mathcal{E}}}$ requires to perform the generalized thresholding 
operation and the matrix multiplication with a complexity of
$\boldsymbol{\mathcal{O}}\big( R^3 N^3 \big)$;
\textbf{5)}
 the time complexity of computing the fifth  term is $\boldsymbol{\mathcal{O}}\big( R^2 N^2 \big)$;
\textbf{6)} ${\boldsymbol{\mathcal{Y}}}$
can be updated by a low consumed algebraic computation.

\vspace{-0.3cm}

\subsection{\textcolor[rgb]{0.00,0.00,0.00}{\textbf{Convergence Guarantee}}}
\textcolor[rgb]{0.00,0.00,0.00} {In this subsection, we   provide
the convergence analysis of 
Algorithm \ref{algorithm1had}. 
The detailed
proofs of relevant theories and lemmas  
can be found
 in our Github:
 \url{https://github.com/Qinwenjinswu/QWJSWU-Convergence-Theory-Proof}.}

\begin{Lemma}\label{yy1}
\textcolor[rgb]{0.00,0.00,0.00}{The sequences
$\big \{{\boldsymbol{\mathcal{Y}}}^{\{\nu\}} \big \}$ and  
$\big  \{  {\boldsymbol{\mathcal{Q}}^{(n,k)} }  ^{\{ \nu \}} \big \}$ ($n,k=1,2,3$)
generated by      Algorithm \ref{algorithm1had} 
are bounded.}
\end{Lemma}

\begin{Lemma}\label{yy15559999999}
\textcolor[rgb]{0.00,0.00,0.00}{Suppose that the   sequences
$\big \{{\boldsymbol{\mathcal{Y}}}^{\{\nu\}} \big \}$ and  
$\big  \{  {\boldsymbol{\mathcal{Q}}^{(n,k)} }  ^{\{ \nu \}} \big \}$ ($n,k=1,2,3$)
generated by  Algorithm \ref{algorithm1had} 
are   bounded,
then
the sequences
$\big  \{ ( {\boldsymbol{\mathcal{L}}^{(n,k)} } ) ^{\{ \nu \}} \big \}$,  
$\big  \{ ( {\boldsymbol{\mathcal{S}}^{(n,k)} } ) ^{\{ \nu \}} \big \}$,
%
$\big  \{ ( {\boldsymbol{\mathcal{G}}^{(n)} } ) ^{\{ \nu \}} \big \}$,
  and $ {\boldsymbol{\mathcal{E}}}^{\{ \nu \}} $
  are bounded.}
\end{Lemma}

\begin{Theorem}\label{conver2222} 
\textcolor[rgb]{0.00,0.00,0.00}{
Suppose that the   sequences
$\big \{{\boldsymbol{\mathcal{Y}}}^{\{\nu\}} \big \}$ and  
$\big  \{  {\boldsymbol{\mathcal{Q}}^{(n,k)} }  ^{\{ \nu \}} \big \}$ ($n,k=1,2,3$)
generated by  Algorithm \ref{algorithm1had} 
are   bounded.
Then, for any   $n,k \in \{1,2,3\}$,
the sequences
$\big  \{ ( {\boldsymbol{\mathcal{L}}^{(n,k)} } ) ^{\{ \nu \}} \big \}$,  
$\big  \{ ( {\boldsymbol{\mathcal{S}}^{(n,k)} } ) ^{\{ \nu \}} \big \}$,
%
$\big  \{ ( {\boldsymbol{\mathcal{G}}^{(n)} } ) ^{\{ \nu \}} \big \}$,
  and $ {\boldsymbol{\mathcal{E}}}^{\{ \nu \}} $ satisfy:}
\begin{align*}
& \textcolor[rgb]{0.00,0.00,0.00}{ 1)\lim_{\nu \rightarrow \infty}{
{
\|\boldsymbol{\mathcal{M}}}-
\Re
(\textbf{[}{\boldsymbol{\mathcal{G}}}\textbf{]})
-{\boldsymbol{\mathcal{E}}}^{\{\nu+1\}}  \| {_{\mathnormal{F}}}   = \| {\boldsymbol{\mathcal{N}}} ^{\{\nu\}}\| {_{\mathnormal{F}}}
=0;}}
 \\
%
&
\textcolor[rgb]{0.00,0.00,0.00}
{2)\lim_{\nu   \rightarrow \infty} {\| ( {\boldsymbol{\mathcal{L}}}^{(n,k)} ) ^{\{\nu+1\}} 
- 
( {\boldsymbol{\mathcal{L}}}^{(n,k)} ) ^{\{\nu\}}
\|{_{\mathnormal{F}}}}
=0;
}
\\
&
\textcolor[rgb]{0.00,0.00,0.00}
{3)
\lim_{\nu  \rightarrow \infty} { \|{\boldsymbol{\mathcal{E}}}^{\{\nu+1\}}-{\boldsymbol{\mathcal{E}}}^{\{\nu\}}
\| {_{\mathnormal{F}}}=0;}}
\\
& \textcolor[rgb]{0.00,0.00,0.00}{ 4)
\lim_{\nu   \rightarrow \infty} {\| ( {\boldsymbol{\mathcal{S}}}^{(n,k)} ) ^{\{\nu+1\}} 
- 
( {\boldsymbol{\mathcal{S}}}^{(n,k)} ) ^{\{\nu\}}
\|{_{\mathnormal{F}}}}
=0;}
\\ &
\textcolor[rgb]{0.00,0.00,0.00}
{5) {\lim_{\nu \rightarrow \infty}{
\| ( {\boldsymbol{\mathcal{G}}}^{(n)}  ) ^{\{\nu+1\}} 
-
 ( {\boldsymbol{\mathcal{G}}}^{(n)}  ) ^{\{\nu\}} 
\|{_{\mathnormal{F}}}}=0.
}}
%
\end{align*}
\end{Theorem}

\begin{Theorem} \label{conver1111}
\textcolor[rgb]{0.00,0.00,0.00}{For any   $n,k \in \{1,2,3\}$,
let
$\big \{{\boldsymbol{\mathcal{Y}}}^{\{\nu\}} \big \}$,   
$\big  \{  {\boldsymbol{\mathcal{Q}}^{(n,k)} }  ^{\{ \nu \}} \big \}$,
$\big  \{ ( {\boldsymbol{\mathcal{L}}^{(n,k)} } ) ^{\{ \nu \}} \big \}$,  
$\big  \{ ( {\boldsymbol{\mathcal{S}}^{(n,k)} } ) ^{\{ \nu \}} \big \}$,
%
$\big  \{ ( {\boldsymbol{\mathcal{G}}^{(n)} } ) ^{\{ \nu \}} \big \}$,
  and $ {\boldsymbol{\mathcal{E}}}^{\{ \nu \}} $
  be  the  sequences
generated by  Algorithm \ref{algorithm1had}.  
Suppose that
the sequences
$\big \{{\boldsymbol{\mathcal{Y}}}^{\{\nu\}} \big \}$ and
$\big  \{  {\boldsymbol{\mathcal{Q}}^{(n,k)} }  ^{\{ \nu \}} \big \}$ 
are bounded.
Then,  any accumulation point of 
the sequence
$\Big \{ \big \{{\boldsymbol{\mathcal{Y}}}^{\{\nu\}} \big \}$,   
$\big  \{  {\boldsymbol{\mathcal{Q}}^{(n,k)} }  ^{\{ \nu \}} \big \}$, 
$\big  \{ ( {\boldsymbol{\mathcal{L}}^{(n,k)} } ) ^{\{ \nu \}} \big \}$,  
$\big  \{ ( {\boldsymbol{\mathcal{S}}^{(n,k)} } ) ^{\{ \nu \}} \big \}$,
%
$\big  \{ ( {\boldsymbol{\mathcal{G}}^{(n)} } ) ^{\{ \nu \}} \big \}$,
   $ {\boldsymbol{\mathcal{E}}}^{\{ \nu \}} \Big\}$
 is a \textit{Karush-Kuhn-Tucker} (KKT) point of 
 (\ref {lag-tttt11RRR}).} 
\end{Theorem}

\vspace{-0.43cm}

\section{\textbf{EXPERIMENTAL RESULTS}}\label{experiments}

In this section, we  perform  extensive experiments    on  
several
HSI datasets
to substantiate the superiority and effectiveness of the proposed HAD approach.
Some of the experimental results are presented in the supplementary materials.
\textcolor[rgb]{0.00,0.00,0.00}{All the experiments are run on the following platforms: 1) Windows 11 and Matlab (R2023b) with 12th Gen Intel(R) Core(TM) i7-12700 2.1GHz CPU and 20GB memory; 2) Windows 11 and Python 3.13//Matlab (R2020b) with NVIDIA GeForce RTX 4090//Intel(R) Core(TM) i9-14900KF 4.56GHz CPU.}

\vspace{-0.4cm}
\subsection
{\textcolor[rgb]{0.00,0.00,0.00}{\textbf{Experimental Datasets}}}

The first Salinas ($126 \times 150 \times 204$) is a synthetic dataset offered by Verdoja and Grangetto \cite{verdoja2020graph}, 
which is generated based on the Salinas real HSI dataset.
The second Pavia dataset ($105 \times 120 \times 102$) is captured by the \textit{Reflection Optical System Imaging Spectroradiometer} (ROSIS) above
the center of Pavia, Italy  \cite{wu2022hyperspectral}.
The third Hyperion dataset ($100 \times 100 \times 145$) is taken by the Hyperion sensor carried by EO-$1$ in the Okavango Delta in
2001 
 \cite{
 xiao2024hyperspectral33
 }.    
The fourth  HYDICE-Urban dataset ($80 \times 100 \times 175$)   is taken by the \textit{Hyperspectral Digital Imagery Collection Experiment}
(HYDICE) sensor  \cite{li2014collaborative}.
The fifth  San Diego dataset ($100 \times 100 \times 189$)  is  collected by the \textit{Airborne Visible/Infrared Imaging Spectrometer} (AVIRIS) sensor over the
State of California, USA  \cite{xu2015anomaly}.
The sixth to the eleventh
HSI datasets (i.e., Airport-4: $100 \times 100 \times 191$,
 Urban-3: $100 \times 100 \times 191$,  Urban-4:  $100 \times 100 \times 205$,
 Urban-5: $100 \times 100 \times 205$,
Beach-3: $100 \times 100 \times 188$, Beach-4: $150 \times 150 \times 102$)
 are from the Airport-Beach-Urban (ABU \cite{kang2017hyperspectral} \footnote{\url{https://ehu.eus/ccwintco/index.php?title=Hyperspectral_Remote_Sensing_Scenes}}) dataset.
 \textcolor[rgb]{0.00,0.00,0.00}{The remaining ones are several larger-scale datasets:
  Qingpu-I dataset ($740 \times 400 \times 250$),  Qingpu-II dataset ($400 \times 600 \times 250$),
  Avon dataset ($400 \times 400 \times 360$), Cri dataset ($400 \times 400 \times 46$),
 beach dataset ($ 452 \times  3072 \times  108 $),
   Sequoia National Park (SNP) dataset ($1116 \times  2499 \times  13$),  synthetic dataset ($ 600 \times 2400 \times 90 $).
   Detailed descriptions of these large-scale datasets can be found in references \cite{wang2025leveraging,garske2025erx55}.}

\begin{figure*}[!htbp]
\renewcommand{\arraystretch}{0.2}
\setlength\tabcolsep{0.5pt}
\centering
\begin{tabular}{ccc ccc  ccc  cc}
\centering
\includegraphics[width=0.60in, height=0.585in,angle=0]{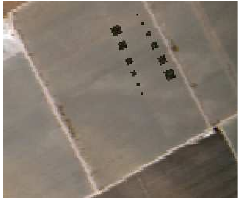}&
\includegraphics[width=0.60in, height=0.585in]{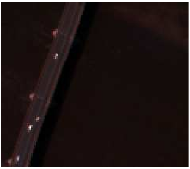}&
\includegraphics[width=0.60in, height=0.585in]{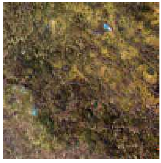}&
\includegraphics[width=0.60in, height=0.585in]{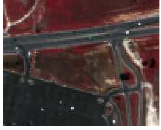}&
\includegraphics[width=0.60in, height=0.585in]{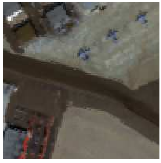}&
\includegraphics[width=0.60in, height=0.585in]{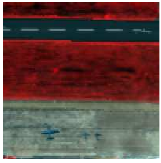}&
\includegraphics[width=0.60in, height=0.585in]{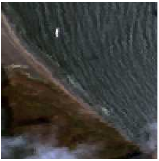}&
\includegraphics[width=0.60in, height=0.585in]{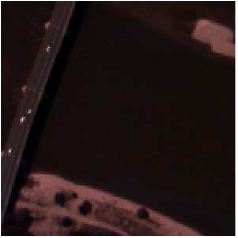}&
\includegraphics[width=0.60in, height=0.585in]{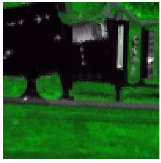}&
\includegraphics[width=0.60in, height=0.585in]{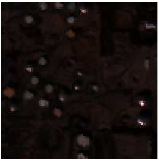}&
\includegraphics[width=0.60in, height=0.585in]{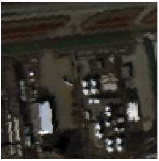}
\\

\includegraphics[width=0.60in, height=0.585in,angle=0]{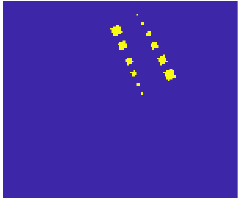}&
\includegraphics[width=0.60in, height=0.585in]{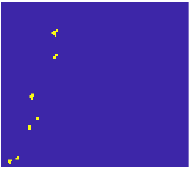}&
\includegraphics[width=0.60in, height=0.585in]{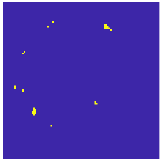}&
\includegraphics[width=0.60in, height=0.585in]{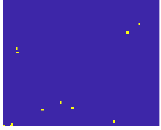}&
\includegraphics[width=0.60in, height=0.585in]{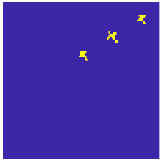}&
\includegraphics[width=0.60in, height=0.585in]{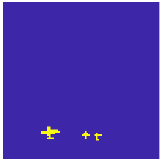}&
\includegraphics[width=0.60in, height=0.585in]{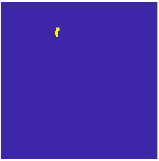}&
\includegraphics[width=0.60in, height=0.585in]{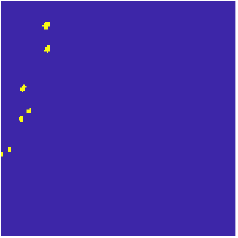}&
\includegraphics[width=0.60in, height=0.585in]{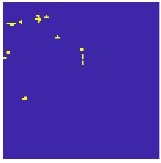}&  %
\includegraphics[width=0.60in, height=0.585in]{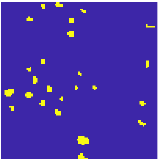}&
\includegraphics[width=0.60in, height=0.585in]{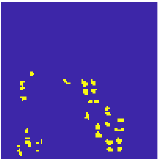}
\\

{{{(a)}}}  &
  {(b)}  & {(c)}
&{(d)} & {(e)}
 &{(f)}& {(g)}&
 {(h)}&
 {(i)}
 &  {(j)}
  &
 {(k)}
\end{tabular}
\caption{
Pseudocolor images and ground-truth maps of eleven 
 HSI datasets.
(a) Salinas.
(b) Pavia.
(c) Hyperion.
(d) HYDICE.
(e) San-Diego.
(f) Airport-4.
(g) Beach-3.
(h) Beach-4.
(i) Urban-3.
(j) Urban-4.
(k)   Urban-5.
}
\vspace{-0.5cm}
\label{fig_visual_hsi}
\end{figure*}

The pseudocolor map and the corresponding ground-truth
map of the afore-mentioned HSIs 
 are shown in Figure \ref{fig_visual_hsi}.

%

\begin{figure*}[!htbp]
\renewcommand{\arraystretch}{0.4}
\setlength\tabcolsep{0.43pt}
\centering
\begin{tabular}{ccc  ccc cc c ccc c  cc}
\centering
\includegraphics[width=0.4631in, height=0.4631in]{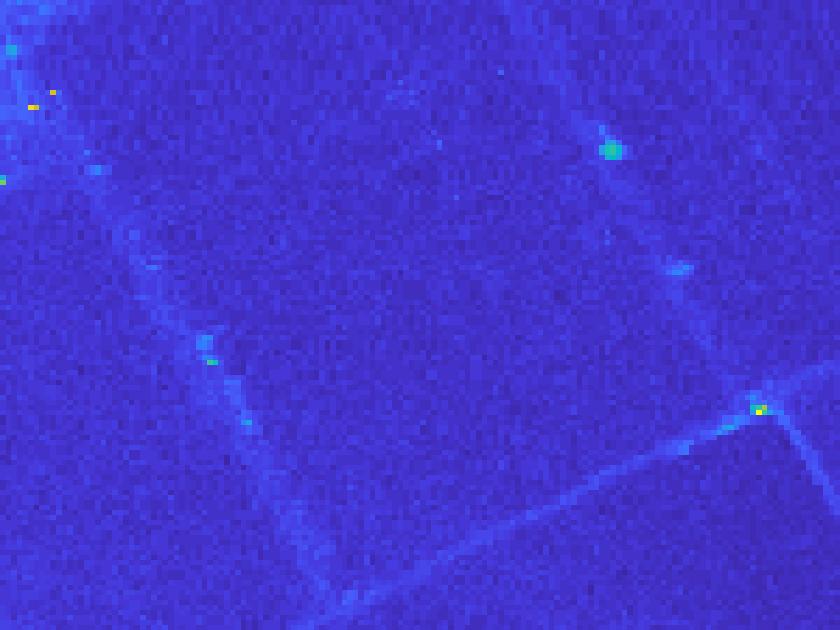}&
\includegraphics[width=0.4631in, height=0.4631in]{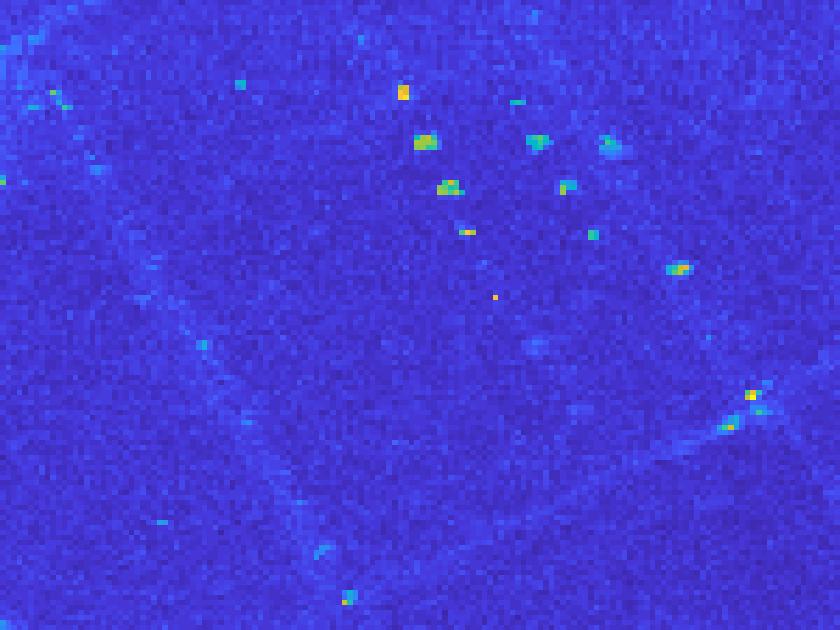}&
\includegraphics[width=0.4631in, height=0.4631in]{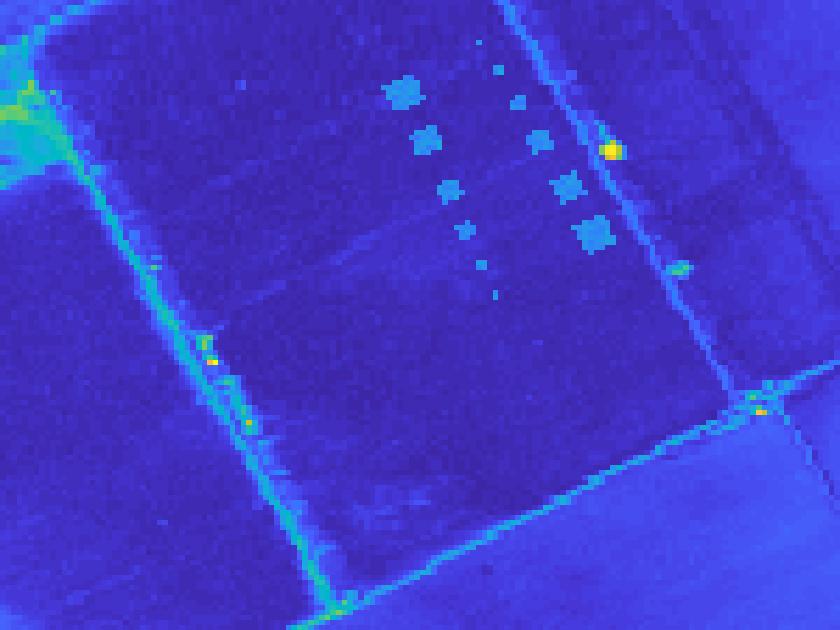}&
\includegraphics[width=0.4631in, height=0.4631in]{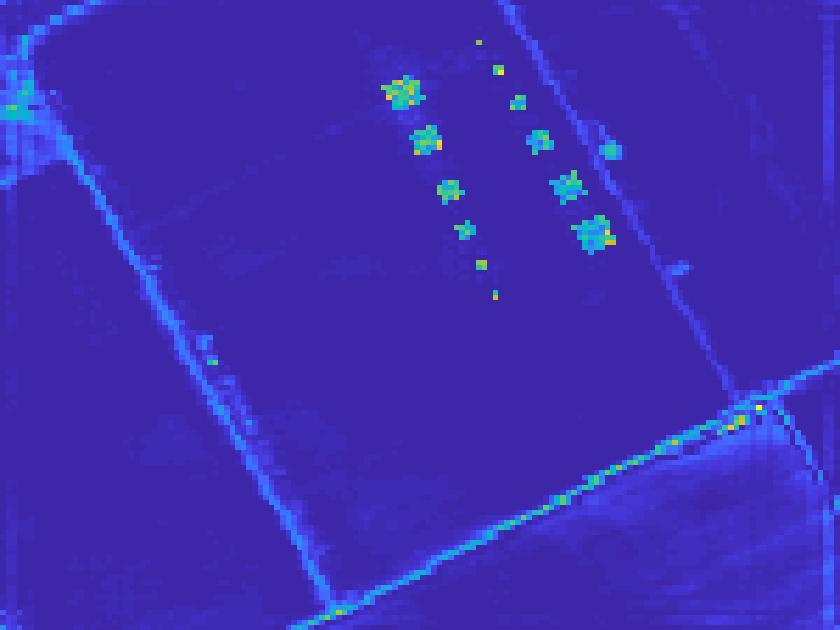}&
\includegraphics[width=0.4631in, height=0.4631in]{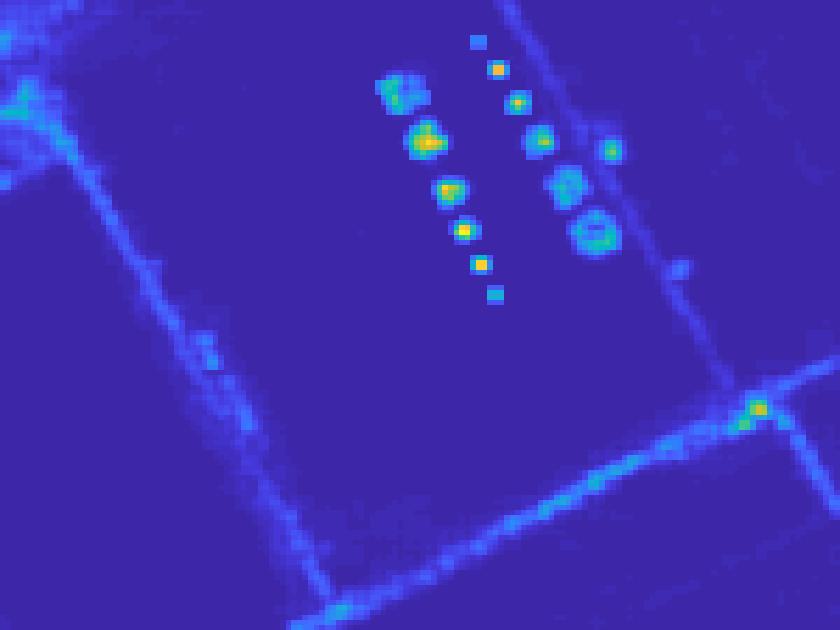}&
\includegraphics[width=0.4631in, height=0.4631in]{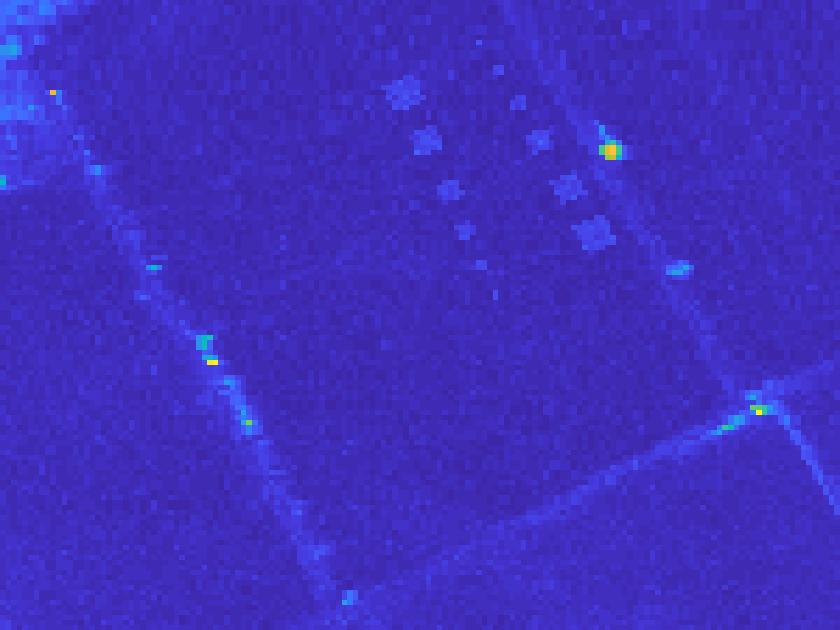}&
\includegraphics[width=0.4631in, height=0.4631in]{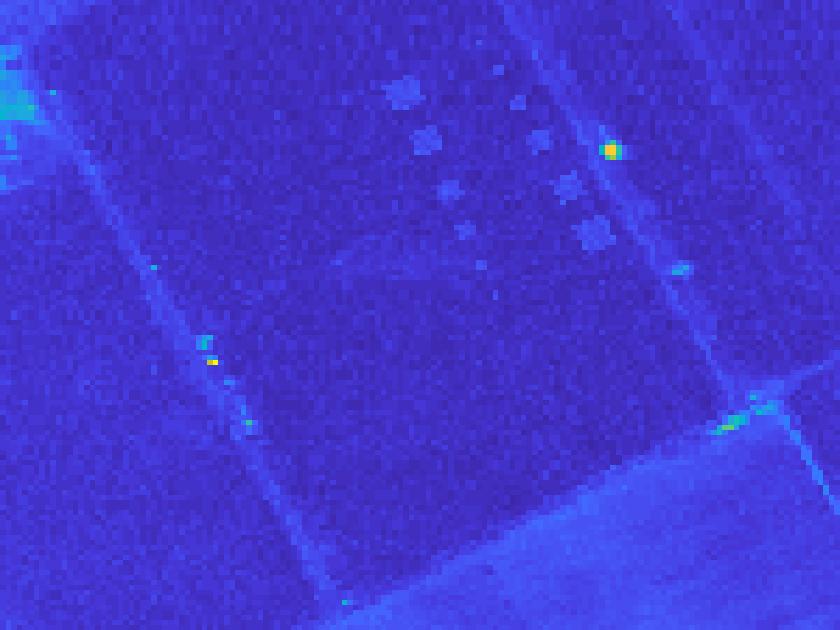}&
\includegraphics[width=0.4631in, height=0.4631in]{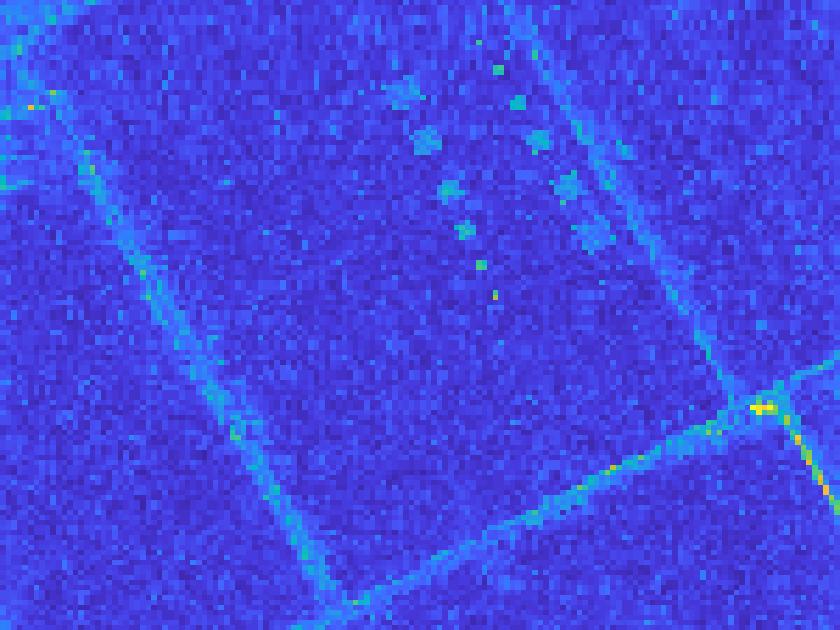}&  %
\includegraphics[width=0.4631in, height=0.4631in]{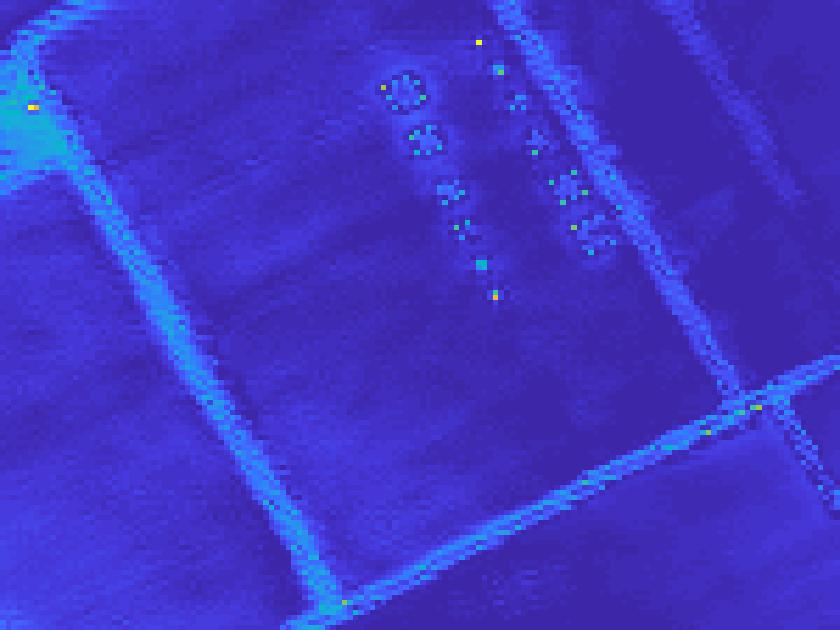} &
\includegraphics[width=0.4631in, height=0.4631in]{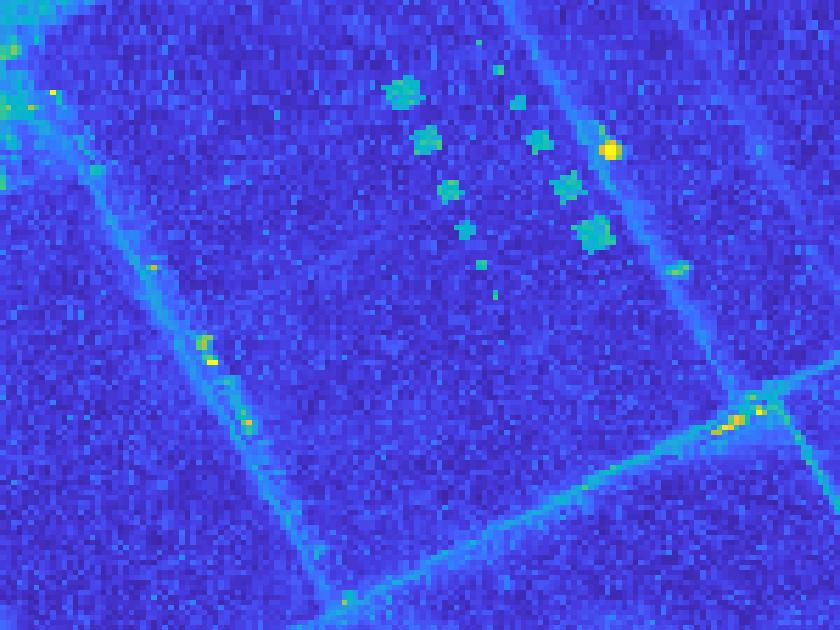} &
\includegraphics[width=0.4631in, height=0.4631in]{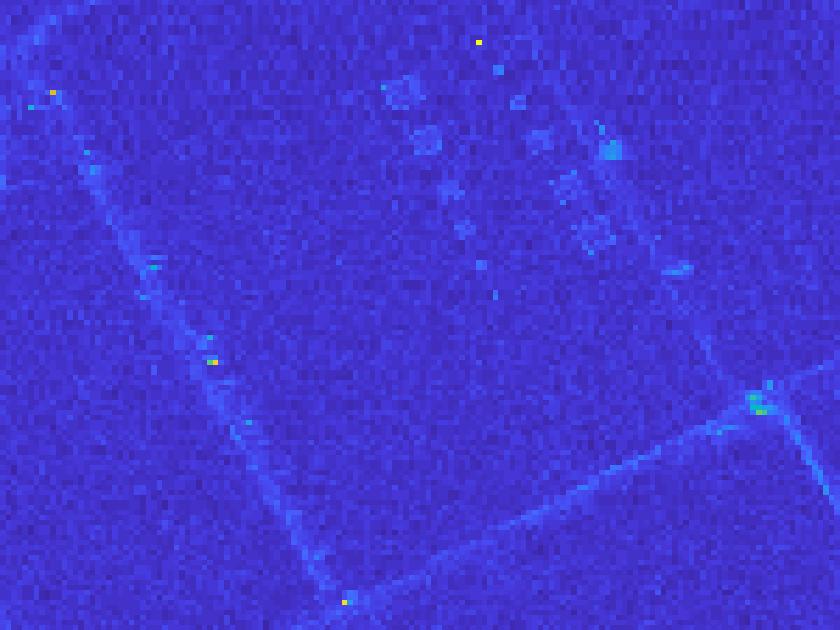}&
\includegraphics[width=0.4631in, height=0.4631in]{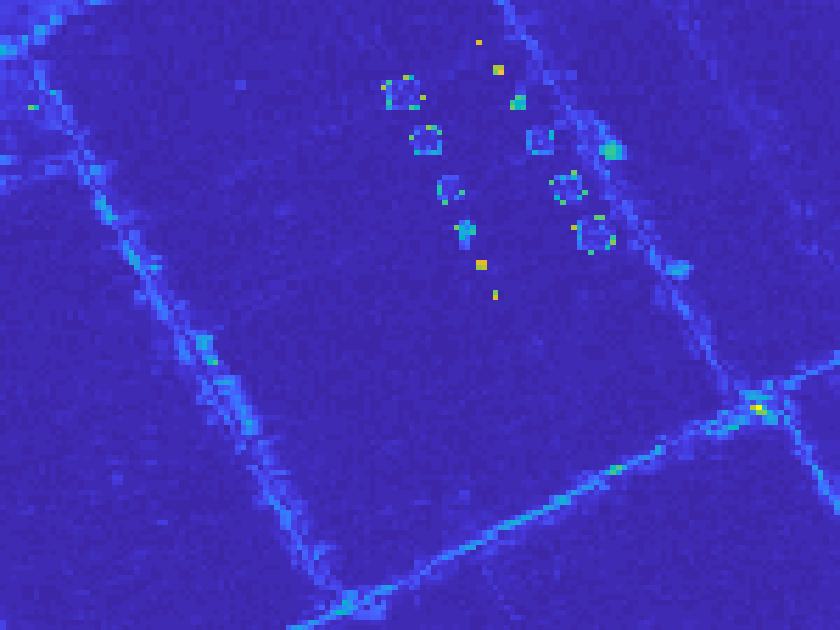}&
\includegraphics[width=0.4631in, height=0.4631in]{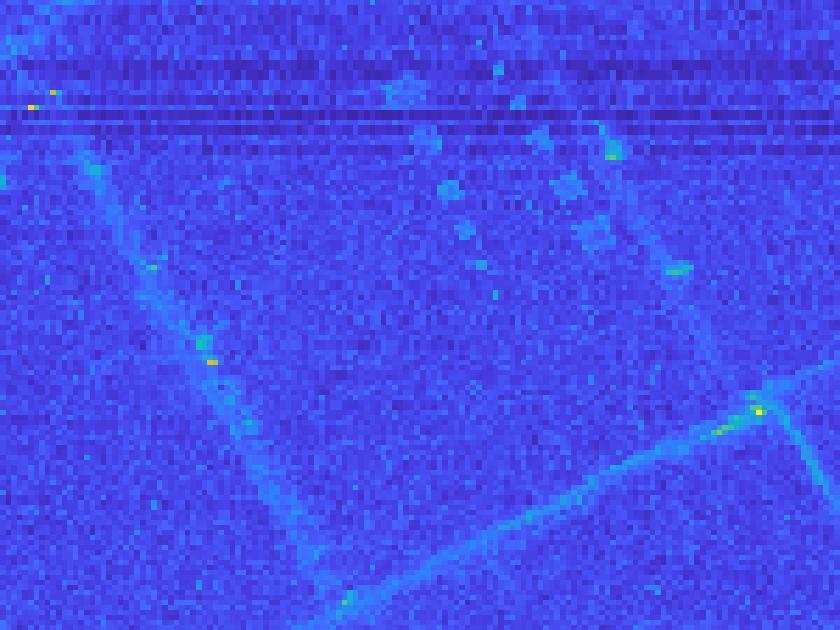} &
\includegraphics[width=0.4631in, height=0.4631in]{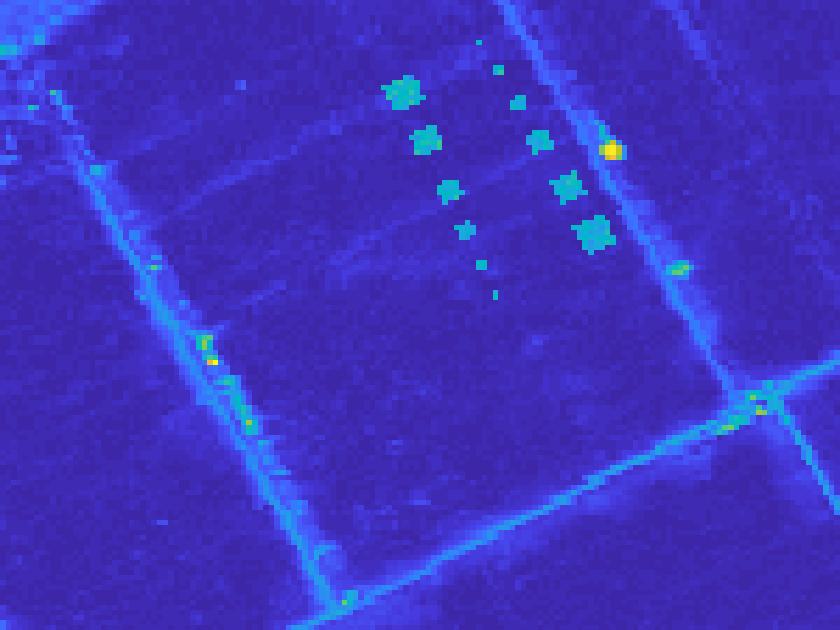}&
\includegraphics[width=0.4631in, height=0.4631in]{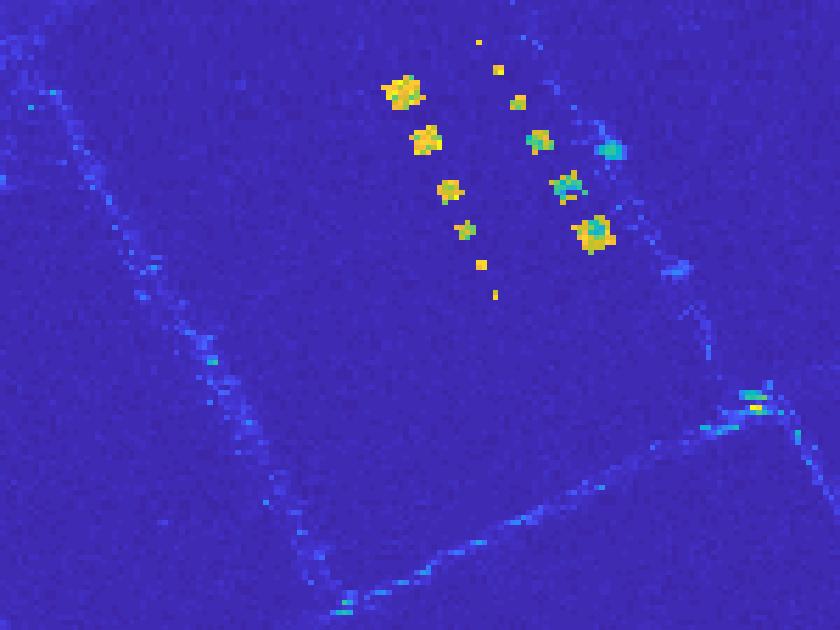}
\\
\includegraphics[width=0.4631in, height=0.4631in]{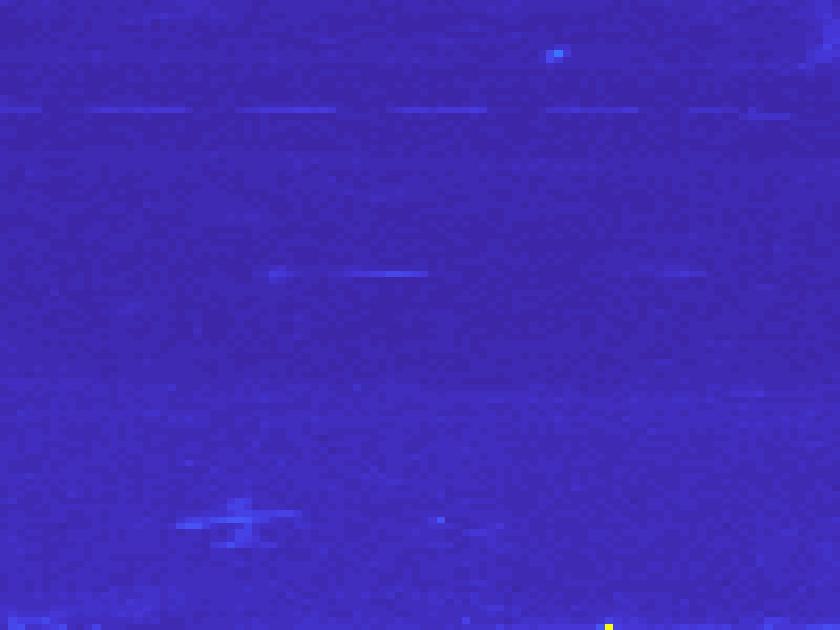}&
\includegraphics[width=0.4631in, height=0.4631in]{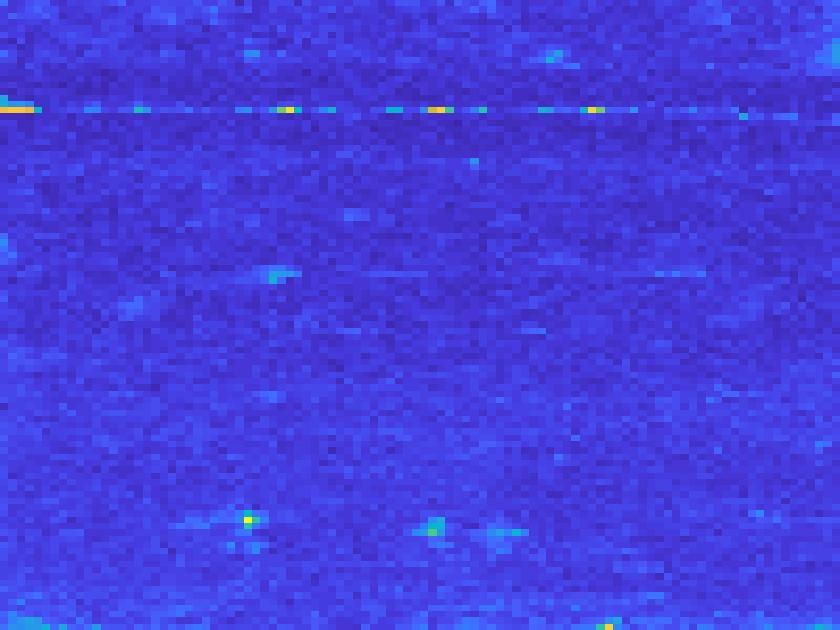}&
\includegraphics[width=0.4631in, height=0.4631in]{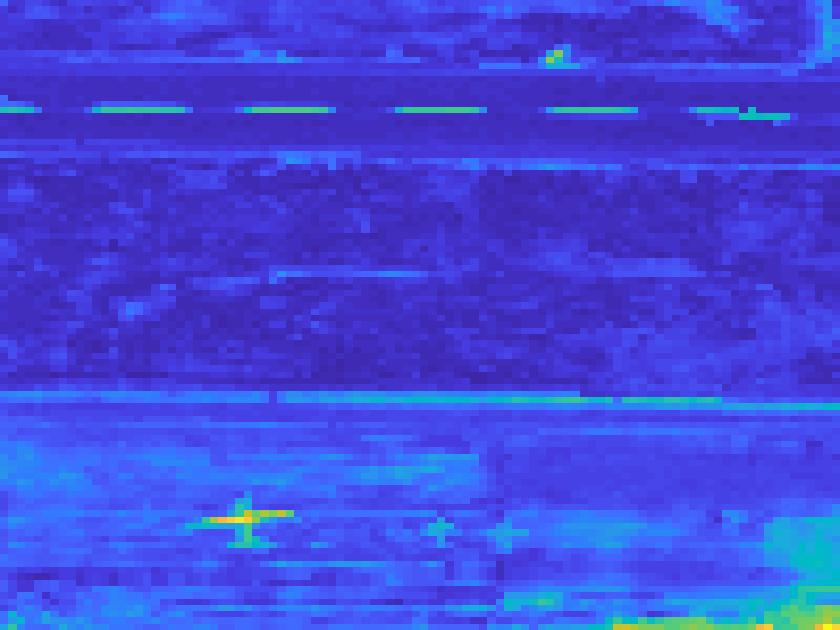}&
\includegraphics[width=0.4631in, height=0.4631in]{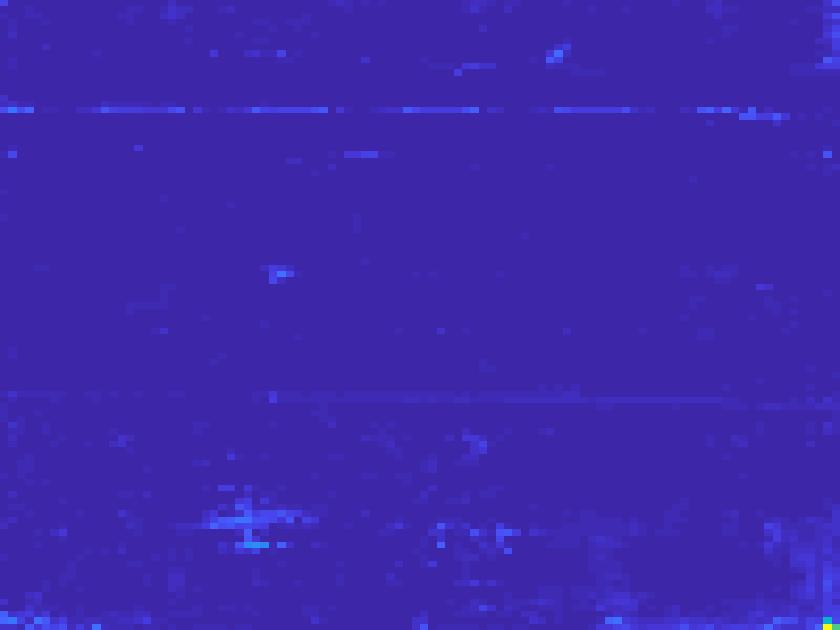}&
\includegraphics[width=0.4631in, height=0.4631in]{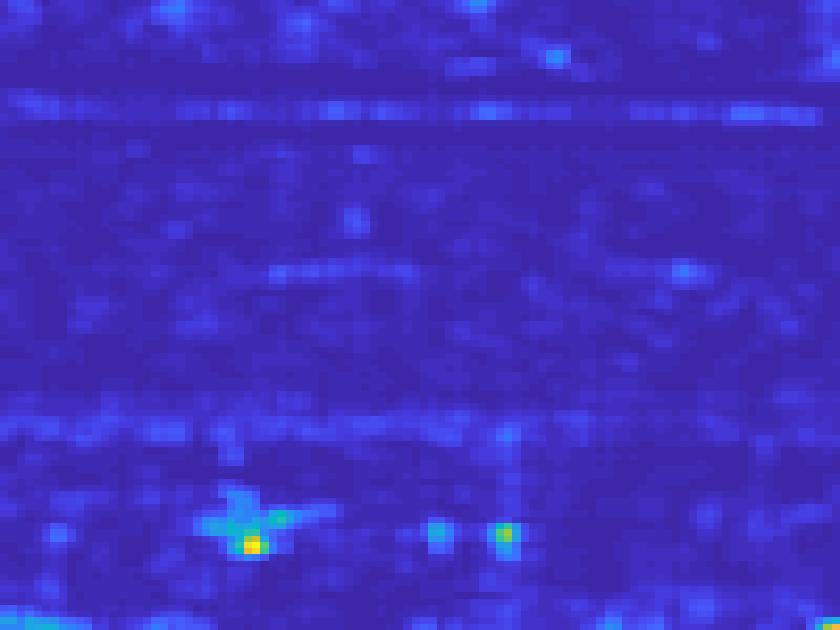}&
\includegraphics[width=0.4631in, height=0.4631in]{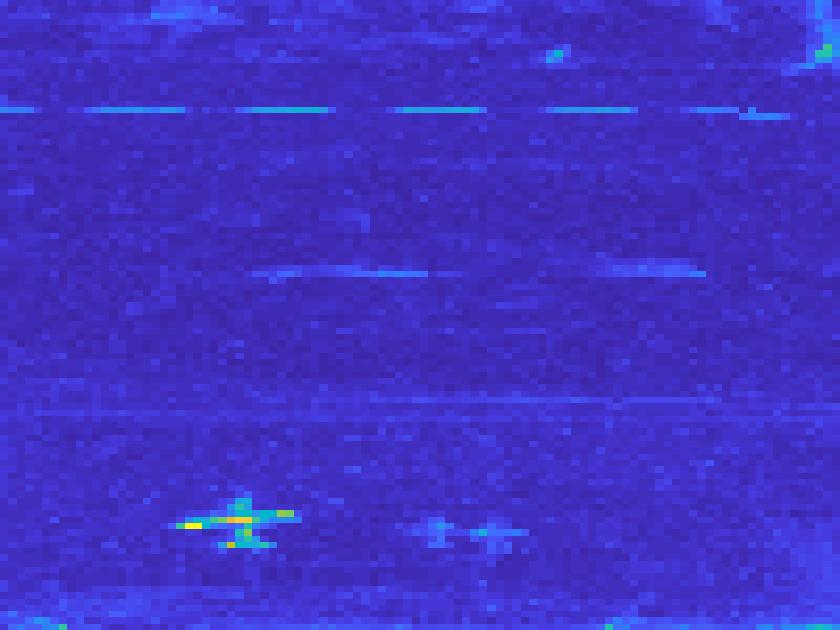}&
\includegraphics[width=0.4631in, height=0.4631in]{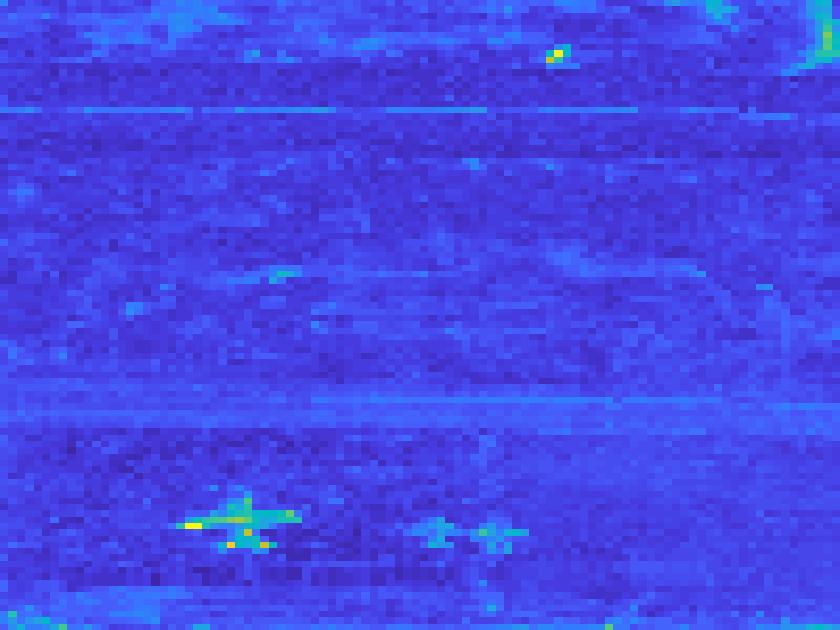}&
\includegraphics[width=0.4631in, height=0.4631in]{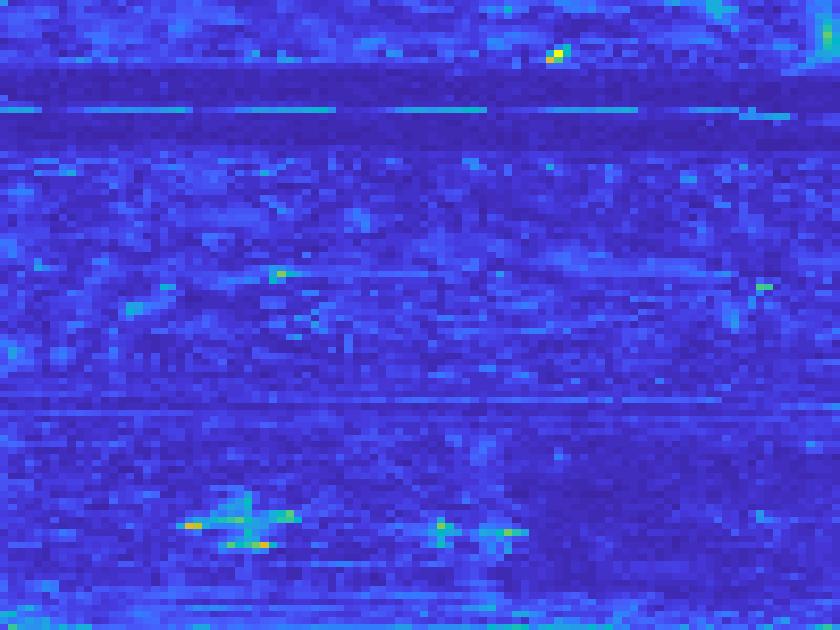}&  %
\includegraphics[width=0.4631in, height=0.4631in]{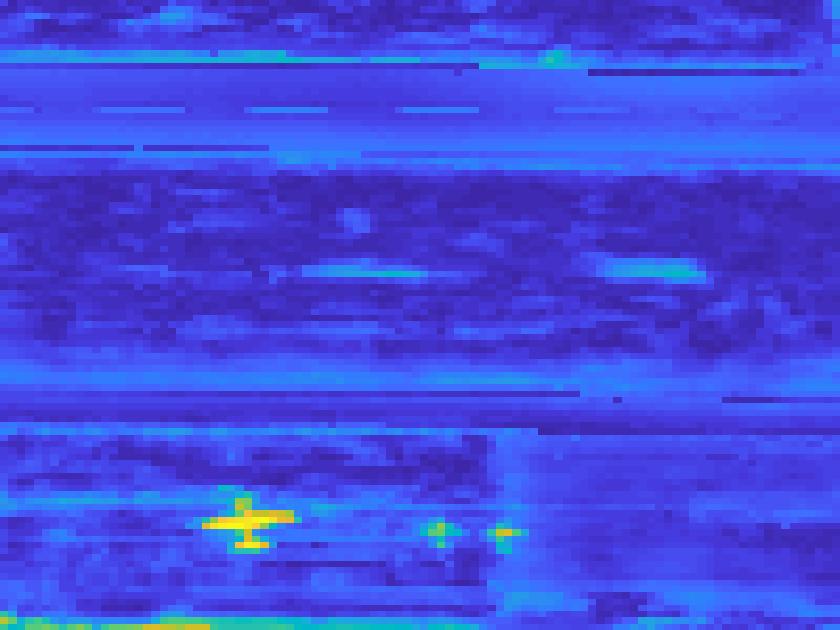} &
\includegraphics[width=0.4631in, height=0.4631in]{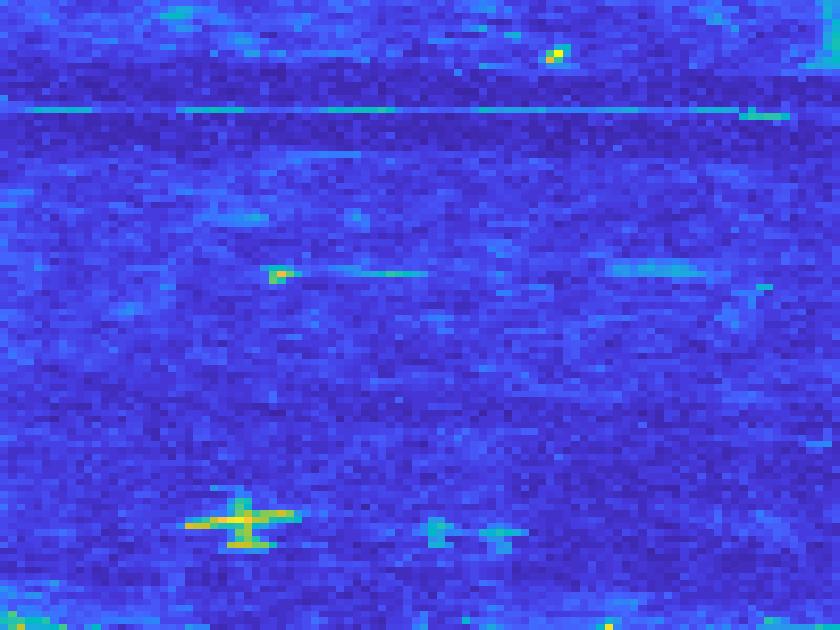} &
\includegraphics[width=0.4631in, height=0.4631in]{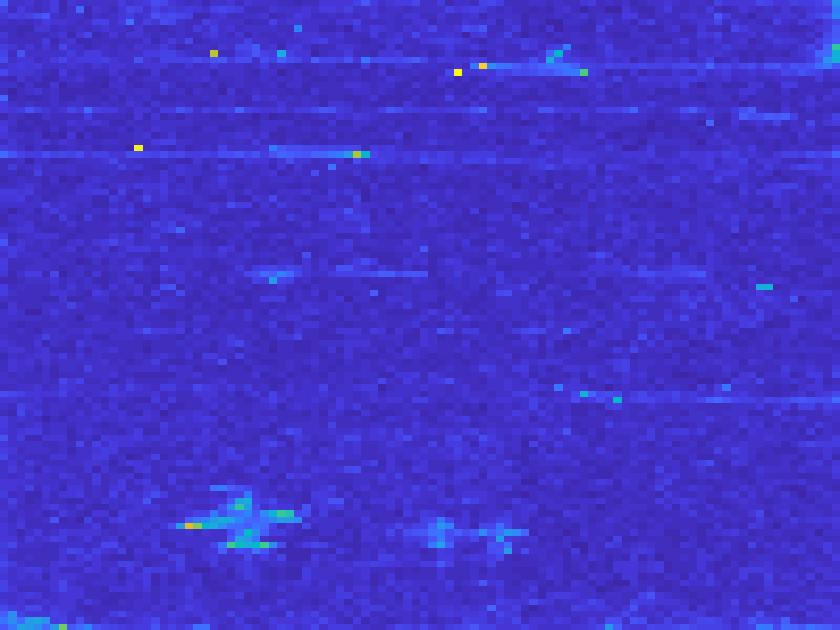}&
\includegraphics[width=0.4631in, height=0.4631in]{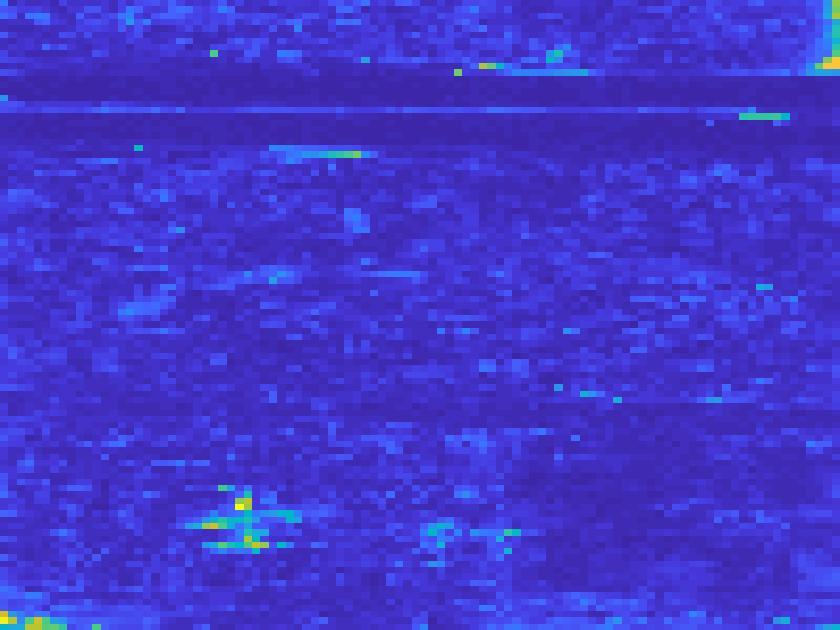}&
\includegraphics[width=0.4631in, height=0.4631in]{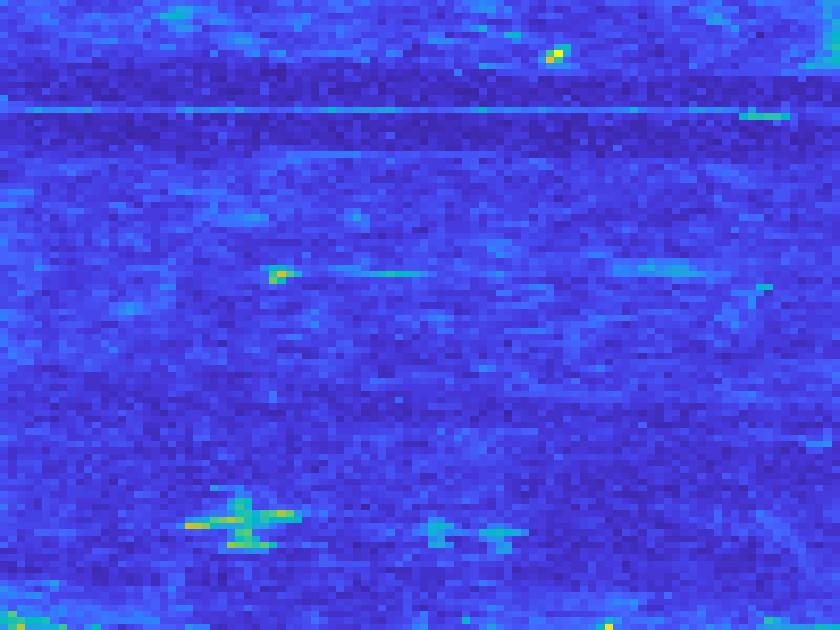} &
\includegraphics[width=0.4631in, height=0.4631in]{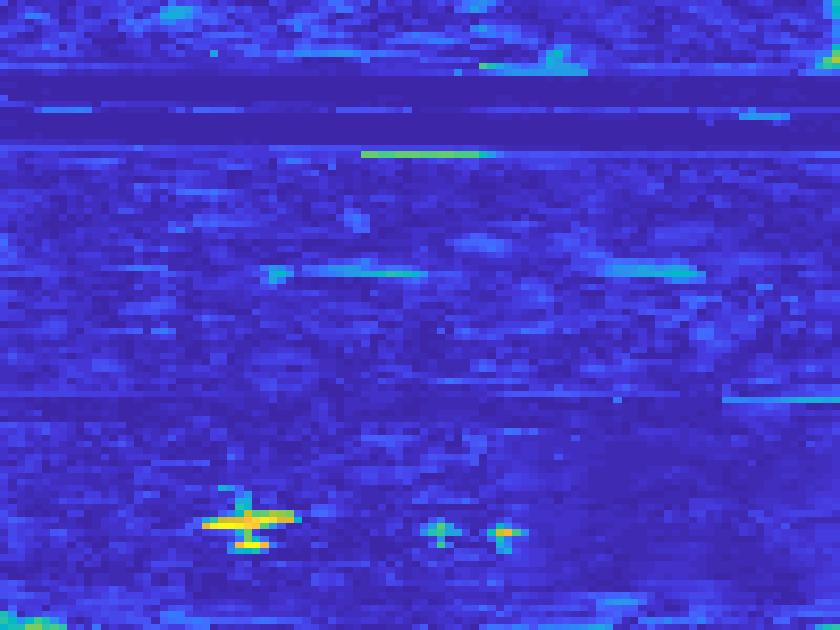}&
\includegraphics[width=0.4631in, height=0.4631in]{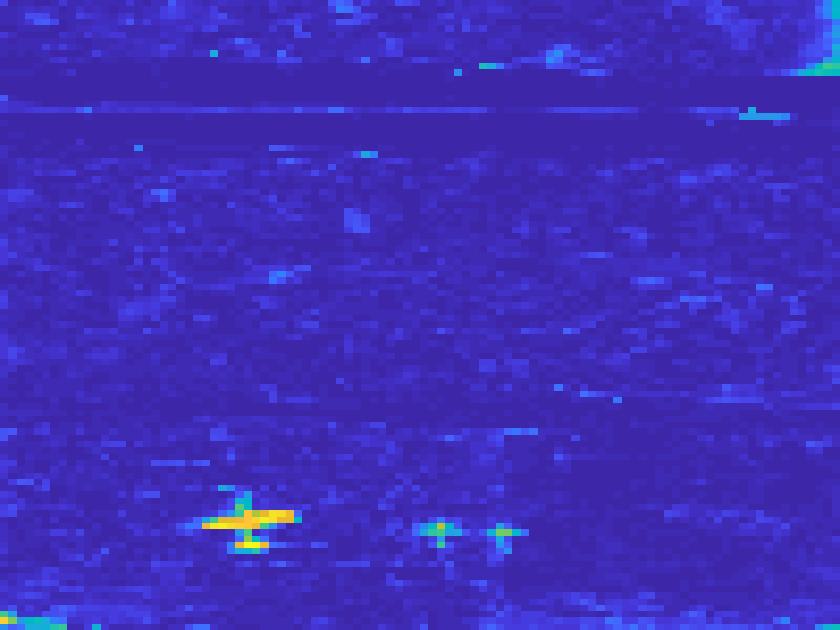}
\\
\includegraphics[width=0.4631in, height=0.4631in]{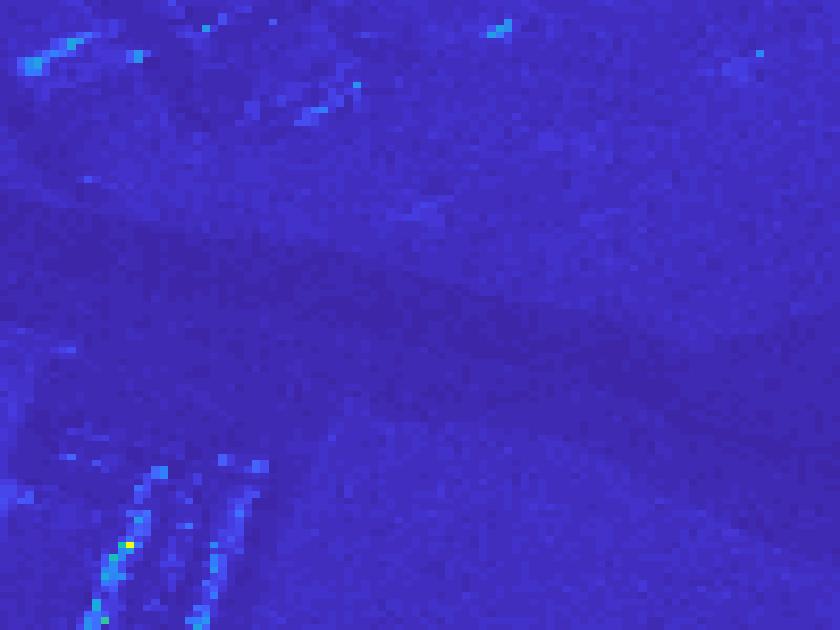}&
\includegraphics[width=0.4631in, height=0.4631in]{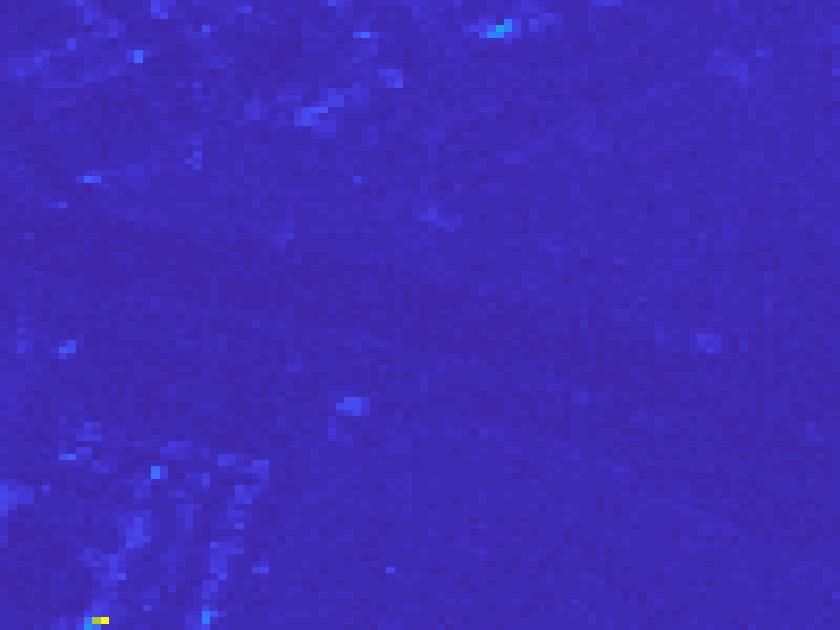}&
\includegraphics[width=0.4631in, height=0.4631in]{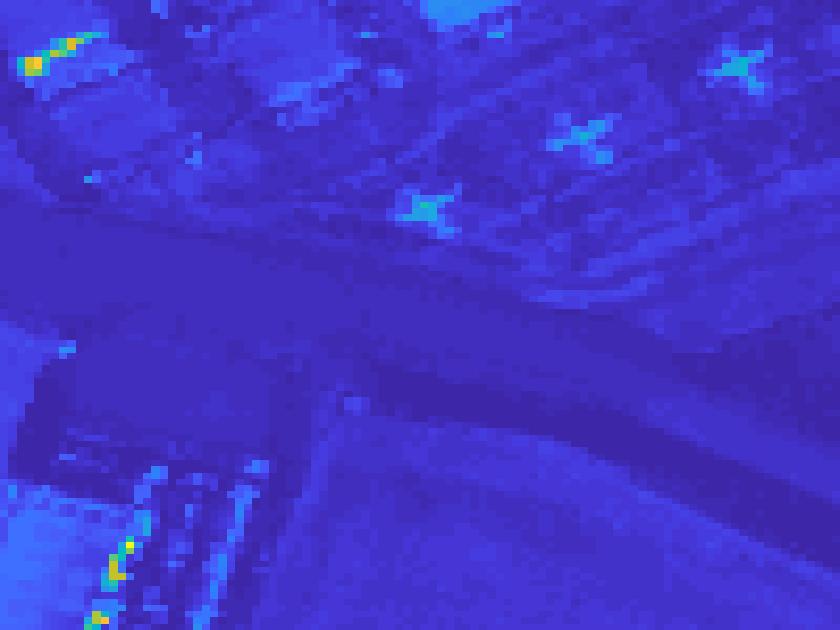}&
\includegraphics[width=0.4631in, height=0.4631in]{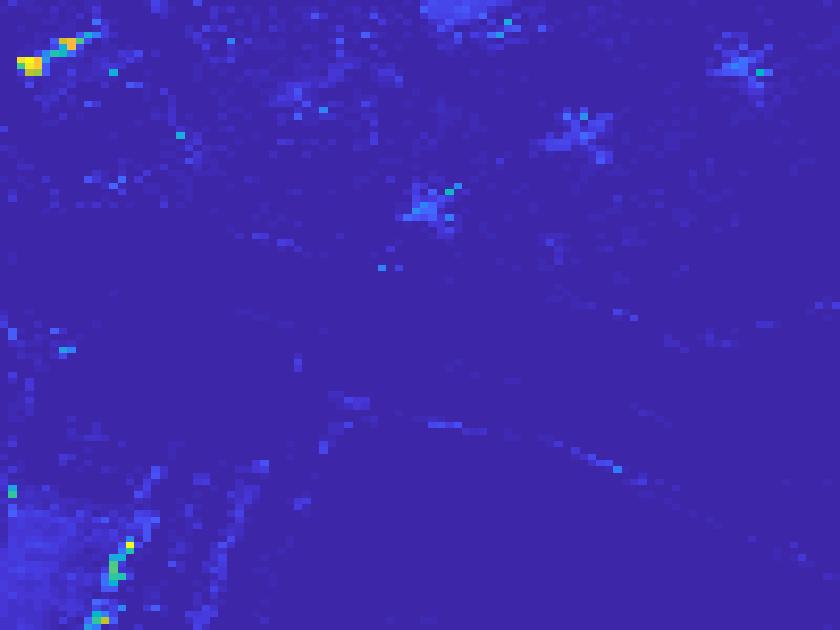}&
\includegraphics[width=0.4631in, height=0.4631in]{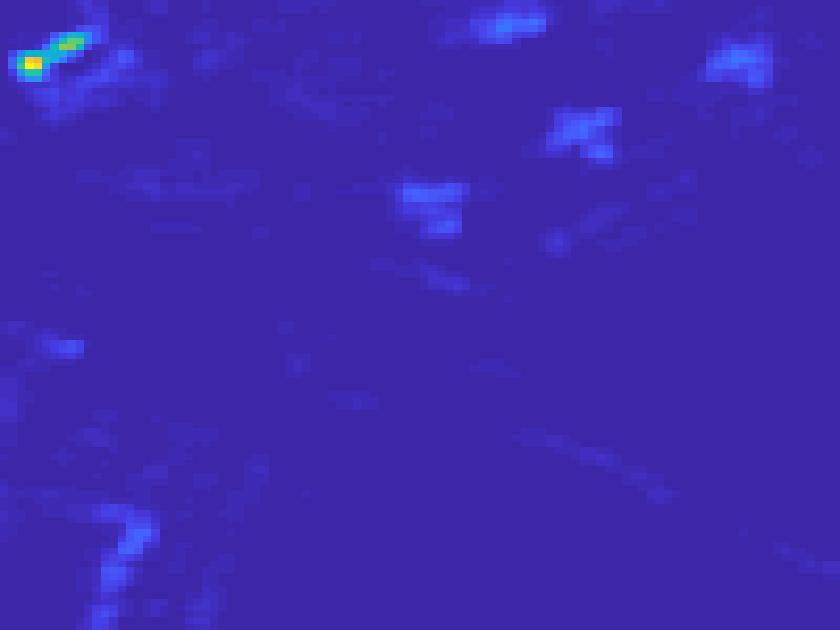}&
\includegraphics[width=0.4631in, height=0.4631in]{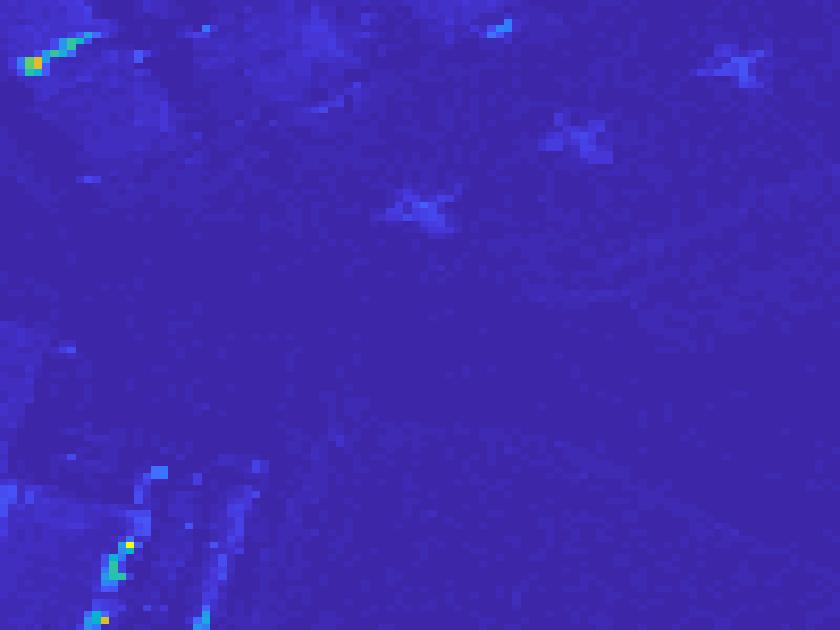}&
\includegraphics[width=0.4631in, height=0.4631in]{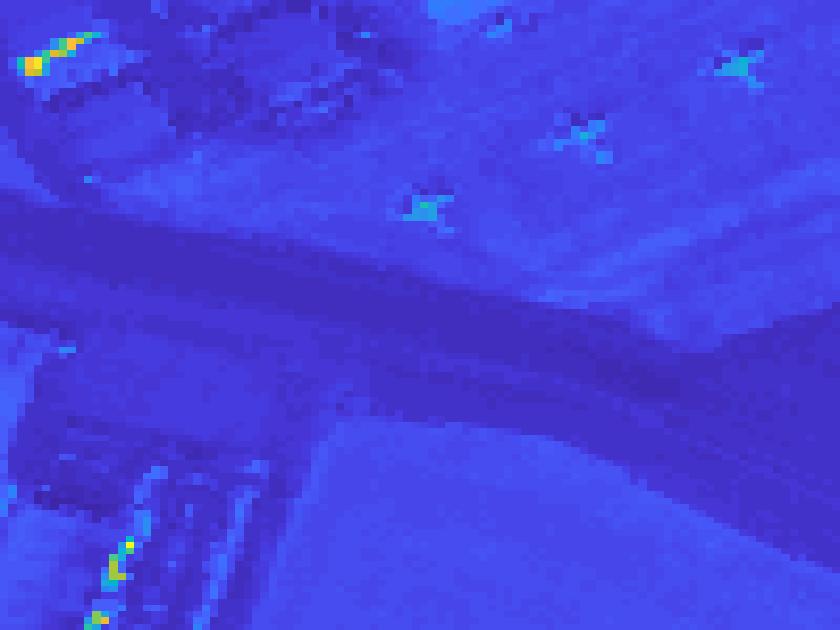}&
\includegraphics[width=0.4631in, height=0.4631in]{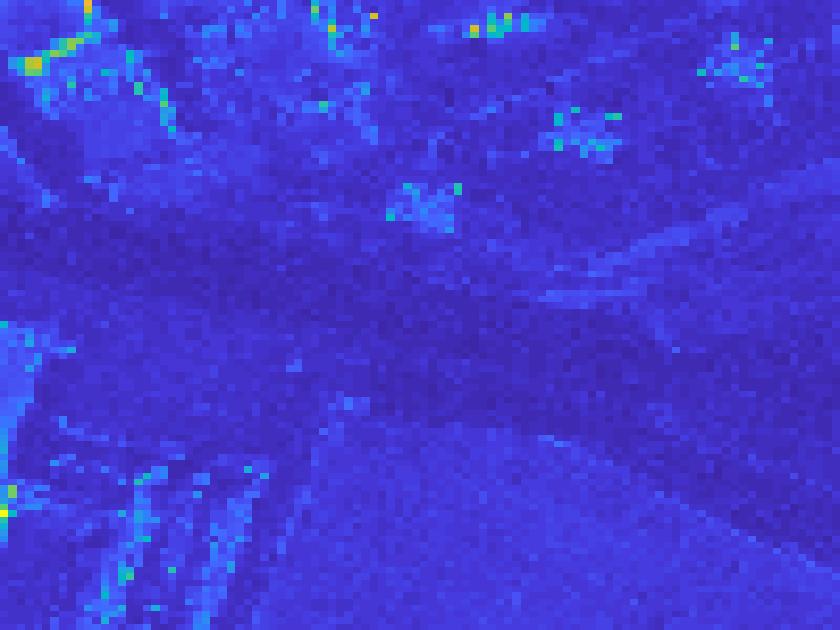}&  %
\includegraphics[width=0.4631in, height=0.4631in]{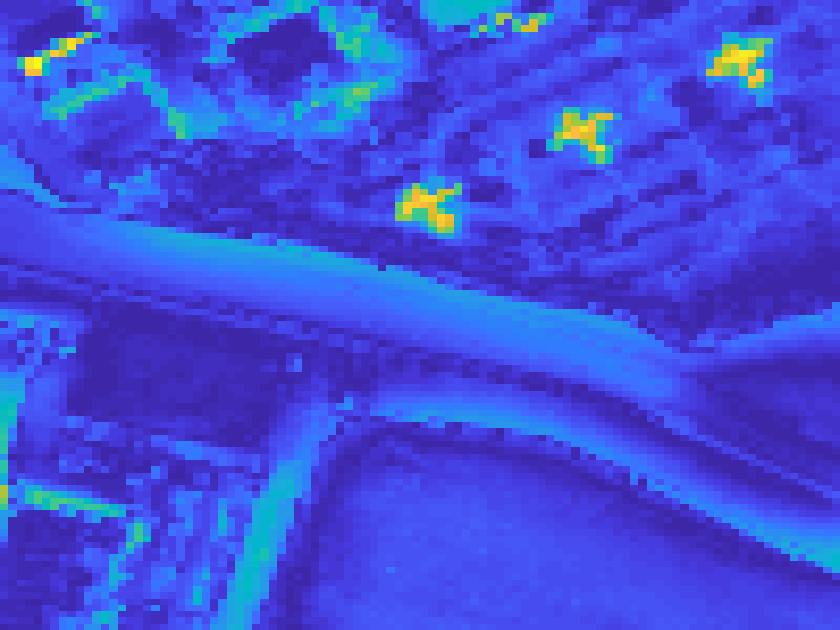} &
\includegraphics[width=0.4631in, height=0.4631in]{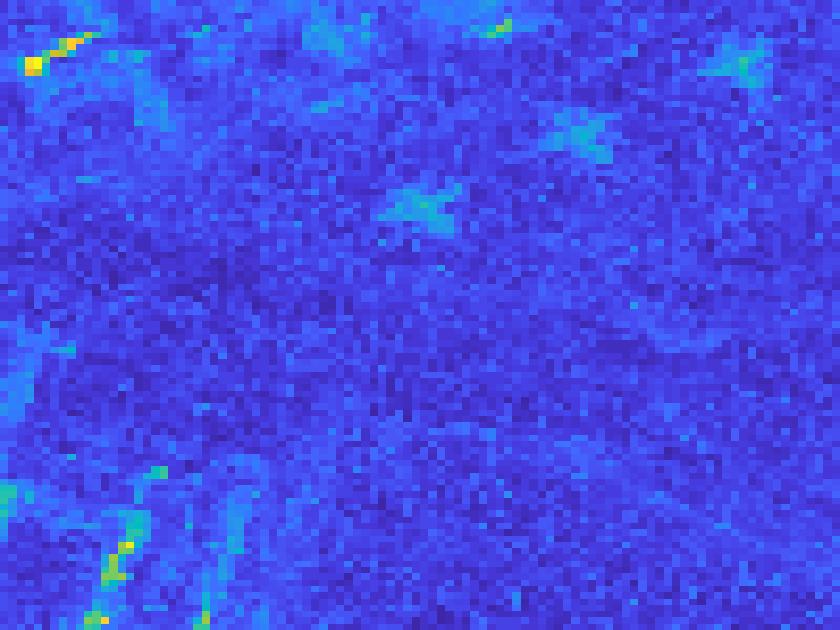} &
\includegraphics[width=0.4631in, height=0.4631in]{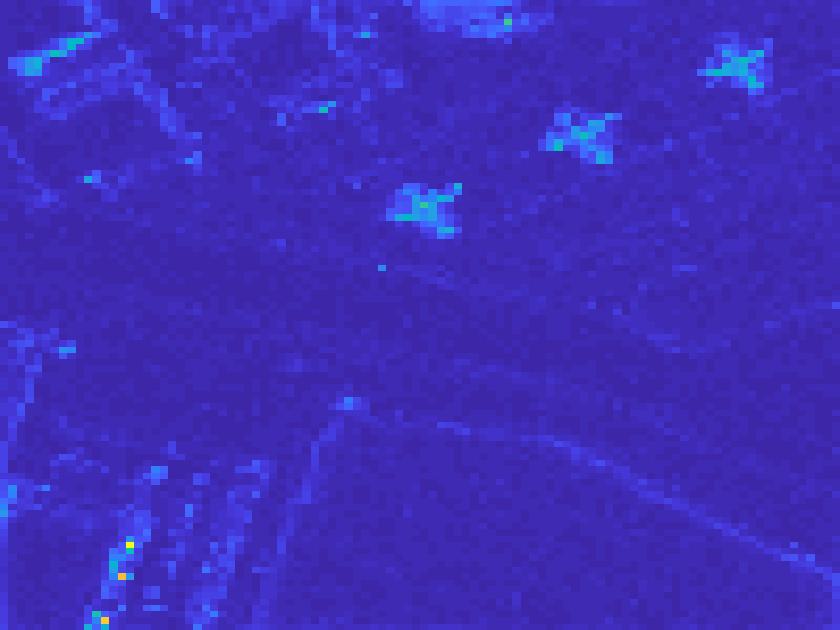}&
\includegraphics[width=0.4631in, height=0.4631in]{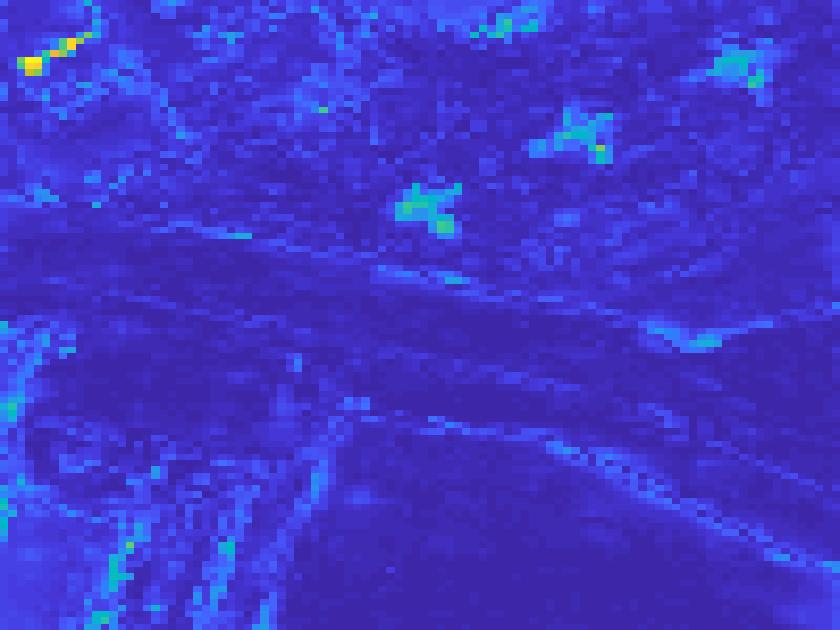}&
\includegraphics[width=0.4631in, height=0.4631in]{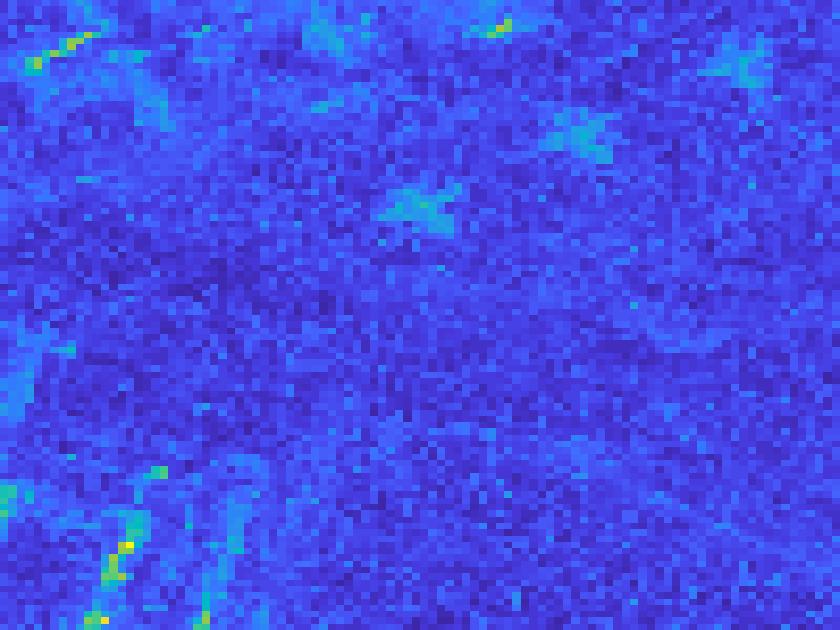} &
\includegraphics[width=0.4631in, height=0.4631in]{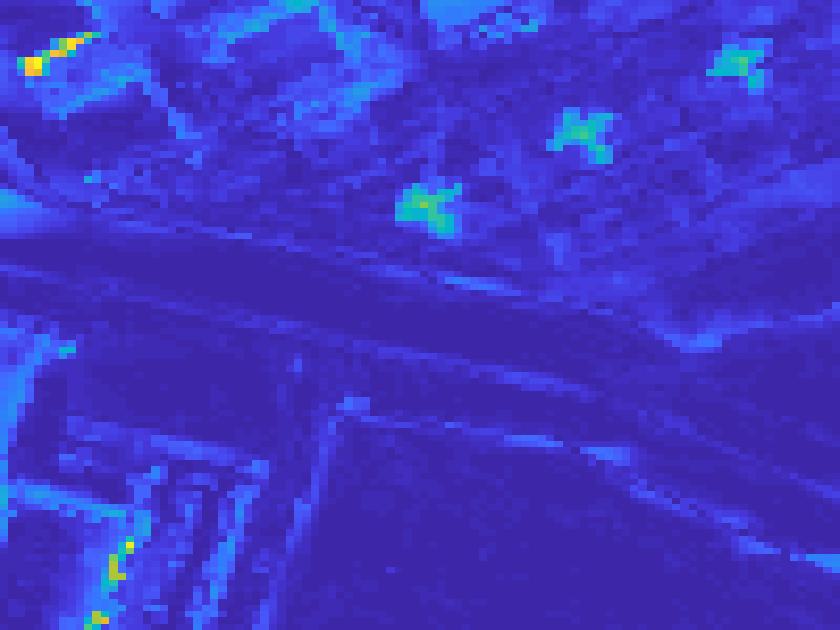}&
\includegraphics[width=0.4631in, height=0.4631in]{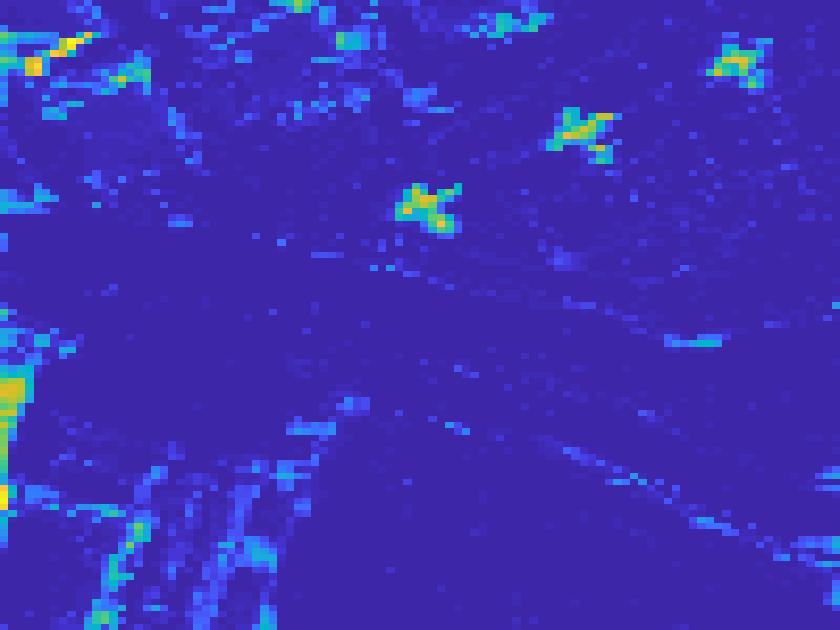}
\\

\includegraphics[width=0.4631in, height=0.4631in]{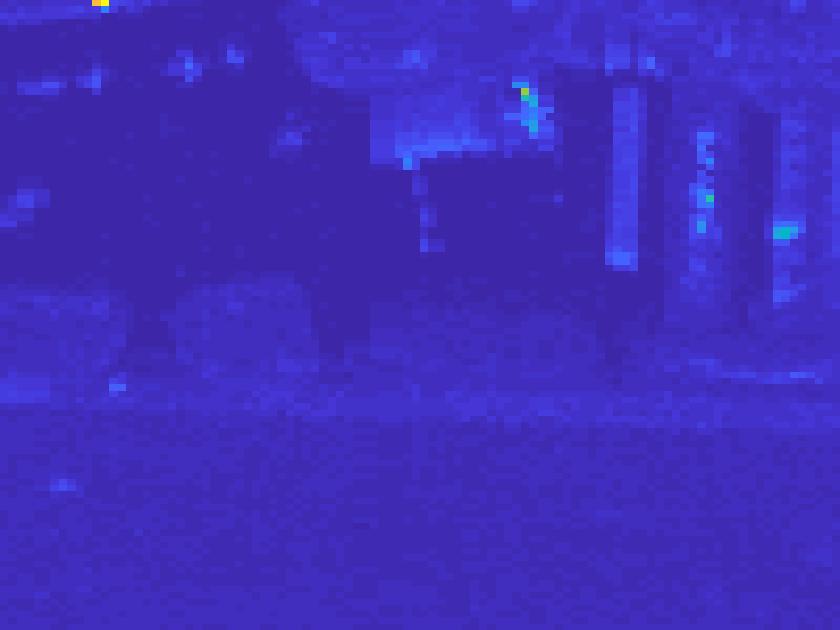}&
\includegraphics[width=0.4631in, height=0.4631in]{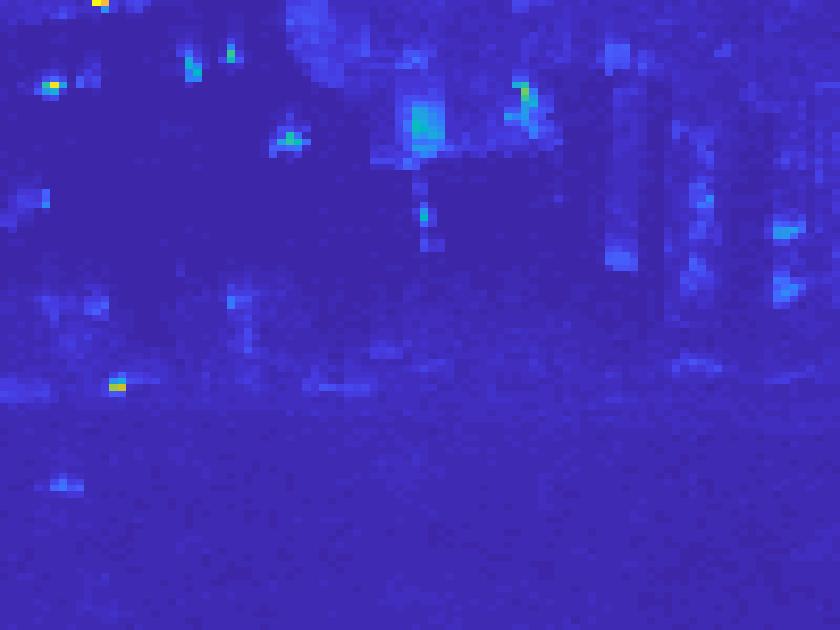}&
\includegraphics[width=0.4631in, height=0.4631in]{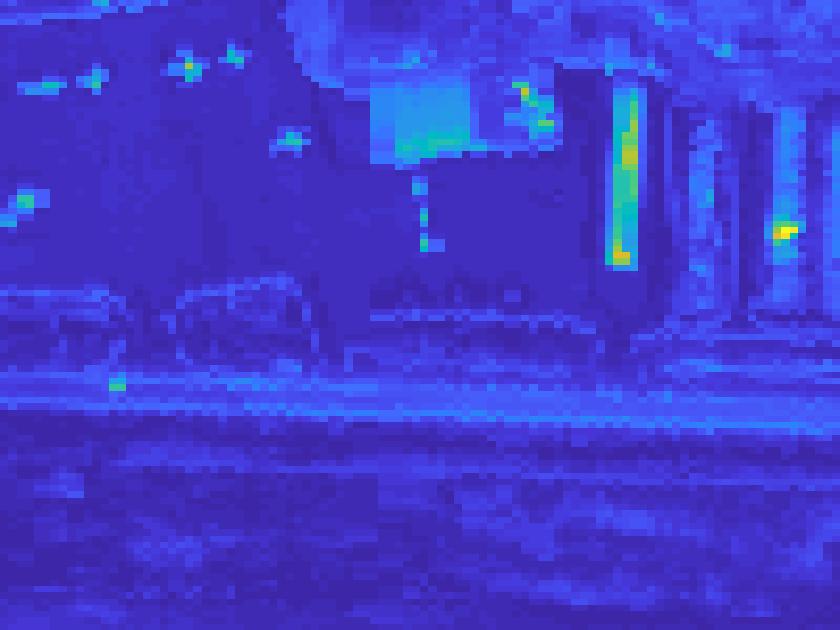}&
\includegraphics[width=0.4631in, height=0.4631in]{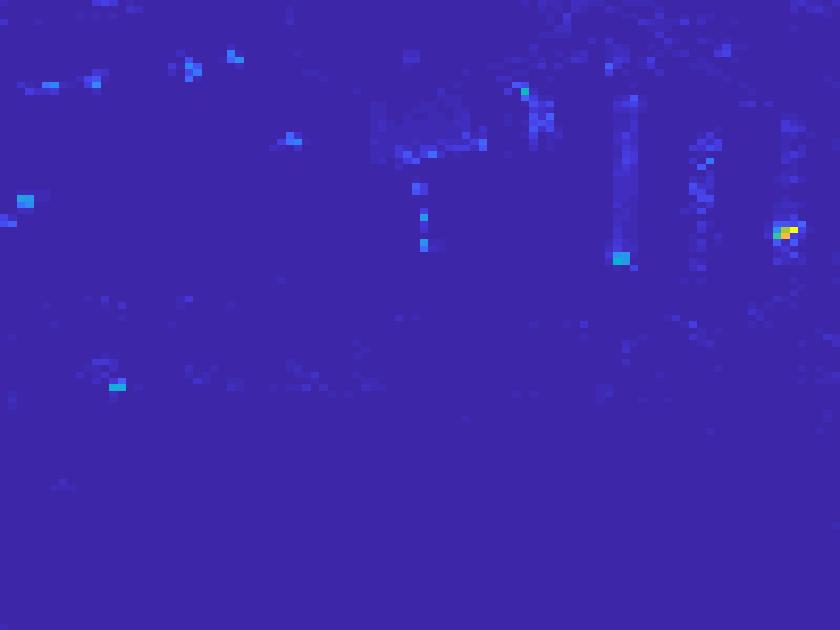}&
\includegraphics[width=0.4631in, height=0.4631in]{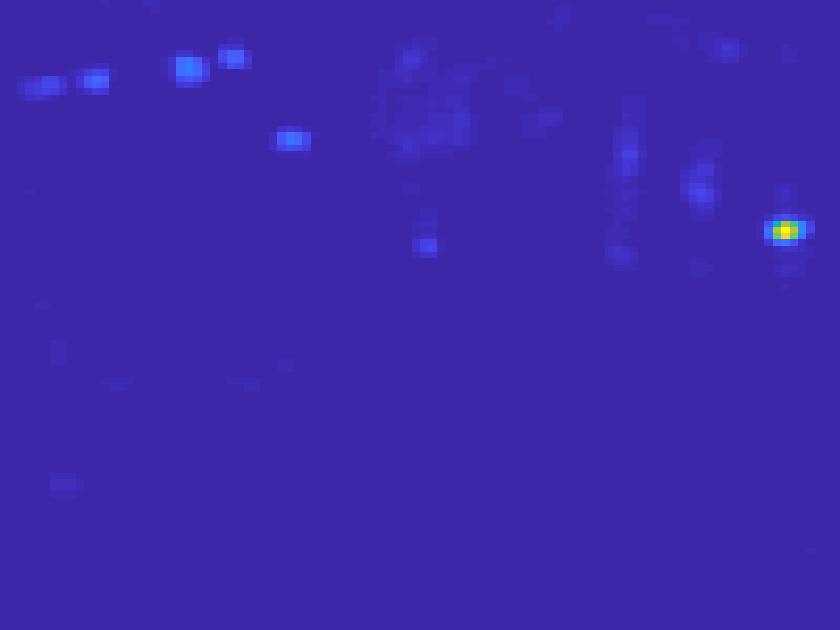}&
\includegraphics[width=0.4631in, height=0.4631in]{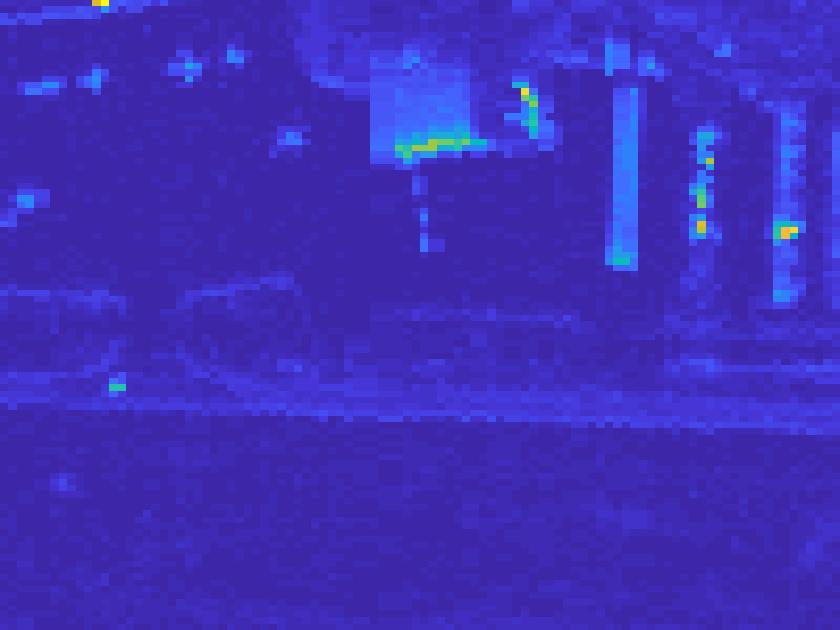}&
\includegraphics[width=0.4631in, height=0.4631in]{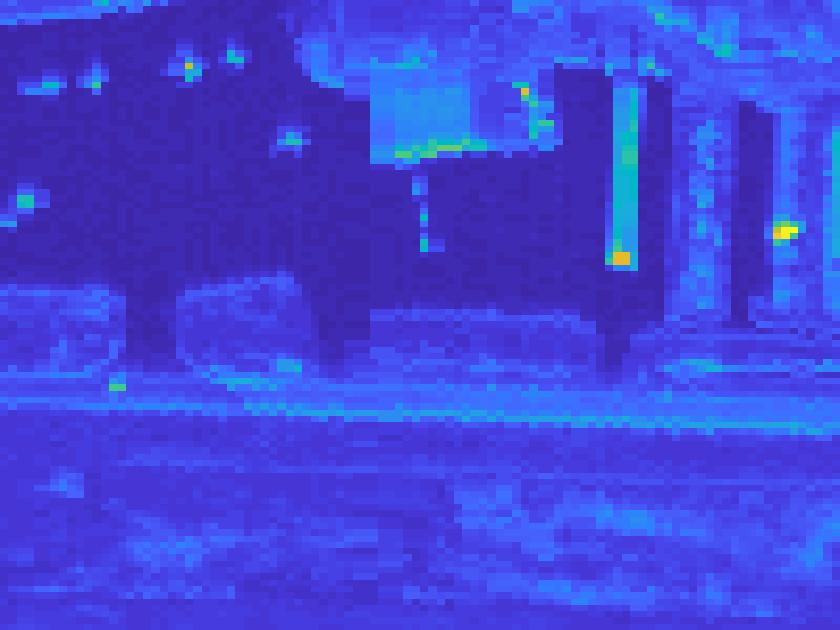}&
\includegraphics[width=0.4631in, height=0.4631in]{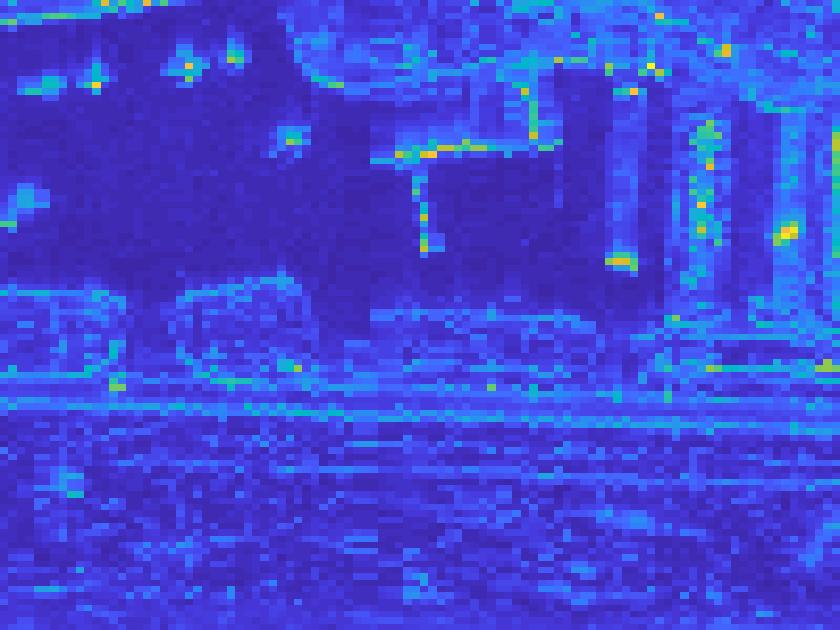}&  %
\includegraphics[width=0.4631in, height=0.4631in]{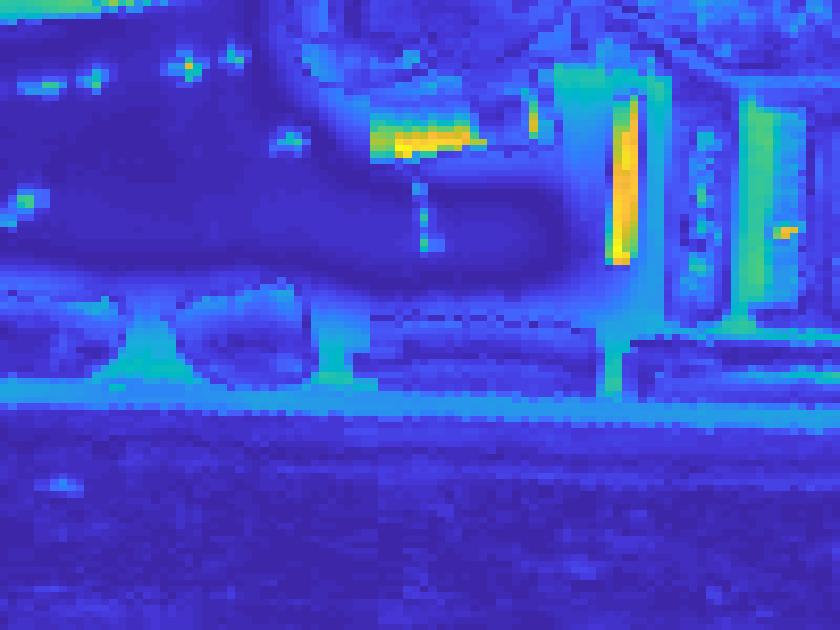} &
\includegraphics[width=0.4631in, height=0.4631in]{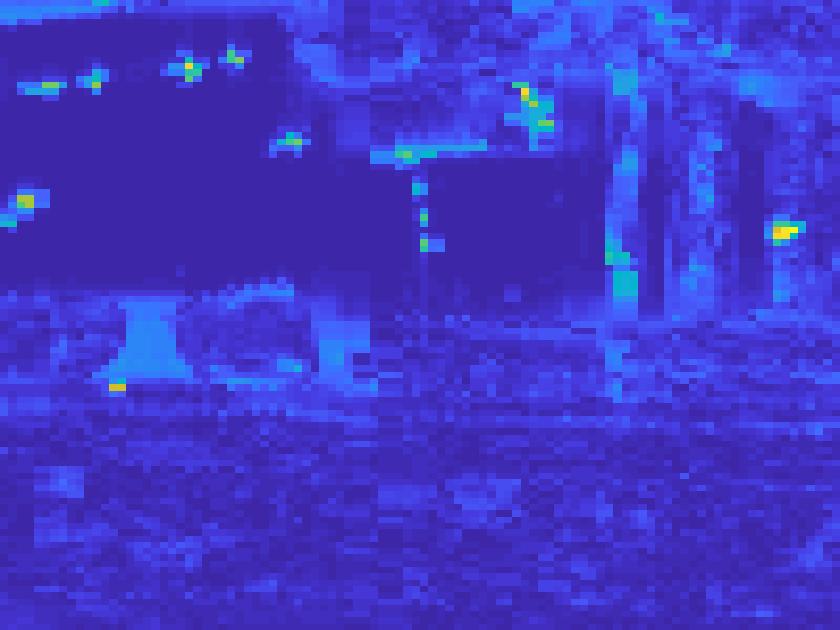} &
\includegraphics[width=0.4631in, height=0.4631in]{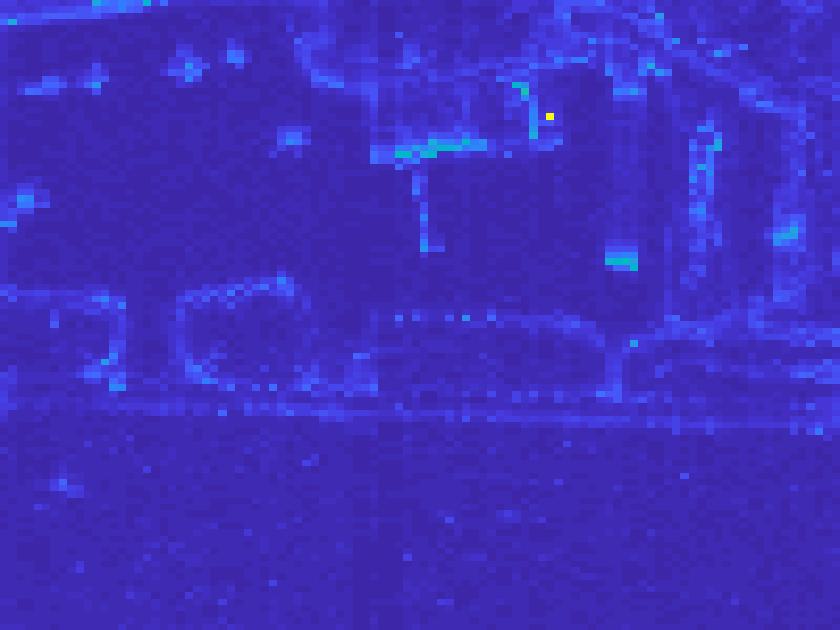}&
\includegraphics[width=0.4631in, height=0.4631in]{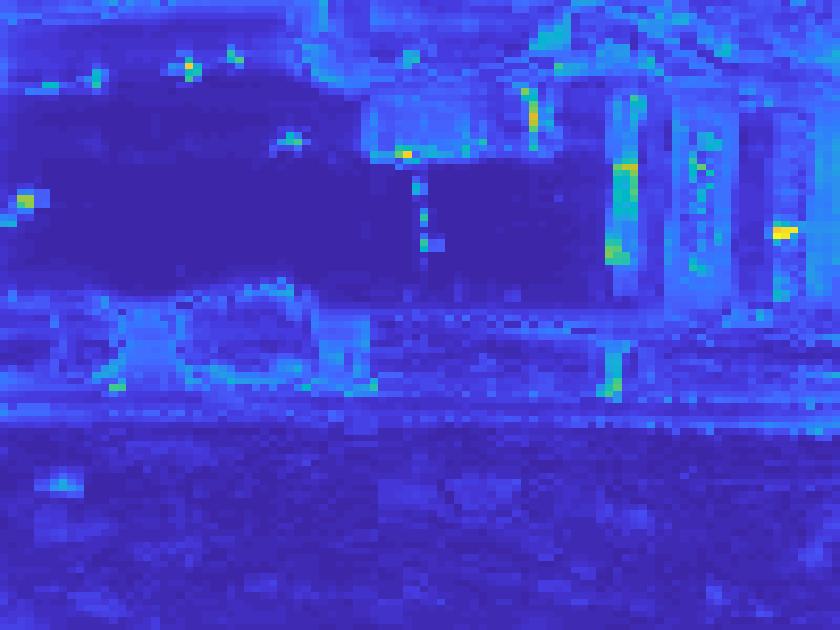}&
\includegraphics[width=0.4631in, height=0.4631in]{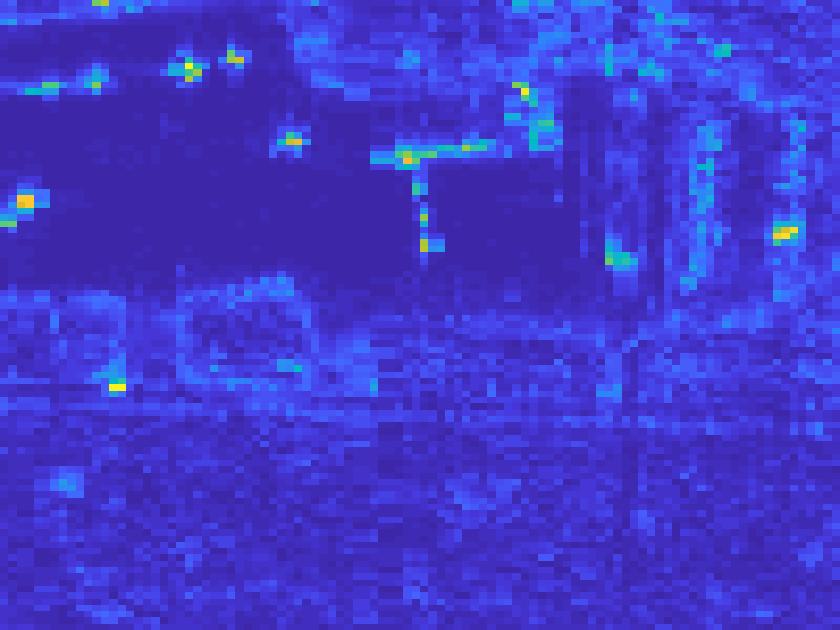} &
\includegraphics[width=0.4631in, height=0.4631in]{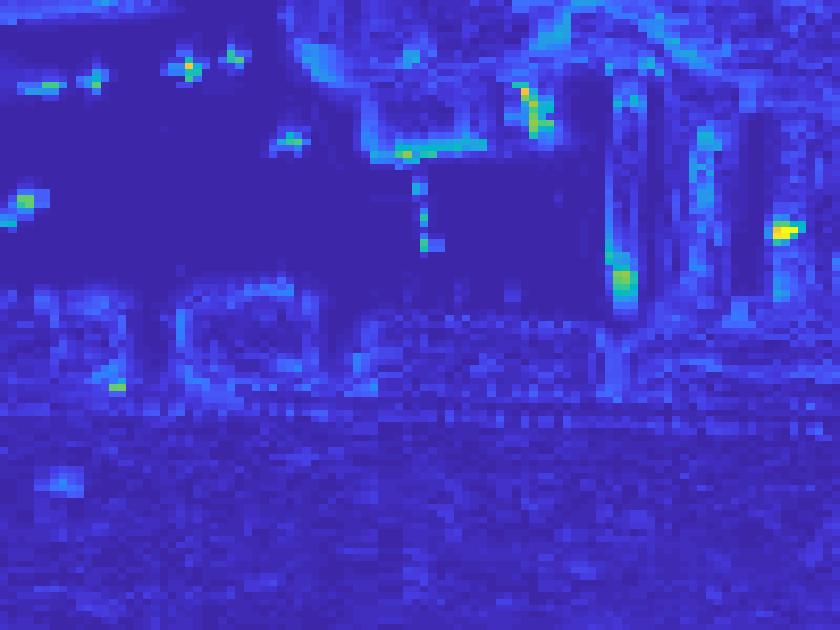}&
\includegraphics[width=0.4631in, height=0.4631in]{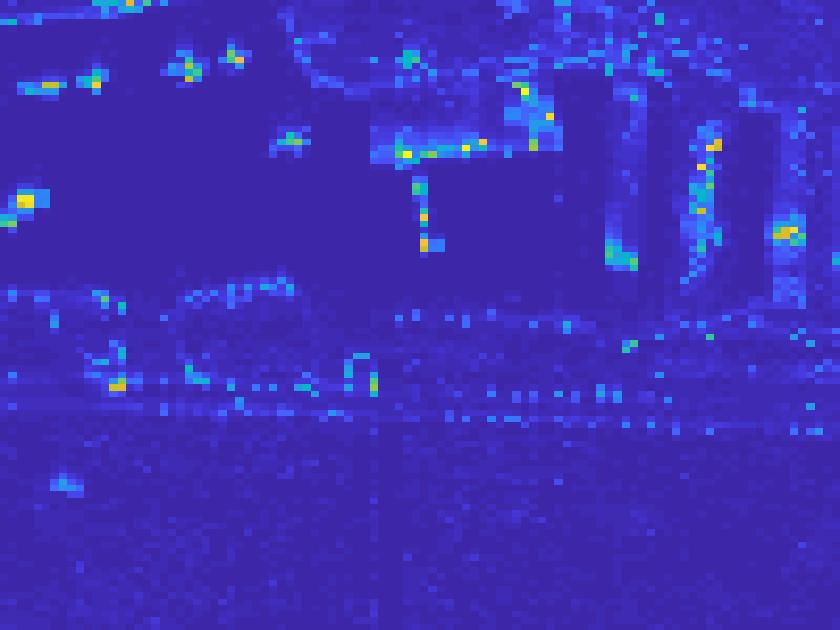}
\\
\includegraphics[width=0.4631in, height=0.4631in]{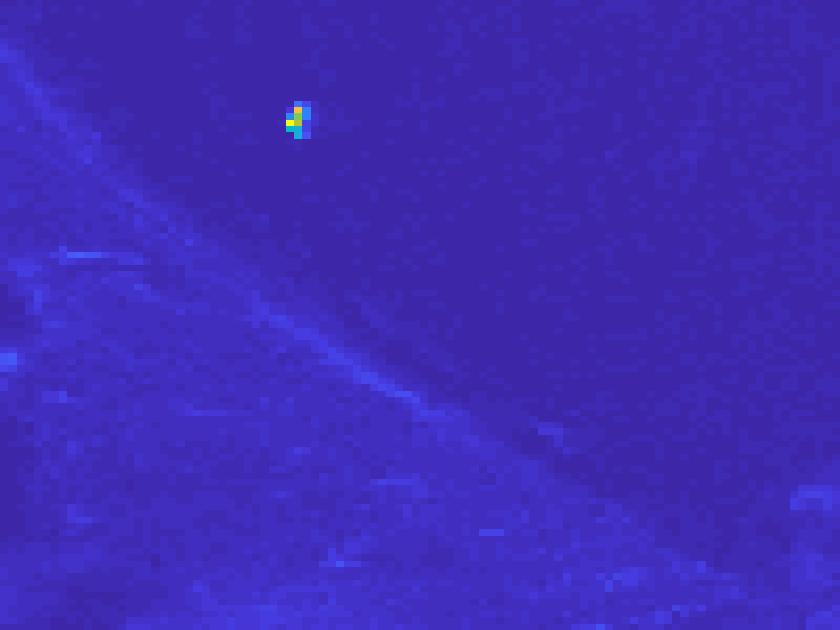}&
\includegraphics[width=0.4631in, height=0.4631in]{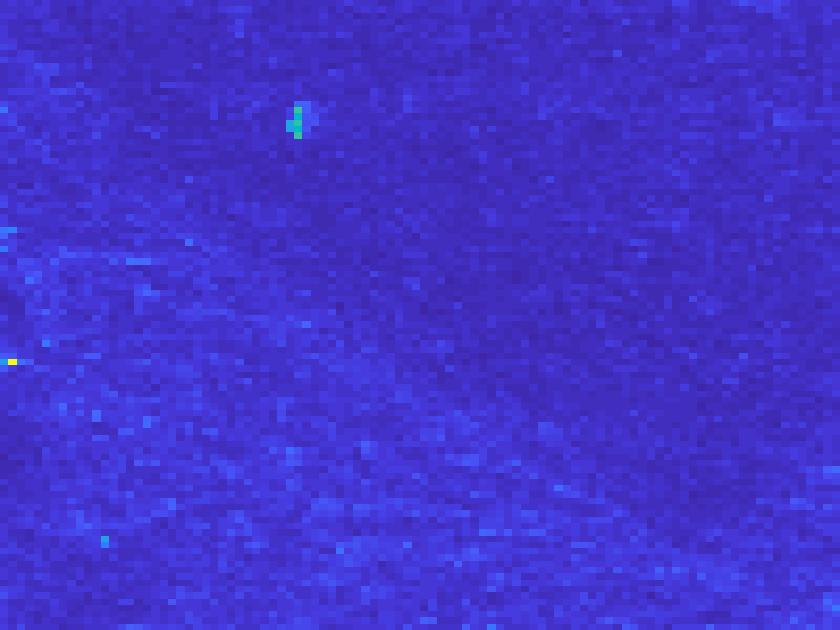}&
\includegraphics[width=0.4631in, height=0.4631in]{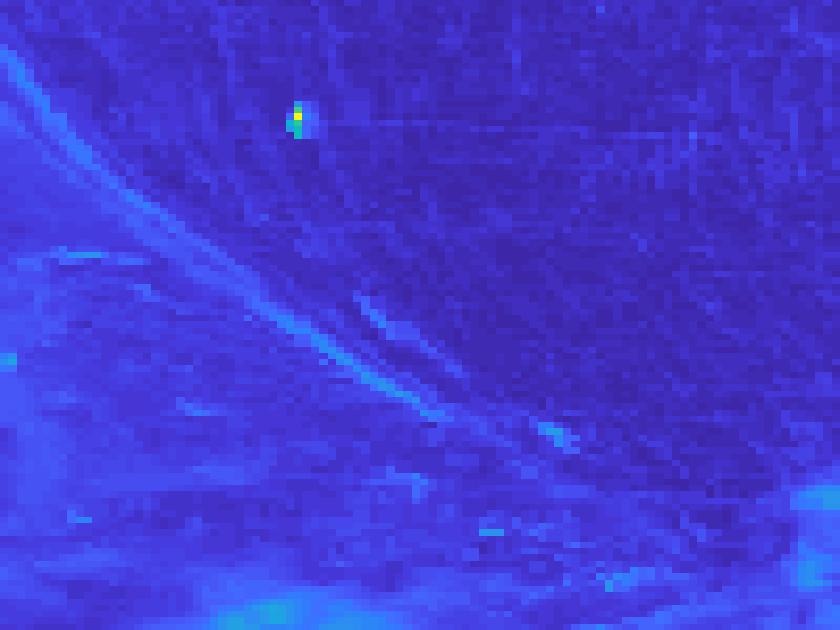}&
\includegraphics[width=0.4631in, height=0.4631in]{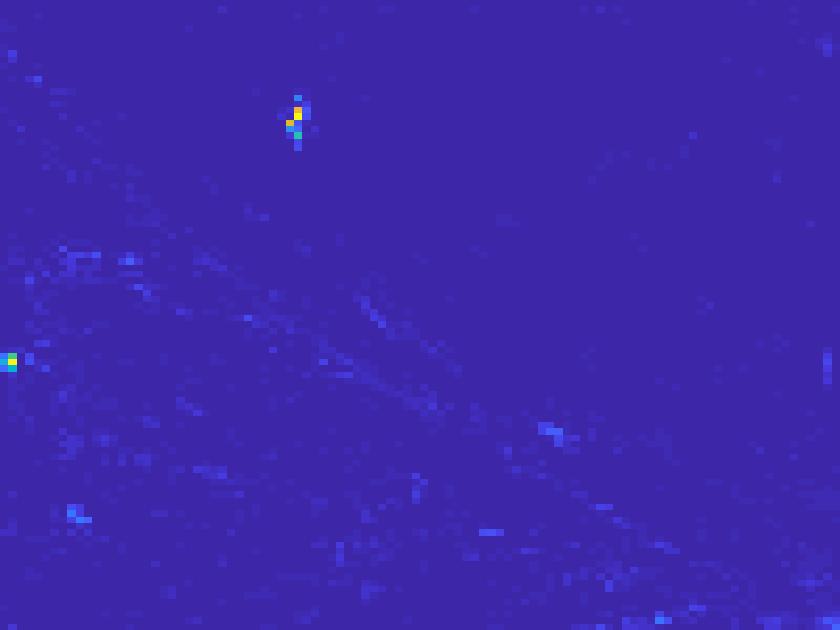}&
\includegraphics[width=0.4631in, height=0.4631in]{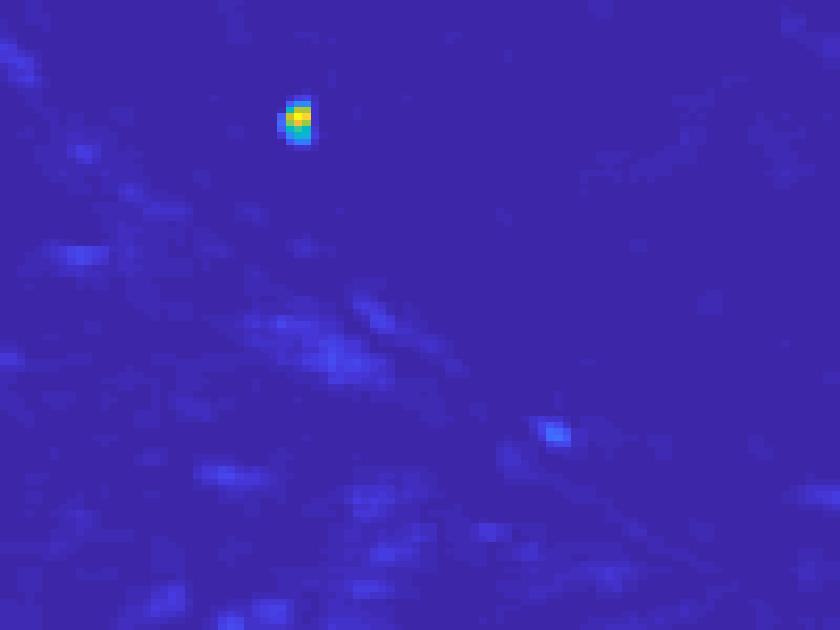}&
\includegraphics[width=0.4631in, height=0.4631in]{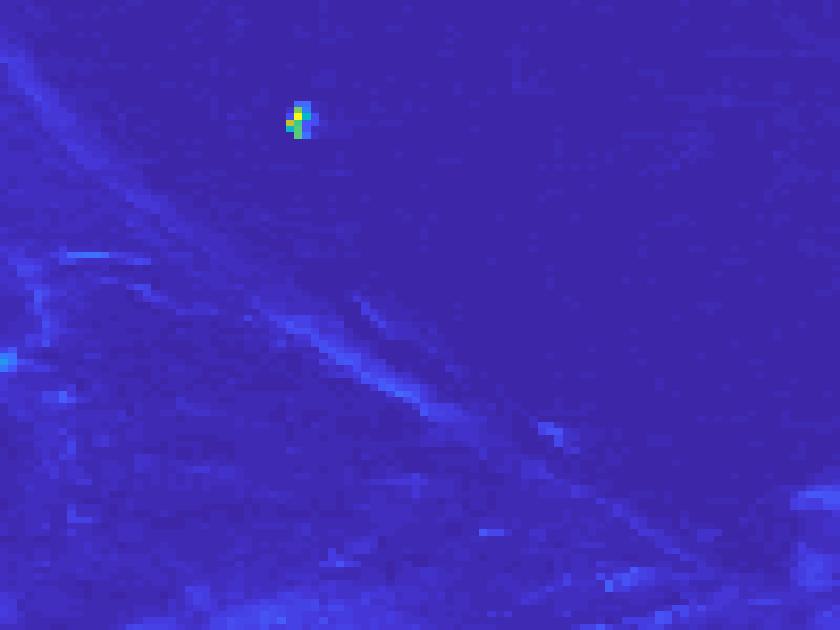}&
\includegraphics[width=0.4631in, height=0.4631in]{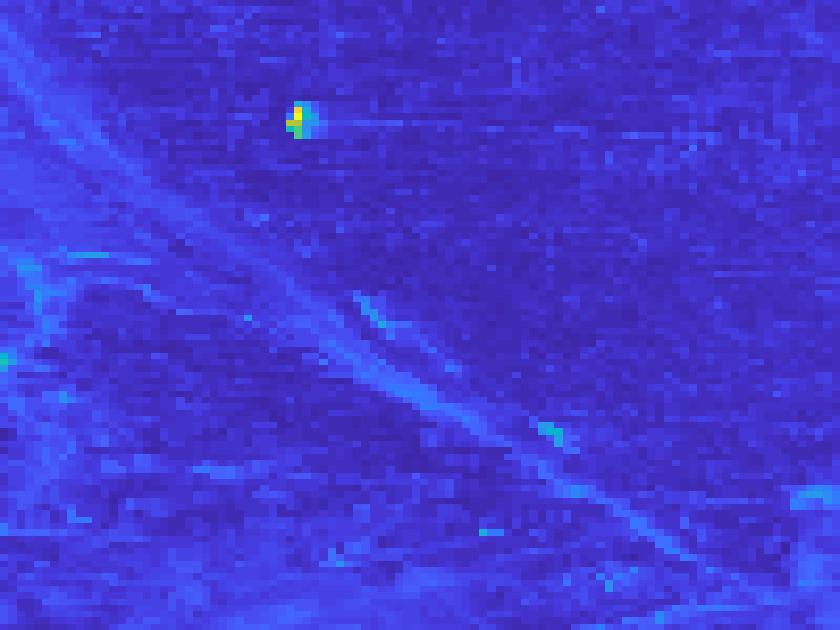}&
\includegraphics[width=0.4631in, height=0.4631in]{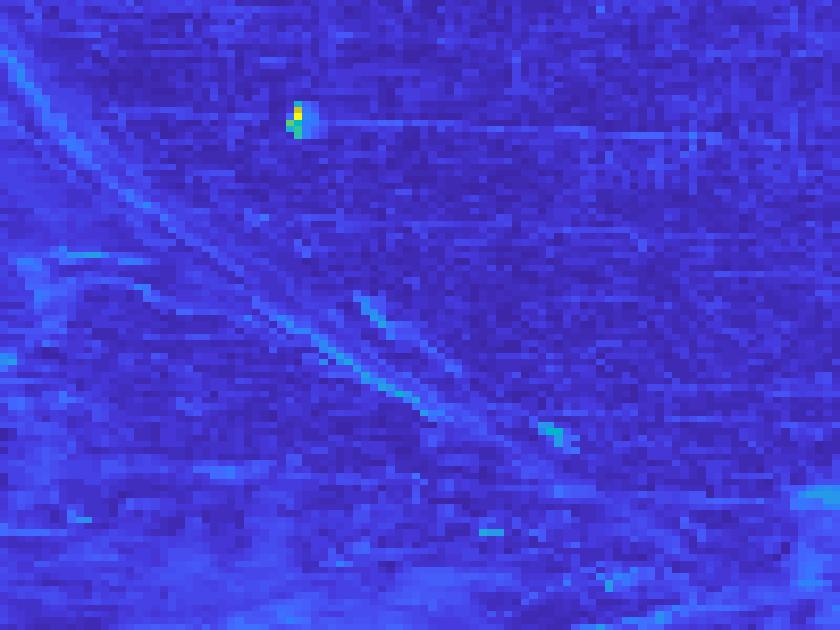}&  %
\includegraphics[width=0.4631in, height=0.4631in]{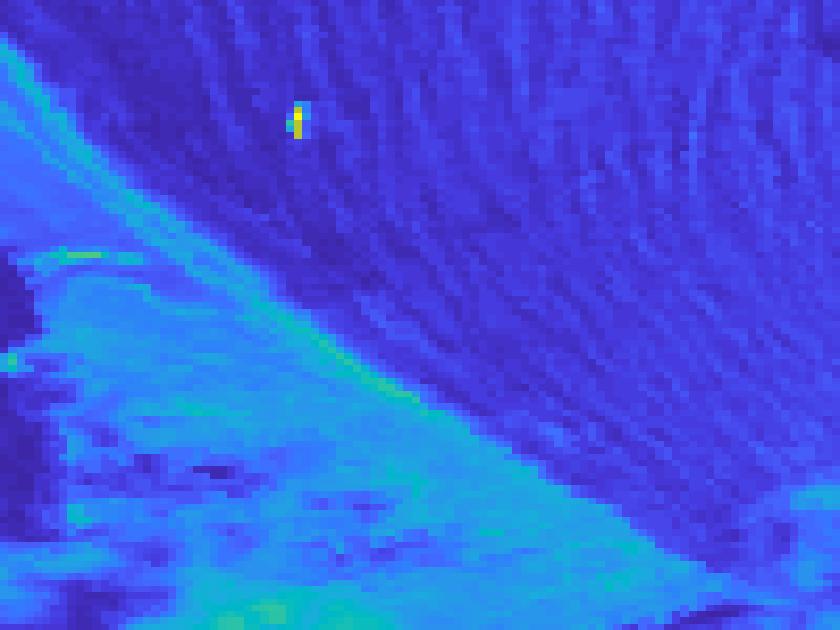} &
\includegraphics[width=0.4631in, height=0.4631in]{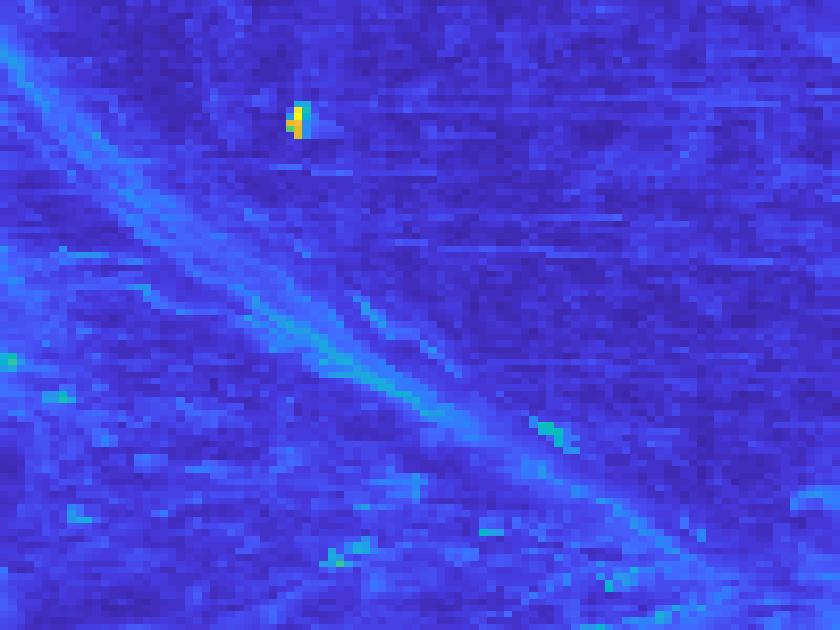} &
\includegraphics[width=0.4631in, height=0.4631in]{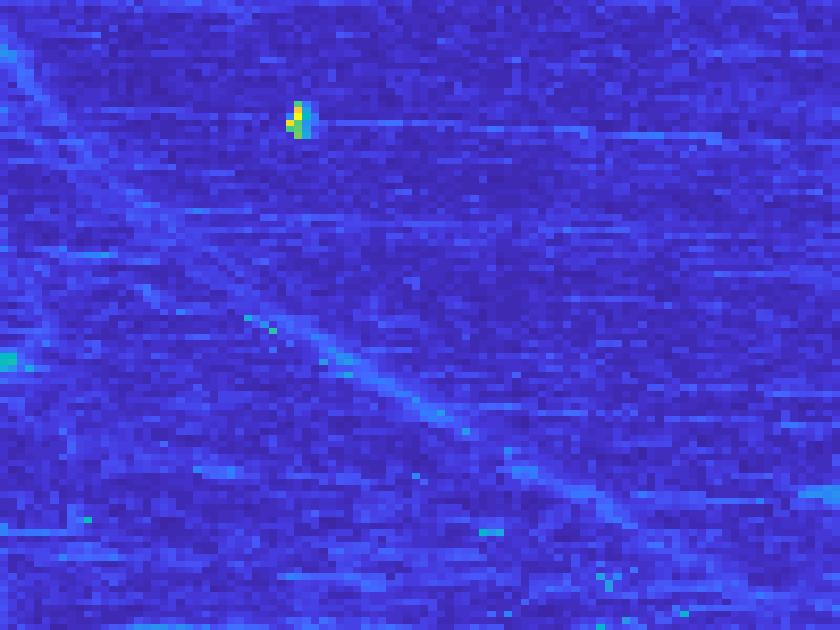}&
\includegraphics[width=0.4631in, height=0.4631in]{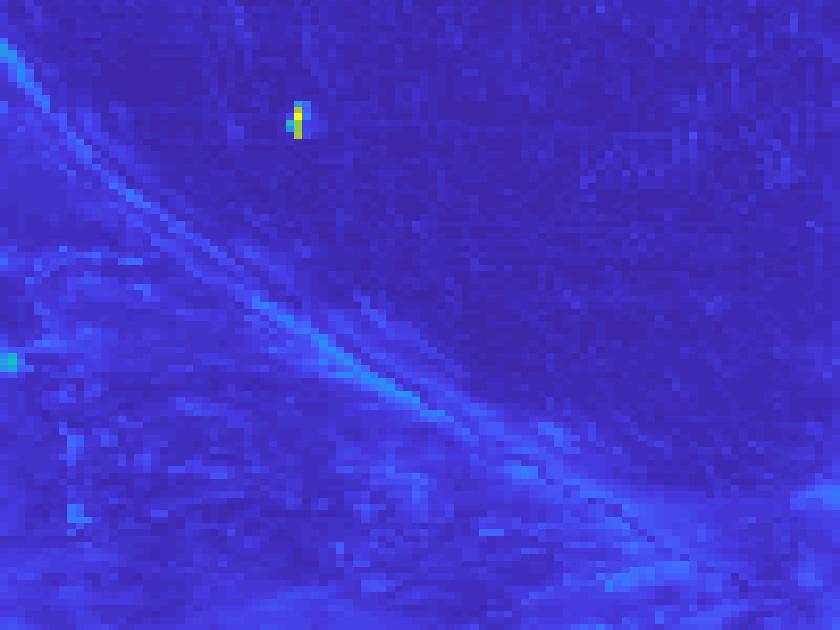}&
\includegraphics[width=0.4631in, height=0.4631in]{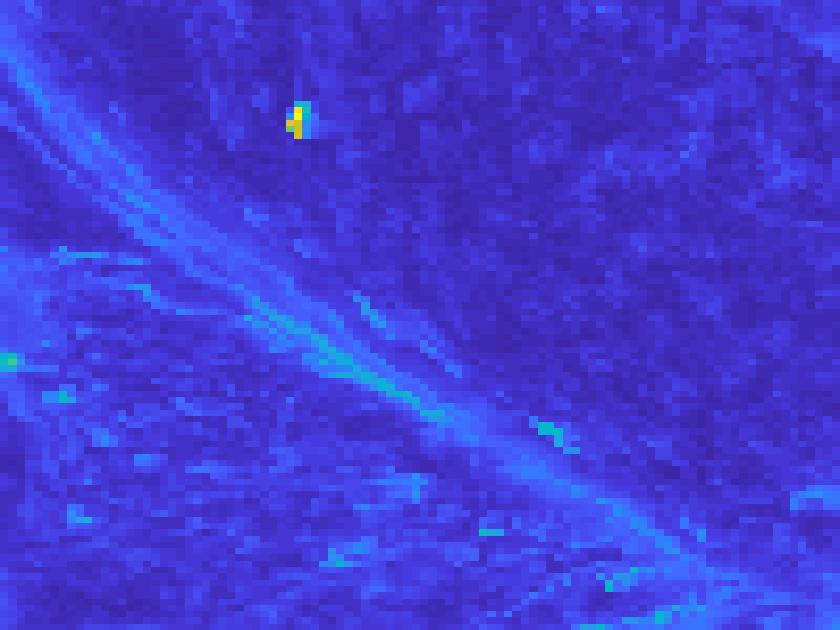} &
\includegraphics[width=0.4631in, height=0.4631in]{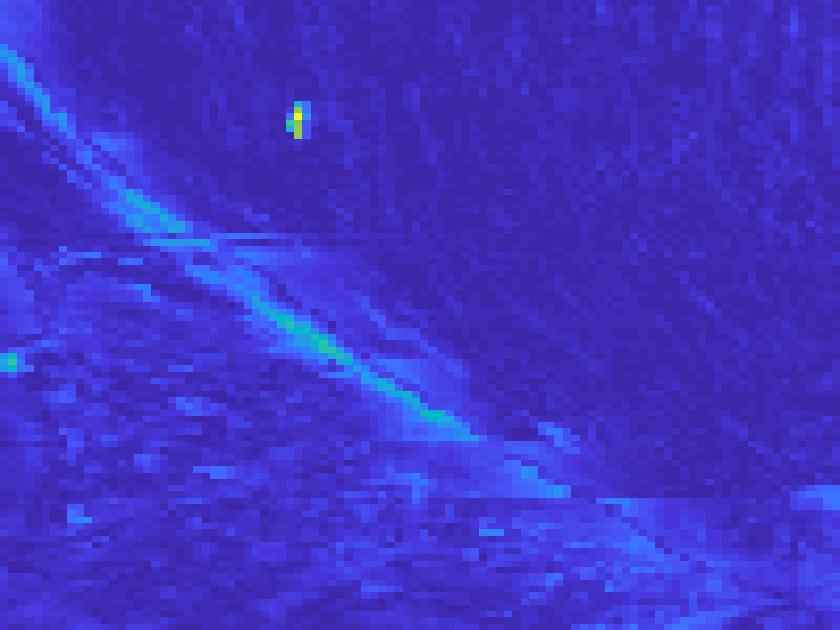}&
\includegraphics[width=0.4631in, height=0.4631in]{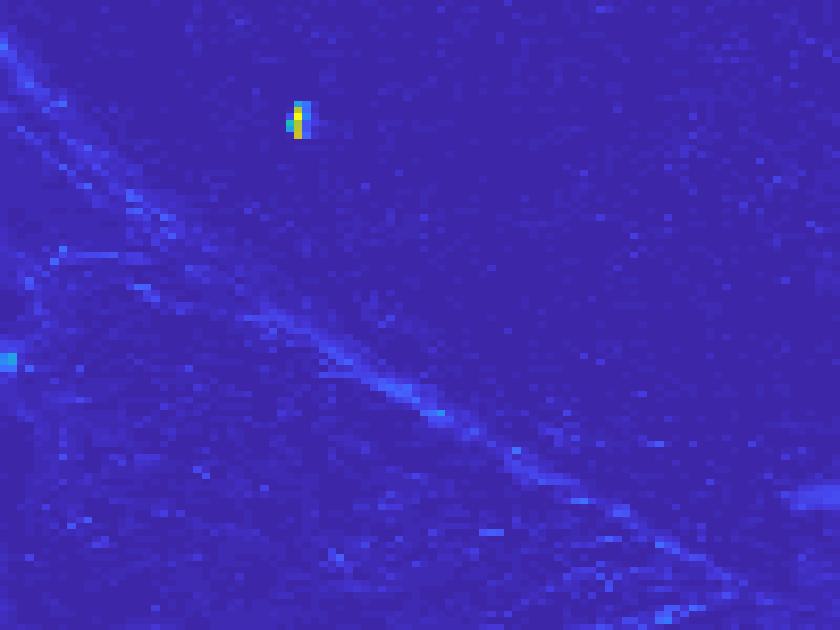}
\\

\includegraphics[width=0.4631in, height=0.4631in]{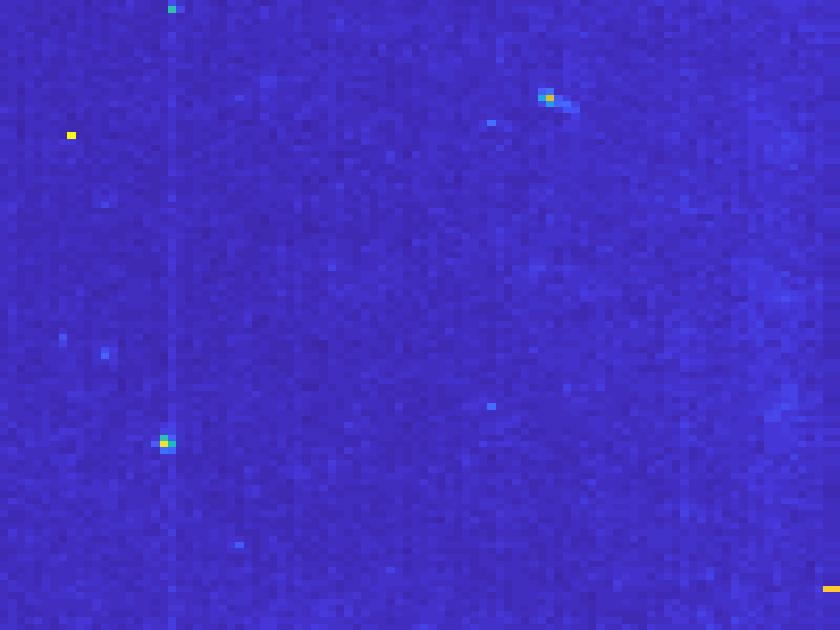}&
\includegraphics[width=0.4631in, height=0.4631in]{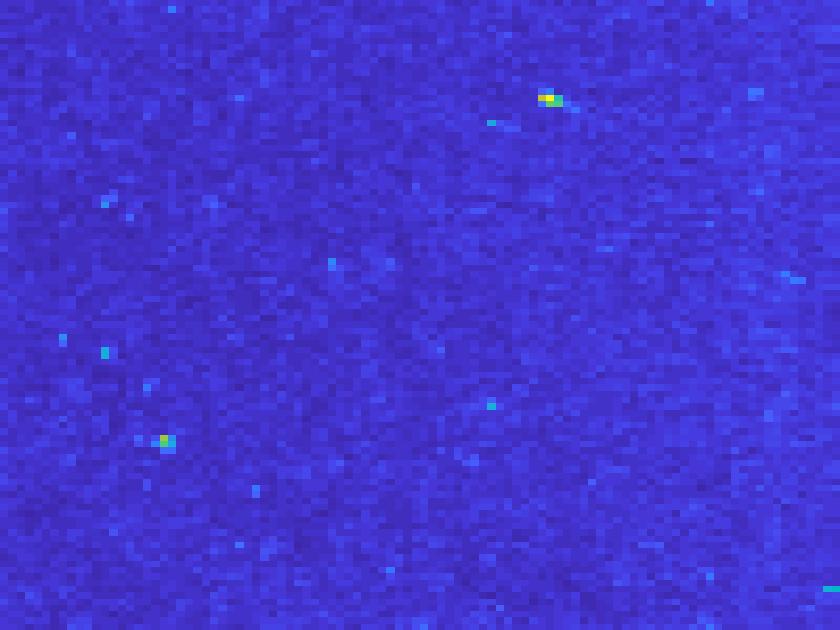}&
\includegraphics[width=0.4631in, height=0.4631in]{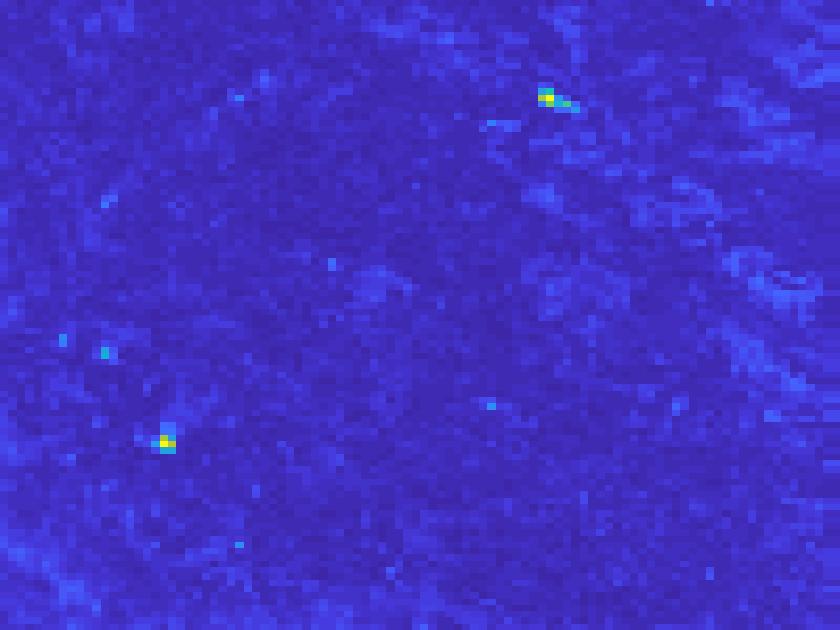}&
\includegraphics[width=0.4631in, height=0.4631in]{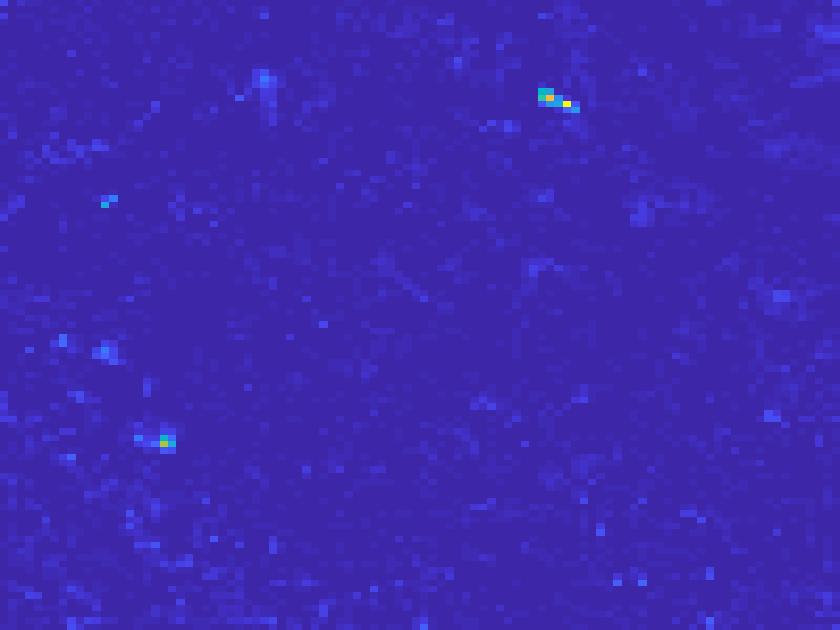}&
\includegraphics[width=0.4631in, height=0.4631in]{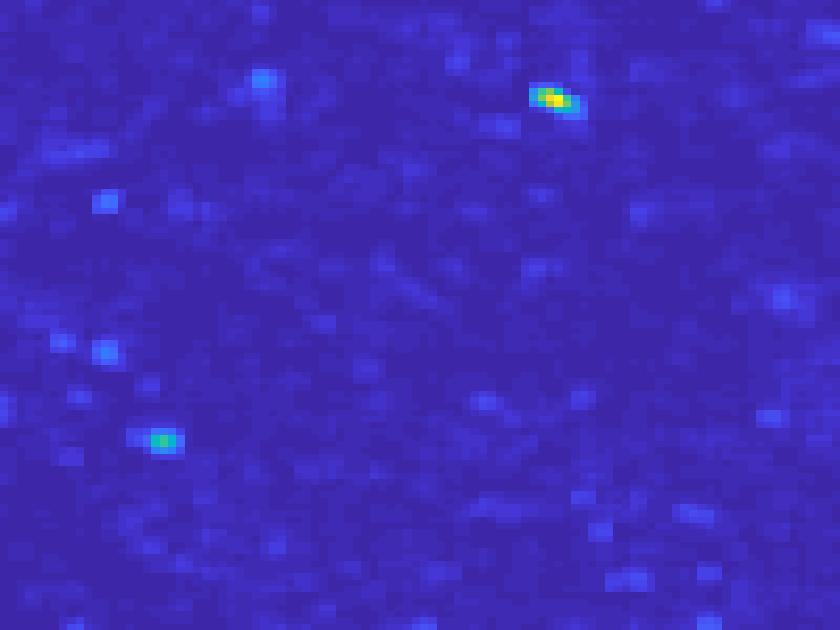}&
\includegraphics[width=0.4631in, height=0.4631in]{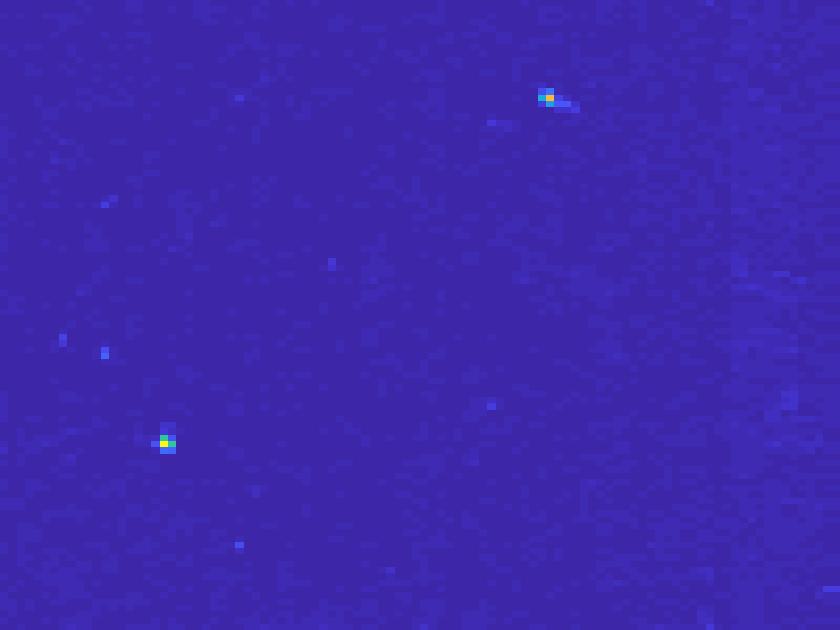}&
\includegraphics[width=0.4631in, height=0.4631in]{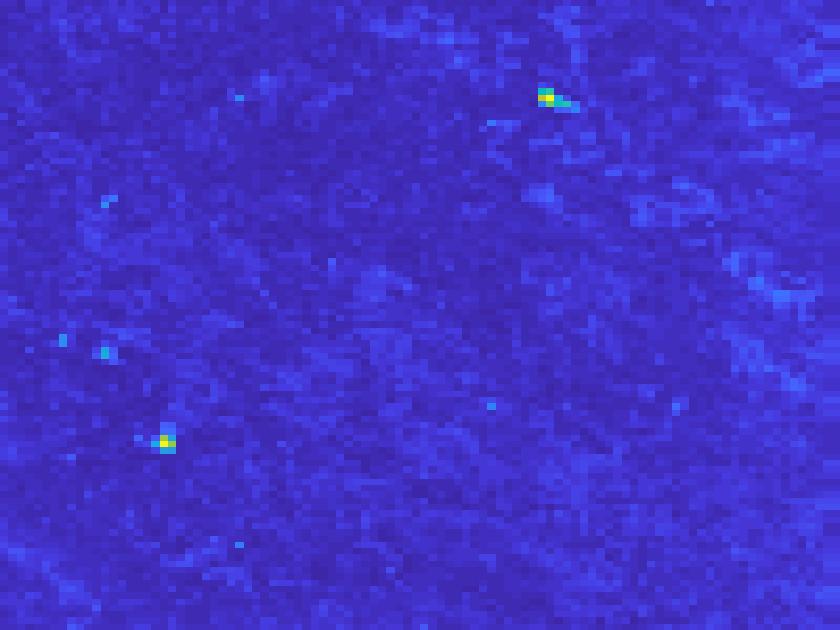}&
\includegraphics[width=0.4631in, height=0.4631in]{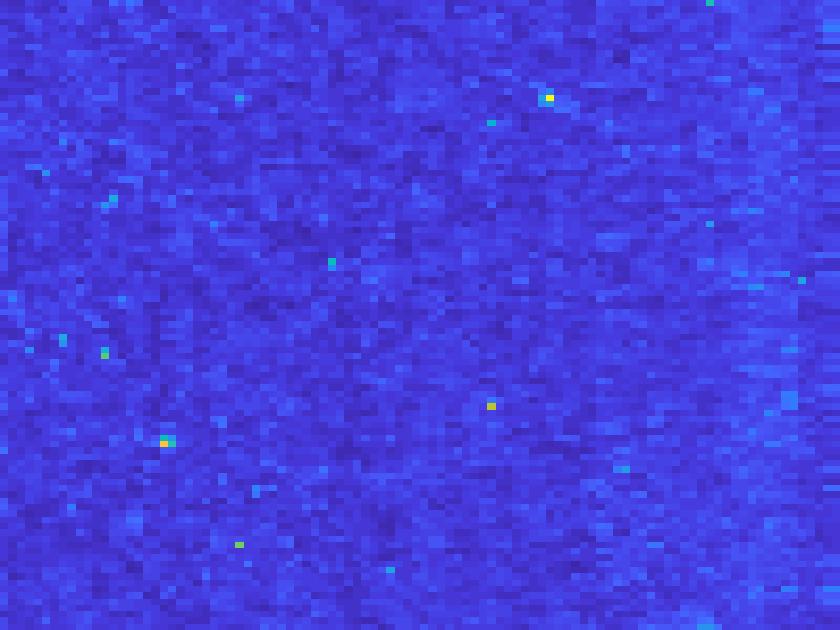}&  %
\includegraphics[width=0.4631in, height=0.4631in]{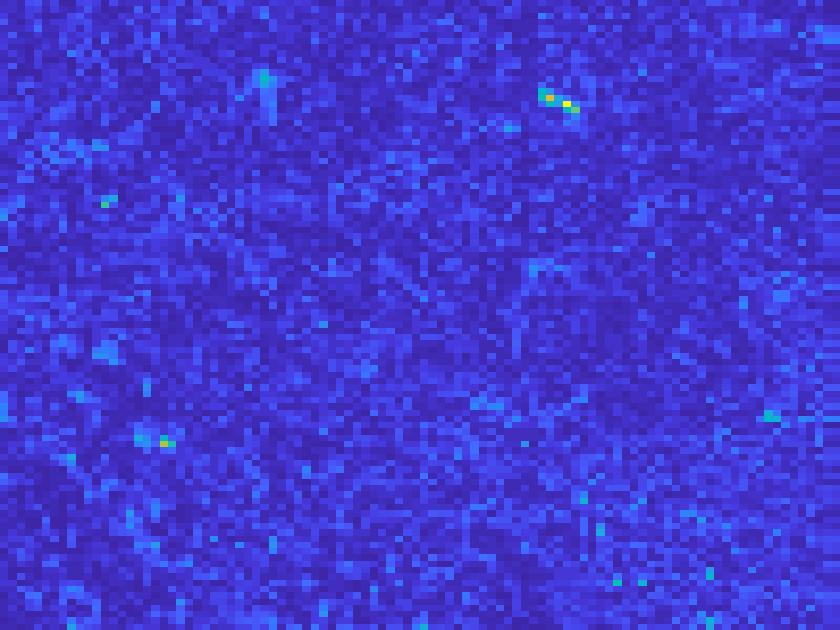} &
\includegraphics[width=0.4631in, height=0.4631in]{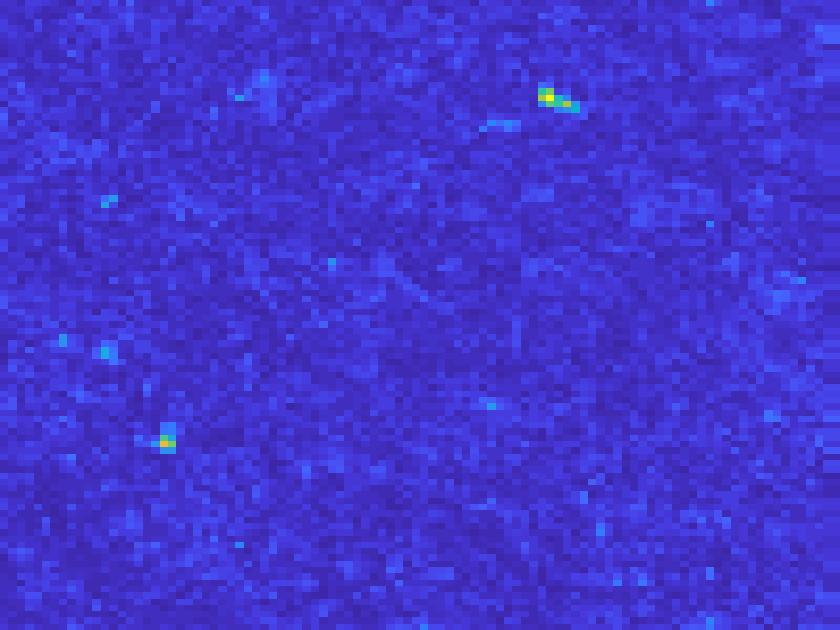} &
\includegraphics[width=0.4631in, height=0.4631in]{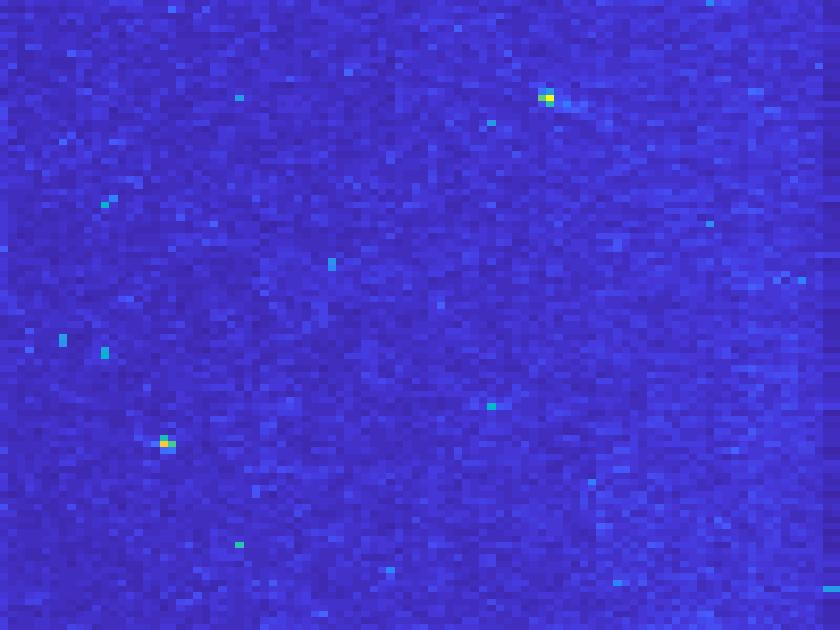}&
\includegraphics[width=0.4631in, height=0.4631in]{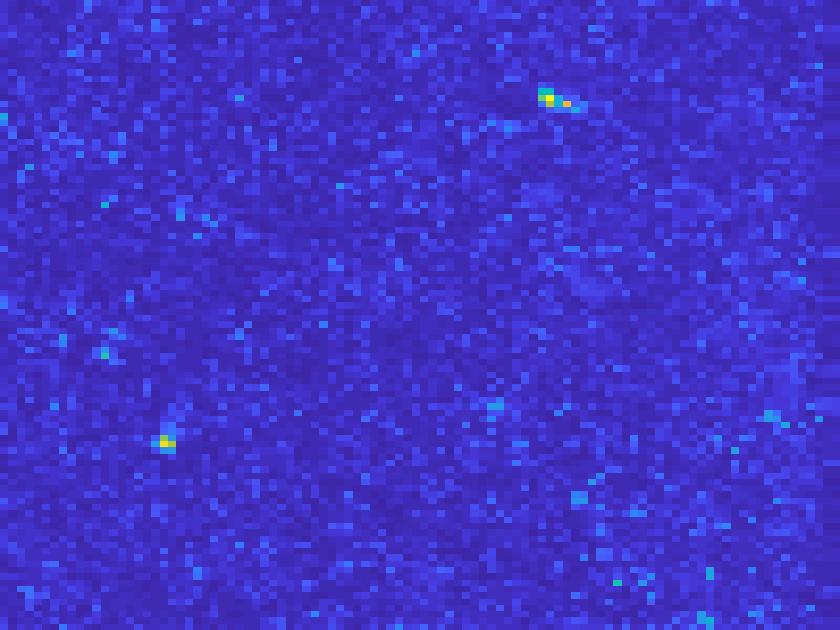}&
\includegraphics[width=0.4631in, height=0.4631in]{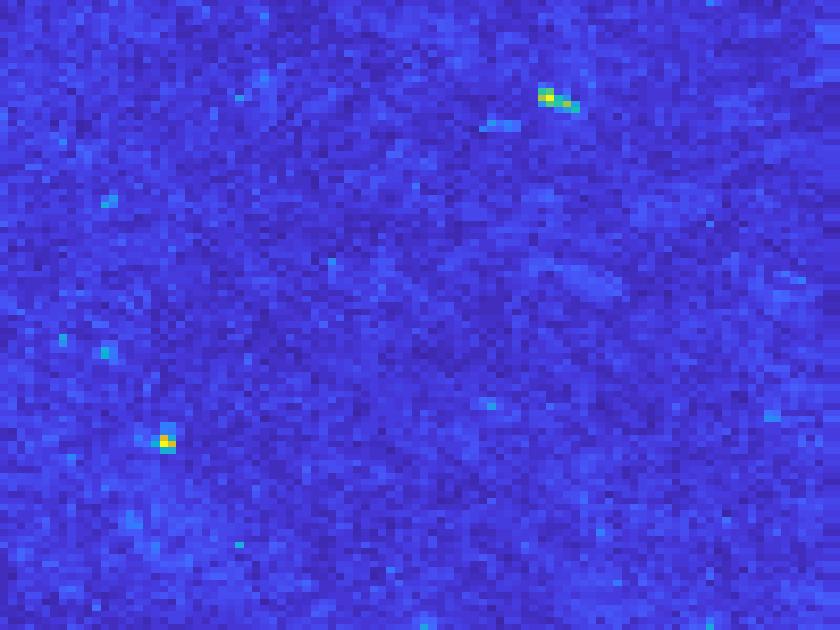} &
\includegraphics[width=0.4631in, height=0.4631in]{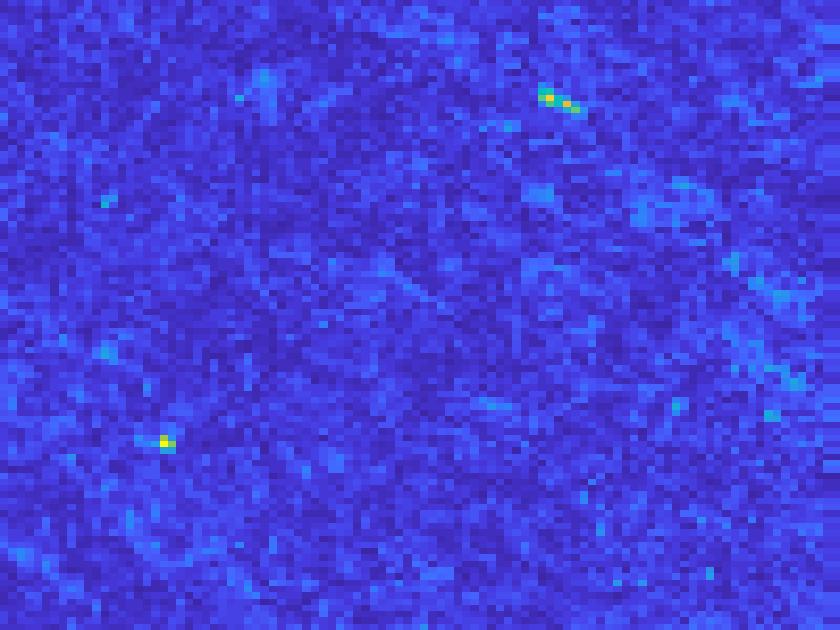}&
\includegraphics[width=0.4631in, height=0.4631in]{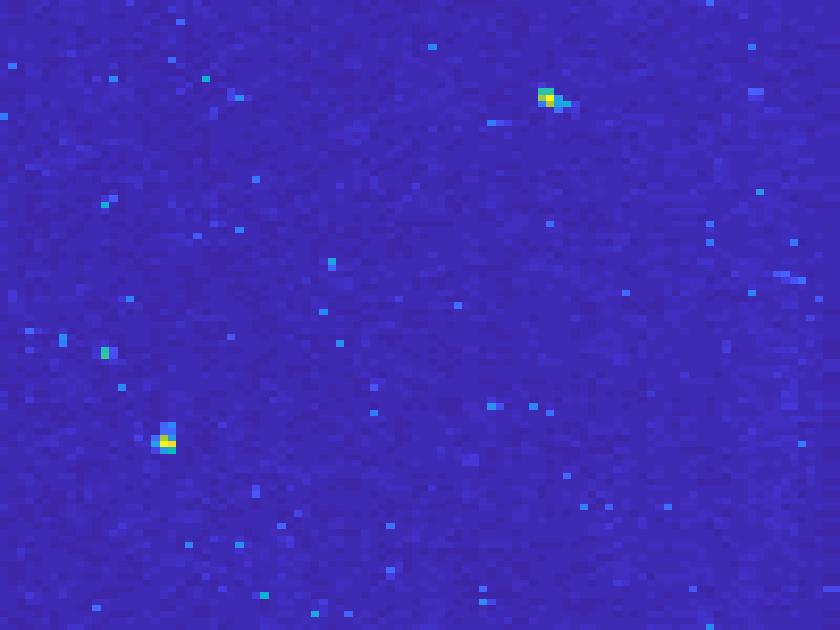}

\\

{{{(a)}}}  &
  {(b)}  & {(c)}
&{(d)} & {(e)}
 &{(f)}& {(g)}&
 {(h)}&
 {(i)}
 &  {(j)}
  &
 {(k)} &(l)  &(m)  &(n) &(o)
\end{tabular}
\caption{
\textcolor[rgb]{0.00,0.00,0.00}{Anomaly detection map of various  HAD  methods on 
five 
 HSI datasets: Salinas,
 Airport-4,   San-Diego,   Urban-3,   Beach-3, Hyperion
 %
(from top to bottom).
%
%
(a)  RX.
(b)  CRD.
(c)  GAED.
(d) PDBSNet.
(e) GT-HAD.
(f)  LSMAD.
(g)  LRASR.
(h)  GTVLRR.
(i)  PTA.
(j)  PCA-TLRSR.
(k)  T-CTV.
(l)  TRDFTVAD.
(m) GNLTR. (n)  GNBRL. 
 (o) Proposed.
}}
\vspace{-0.43cm}
\label{fig_visual_dectionmap}
\end{figure*}

%
\begin{figure*}[!htbp]
\renewcommand{\arraystretch}{0.8}
\setlength\tabcolsep{2.8pt}
\centering
\begin{tabular}{ccc c}
\centering

\footnotesize{{{(a)}}}  &
  \footnotesize{(b)}  & \footnotesize{(c)}
&\footnotesize{(d)}
\\
\includegraphics[width=1.6758in, height=1.51435in]{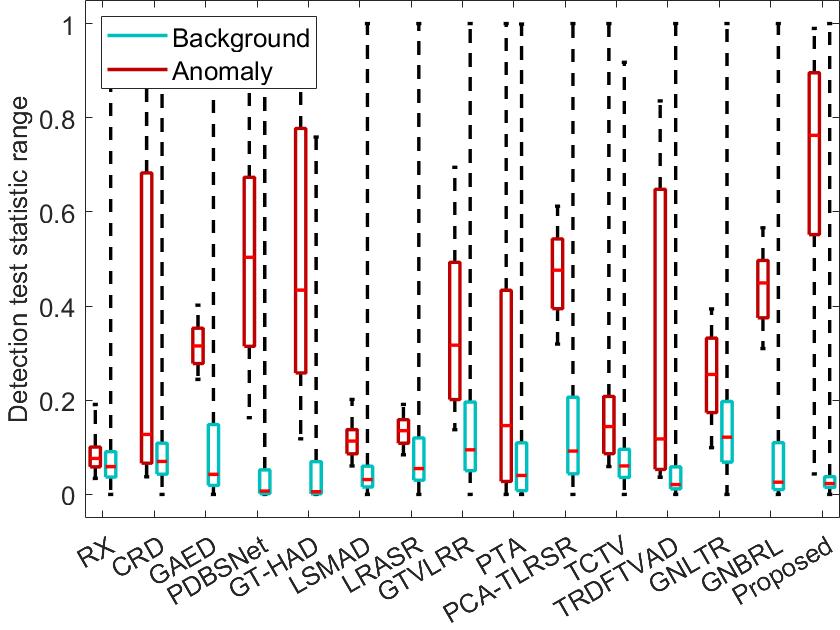}&
\includegraphics[width=1.6758in, height=1.51435in]{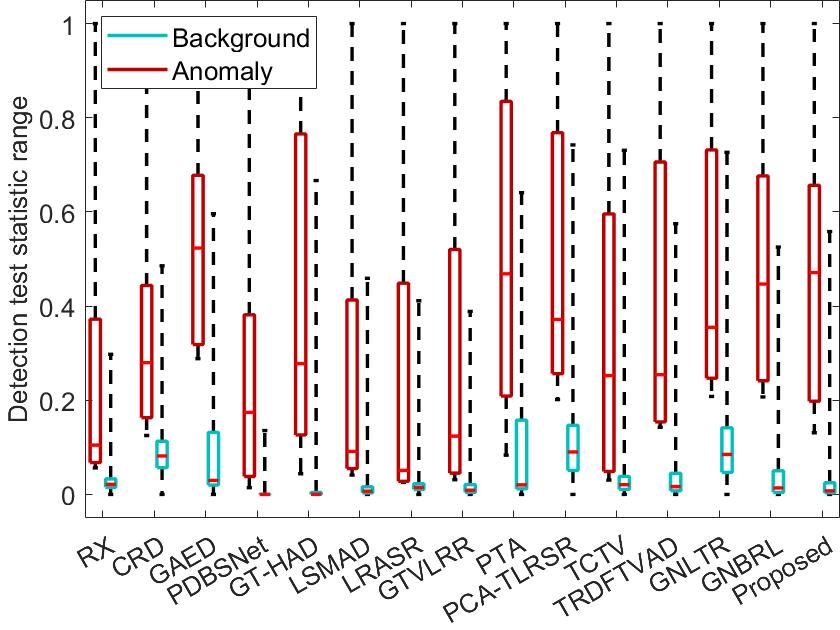}&
\includegraphics[width=1.6758in, height=1.51435in]{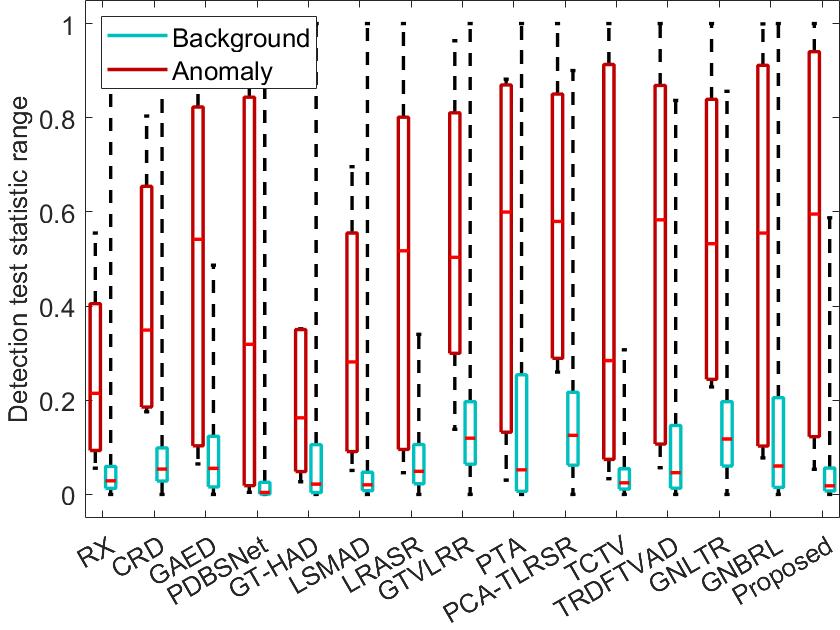}&
\includegraphics[width=1.6758in, height=1.51435in]{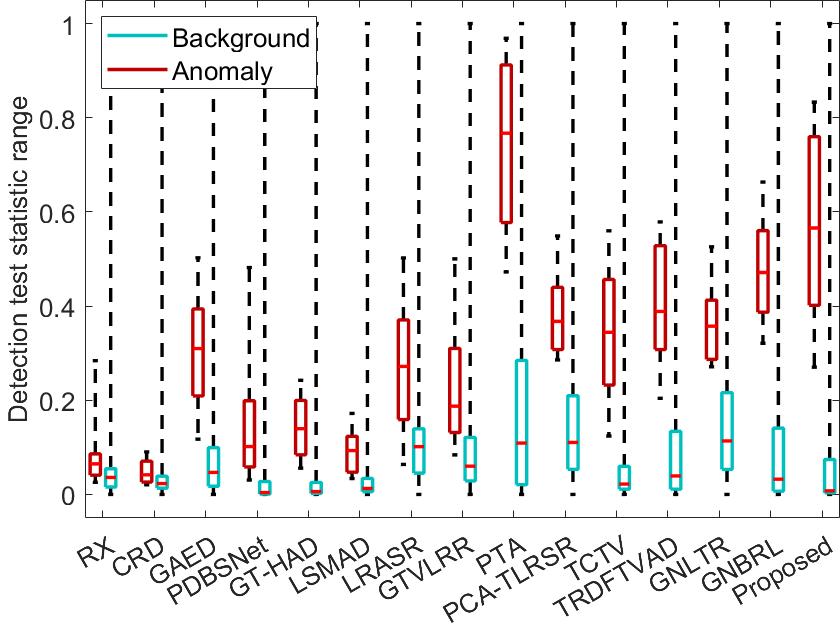}
%
\\

 \footnotesize{(e)}
 &\footnotesize{(f)}& \footnotesize{(g)}&
 \footnotesize{(h)}
\\
\includegraphics[width=1.6758in, height=1.51435in]{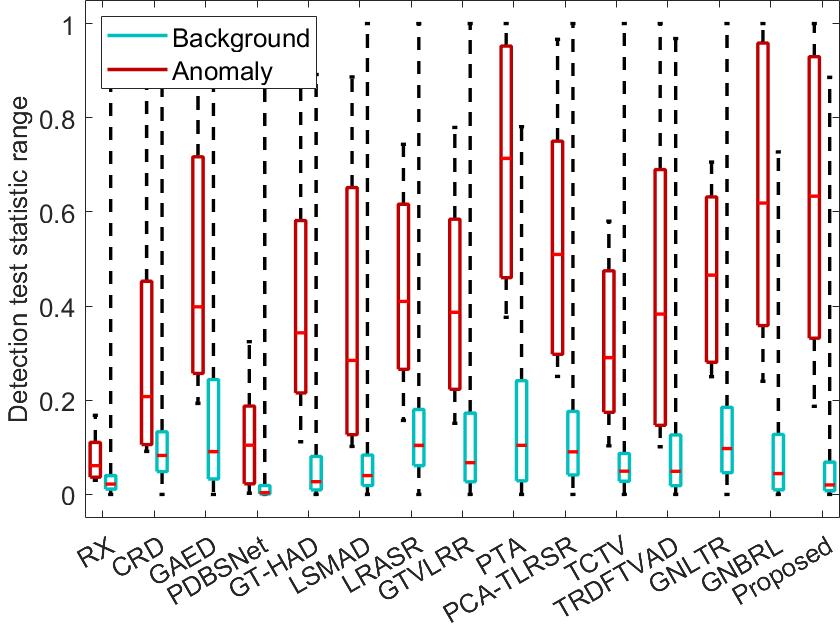}&
\includegraphics[width=1.6758in, height=1.51435in]{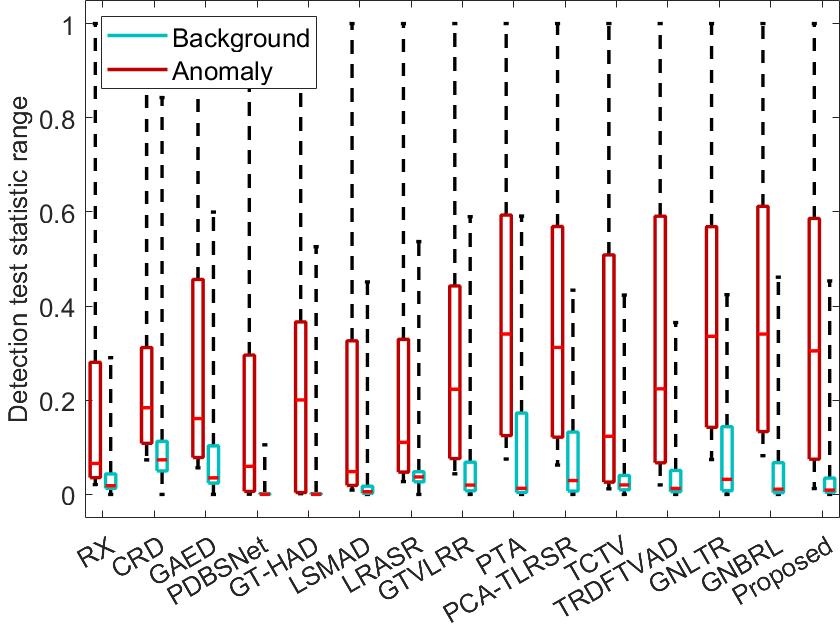}&
\includegraphics[width=1.6758in, height=1.51435in]{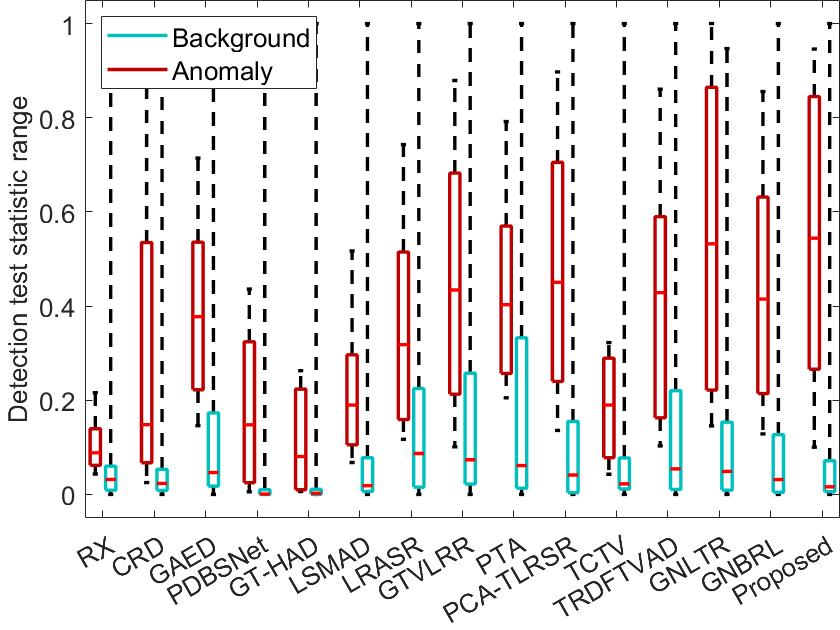}&
\includegraphics[width=1.6758in, height=1.51435in]{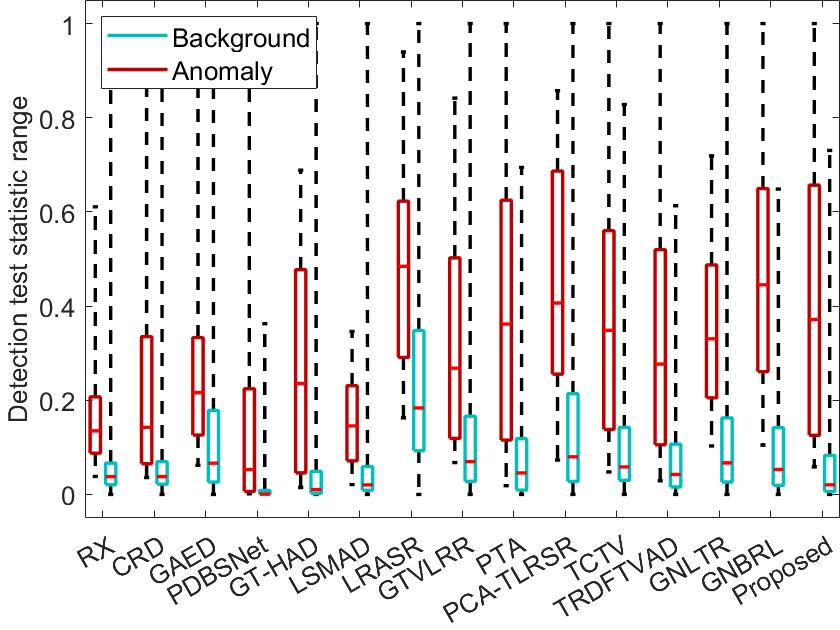}\\

\end{tabular}
\caption{
 \textcolor[rgb]{0.00,0.00,0.00}{Separability maps of various HAD methods for different HSI datasets.
(a)  Salinas.
(b) Pavia.
(c) HYDICE.
(d)  San-Diego.
(e) Airport-4.
(f)  Beach-4.
(g) Urban-3.
(h) Urban-5.
}}
\vspace{-0.44cm}
\label{fig_sep_box}
\end{figure*}

\begin{table*}[tp]
\renewcommand{\arraystretch}{1.15}
\setlength\tabcolsep{2pt}
  \centering
  \caption{\textcolor[rgb]{0.00,0.00,0.00}{The  AUC  values  and average running time output by various HAD  methods on 
  twelve
  HSI  datasets.}
   \textcolor[rgb]{0.00,0.00,0.00}{The best  and the second-best results are highlighted in boldface
and blue, respectively.}
  }
  \label{hsi_various_auc}
  \scriptsize
  \begin{threeparttable}
    \begin{tabular}{cl  c cc ccc ccc ccc  c    cc }
    \hline
   \multicolumn{1}{c}{HSI-Name}&\multicolumn{1}{c}{AUC-Metrics}
   &\multicolumn{1}{c}{RX}&\multicolumn{1}{c}{CRD}&\multicolumn{1}{c}{GAED} 
  & \multicolumn{1}{c}{PDBSNet}   & \tabincell{c} {GT-\\HAD} 
    &\multicolumn{1}{c}{LSMAD}& \multicolumn{1}{c}{LRASR}&  
    \tabincell{c}{GTV-\\LRR}&
    \multicolumn{1}{c}{PTA}&  
    \tabincell{c}{PCA-\\TLRSR}&
      \multicolumn{1}{c}{\text{TCTV}} &
    \tabincell{c} {TRD-\\FTVAD}&
    \multicolumn{1}{c}{\text{GNLTR}}& \multicolumn{1}{c}{\text{GNBRL}}&
    \multicolumn{1}{c}{\text{Proposed}}
    \cr

    \hline
    \hline

   \multirow{6}{*}{Salinas}& $\operatorname{AUC} _{({\operatorname{P}}_{\mathnormal{D}}, {\operatorname{P}}_{\mathnormal{F}})}$
 &    0.7437   & 0.8343&    0.9660 &
   \textcolor[rgb]{0.00,0.00,1.00}{0.9957}  &  0.9936 &
   0.9746  &  0.9227 &   0.9618  &  0.8230  &  0.9916   & 0.9438   & 0.9482 &   0.9369   & \textcolor[rgb]{0.00,0.00,1.00}{0.9957}& \textbf{0.9998}
       \cr
   \qquad	&  $\operatorname{AUC} _{({\operatorname{P}}_{\mathnormal{D}}, \tau)}$
   &0.0805  &  0.2631 &   0.3161 &
      \textcolor[rgb]{0.00,0.00,1.00}{0.5028} &0.4675 &
     0.1134   & 0.1337  &  0.3303  &  0.1943 &   {0.4715}  &  0.1503   & 0.2437  &  0.2554  &  0.4435&\textbf{0.7654}
   \cr
    \qquad	&   $\operatorname{AUC} _{({\operatorname{P}}_{\mathnormal{F}}, \tau)}$
  & 0.0638&0.0754&0.0727&
   \textbf{0.0248}  &\textcolor[rgb]{0.00,0.00,1.00}{0.0252}&
  0.0383&0.0668&0.1136&0.0575&0.1141&0.0653&{0.0344}&0.1295&0.0489& {0.0292}

     \cr
   \qquad	&  $\operatorname{AUC} _{(\operatorname{ODP})}$
   &  0.7604&1.0220&1.2094&
   \textcolor[rgb]{0.00,0.00,1.00}{1.4737} & 1.4359 &
   1.0496&0.9897&1.1785&0.9597&1.3489&1.0288&1.1575&1.0629&{1.3902}&\textbf{1.7360}

      \cr
   \qquad	&  $\operatorname{AUC} _{(\operatorname{SNPR})}$
   &1.2623&3.4909&4.3467&
    \textcolor[rgb]{0.00,0.00,1.00}{20.2521}  & 18.5308 &
   2.9582&2.0028&2.9077&3.3777&4.1321&2.3012&7.0921&1.9727&{9.0624}&\textbf{26.1976}

    \cr
 \qquad	& $\operatorname{AUC} _{(\operatorname{TDBS})}$
 &0.0167&0.1878&0.2434&
 \textcolor[rgb]{0.00,0.00,1.00}{0.4781}&0.4422&
 0.0750&0.0670&0.2167&0.1368&0.3574&0.0850&0.2094&0.1259&{0.3946}&\textbf{0.7362}

   \cr

   \hline
   \multirow{6}{*}{Pavia}& $\operatorname{AUC} _{({\operatorname{P}}_{\mathnormal{D}}, {\operatorname{P}}_{\mathnormal{F}})}$
     & 0.9981&0.9963&\textcolor[rgb]{0.00,0.00,1.00} {0.9993}&
     \textbf{0.9995}  & 0.9986   &
     0.9978&0.9600&0.9907&0.9779&0.9969&0.9867&0.9961&0.9970&{0.9985}&0.9976

       \cr  
   \qquad	&  $\operatorname{AUC} _{({\operatorname{P}}_{\mathnormal{D}}, \tau)}$
  &0.1884&0.3122&\textbf{0.5313}&
  0.2260&  0.3252  &
  0.1859&
  0.1449&0.1931&0.4872&0.4375&0.3019&0.3568&0.4242&0.4614&  \textcolor[rgb]{0.00,0.00,1.00}{0.4901}

    \cr
    \qquad	&   $\operatorname{AUC} _{({\operatorname{P}}_{\mathnormal{F}}, \tau)}$
    &0.0233&0.0844&0.0472&
    \textbf{0.0004}  & \textcolor[rgb]{0.00,0.00,1.00}{0.0026} &
    {0.0091}&0.0163&{0.0123}&0.0475&0.0963&0.0242&0.0247&0.0914&0.0248& 0.0161

     \cr
   \qquad	&  $\operatorname{AUC} _{(\operatorname{ODP})}$
    &1.1632&1.2241&\textbf{1.4834}&
     1.2251 &  1.3212 &
    1.1747&1.0886&1.1715&1.4176&1.3381&1.2644&1.3282&1.3298&1.4351&\textcolor[rgb]{0.00,0.00,1.00}{1.4716}

     \cr
   \qquad	&  $\operatorname{AUC} _{(\operatorname{SNPR})}$
   &8.0852&3.6981 &  11.2480  &
   \textbf{490.1876}   &\textcolor[rgb]{0.00,0.00,1.00}{123.7455} &
   {20.4287}&8.8862  & 15.6551  & 10.2628&4.5422 &  12.4909 &  14.4401&4.6406 &  18.6028&{30.4869}

    \cr
 \qquad	& $\operatorname{AUC} _{(\operatorname{TDBS})}$
  &0.1651&0.2278&\textbf{0.4841}&
   0.2255&0.3226 &
  0.1768&0.1286&0.1808&0.4398&0.3412&0.2778&0.3320&0.3328&0.4366&\textcolor[rgb]{0.00,0.00,1.00}{0.4740}
  \cr


   \hline

   \multirow{6}{*}{Hyperion}& $\operatorname{AUC} _{({\operatorname{P}}_{\mathnormal{D}}, {\operatorname{P}}_{\mathnormal{F}})}$
   &0.9828&0.9555&0.9952&
   0.9879
&0.9860  &
   \textbf{0.9979}&
   0.9927&0.9602&0.9505&\textcolor[rgb]{0.00,0.00,1.00}{0.9975}&0.9239&0.9900&
   \textcolor[rgb]{0.00,0.00,1.00}{0.9975}&0.9888&0.9972
      \cr  
   \qquad	&  $\operatorname{AUC} _{({\operatorname{P}}_{\mathnormal{D}}, \tau)}$
  &0.2183&0.3280&0.3849&
   0.2899 &   0.3947 &
  0.1976&0.3928&0.3287&0.3566&0.4242&0.3099&0.4116&\textbf{0.4566}&\textcolor[rgb]{0.00,0.00,1.00}{0.4262}& 0.4065
   \cr
    \qquad	&   $\operatorname{AUC} _{({\operatorname{P}}_{\mathnormal{F}}, \tau)}$
  &0.0434&0.0621&0.0448&
  \textcolor[rgb]{0.00,0.00,1.00}{0.0134} & 0.0259 &
  \textbf{0.0132}&0.0500&0.0896&0.0708&0.0614&0.0574&0.0531&0.0821&0.0904& {0.0238}

    \cr
   \qquad	&  $\operatorname{AUC} _{(\operatorname{ODP})}$
    &1.1577&1.2214&1.3353&
    1.2644 &  1.3547  &
    1.1823&1.3355&1.1993&1.2363&1.3603&1.1764&1.3484&\textcolor[rgb]{0.00,0.00,1.00}{1.3721}&1.3245&\textbf{1.3799}

     \cr
   \qquad	&  $\operatorname{AUC} _{(\operatorname{SNPR})}$
  &5.0321&5.2810&8.5907  &
  \textbf{21.5669}
  &15.2203  &
   {14.9777}&7.8601&3.6697&5.0364&6.9071&5.3977&7.7457&5.5649&4.7127&\textcolor[rgb]{0.00,0.00,1.00}{17.0651}

    \cr
 \qquad	& $\operatorname{AUC} _{(\operatorname{TDBS})}$
  &0.1749&0.2659&0.3401&
    0.2765&0.3687&
  0.1844&0.3429&0.2391&0.2858&0.3628&0.2525&0.3585&\textcolor[rgb]{0.00,0.00,1.00}{0.3746}&0.3358&\textbf{0.3827}

  \cr

   \hline

   \multirow{6}{*}{HYDICE}& $\operatorname{AUC} _{({\operatorname{P}}_{\mathnormal{D}}, {\operatorname{P}}_{\mathnormal{F}})}$
     &0.9855&\textbf{0.9956}&0.9600&
     0.9578 & 0.9089  &
     0.9925&0.9580&0.9808&0.9236&\textcolor[rgb]{0.00,0.00,1.00}{0.9938}&0.9777&0.9564&0.9931&0.9403&0.9921

      \cr  
   \qquad	&  $\operatorname{AUC} _{({\operatorname{P}}_{\mathnormal{D}}, \tau)}$
   &0.2339&0.4091&0.5020&
   0.3819&  0.1870 &
   0.3032&0.4741&0.5156&0.5375&\textcolor[rgb]{0.00,0.00,1.00}{0.5857}&0.3976&0.5347&0.5606&0.5490&\textbf{0.6026}
   \cr
    \qquad	&   $\operatorname{AUC} _{({\operatorname{P}}_{\mathnormal{F}}, \tau)}$
 &0.0351&0.0618&0.0644&
 \textbf{0.0126} &    0.0447  &
 \textcolor[rgb]{0.00,0.00,1.00}{0.0259}&0.0577&0.1269&0.0983&0.1361&0.0309&0.0673&0.1253&0.0922& {0.0289}

     \cr
   \qquad	&  $\operatorname{AUC} _{(\operatorname{ODP})}$
   &1.1843&1.3428&1.3977&
   1.3271 &1.0513  &
   1.2698&1.3744&1.3696&1.3628&\textcolor[rgb]{0.00,0.00,1.00}{1.4435}&1.3444&1.4238&1.4283&1.3970& \textbf{1.5658}

     \cr
   \qquad	&  $\operatorname{AUC} _{(\operatorname{SNPR})}$
    &6.6667&6.6153&7.7972 &
      \textbf{30.1095} & 4.1815
     &11.6882&8.2154&4.0633&5.4662&4.3047  & {12.8552}&7.9402&4.4720&5.9523& \textcolor[rgb]{0.00,0.00,1.00}{20.8786}
    \cr
 \qquad	& $\operatorname{AUC} _{(\operatorname{TDBS})}$
  &0.1988&0.3472&0.4376&
  0.3692& 0.1423&
  0.2772&0.4164&0.3887&0.4392&0.4497&0.3666&\textcolor[rgb]{0.00,0.00,1.00}{0.4674}&0.4352&0.4568& \textbf{0.5737}
  \cr

   \hline

   \multirow{6}{*}{San-Diego}& $\operatorname{AUC} _{({\operatorname{P}}_{\mathnormal{D}}, {\operatorname{P}}_{\mathnormal{F}})}$
   &0.8884&0.8577&0.9896&
   0.9815&  0.9867 &
   0.9788&0.9670&0.9529&\textcolor[rgb]{0.00,0.00,1.00}{0.9959}&0.9862&\textbf{0.9963}&0.9907&0.9828&{0.9953}&0.9916

      \cr  
   \qquad	&  $\operatorname{AUC} _{({\operatorname{P}}_{\mathnormal{D}}, \tau)}$
  &0.0685&0.0447&0.3081&
   0.1179 &  0.1395&
  0.0914&0.2717&0.2113&\textbf{0.7454}&0.3759&0.3417&0.3989&0.3597&0.4755& \textcolor[rgb]{0.00,0.00,1.00}{0.5642}

   \cr
    \qquad	&   $\operatorname{AUC} _{({\operatorname{P}}_{\mathnormal{F}}, \tau)}$
 &0.0381&{0.0263}&0.0589&
  \textbf{0.0125}&     \textcolor[rgb]{0.00,0.00,1.00}{0.0140} &
 {0.0193}&0.0974&0.0742&0.1341&0.1257&0.0325&0.0612&0.1283&0.0593& 0.0345

    \cr
   \qquad	&  $\operatorname{AUC} _{(\operatorname{ODP})}$
   &0.9188&0.8761&1.2388&
   1.0869 & 1.1121 &
   1.0509&
   1.1412&1.0900&\textbf{1.6072}&1.2363&1.3054&1.3284&1.2142&1.4116&\textcolor[rgb]{0.00,0.00,1.00}{1.5213}

     \cr
   \qquad	&  $\operatorname{AUC} _{(\operatorname{SNPR})}$
   &1.7994&1.7001&5.2289&
     9.4199 & 9.9285 &
   4.7259&2.7879&2.8491&5.5591&2.9901  & \textcolor[rgb]{0.00,0.00,1.00}{10.5072}&6.5192&2.8036&8.0227&\textbf{16.3419}

    \cr
 \qquad	& $\operatorname{AUC} _{(\operatorname{TDBS})}$
 &0.0304&0.0184&0.2492&
  0.1053&0.1254&
 0.0720&0.1742&0.1371&\textbf{0.6113}&0.2502&0.3092&0.3377&0.2314&0.4162&\textcolor[rgb]{0.00,0.00,1.00}{0.5297}

   \cr

   \hline

   \multirow{6}{*}{Airport-4}& $\operatorname{AUC} _{({\operatorname{P}}_{\mathnormal{D}}, {\operatorname{P}}_{\mathnormal{F}})}$
    &0.9525&0.9318&0.9660&
    0.9644  &0.9951 &
    0.9888&0.9889&0.9838&\textcolor[rgb]{0.00,0.00,1.00}{0.9972}&0.9931&0.9928&0.9787&0.9925&0.9964&\textbf{0.9982}

      \cr  
   \qquad	&  $\operatorname{AUC} _{({\operatorname{P}}_{\mathnormal{D}}, \tau)}$
  &0.0727&0.2449&0.4579&
  0.1118  &0.3765 &
  0.3545&0.4260&0.4059&\textbf{0.7045}&0.5411&0.3049&0.3950&0.4640&0.6266&\textcolor[rgb]{0.00,0.00,1.00}{0.6314}

   \cr
    \qquad	&   $\operatorname{AUC} _{({\operatorname{P}}_{\mathnormal{F}}, \tau)}$
 &\textcolor[rgb]{0.00,0.00,1.00}{0.0247}&0.0899&0.1197&
 \textbf{0.0091}  &0.0391 &
 0.0495&0.1161&0.0885&0.1244&0.1048&0.0569&0.0658&0.1105&0.0621& {0.0331}

    \cr
   \qquad	&  $\operatorname{AUC} _{(\operatorname{ODP})}$
   &1.0005&1.0868&1.3043&
   1.0672  &1.3325&
   1.2938&1.2988&1.3012&\textcolor[rgb]{0.00,0.00,1.00}{1.5773}&1.4294&1.2408&1.3079&1.3460&1.5609&\textbf{1.5965}

     \cr
   \qquad	&  $\operatorname{AUC} _{(\operatorname{SNPR})}$
  &2.9409&2.7240&3.8270&
  \textcolor[rgb]{0.00,0.00,1.00}{12.3651}  & 9.6271  &
  7.1568&3.6697&4.5878&5.6614&5.1652&5.3619&6.0057&4.1997  & {10.0830}& \textbf{19.0670}

    \cr
 \qquad	& $\operatorname{AUC} _{(\operatorname{TDBS})}$
  &0.0480&0.1550&0.3383&
  0.1027&  0.3374 &
  0.3050&0.3099&0.3175&\textcolor[rgb]{0.00,0.00,1.00}{0.5801}&0.4363&0.2480&0.3293&0.3535&{0.5645}&\textbf{0.5983}
 \cr

   \hline

   \multirow{6}{*}{Beach-3}& $\operatorname{AUC} _{({\operatorname{P}}_{\mathnormal{D}}, {\operatorname{P}}_{\mathnormal{F}})}$
       &\textbf{0.9997}&0.9888&0.9863&
       0.9919  &\textcolor[rgb]{0.00,0.00,1.00}{0.9996} &
       {0.9992}&0.9989&0.9944&0.9109&0.9984&0.9985&0.9910&0.9985&0.9928&0.9991
       \cr  
   \qquad	&  $\operatorname{AUC} _{({\operatorname{P}}_{\mathnormal{D}}, \tau)}$

&0.4937&0.3590&0.4357&
 0.3706  &  0.6138 &
0.5011&0.5811&0.4945&0.5397&\textbf{0.6670}&0.6158&0.4962&\textcolor[rgb]{0.00,0.00,1.00}{0.6597}&0.5297&0.5271
    \cr
    \qquad	&   $\operatorname{AUC} _{({\operatorname{P}}_{\mathnormal{F}}, \tau)}$

&0.0259&0.0559&0.0733&
 \textbf{0.0077}  &\textcolor[rgb]{0.00,0.00,1.00}{0.0138} &
{0.0237}&0.0702&0.0739&0.1656&0.0947&0.0695&0.0540&0.0804&0.0535&{0.0197}

    \cr
   \qquad	&  $\operatorname{AUC} _{(\operatorname{ODP})}$
   &1.4675&1.2919&1.3486&
   1.3548  & \textbf{1.5996} &
   1.4766&{1.5098}&1.4150&1.2850&{1.5707}&1.5449&1.4332&\textcolor[rgb]{0.00,0.00,1.00}{1.5778}&1.4691
   & {1.5065}

     \cr
   \qquad	&  $\operatorname{AUC} _{(\operatorname{SNPR})}$

& 19.0477&6.4267&5.9408 &
 \textbf{47.9560} & \textcolor[rgb]{0.00,0.00,1.00}{44.2937} 
 & {21.1441}&8.2831&6.6931&3.2596&7.0459&8.8660&9.1903&8.2065&9.9065&{26.8211}

    \cr
 \qquad	& $\operatorname{AUC} _{(\operatorname{TDBS})}$
  &0.4678&0.3032&0.3624&
    0.3628&\textbf{0.5999}&
  0.4774&0.5110&0.4206&0.3741&{0.5723}&0.5464&0.4422&\textcolor[rgb]{0.00,0.00,1.00}{0.5793}&0.4762&0.5074

  \cr

   \hline

   \multirow{6}{*}{Beach-4}& $\operatorname{AUC} _{({\operatorname{P}}_{\mathnormal{D}}, {\operatorname{P}}_{\mathnormal{F}})}$

 &0.9538&0.9571&0.9362&
 0.9870 &    \textcolor[rgb]{0.00,0.00,1.00}{0.9906}  &
 0.9702&0.9341&0.9757&0.9515&0.9643&0.9291&0.9799&0.9696&\textbf{0.9912}& {0.9863}
      \cr  
   \qquad	&  $\operatorname{AUC} _{({\operatorname{P}}_{\mathnormal{D}}, \tau)}$

&0.1284&0.2121&0.2235&
0.1270 & 0.2169  &
0.1133&0.1728&0.2542&\textbf{0.3732}&0.3328&0.2059&0.2791&0.3511&\textcolor[rgb]{0.00,0.00,1.00}{0.3706}& 0.3290
   \cr
    \qquad	&   $\operatorname{AUC} _{({\operatorname{P}}_{\mathnormal{F}}, \tau)}$

&0.0233&0.0787&0.0540&
\textbf{0.0011} &   \textcolor[rgb]{0.00,0.00,1.00}{0.0013} &
{0.0087}&0.0380&0.0311&0.0586&0.0540& 0.0236&0.0226&0.0580&0.0260& {0.0180}

    \cr
   \qquad	&  $\operatorname{AUC} _{(\operatorname{ODP})}$

&1.0589&1.0905&1.1058&
1.1131 & 1.2063  &
1.0749&1.0688&1.1988&1.2660&1.2431&1.1113&1.2364&1.2626&\textbf{1.3358}& \textcolor[rgb]{0.00,0.00,1.00}{1.2973}
     \cr
   \qquad	&  $\operatorname{AUC} _{(\operatorname{SNPR})}$

&5.5066&2.6949&4.1404  &
 \textcolor[rgb]{0.00,0.00,1.00}{125.7441}  &\textbf{164.1470}  &
 13.0048&4.5427&8.1761&6.3651&6.1572&8.7213  & 12.3738&6.0506  & {14.2684}&{18.3170}

    \cr
 \qquad	& $\operatorname{AUC} _{(\operatorname{TDBS})}$

&0.1051&0.1334&0.1695&
0.1261&0.2156&
0.1046&0.1348&0.2231&\textcolor[rgb]{0.00,0.00,1.00}{0.3146}&0.2787&0.1822&0.2565&0.2930&\textbf{0.3446}&0.3110

  \cr

   \hline
\multirow{6}{*}{Urban-3}& $\operatorname{AUC} _{({\operatorname{P}}_{\mathnormal{D}}, {\operatorname{P}}_{\mathnormal{F}})}$
   & 0.9512&0.9611&0.9785&
   0.9846 &0.9689 &
   0.9656&0.9210&0.9423&0.9227&0.9849&0.9650&0.9447&0.9829&\textcolor[rgb]{0.00,0.00,1.00}{0.9862}& \textbf{0.9916}

      \cr  
   \qquad	&  $\operatorname{AUC} _{({\operatorname{P}}_{\mathnormal{D}}, \tau)}$

&0.0961&0.2581&0.3887&
0.1620 &    0.0995 &
0.2070&0.3373&0.4393&0.4118&0.4665&0.1856&0.3988&\textcolor[rgb]{0.00,0.00,1.00}{0.5311}&0.4255
& \textbf{0.5342}

   \cr
    \qquad	&   $\operatorname{AUC} _{({\operatorname{P}}_{\mathnormal{F}}, \tau)}$

&{0.0350}&{0.0312}&0.0784&
\textbf{0.0049} & \textcolor[rgb]{0.00,0.00,1.00}{0.0061}  &
0.0369&0.1071&0.1123&0.1228&0.0644&0.0369&0.0892&0.0696&0.0547
&
 0.0363
     \cr
   \qquad	&  $\operatorname{AUC} _{(\operatorname{ODP})}$
   &1.0123&1.1881&1.2889&
   1.1417 &  1.0624
   &1.1358
   &1.1512&1.2693&1.2116&1.3871&1.1137&1.2542&\textcolor[rgb]{0.00,0.00,1.00}{1.4444}&1.3570&\textbf{1.4895}

      \cr
   \qquad	&  $\operatorname{AUC} _{(\operatorname{SNPR})}$
   &2.7448&{8.2811}&4.9608&
    \textbf{32.5299} &   \textcolor[rgb]{0.00,0.00,1.00}{16.5531}  &
   5.6121&3.1500&3.9137&3.3542&7.2451&5.0274&4.4690&7.6290&{7.7790}
&{14.7299}

    \cr
 \qquad	& $\operatorname{AUC} _{(\operatorname{TDBS})}$
 &0.0611&0.2269&0.3104&
 0.1571& 0.0935&
 0.1701&0.2302&0.3271&0.2890&0.4021&0.1487&0.3096&\textcolor[rgb]{0.00,0.00,1.00}{0.4615}&0.3708&\textbf{0.4979}
  \cr

   \hline
\multirow{6}{*}{Urban-4}& $\operatorname{AUC} _{({\operatorname{P}}_{\mathnormal{D}}, {\operatorname{P}}_{\mathnormal{F}})}$
    &0.9886&0.9832&0.9920&
    0.9375 &0.9885  &
    0.9815&0.9862&0.9401&\textcolor[rgb]{0.00,0.00,1.00}{0.9955}&0.9764&0.9747&0.9953&0.9876&0.9743&  \textbf{0.9958}

       \cr  
   \qquad	&  $\operatorname{AUC} _{({\operatorname{P}}_{\mathnormal{D}}, \tau)}$

&0.0874&0.0295&0.1417&
 0.0213  &0.0828  &
0.0312&\textcolor[rgb]{0.00,0.00,1.00}{0.1301}&0.0333&0.1161&0.1153&0.0870&{0.1237}&0.1187&0.1169& \textbf{0.1355}

   \cr
    \qquad	&   $\operatorname{AUC} _{({\operatorname{P}}_{\mathnormal{F}}, \tau)}$

&0.0114&{0.0013}&0.0185&
\textbf{0.0004} &0.0033 &
\textcolor[rgb]{0.00,0.00,1.00}  {0.0012}&
0.0169&0.0038&0.0074&0.0152&0.0062&0.0139&0.0171&0.0253& 0.0055

    \cr
   \qquad	&  $\operatorname{AUC} _{(\operatorname{ODP})}$

&1.0647&1.0114&\textcolor[rgb]{0.00,0.00,1.00}{1.1152}&
0.9584 & 1.0680  &
1.0116&1.0994&0.9696&1.1042&1.0764&1.0555&1.1051&1.0892&1.0659&\textbf{1.1258}

      \cr
   \qquad	&  $\operatorname{AUC} _{(\operatorname{SNPR})}$

&7.6677  & 22.6400&7.6387
&   \textbf{50.4986}  & 24.5451
&\textcolor[rgb]{0.00,0.00,1.00}{26.1749}&7.6901&8.6572   &15.7677&7.5638 &  14.0510&8.9005&6.9238&4.6225&{24.6349}

    \cr
 \qquad	& $\operatorname{AUC} _{(\operatorname{TDBS})}$
  &0.0760&0.0282&\textcolor[rgb]{0.00,0.00,1.00}{0.1231}&
  0.0208&   0.0795&
  0.0300&0.1132&
  0.0295&0.1087&0.1000&0.0808&0.1098&0.1016&0.0916&\textbf{0.1300}
  \cr

   \hline
\multirow{6}{*}{Urban-5}& $\operatorname{AUC} _{({\operatorname{P}}_{\mathnormal{D}}, {\operatorname{P}}_{\mathnormal{F}})}$
   &0.9691&0.9493&0.9075&
0.9550 &0.9521  &
   0.9612&0.9263&0.9317&0.9510&0.9700&0.9587&0.9626&0.9812&\textbf{0.9907}&\textcolor[rgb]{0.00,0.00,1.00}{0.9847}

       \cr  
   \qquad	&  $\operatorname{AUC} _{({\operatorname{P}}_{\mathnormal{D}}, \tau)}$
   &0.1456&0.1785&0.2300&
    0.0929 & 0.2453 &
   0.1475&\textbf{0.4686}&0.2914&0.3690&0.4410&0.3568&0.3057&0.3425&\textcolor[rgb]{0.00,0.00,1.00}{0.4576}&0.3773

   \cr
    \qquad	&   $\operatorname{AUC} _{({\operatorname{P}}_{\mathnormal{F}}, \tau)}$

    &0.0437&0.0474&0.0881&
      \textbf{0.0041}  &  \textcolor[rgb]{0.00,0.00,1.00}{0.0300} &
   {0.0318}&0.2118&0.0888&0.0613&0.1054&0.0770&0.0563&0.0841&0.0713&{0.0366}

     \cr
   \qquad	&  $\operatorname{AUC} _{(\operatorname{ODP})}$
   &1.0710&1.0804&1.0495&
    1.0439  & 1.1674  &
   1.0769&1.1831&1.1343&1.2587&1.3056&1.2385&1.2120&1.2395&\textbf{1.3770}&\textcolor[rgb]{0.00,0.00,1.00}{1.3254}

     \cr
   \qquad	&  $\operatorname{AUC} _{(\operatorname{SNPR})}$

&3.3311&3.7640&2.6111&
\textbf{22.8410}  &8.1646  &
4.6403&2.2128&3.2816&6.0177&4.1850&4.6356&5.4297&4.0704&{6.4191}& \textcolor[rgb]{0.00,0.00,1.00}{10.3161}

    \cr
 \qquad	& $\operatorname{AUC} _{(\operatorname{TDBS})}$
  &0.1019&0.1311&0.1419&
  0.0888&0.2153 & 
  0.1157&0.2568&0.2026&0.3077&0.3357&0.2798&0.2494&0.2583&\textbf{0.3863}& \textcolor[rgb]{0.00,0.00,1.00}{0.3407}
  \cr


   \hline
    \multicolumn{2}{c}{Average Time (second)} 
  &\textbf{0.0853} & {4.7598}& 109.5701&
  \textcolor[rgb]{0.00,0.00,1.00}{1.0204}& 43.4618&
  8.6138 &61.9130 & 121.0873&22.4573 &{3.5494} &131.0160 &67.7740 &5.1300 &65.8005  & 53.62
\cr
   \hline

    \end{tabular}
    \end{threeparttable}
    \vspace{-0.2cm}
\end{table*}


\begin{table*}[tp]
\renewcommand{\arraystretch}{1.0}
\setlength\tabcolsep{2.7pt}
  \centering
  \caption{\textcolor[rgb]{0.00,0.00,0.00}{The  AUC  values  and average running time output by various HAD  methods on 
 large-scale 
  HSI  datasets.}
   \textcolor[rgb]{0.00,0.00,0.00}{The best  and the second-best results are highlighted in boldface
and blue, respectively.}
  }
  \label{hsi_various_auc111117777}
  \scriptsize
  \begin{threeparttable}
    \begin{tabular}{cl   c  c c   c cc   c cc  c cc   c  c    c }
    \hline
   \multicolumn{1}{c}{HSI-Name}&   \multicolumn{1}{c}{AUC-Metrics}   &
   \multicolumn{1}{c}{RX}&\multicolumn{1}{c}{CRD}
   &  \multicolumn{1}{c}{GAED}   & \multicolumn{1}{c}{PDBSNet}
    & \tabincell{c} {LRASR}   &  \tabincell{c} {GTV-\\LRR}
    &  \multicolumn{1}{c}{PTA}&     \tabincell{c}{PCA-\\TLRSR}& \multicolumn{1}{c}{TCTV}
    &  \multicolumn{1}{c}{\text{GCS}} &  \tabincell{c}{MERA-\\ETV}&
    \tabincell{c} {TRD-\\FTVAD}&  \multicolumn{1}{c}{\text{GNLTR}}& \multicolumn{1}{c}{\text{GNBRL}}&
    \multicolumn{1}{c}{\text{Proposed}}
    \cr

    \hline
    \hline


   \multirow{6}{*}{Qingpu-I} 
   & $\operatorname{AUC} _{({\operatorname{P}}_{\mathnormal{D}}, {\operatorname{P}}_{\mathnormal{F}})}$
 &0.9958&0.9856&0.9968&0.9410&0.9576&\textcolor[rgb]{0.00,0.00,1.00}{0.9993}&0.8750&0.9969&0.9898&0.9968&0.9003&0.9976&0.9961&0.9948&\textbf{0.9997}

       \cr
   \qquad	&  $\operatorname{AUC} _{({\operatorname{P}}_{\mathnormal{D}}, \tau)}$
   &0.3375&0.2300&0.4680&0.0286&0.4053&0.5736&0.1626&0.5941&0.5562&\textbf{0.6185}&0.4530&0.4995&0.5813&0.5745&\textcolor[rgb]{0.00,0.00,1.00}{0.5943}

   \cr
    \qquad	&   $\operatorname{AUC} _{({\operatorname{P}}_{\mathnormal{F}}, \tau)}$
  &
0.0402&\textcolor[rgb]{0.00,0.00,1.00}{0.0315}&0.0802&\textbf{0.0047}&0.1407&0.0556&0.0676&0.1315&0.1553&0.1472&0.1259&0.0665&0.1376&0.1327&{0.0363}

     \cr
   \qquad	&  $\operatorname{AUC} _{(\operatorname{ODP})}$
  &  1.2932&1.1842&1.3846&0.9648&1.2222&\textcolor[rgb]{0.00,0.00,1.00}{1.5173}&0.9700&1.4596&1.3908&1.4681&1.2275&1.4306&1.4398&1.4365&\textbf{1.5576}

      \cr
   \qquad	&  $\operatorname{AUC} _{(\operatorname{SNPR})}$
   &8.4002&7.3067&5.8323&6.0404&2.8811&\textcolor[rgb]{0.00,0.00,1.00}{10.3144}&2.4039&4.5193&3.5819&4.2016&3.5993&7.5084&4.2259&4.3287&\textbf{16.3678}

    \cr
 \qquad	& $\operatorname{AUC} _{(\operatorname{TDBS})}$
 &0.2974&0.1985&0.3878&0.0238&0.2646&\textcolor[rgb]{0.00,0.00,1.00}{0.5180}&0.0950&0.4627&0.4009&0.4713&0.3272&0.4330&0.4437&0.4417&\textbf{0.5580}

   \cr

    \hline

    \multirow{6}{*}{Qingpu-II}& $\operatorname{AUC} _{({\operatorname{P}}_{\mathnormal{D}}, {\operatorname{P}}_{\mathnormal{F}})}$
 &
 0.9990&0.9997&\textbf{1.0000}&0.9982&\textbf{1.0000}&
 \textcolor[rgb]{0.00,0.00,1.00}{0.9999}&0.9951&\textcolor[rgb]{0.00,0.00,1.00}{0.9999}&\textcolor[rgb]{0.00,0.00,1.00}{0.9999}&0.9994&0.9988&0.9977&0.9996&0.9856&\textbf{1.0000}

       \cr
   \qquad	&  $\operatorname{AUC} _{({\operatorname{P}}_{\mathnormal{D}}, \tau)}$
   &
   0.1175&0.1708&\textcolor[rgb]{0.00,0.00,1.00}{0.5861}&0.2988&\textbf{0.6506}&0.5501&0.2728&0.3606&0.3518&0.3490&0.2673&0.1646&0.3820&0.3648&0.3743

   \cr
    \qquad	&   $\operatorname{AUC} _{({\operatorname{P}}_{\mathnormal{F}}, \tau)}$
  &\textcolor[rgb]{0.00,0.00,1.00}{0.0158}&0.0182&0.0365&\textbf{0.0048}&0.0746&0.0360&0.0272&0.0586&0.0551&0.0658&0.0551&0.0278&0.0707&0.1800&{0.0229}

     \cr
   \qquad	&  $\operatorname{AUC} _{(\operatorname{ODP})}$
   &
    1.1008&1.1524&\textcolor[rgb]{0.00,0.00,1.00}{1.5496}&1.2923&\textbf{1.5759}&
    1.5141&1.2407&1.3019&1.2966&1.2826&1.2110&1.1344&1.3110&1.1704&1.3514
      \cr
   \qquad	&  $\operatorname{AUC} _{(\operatorname{SNPR})}$
   &

7.4418&9.4060&16.0620&\textbf{62.0467}&8.7183&15.2814&10.0268&6.1560&6.3875&5.3052&4.8498&5.9125&5.4069&2.0265&\textcolor[rgb]{0.00,0.00,1.00}{16.3476}
    \cr
 \qquad	& $\operatorname{AUC} _{(\operatorname{TDBS})}$
 &

0.1017&0.1527&\textcolor[rgb]{0.00,0.00,1.00}{0.5496}&0.2940&\textbf{0.5759}&0.5141&0.2456&0.3020&0.2967&0.2832&0.2122&0.1367&0.3114&0.1848&0.3514
   \cr

\hline
    \multirow{6}{*}{AVON}& $\operatorname{AUC} _{({\operatorname{P}}_{\mathnormal{D}}, {\operatorname{P}}_{\mathnormal{F}})}$
 &

0.8358&0.8047&0.9432&0.8233&0.8884&0.9332&0.8948&0.8994&0.9179&0.8792&0.9028&\textcolor[rgb]{0.00,0.00,1.00}{0.9797}&0.9108&0.9128&\textbf{0.9916}

       \cr
   \qquad	&  $\operatorname{AUC} _{({\operatorname{P}}_{\mathnormal{D}}, \tau)}$
   &
   0.0274&0.0428&0.1037&0.0272&0.1231&0.1896&0.1062&0.1823&0.1813&0.1293&0.1814&0.1521&\textcolor[rgb]{0.00,0.00,1.00}{0.1966}&\textbf{0.1970}&0.1715

   \cr
    \qquad	&   $\operatorname{AUC} _{({\operatorname{P}}_{\mathnormal{F}}, \tau)}$
  &
\textcolor[rgb]{0.00,0.00,1.00}{0.0033} &
\textbf{0.0015}&0.0209&0.0034&0.0268&0.0309&0.0276&0.0403&0.0410&0.0265&0.0373&0.0131&0.0521&0.0517&0.0092

     \cr
   \qquad	&  $\operatorname{AUC} _{(\operatorname{ODP})}$
   &
   0.8599&0.8460&1.0260&0.8471&0.9847&1.0919&0.9734&1.0414&1.0583&0.9821&1.0469&\textcolor[rgb]{0.00,0.00,1.00}{1.1187}&1.0553&1.0580&\textbf{1.1539}

      \cr
   \qquad	&  $\operatorname{AUC} _{(\operatorname{SNPR})}$
   &
   8.3872&\textbf{28.0867}&4.9604&8.0295&4.5876&6.1452&3.8444&4.5237&4.4251&4.8847&4.8608&{11.6118}&3.7733&3.8065&\textcolor[rgb]{0.00,0.00,1.00}{18.6560}

    \cr
 \qquad	& $\operatorname{AUC} _{(\operatorname{TDBS})}$
 &
0.0241&0.0413&0.0828&0.0238&0.0963&\textcolor[rgb]{0.00,0.00,1.00}{0.1588}&0.0786&0.1420&0.1404&0.1028&0.1441&0.1390&0.1445&{0.1452}&\textbf{0.1623}
   \cr
\hline
    \multirow{6}{*}{CRI}& $\operatorname{AUC} _{({\operatorname{P}}_{\mathnormal{D}}, {\operatorname{P}}_{\mathnormal{F}})}$
 &

0.9781&0.9803&\textcolor[rgb]{0.00,0.00,1.00}{0.9843}&\textbf{0.9876}&0.9101&0.9333&0.9750&0.8527&0.9173&0.9517&0.7122&0.8887&0.8254&0.9760&0.9761

       \cr
   \qquad	&  $\operatorname{AUC} _{({\operatorname{P}}_{\mathnormal{D}}, \tau)}$
   &
   0.0888&0.1115&\textcolor[rgb]{0.00,0.00,1.00}{0.5339}&0.3168&0.4329&0.4264&\textbf{0.7723}&0.2607&0.3767&0.4828&0.2201&0.3778&0.3805&0.5042&{0.5046}

   \cr
    \qquad	&   $\operatorname{AUC} _{({\operatorname{P}}_{\mathnormal{F}}, \tau)}$
  &
0.0236&\textbf{0.0037}&0.0916&\textcolor[rgb]{0.00,0.00,1.00}{0.0172}&0.1286&0.1785&0.2456&0.0831&0.0993&0.2052&0.1160&0.2120&0.2293&0.1675&0.1717

     \cr
   \qquad	&  $\operatorname{AUC} _{(\operatorname{ODP})}$
   &
   1.0432&1.0881&\textcolor[rgb]{0.00,0.00,1.00}{1.4266}&1.2872&1.2144&1.1813&\textbf{1.5018}&1.0303&1.1947&1.2293&0.8163&1.0546&0.9766&1.3128&1.3089

      \cr
   \qquad	&  $\operatorname{AUC} _{(\operatorname{SNPR})}$
   &
   3.7571&\textbf{30.0465}&5.8285&\textcolor[rgb]{0.00,0.00,1.00}{18.4298}&3.3663&2.3892&3.1452&3.1383&3.7925&2.3526&1.8974&1.7824&1.6595&3.0107&2.9379

    \cr
 \qquad	& $\operatorname{AUC} _{(\operatorname{TDBS})}$
 &
0.0651&0.1078&\textcolor[rgb]{0.00,0.00,1.00}{0.4423}&0.2996&0.3043&0.2479&\textbf{0.5268}&0.1776&0.2774&0.2776&0.1041&0.1658&0.1512&0.3367&0.3328
   \cr
\hline

    \multirow{6}{*}{Beach}& $\operatorname{AUC} _{({\operatorname{P}}_{\mathnormal{D}}, {\operatorname{P}}_{\mathnormal{F}})}$
 &

0.8106&0.7462&0.8522&0.9236&0.7452&0.8110&0.9002&0.8048&0.8712&0.9015&0.8321&0.9194&0.9487&\textbf{0.9715}&\textcolor[rgb]{0.00,0.00,1.00}{0.9709}

       \cr
   \qquad	&  $\operatorname{AUC} _{({\operatorname{P}}_{\mathnormal{D}}, \tau)}$
   &
   0.0157&0.0024&0.1993&0.0143&0.1751&0.0716&\textbf{0.6674}&0.0669&0.1160&0.1744&0.0677&0.1624&0.1442&\textcolor[rgb]{0.00,0.00,1.00}{0.2376}&0.2182

   \cr
    \qquad	&   $\operatorname{AUC} _{({\operatorname{P}}_{\mathnormal{F}}, \tau)}$
  &
0.0032&\textcolor[rgb]{0.00,0.00,1.00}{0.0007}&0.0644&\textbf{0.0006}&0.0635&0.0298&0.6171&0.0286&0.0252&0.0578&0.0264&0.0277&0.0318&0.0358&0.0368

     \cr
   \qquad	&  $\operatorname{AUC} _{(\operatorname{ODP})}$
   &
   0.8231&0.7479&0.9871&0.9373&0.8568&0.8528&0.9506&0.8430&0.9620&1.0182&0.8734&1.0541&1.0611&\textbf{1.1733}&\textcolor[rgb]{0.00,0.00,1.00}{1.1523}

      \cr
   \qquad	&  $\operatorname{AUC} _{(\operatorname{SNPR})}$
   &
   4.8724&3.3326&3.0933&\textbf{24.8677}&2.7562&2.4034&1.0816&2.3390&4.5990&3.0191&2.5603&5.8697&4.5378&\textcolor[rgb]{0.00,0.00,1.00}{6.6310}&5.9277

    \cr
 \qquad	& $\operatorname{AUC} _{(\operatorname{TDBS})}$
 &0.0125&0.0017&0.1349&0.0137&0.1115&0.0418&0.0504&0.0383&0.0908&0.1167&0.0413&0.1347&0.1124&\textbf{0.2018}&\textcolor[rgb]{0.00,0.00,1.00}{0.1814}

   \cr
\hline
    \multirow{6}{*}{SNP}& $\operatorname{AUC} _{({\operatorname{P}}_{\mathnormal{D}}, {\operatorname{P}}_{\mathnormal{F}})}$
 &

\textbf{0.9969}&0.9076&0.9732&\textcolor[rgb]{0.00,0.00,1.00}{0.9966}&0.9944&0.9928&0.9879&0.9654&0.9904&0.9936&0.9643&0.9780&0.9922&0.9842&\textcolor[rgb]{0.00,0.00,0.00}{0.9948}

       \cr
   \qquad	&  $\operatorname{AUC} _{({\operatorname{P}}_{\mathnormal{D}}, \tau)}$
   &
   0.0197&0.0102&\textcolor[rgb]{0.00,0.00,1.00}{0.4965}&0.4385&\textbf{0.5239}&0.3512&0.4159&0.2480&0.3096&0.3793&0.1605&0.2943&0.3491&0.4263&0.3405

   \cr
    \qquad	&   $\operatorname{AUC} _{({\operatorname{P}}_{\mathnormal{F}}, \tau)}$
  &
\textcolor[rgb]{0.00,0.00,1.00}{0.0002}&\textbf{0.0001}&0.1296&0.0032&0.0491&0.0486&0.0745&0.0459&0.0248&0.0452&0.0171&0.0550&0.0544&0.0741&0.0226

     \cr
   \qquad	&  $\operatorname{AUC} _{(\operatorname{ODP})}$
   &1.0163&0.9176&1.3402&\textcolor[rgb]{0.00,0.00,1.00}{1.4319}&\textbf{1.4692}&1.2953&1.3293&1.1675&1.2752&1.3277&1.1078&1.2173&1.2869&1.3364&1.3126

      \cr
   \qquad	&  $\operatorname{AUC} _{(\operatorname{SNPR})}$
   &
   \textcolor[rgb]{0.00,0.00,1.00}{88.5452}&72.7690&3.8319&\textbf{137.4273}&10.6779&7.2218&5.5829&5.4009&12.4751&8.3901&9.4019&5.3515&6.4163&5.7532&15.0538

    \cr
 \qquad	& $\operatorname{AUC} _{(\operatorname{TDBS})}$
 &
0.0194&0.0100&0.3670&\textcolor[rgb]{0.00,0.00,1.00}{0.4353}&\textbf{0.4748}&0.3026&0.3414&0.2021&0.2847&0.3341&0.1434&0.2393&0.2947&0.3522&0.3179
   \cr
   \hline

    \multirow{6}{*}{Synthetic}& $\operatorname{AUC} _{({\operatorname{P}}_{\mathnormal{D}}, {\operatorname{P}}_{\mathnormal{F}})}$
 &

0.7118&0.4997&0.8429&\textbf{0.9506}&0.7032&0.7074&0.6116&0.9078&0.9160&0.6826&0.7917&0.7814&0.7993&0.9192&\textcolor[rgb]{0.00,0.00,1.00}{0.9210}

       \cr
   \qquad	&  $\operatorname{AUC} _{({\operatorname{P}}_{\mathnormal{D}}, \tau)}$
   &
   0.0362&0.0001&\textbf{0.3668}&0.1463&0.2515&0.1936&0.1219&0.2739&0.2812&0.1697&0.1680&0.1095&0.2165&\textcolor[rgb]{0.00,0.00,1.00}{0.2823}&0.2813

   \cr
    \qquad	&   $\operatorname{AUC} _{({\operatorname{P}}_{\mathnormal{F}}, \tau)}$
  &
0.0147&\textbf{0.0001}&0.1855&\textcolor[rgb]{0.00,0.00,1.00}{0.0097}&0.2078&0.1382&0.0846&0.1225&0.1236&0.1146&0.0881&0.0511&0.1256&0.1236&0.1222

     \cr
   \qquad	&  $\operatorname{AUC} _{(\operatorname{ODP})}$
   &0.7332&0.4997&1.0242&\textbf{1.0871}&0.7469&0.7628&0.6489&1.0592&1.0737&0.7377&0.8716&0.8397&0.8902&1.0778&\textcolor[rgb]{0.00,0.00,1.00}{1.0801}

      \cr
   \qquad	&  $\operatorname{AUC} _{(\operatorname{SNPR})}$
   &
   2.4568&0.9921&1.9776&\textbf{15.0512}&1.2104&1.4012&1.4410&2.2355&2.2752&1.4810&1.9073&2.1402&1.7242&2.2835&\textcolor[rgb]{0.00,0.00,1.00}{2.3022}

    \cr
 \qquad	& $\operatorname{AUC} _{(\operatorname{TDBS})}$
 &
0.0214&0.0000&\textbf{0.1813}&0.1365&0.0437&0.0554&0.0373&0.1514&0.1576&0.0551&0.0799&0.0583&0.0909&0.1587&\textcolor[rgb]{0.00,0.00,1.00}{0.1591}
   \cr


   \hline

    \multicolumn{2}{c}{Average Time (second)}
  &\textbf{3.13} 
  &3349.89
  & 2962.81
  & 834.38
  &  3342.15
  & 5710.69&\textcolor[rgb]{0.00,0.00,1.00}{483.88} &2210.92
  &747.78  &{529.65} &850.76
  &709.06
  &
  3072.29
  & 1091.67 & 851.96
\cr
   \hline
    \end{tabular}
    \end{threeparttable}
    \vspace{-0.52cm}
\end{table*}
\begin{figure*}[!htbp]
\renewcommand{\arraystretch}{0.4}
\setlength\tabcolsep{0.43pt}
\centering
\begin{tabular}{cc c  c cc cc }
\centering
\includegraphics[width=0.8683in, height=0.55in]{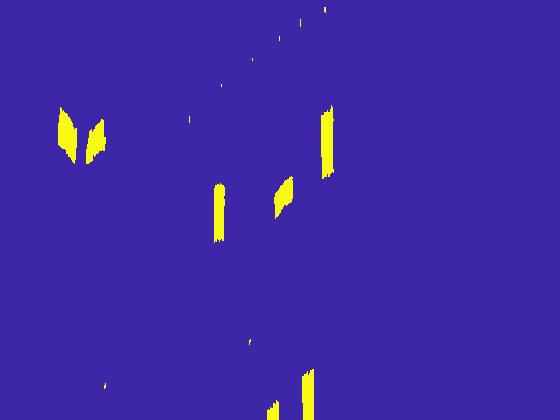}&
\includegraphics[width=0.8683in, height=0.55in]{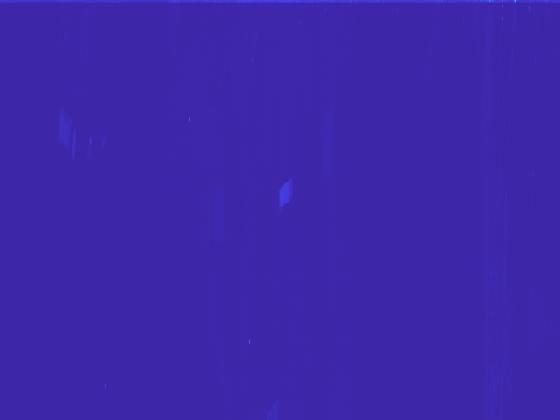}&
\includegraphics[width=0.8683in, height=0.55in]{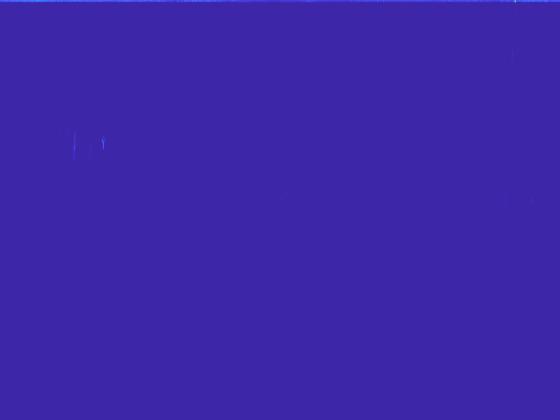}&
\includegraphics[width=0.8683in, height=0.55in]{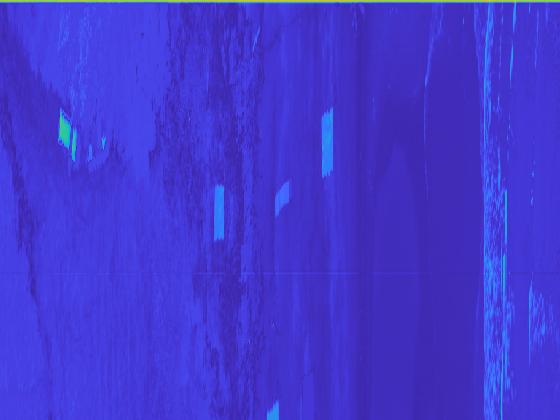}&
\includegraphics[width=0.8683in, height=0.55in]{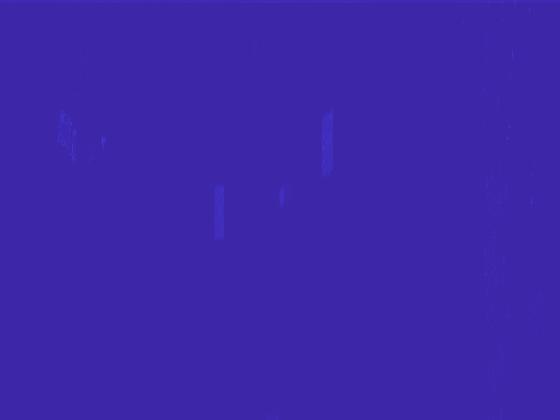}&
\includegraphics[width=0.8683in, height=0.55in]{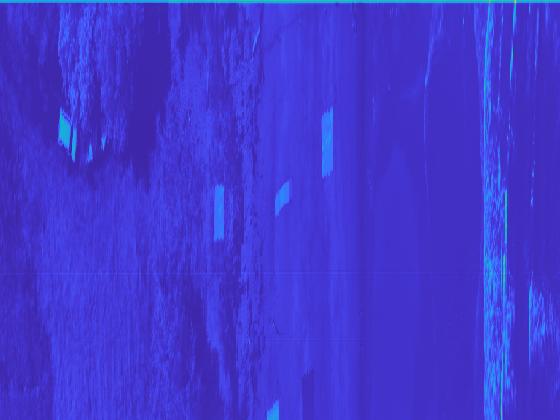}&
\includegraphics[width=0.8683in, height=0.55in]{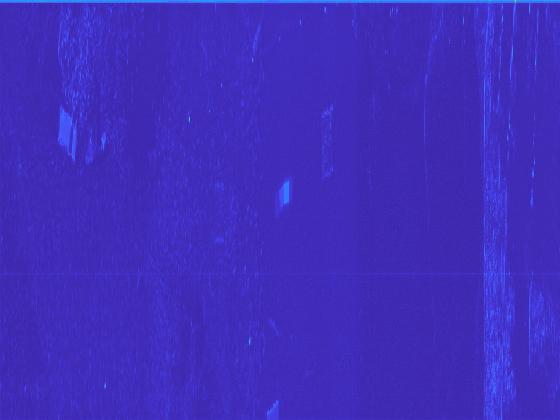}&
\includegraphics[width=0.8683in, height=0.55in]{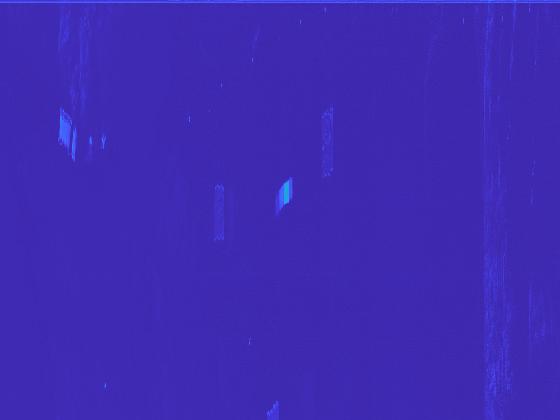} \\
{{{(a)}}}  &   {(b)}  & {(c)} &{(d)} & {(e)}  &{(f)}& {(g)}& {(h)}\\
\includegraphics[width=0.8683in, height=0.55in]{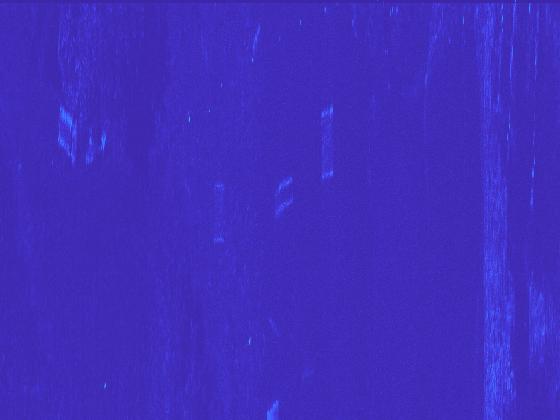}&  %
\includegraphics[width=0.8683in, height=0.55in]{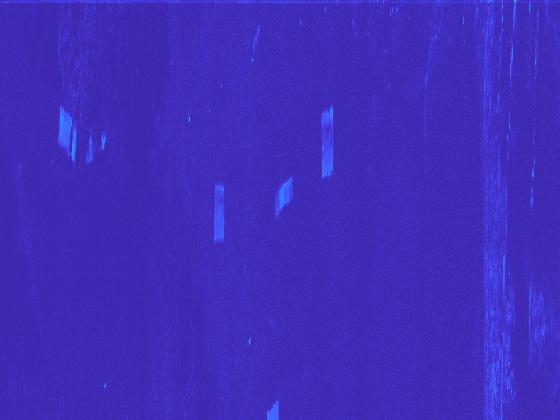} &
\includegraphics[width=0.8683in, height=0.55in]{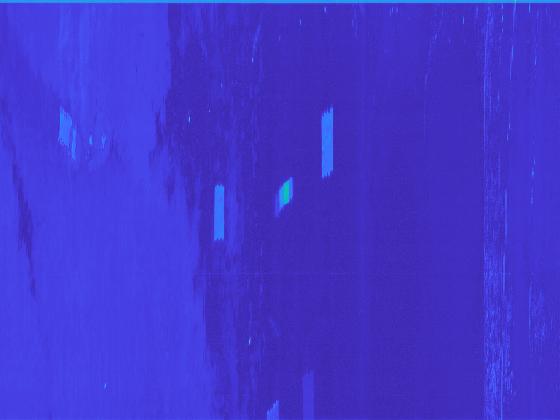} &
\includegraphics[width=0.8683in, height=0.55in]{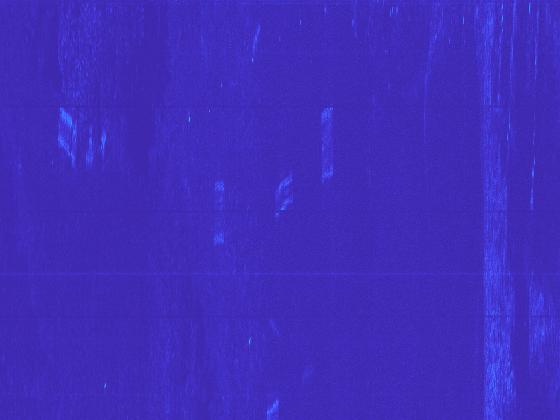}&
\includegraphics[width=0.8683in, height=0.55in]{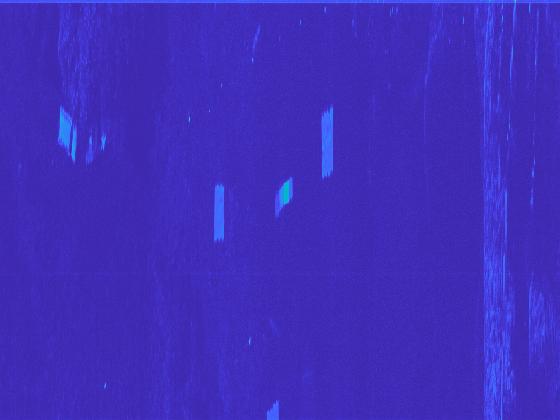}&
\includegraphics[width=0.8683in, height=0.55in]{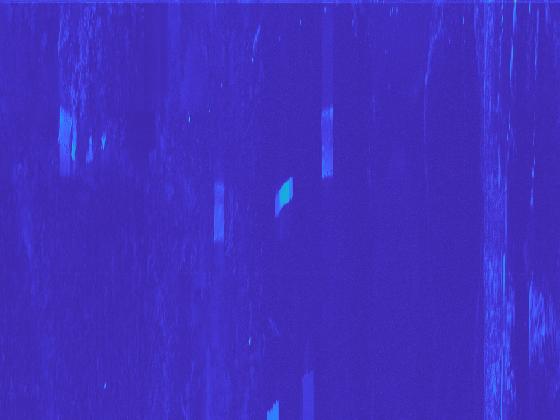} &
\includegraphics[width=0.8683in, height=0.55in]{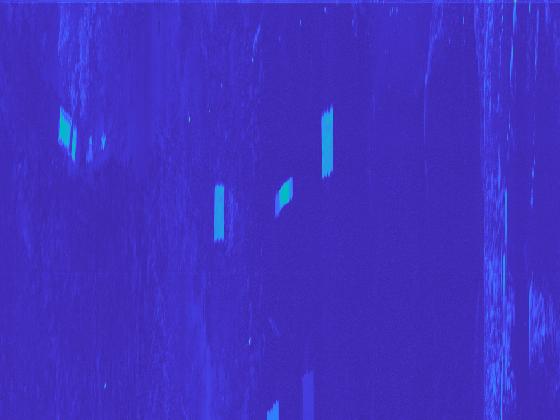}&
\includegraphics[width=0.8683in, height=0.55in]{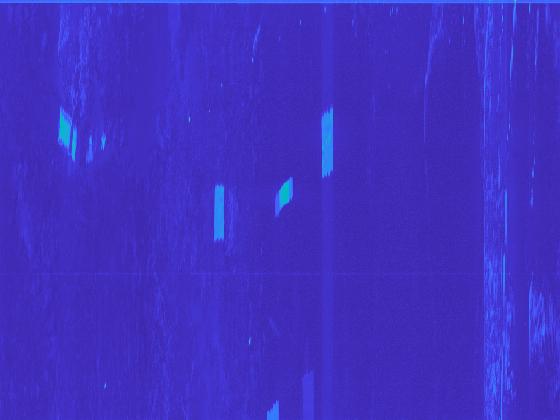}
\\
 {(i)}  &  {(j)} &  {(k)} &(l)  &(m)  &(n) &(o) &(p)\\
 \toprule

 \includegraphics[width=0.8683in, height=0.85in]{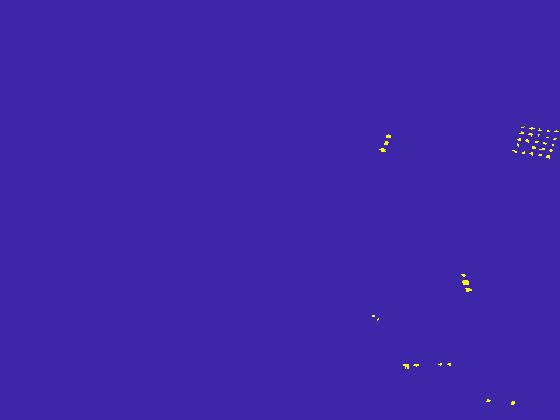}&
\includegraphics[width=0.8683in, height=0.85in]{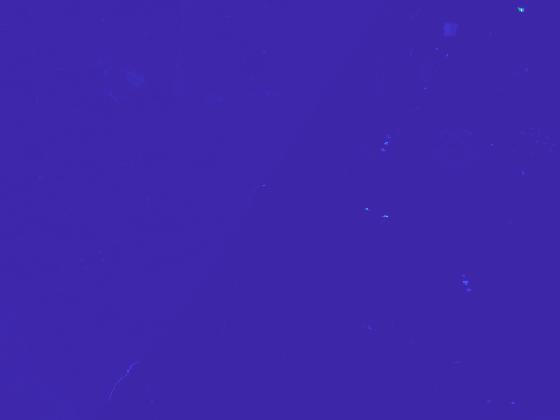}&
\includegraphics[width=0.8683in, height=0.85in]{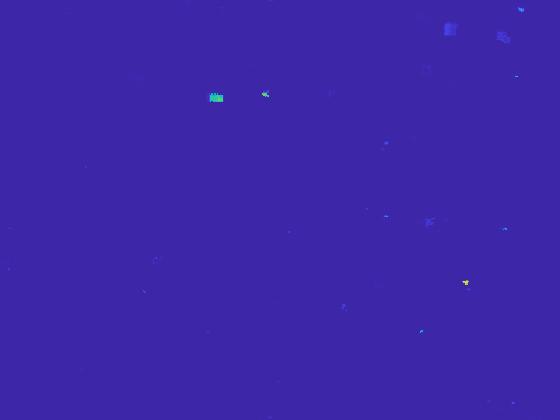}&
\includegraphics[width=0.8683in, height=0.85in]{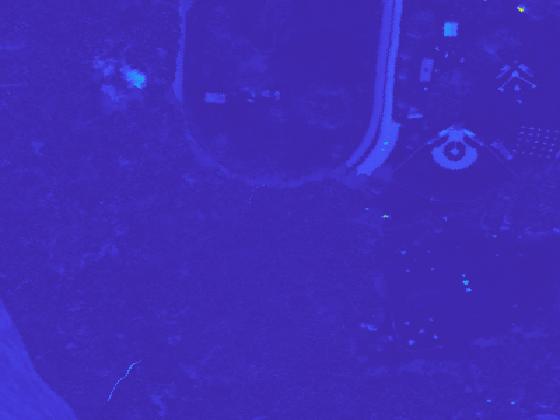}&
\includegraphics[width=0.8683in, height=0.85in]{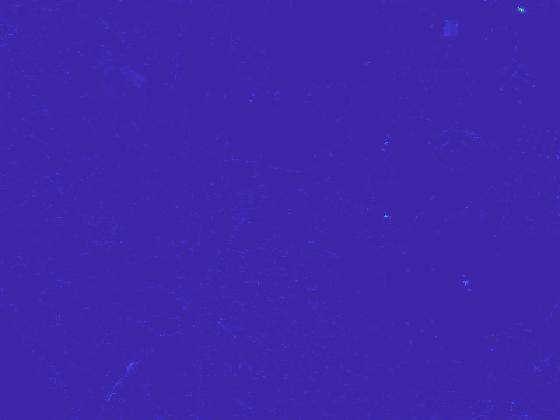}&
\includegraphics[width=0.8683in, height=0.85in]{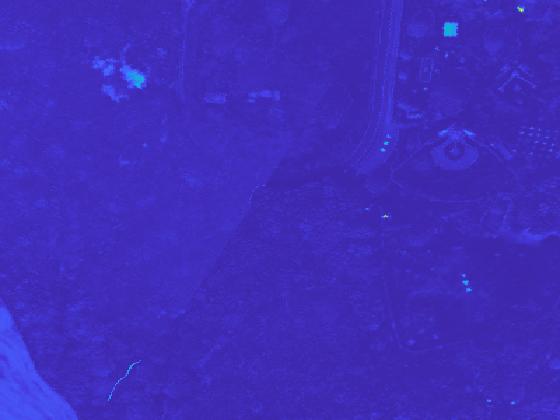}&
\includegraphics[width=0.8683in, height=0.85in]{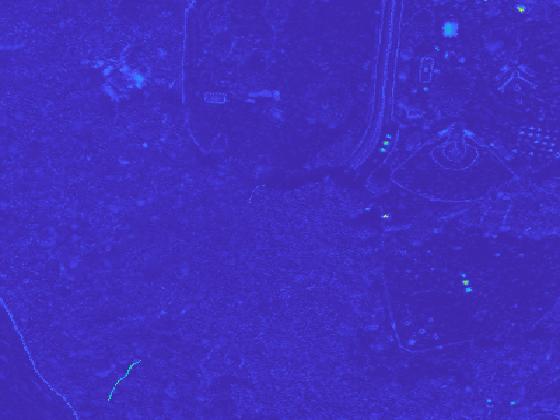}&
\includegraphics[width=0.8683in, height=0.85in]{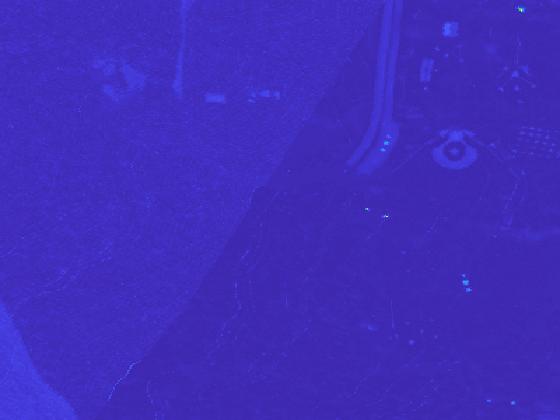} \\
{{{(a)}}}  &   {(b)}  & {(c)} &{(d)} & {(e)}  &{(f)}& {(g)}& {(h)}\\
\includegraphics[width=0.8683in, height=0.85in]{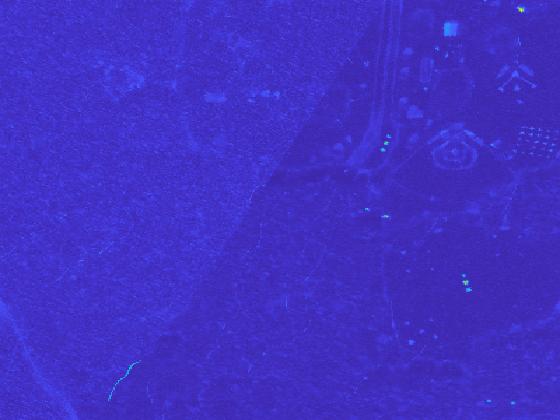}&  %
\includegraphics[width=0.8683in, height=0.85in]{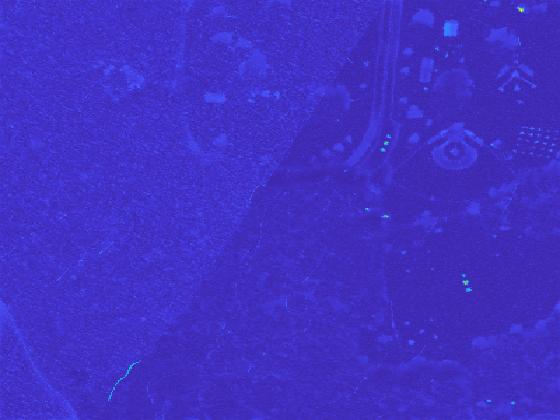} &
\includegraphics[width=0.8683in, height=0.85in]{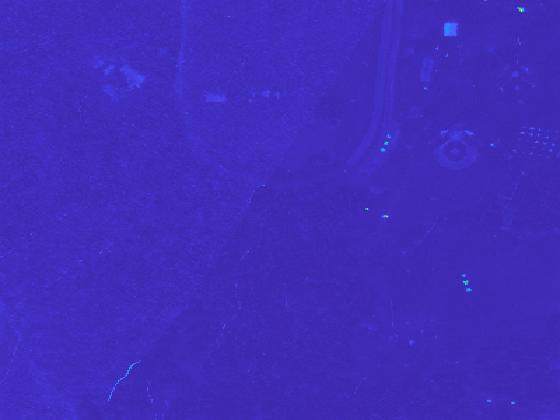} &
\includegraphics[width=0.8683in, height=0.85in]{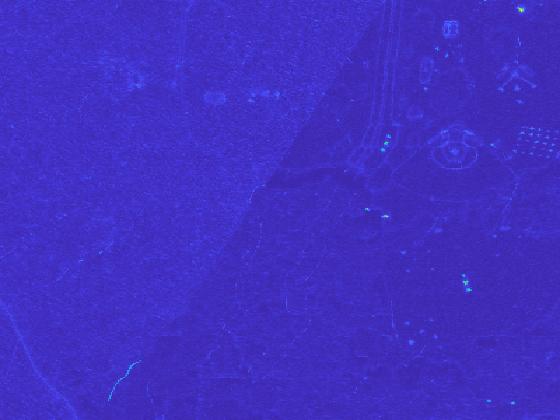}&
\includegraphics[width=0.8683in, height=0.85in]{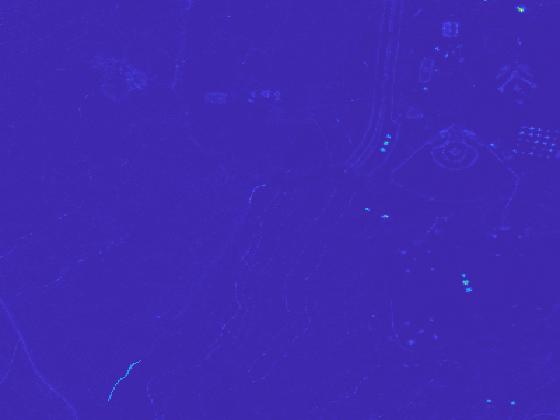}&
\includegraphics[width=0.8683in, height=0.85in]{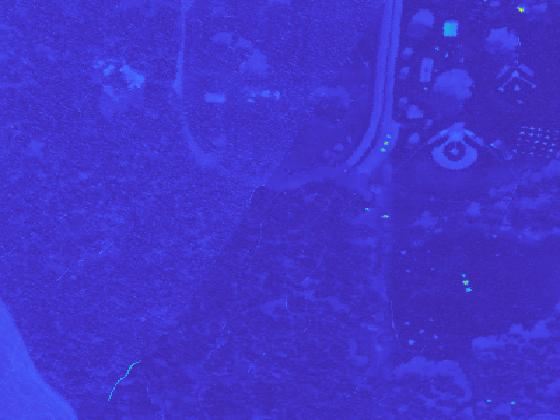} &
\includegraphics[width=0.8683in, height=0.85in]{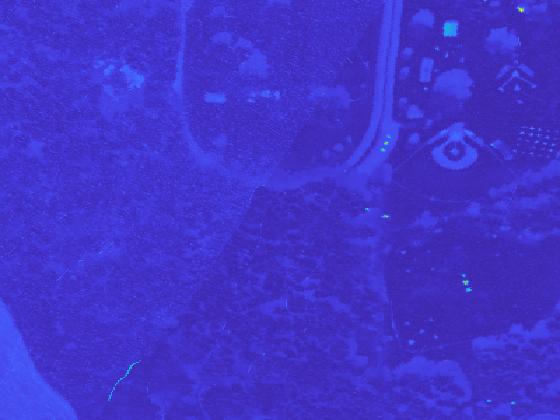}&
\includegraphics[width=0.8683in, height=0.85in]{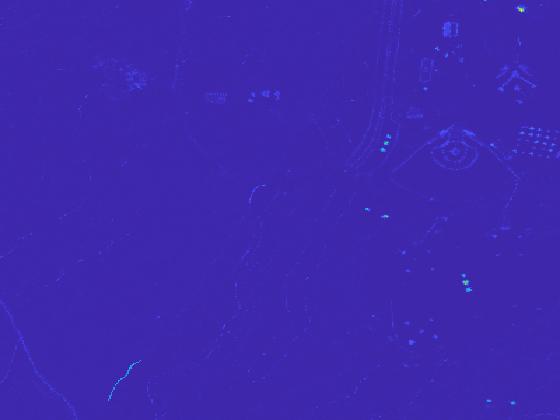}
\\
 {(i)}  &  {(j)} &  {(k)} &(l)  &(m)  &(n) &(o) &(p)\\
 \toprule
 \includegraphics[width=0.8683in, height=0.85in]{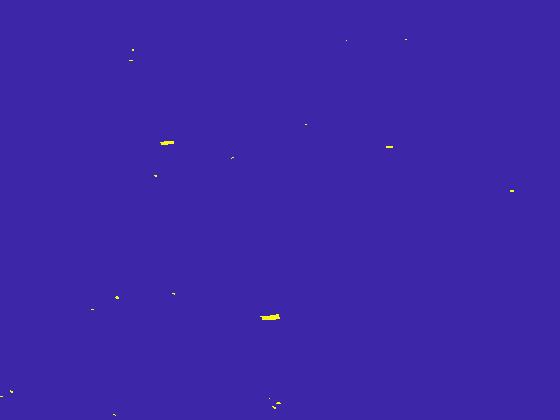}&
\includegraphics[width=0.8683in, height=0.85in]{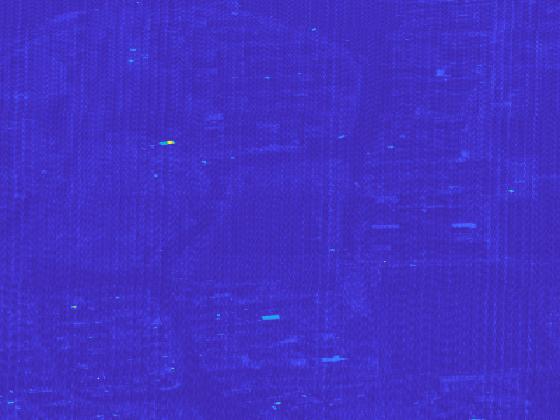}&
\includegraphics[width=0.8683in, height=0.85in]{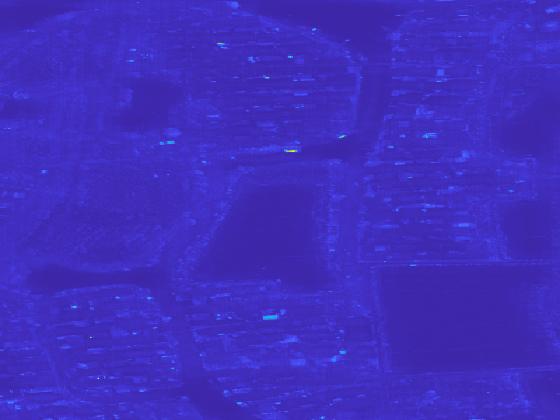}&
\includegraphics[width=0.8683in, height=0.85in]{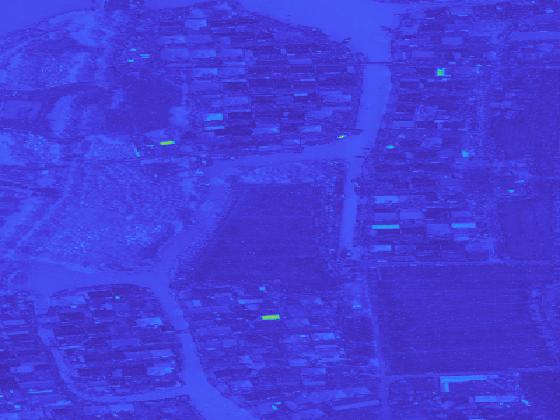}&
\includegraphics[width=0.8683in, height=0.85in]{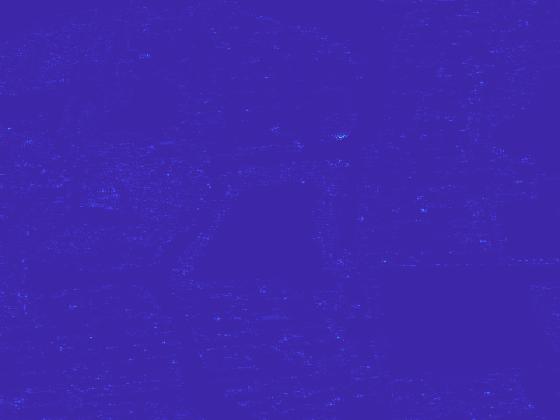}&
\includegraphics[width=0.8683in, height=0.85in]{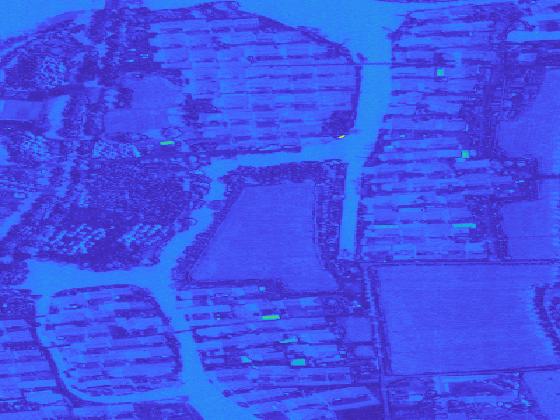}&
\includegraphics[width=0.8683in, height=0.85in]{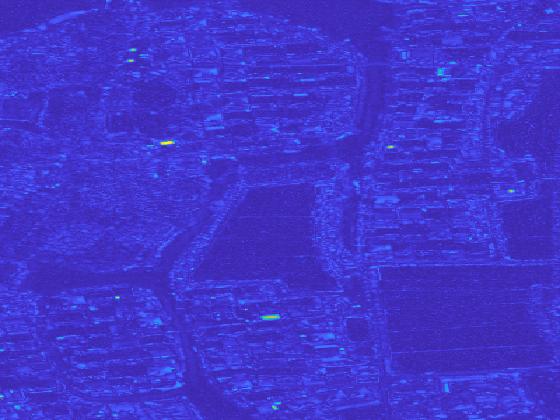}&
\includegraphics[width=0.8683in, height=0.85in]{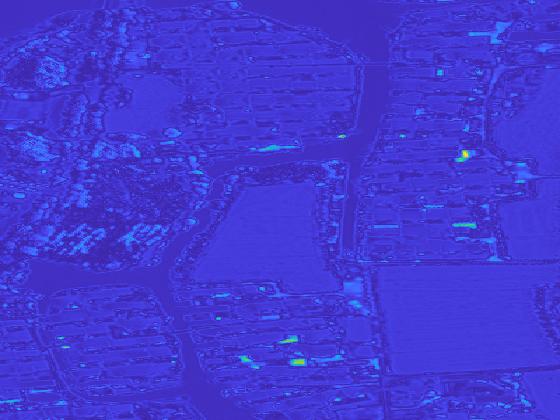} \\
{{{(a)}}}  &   {(b)}  & {(c)} &{(d)} & {(e)}  &{(f)}& {(g)}& {(h)}\\
\includegraphics[width=0.8683in, height=0.85in]{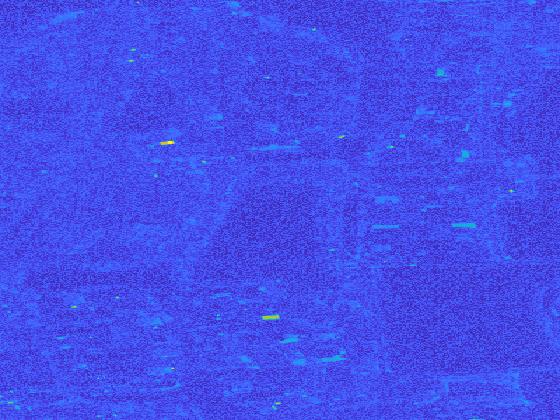}&  %
\includegraphics[width=0.8683in, height=0.85in]{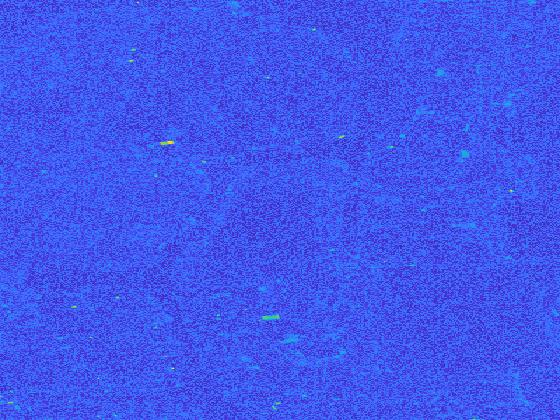} &
\includegraphics[width=0.8683in, height=0.85in]{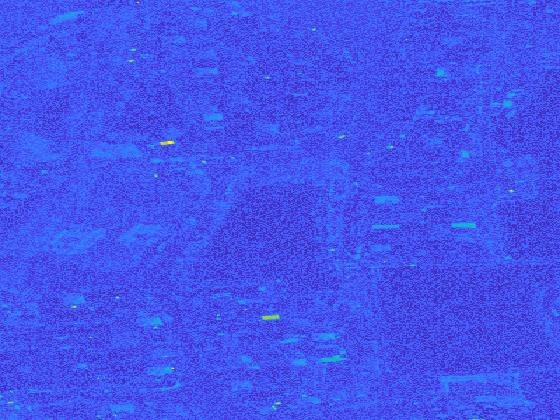} &
\includegraphics[width=0.8683in, height=0.85in]{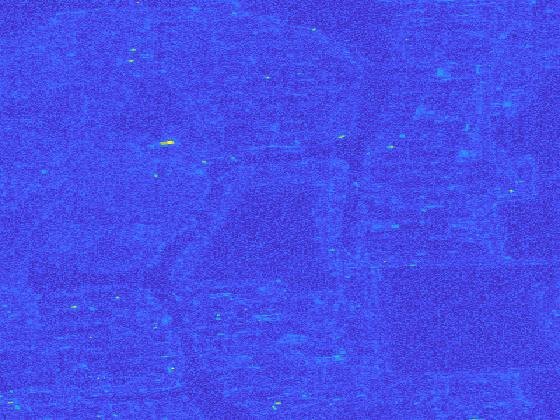}&
\includegraphics[width=0.8683in, height=0.85in]{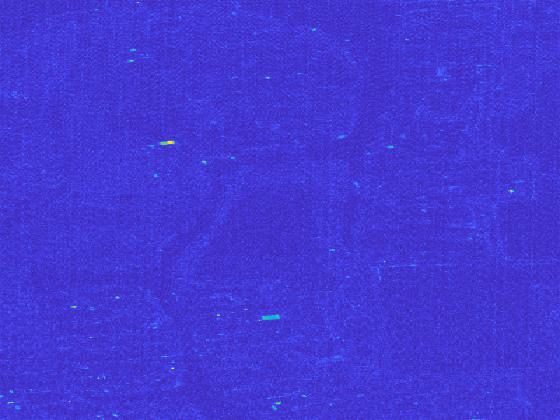}&
\includegraphics[width=0.8683in, height=0.85in]{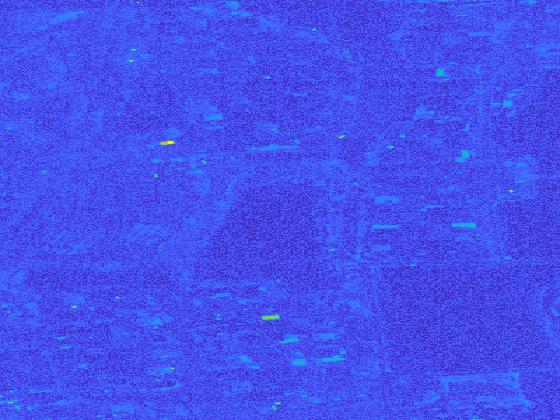} &
\includegraphics[width=0.8683in, height=0.85in]{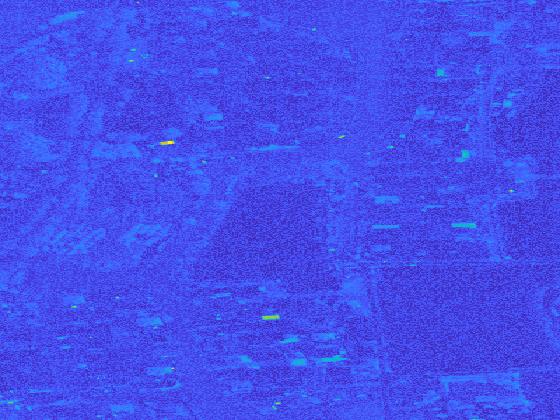}&
\includegraphics[width=0.8683in, height=0.85in]{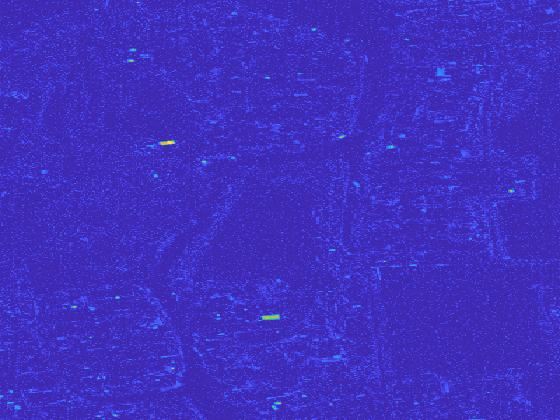}
\\
 {(i)}  &  {(j)} &  {(k)} &(l)  &(m)  &(n) &(o) &(p)\\
 \toprule
\end{tabular}
\caption{
\textcolor[rgb]{0.00,0.00,0.00}{Anomaly detection map of various  HAD  methods on large-scale
 HSI datasets: Beach, AVON,  Qingpu-I
(from top to bottom).
(a) Ground-Truth.
(b)   RX.
(c)  CRD.
(d) GAED.
(e)  PDBSNet.
(f) LRASR.
(g)  GTVLRR.
(h) PTA.
(i)   PCA-TLRSR.
(j)  T-CTV.
(k)  GCS.
(l)  MERAETV
(m) TRDFTVAD.
(n)  GNLTR.
 (o) GNBRL. 
 (p) Proposed.
}}
\vspace{-0.5cm}
\label{fig_visual_dectionmap111}
\end{figure*}

\begin{figure*}[!htbp]
\renewcommand{\arraystretch}{0.8}
\setlength\tabcolsep{0.00pt}
\centering
\begin{tabular}{c|c|c |c}
\centering

\includegraphics[width=1.8758in, height=1.367864in]{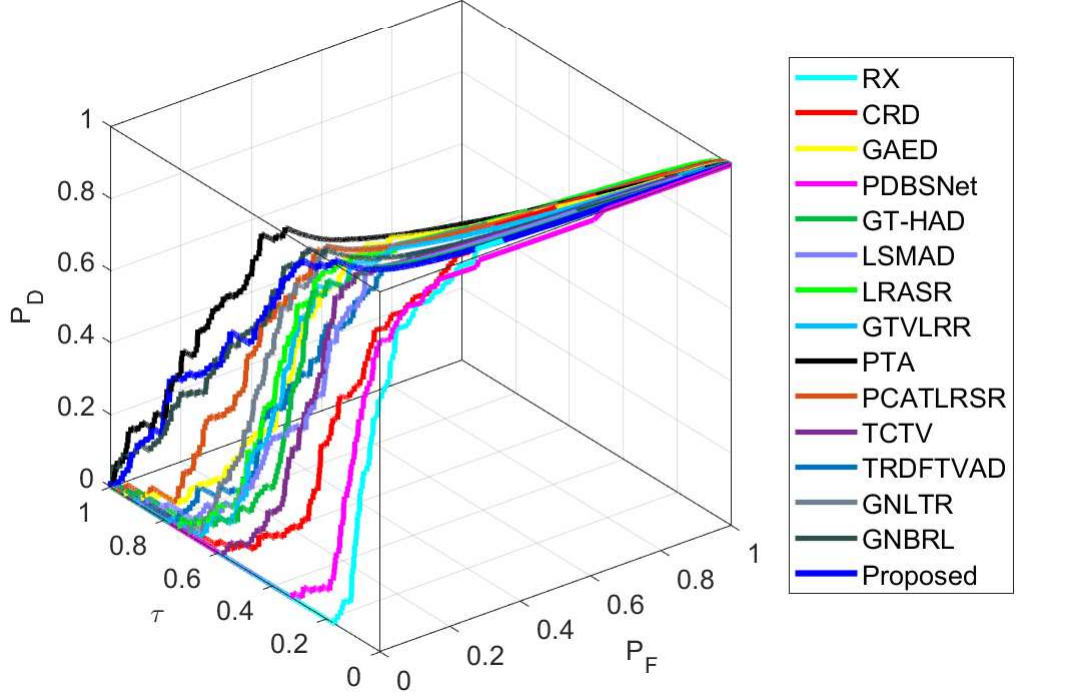}&
\includegraphics[width=1.758in, height=1.367864in]{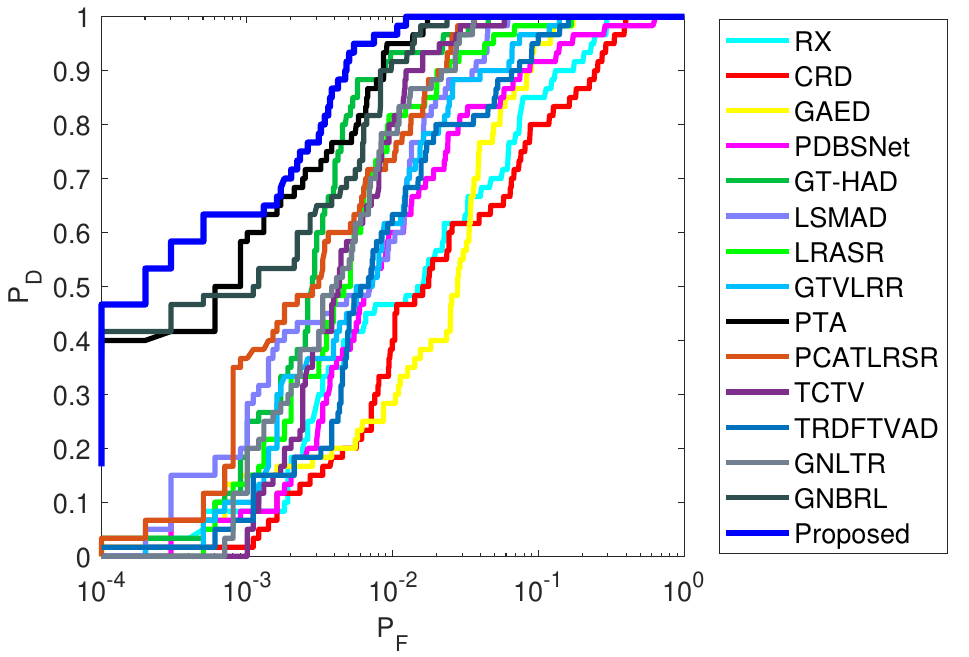}&
\includegraphics[width=1.758in, height=1.367864in]{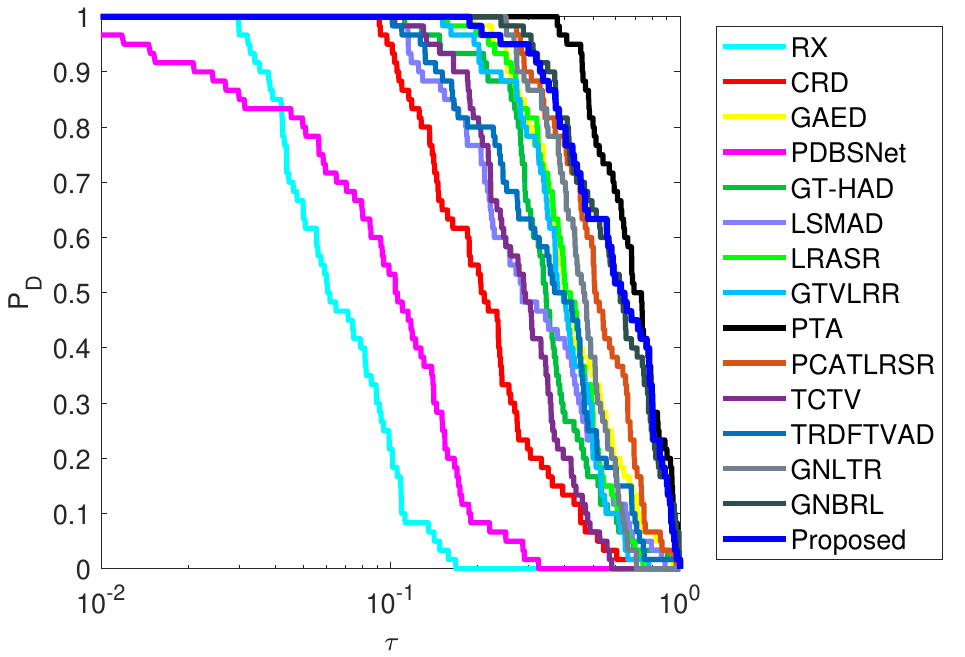}&
\includegraphics[width=1.758in, height=1.367864in]{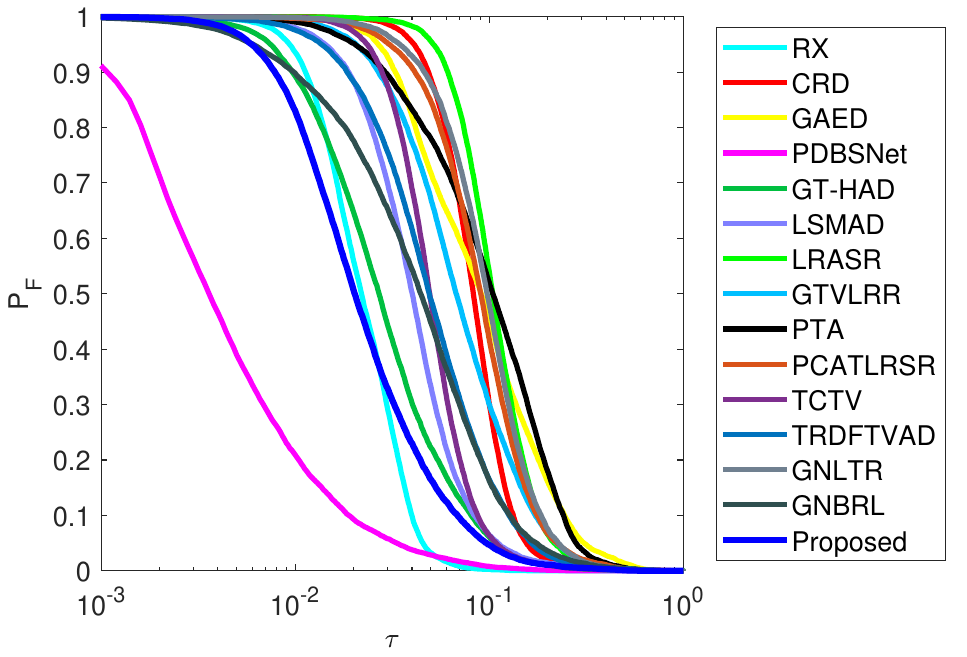}

\\

\includegraphics[width=1.8758in, height=1.367864in]{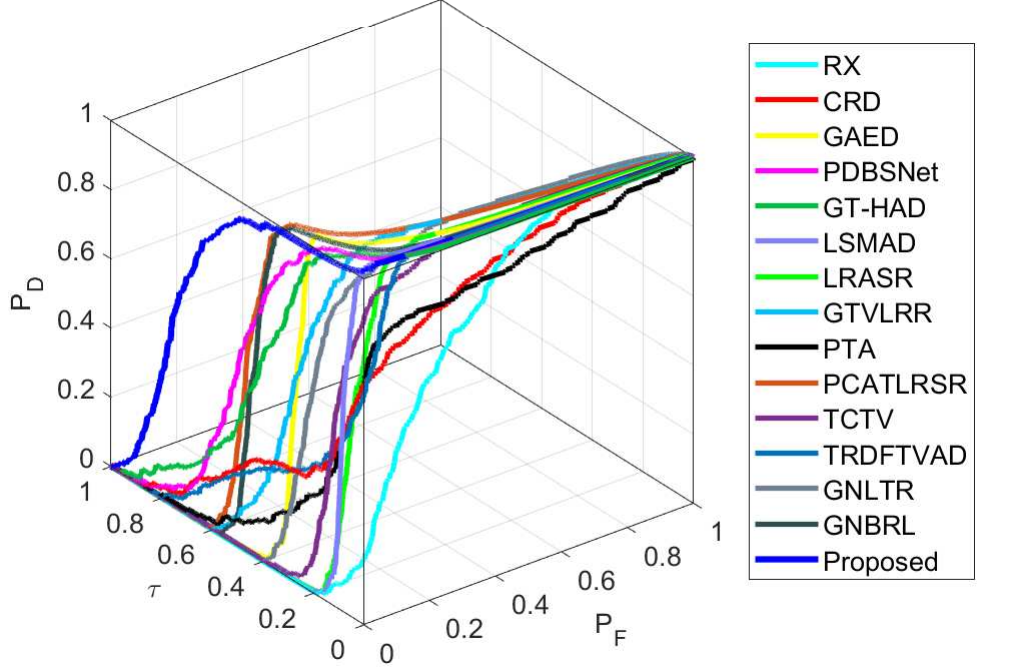}&
\includegraphics[width=1.758in, height=1.367864in]{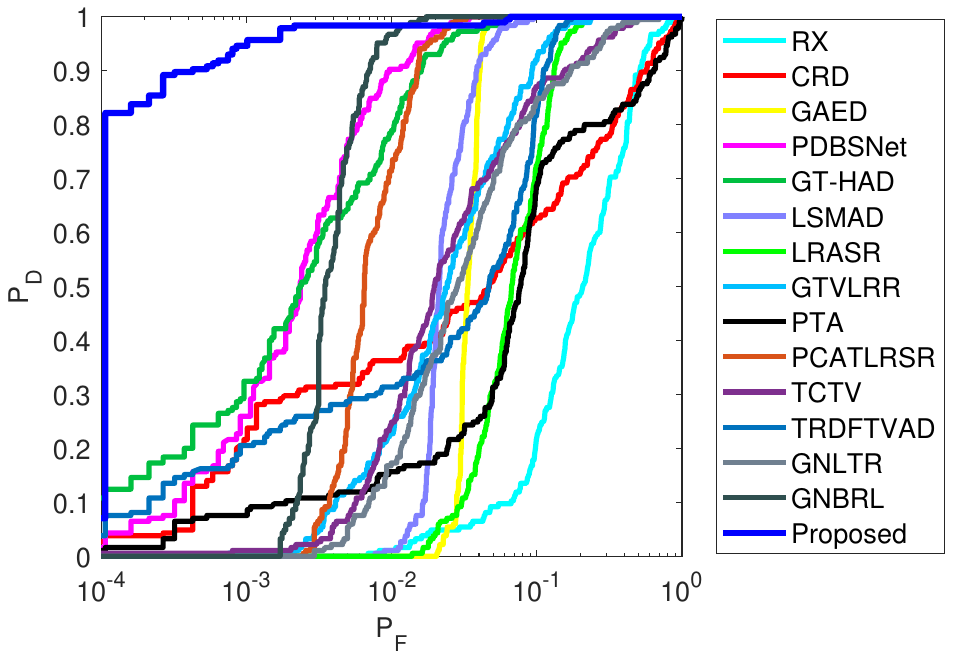}&
\includegraphics[width=1.758in, height=1.367864in]{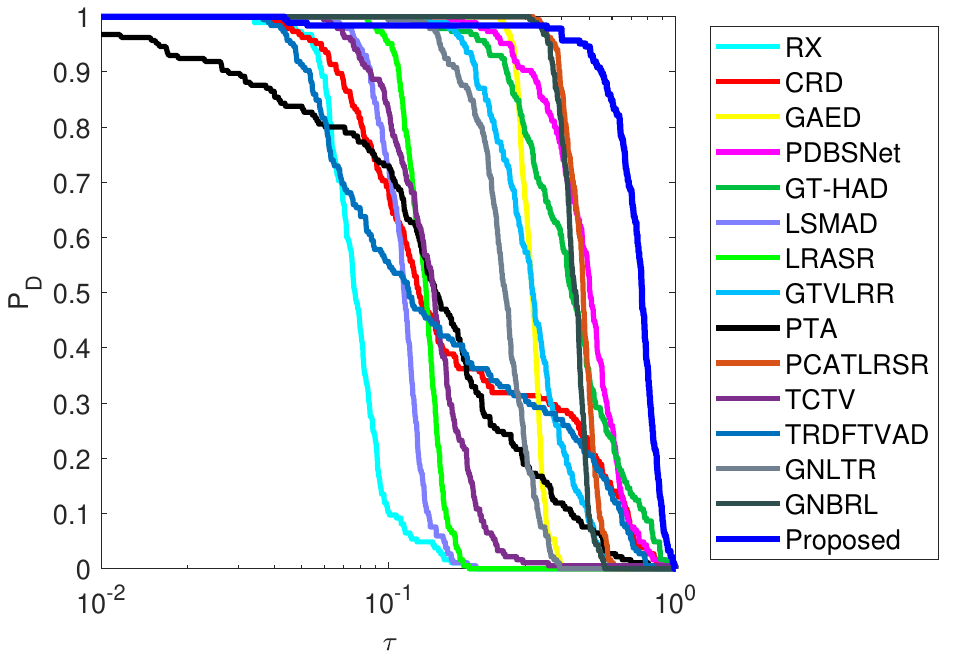}&
\includegraphics[width=1.758in, height=1.367864in]{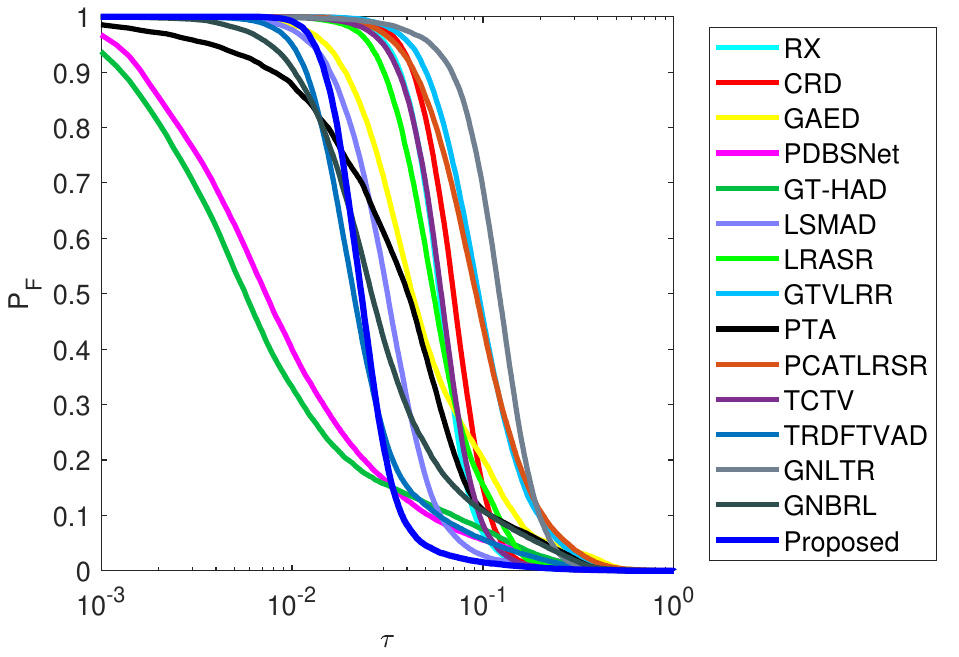}
\\

\includegraphics[width=1.8758in, height=1.367864in]{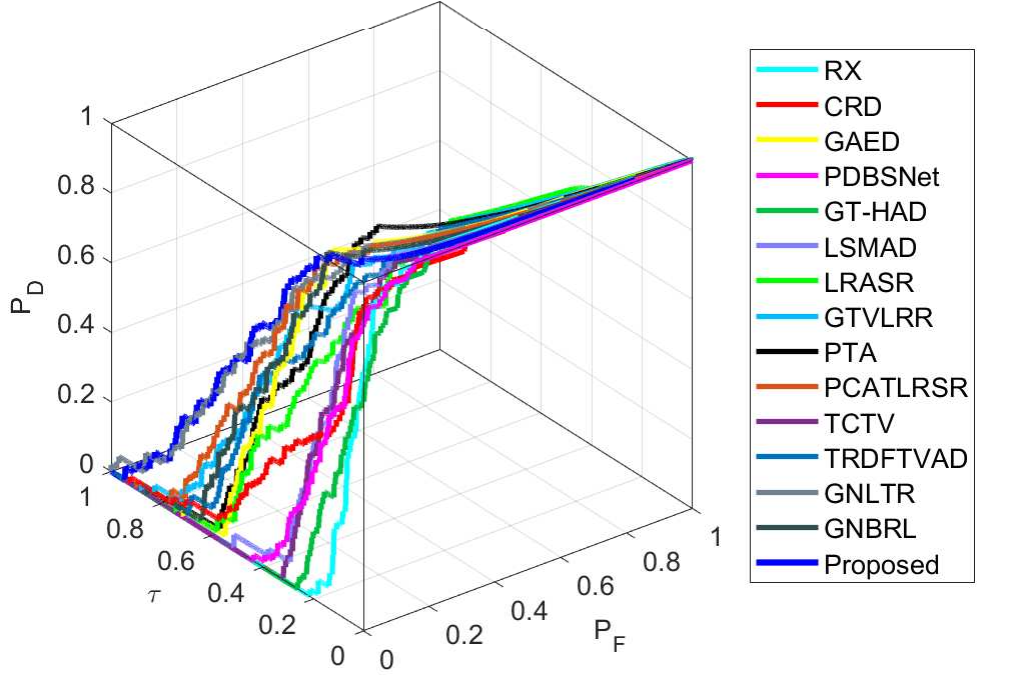}&
\includegraphics[width=1.758in, height=1.367864in]{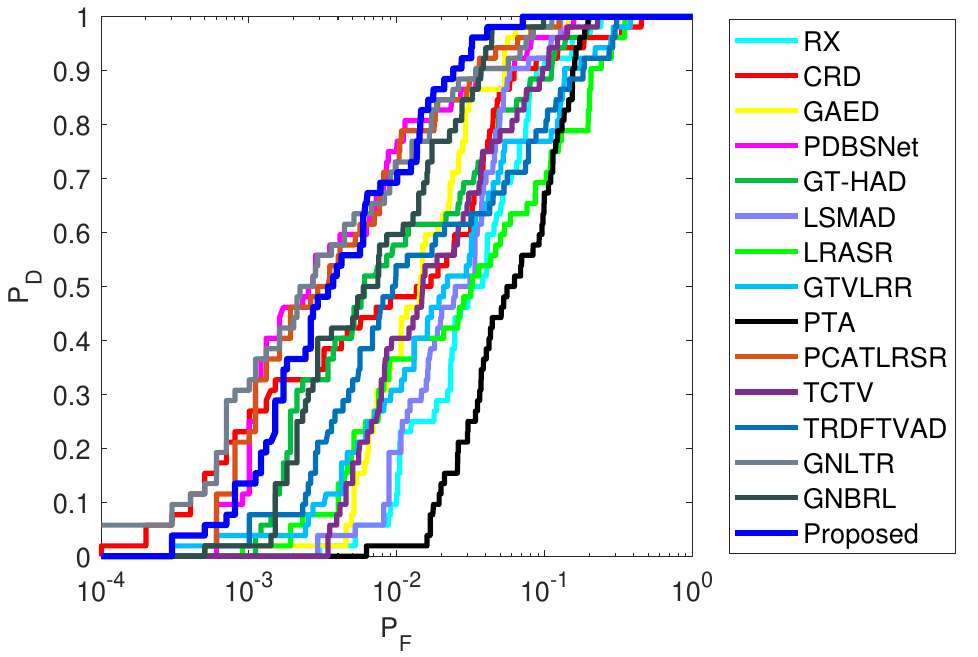}&
\includegraphics[width=1.758in, height=1.367864in]{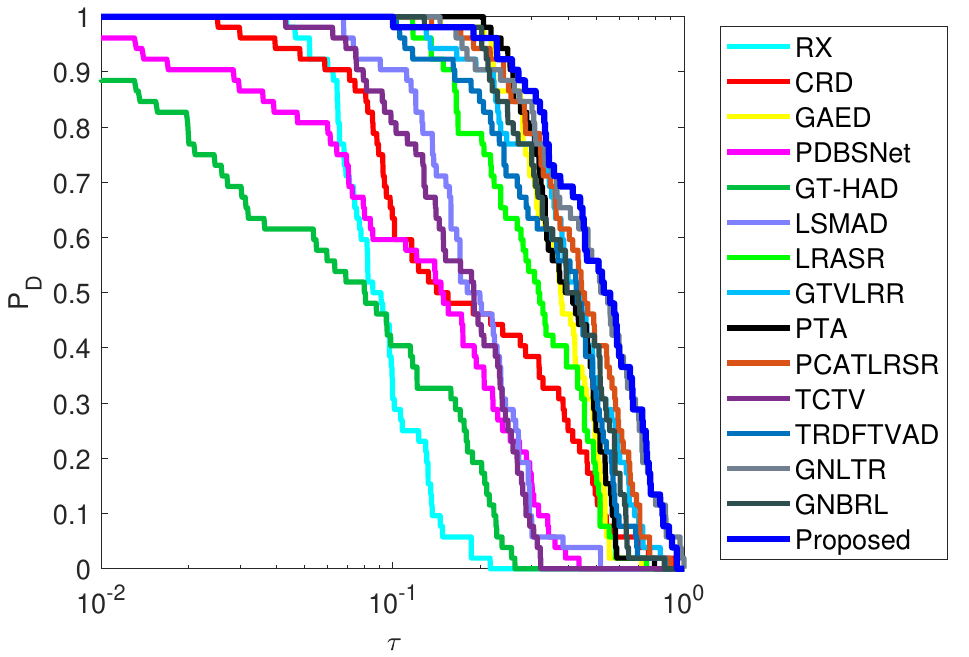}&
\includegraphics[width=1.758in, height=1.367864in]{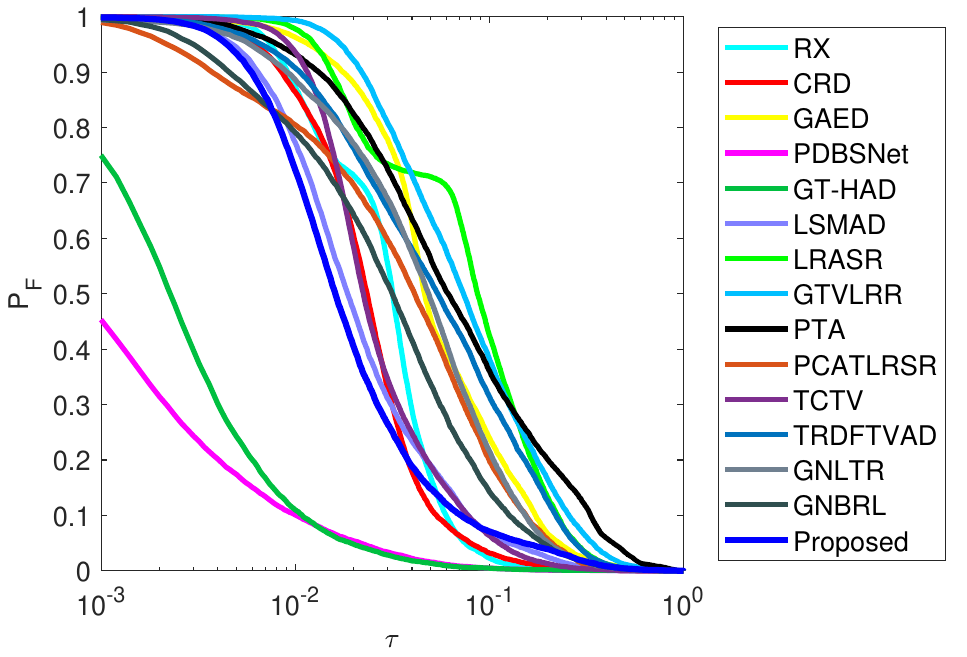} \\

\footnotesize{(a)
 ROC curve of
$({\operatorname{P}}_{\mathnormal{D}}, {\operatorname{P}}_{\mathnormal{F}}, \tau )$
}
 &\footnotesize{(b)
  ROC curves of $({\operatorname{P}}_{\mathnormal{D}}  , {\operatorname{P}}_{\mathnormal{F}})$

 }& \footnotesize{(c) 
 ROC curves of $({\operatorname{P}}_{\mathnormal{D}}, \tau )$
 }&
 \footnotesize{(d) 
 ROC curves of
$({\operatorname{P}}_{\mathnormal{F}}, \tau )$.
 }
\\

\end{tabular}
\caption{
\textcolor[rgb]{0.00,0.00,0.00}{$3$-D and $2$-D ROC curves' performance comparison of different HAD approaches on 
 HSI datasets:
   Airport-4, Salinas 
   and Urban-3 
    (from top to bottom).
}}
\vspace{-0.62cm}
\label{fig_2d3droc}
\end{figure*}

\vspace{-0.4cm}

\subsection{\textbf{Comparative Experiments}}

\subsubsection{\textbf{Comparison Methods}}

To showcase 
 the effectiveness of the proposed HAD algorithm,
\textcolor[rgb]{0.00,0.00,0.00}{some baselines  are introduced as compared detectors in our experiments,
which 
 contain
 one  classical 
 statistics-based
 method:   RX \cite{reed1990adaptive},
 one collaborative representation based  method:  CRD \cite{li2014collaborative},
three deep learning-based  methods: 
 %
GAED \cite{xiang2022hyperspectral}, PDBSNet \cite{wang2023pdbsnet}, GT-HAD \cite{GTHAD2025},
three matrix-based methods: LSMAD \cite{zhang2015low}, LRASR \cite{xu2015anomaly}  and GTVLRR \cite{cheng2019graph},
eight
 tensor-based methods:
 PTA \cite{li2020prior}, 
 PCA-TLRSR \cite{wang2022learning1},  T-CTV \cite{wang2023guaranteed},
  MERAETC \cite{xiao2024hyperspectral33},  GCS \cite{shang2023hyperspectral},
 TRDFTVAD \cite{feng2023hyperspectral}, GNLTR\cite{qin2023generalized1},   and GNBRL \cite{yu2024generalized}.}

\subsubsection{\textbf{Experimental    Setings}}

In our experiments, each  raw HSI data is  conducted with   band-by-band normalization operation.
 \textcolor[rgb]{0.00,0.00,0.00}{The order of the tensor ring is consistent with the dimensions of the raw hyperspectral data.}
For 
the proposed HAD method, the  nonconvex combinations $\Phi(\cdot)$+$\psi(\cdot)$ are set to be the same
($\Phi$=$ \psi$=Capped-Log),  
with the exception of Hyperion ($\Phi$=$ \psi$=MCP), San-Diego ($\Phi$=$ \psi$=Capped-Lp), and Urban-3 ($\Phi$=$ \psi$=MCP).
For the  Pavia, Airport-4, and Urban-5 datasets,
the TR rank 
$(r_1,r_2,r_3)$ is set to be
 $(6,6,6)$. For the other  
 remaining datasets, we assign 
$ (r_1,r_2,r_3)=  (6, 16, 6) $.
 \textcolor[rgb]{0.00,0.00,0.00}{Besides,
the trade-off parameters $\alpha$ and $\beta$
are both searched from 
$\{10^{-6}, 5 \cdot 10^{-6}, 10^{-5}, 5 \cdot 10^{-5},  10^{-4}, 5 \cdot 10^{-4}, 10^{-3}, 5 \cdot 10^{-3},
10^{-2}, 5 \cdot 10^{-2},
0.1, 0.5,  1, 5, 10\}$,
set
$\mu^{\{0\}} =10^{-3}$,
$\mu^{\max}=10^{10}$,
$\vartheta=1.1$,
$\varpi=10^{-5}$,
$\nu_{\max}=500$,
 $\mathfrak{L}=\operatorname{FFT}$.
 \textcolor[rgb]{0.00,0.00,0.00}{For large-scale datasets, the TR rank 
$(r_1,r_2,r_3)$    are chosen to be  $\{  (10,200,10), (10,300,10), (10,500,10), (15,300,15)\}$.}}
To maintain the fairness of our comparative experiments, optimal parameters 
 of 
 competing methods
are given  according to the relevant references.
 Please refer  the supplementary materials for more details.

\subsubsection{\textbf{Evaluation Metrics}}
In our experiments, four commonly used evaluation metrics are adopted, namely, anomaly detection map,
 $3$-D receiver operating characteristic (ROC) curve \cite{
 chang2022comprehensive}, area under the ROC curve (AUC),
 and box-whisker map \cite{williamson1989box}.
 $3$-D ROC describes the relationship between detection probability ${\operatorname{P}}_{\mathnormal{D}}$,  false alarm probability ${\operatorname{P}}_{\mathnormal{F}}$,
  and detection threshold $\tau$.
  Moreover, three $2$-D  ROC curves can be decomposed from the $3$-D ROC curve,
namely,  $({\operatorname{P}}_{\mathnormal{D}}, {\operatorname{P}}_{\mathnormal{F}})$, $({\operatorname{P}}_{\mathnormal{D}}, \tau )$, and $({\operatorname{P}}_{\mathnormal{F}}, \tau )$.
Correspondingly, we can get 
$\operatorname{AUC} _{({\operatorname{P}}_{\mathnormal{D}}, {\operatorname{P}}_{\mathnormal{F}})}$, $\operatorname{AUC} _{({\operatorname{P}}_{\mathnormal{D}}, \tau)}$, and $\operatorname{AUC} _{({\operatorname{P}}_{\mathnormal{F}}, \tau)}$.
By combining these three evaluation indicators with each other, the following 
comprehensive evaluation indicators can be obtained and defined as follows:
\begin{eqnarray*}
\begin{cases}
\operatorname{AUC} _{(\operatorname{ODP})}=
\operatorname{AUC} _{({\operatorname{P}}_{\mathnormal{D}}, {\operatorname{P}}_{\mathnormal{F}})}  +
\operatorname{AUC} _{({\operatorname{P}}_{\mathnormal{D}}, \tau)} -
\operatorname{AUC} _{({\operatorname{P}}_{\mathnormal{F}}, \tau)},
\\
\operatorname{AUC} _{(\operatorname{SNPR})} =\operatorname{AUC} _{({\operatorname{P}}_{\mathnormal{D}}, \tau)} /
\operatorname{AUC} _{({\operatorname{P}}_{\mathnormal{F}}, \tau)},
\\
\operatorname{AUC} _{(\operatorname{TDBS})} = \operatorname{AUC} _{({\operatorname{P}}_{\mathnormal{D}}, \tau)} -
\operatorname{AUC} _{({\operatorname{P}}_{\mathnormal{F}}, \tau)},
\end{cases}
\end{eqnarray*}
where 
$\operatorname{AUC} _{(\operatorname{SNPR})}$ evaluates the signal-to-noise ratio of the detector,
$\operatorname{AUC} _{(\operatorname{TDBS})}$ evaluates the comprehensive target detection and background suppression
capabilities of the detector, and
$\operatorname{AUC} _{(\operatorname{ODP})}$ evaluates the overall detection probability of the detector.
In theory, an excellent detector should have very small
$\operatorname{AUC} _{({\operatorname{P}}_{\mathnormal{F}}, \tau)}$ and very large other AUC values.
 The box-whisker map is primarily utilized for assessing the discriminability between anomalies and background. The greater the gap
between the anomaly box and the background box, the enhanced separation and detection effect of the detector can
be achieved.

\subsubsection{\textbf{Experimental Results and Analysis}} 


\textcolor[rgb]{0.00,0.00,0.00}{The AUC values obtained by  various  HAD detectors on 
extensive HSI datasets are reported 
in Tables \ref{hsi_various_auc}, \ref{hsi_various_auc111117777}.}
\textcolor[rgb]{0.00,0.00,0.00}{Regarding the overall detection performance
index $\operatorname{AUC} _{(\operatorname{ODP})}$, 
we observed 
that the proposed algorithm achieves the optimal $\operatorname{AUC} _{(\operatorname{ODP})}$ on the
Salinas, Hyperion, HYDICE, Airport-4, Urban-3,  Urban-4, Qingpu-I, and AVON  datasets
but achieves suboptimal $\operatorname{AUC} _{(\operatorname{ODP})}$  on the Pavia, San-Diego, Urban-5, Beach-4,
Synthetic, and  large-scale Beach
datasets.}
\textcolor[rgb]{0.00,0.00,0.00}{Regarding the 
 metric $\operatorname{AUC} _{(\operatorname{TDBS})}$, 
the proposed  algorithm outperforms all  competitors  
on the Salinas, Hyperion, HYDICE, Airport-4, Urban-3,  Urban-4, Qingpu-I, and AVON datasets,
while yielding results that are on a par with those of other algorithms on the remaining datasets.}
This indicates that our detector possesses excellent comprehensive capabilities in both target detection and background suppression.
 When it comes to the 
 evaluation indicator 
 $\operatorname{AUC} _{(\operatorname{SNPR})}$,
the proposed method reaches the optimal value on the Salinas,   Airport-4, San-Diego, and  Qingpu-I 
datasets.
 Although the PDBSNet and   GT-HAD methods 
 achieve notably high $\operatorname{AUC} _{(\operatorname{SNPR})}$ 
  values on the remaining datasets,
 their performance in terms of other evaluation metrics (e.g., $\operatorname{AUC} _{(\operatorname{ODP})}$, $\operatorname{AUC} _{(\operatorname{TDBS})}$
 $({\operatorname{P}}_{\mathnormal{D}}, \tau )$)  is suboptimal. %
These  quantitative
results effectively prove the superiority of the
proposed  algorithm compared to other competing algorithms.


\textcolor[rgb]{0.00,0.00,0.00}{Figures \ref{fig_visual_dectionmap} and \ref{fig_visual_dectionmap111} display the two-dimensional anomaly detection maps of various  HAD  methods on 
extensive
 HSI datasets.}  Compared with  other detectors,
 the proposed method strikes a superior balance between accurate anomaly recognition and effective
background removal.
The detection maps from the Salinas and San-Diego datasets  are  utilized 
as illustrative examples to support our conclusion.
For the Salinas's detection maps, 
we find that the PTA detector  fails to detect the abnormal target.
GTVLRR and  PCA-TLRSR misidentify 
more backgrounds as abnormal targets, resulting in poor  visual effect.
The LRASR 
and LSMAD  
detectors have better background suppression, but abnormal
targets can hardly be detected. 
The GAED,
PDBSNet, GT-HAD,
 TRDFTVAD, and GNBRL detectors can distinguish 
the approximate distribution of abnormal targets, yet they suffer from 
a certain degree of background contamination.
For the San-Diego's detection maps, 
we find that RX and CRD completely are unable to 
detect  three airplanes. The LSMAD,
LRASR and GTVLRR can detect three planes but are not in good 
 shape.
The PTA, TRDFTVAD and GNBRL %
can brightly highlight anomalous targets, but it does not suppress the background well.
In stark contrast, GAED, PDBSNet, GT-HAD, and  T-CTV    perform the inverse. Overall,
our method effectively suppressed the background while detecting clear anomalous targets.

The box-whisker maps of different detectors on eight datasets are displayed in Figure \ref{fig_sep_box}.
This figure 
%
%
shows that  
the proposed method 
 has the widest gap between the anomaly box and the background box  in most cases.
\textcolor[rgb]{0.00,0.00,0.00}{Besides,
%
the abnormal box
of our HAD method is not always at the highest position for the
different HSI datasets, but the position of the background box
is almost always at the lowest position.
%
This imply that
the proposed method has a good 
inhibitory effect on the background, which
is consistent with the  above visual 
results.}  
The 3-D ROC curves and 2-D ROC curves of each algorithm
on the eight datasets are shown in Figure \ref{fig_2d3droc}.
As shown in
 Figure \ref{fig_2d3droc} (b), 
 the proposed algorithm  achieves a higher detection rate
than other algorithms under different false alarm rates in most cases.
 As can be observed from Figure \ref{fig_2d3droc} (c),  the proposed  algorithm beats 
most of its competitors in the ROC curves on all HSI 
datasets, exhibiting
better target detection performance, especially on the Salinas and Urban-3 datasets.
As can be seen from Figure \ref{fig_2d3droc} (d), the ROC
curves of the proposed  algorithm are almost all in the lower left corner on all  datasets, which  indicates that
our detector  has a very low false alarm rate under
different threshold $\tau$'s.
Overall, our HAD algorithm ranks third after
PDBSNet and  GT-HAD in terms of 
ROC
curves of
$({\operatorname{P}}_{\mathnormal{F}}, \tau )$.

\vspace{-0.3cm}
\subsection{\textcolor[rgb]{0.00,0.00,0.00}{\textbf{Parameter  Analysis}}} \label{parameter-trtr}

The proposed  HAD  algorithm involves five  parameters that need to be tuned, namely, the TR rank $(r_1,r_2,r_3)$, the
nonconvex functions $\Phi(\cdot)$, $\psi(\cdot)$,  and
two  tradeoff parameters $\alpha$ and $\beta$. In our experiments,
we tune a certain parameter by fixing the  others. 
For brevity, we set the nonconvex functions $\Phi(\cdot)$ and $\psi(\cdot)$ to be the same, 
$\Phi, \psi \in \{\operatorname{L1},$
$\operatorname{Lp},$
$\operatorname{Log},$
$\operatorname{MCP},$
$\operatorname{Capped-L1},$
$\operatorname{Capped-Lp},$
$\operatorname{Capped-Log},$
$\operatorname{Capped-MCP} \}$,  $\mathfrak{L}=\operatorname{FFT}$.
The 
 value
ranges for $\alpha$ and   $\beta$  are  selected from
$\{10^{-6}, 5 \cdot 10^{-6}, 10^{-5}, 5 \cdot 10^{-5},  10^{-4}, 5 \cdot 10^{-4}, 10^{-3}, 5 \cdot 10^{-3},
10^{-2}, 5 \cdot 10^{-2},
0.1, 0.5,  1, 5, 10\}$.

\subsubsection{\textbf{Trade-off Parameters $\alpha$ and   $\beta$}}
\begin{figure*}[!htbp]
\renewcommand{\arraystretch}{0.128}
\setlength\tabcolsep{0.0pt}
\centering
\begin{tabular}{ccc c}
\centering


\includegraphics[width=1.84in, height=1.39843in]{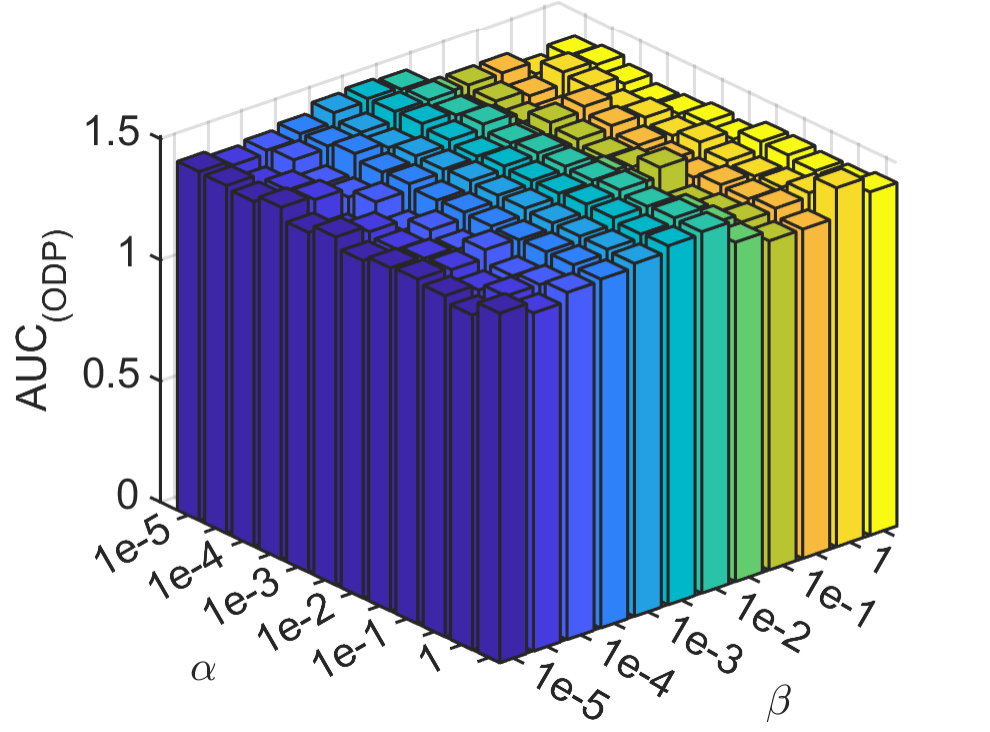}&
\includegraphics[width=1.775in, height=1.39843in]{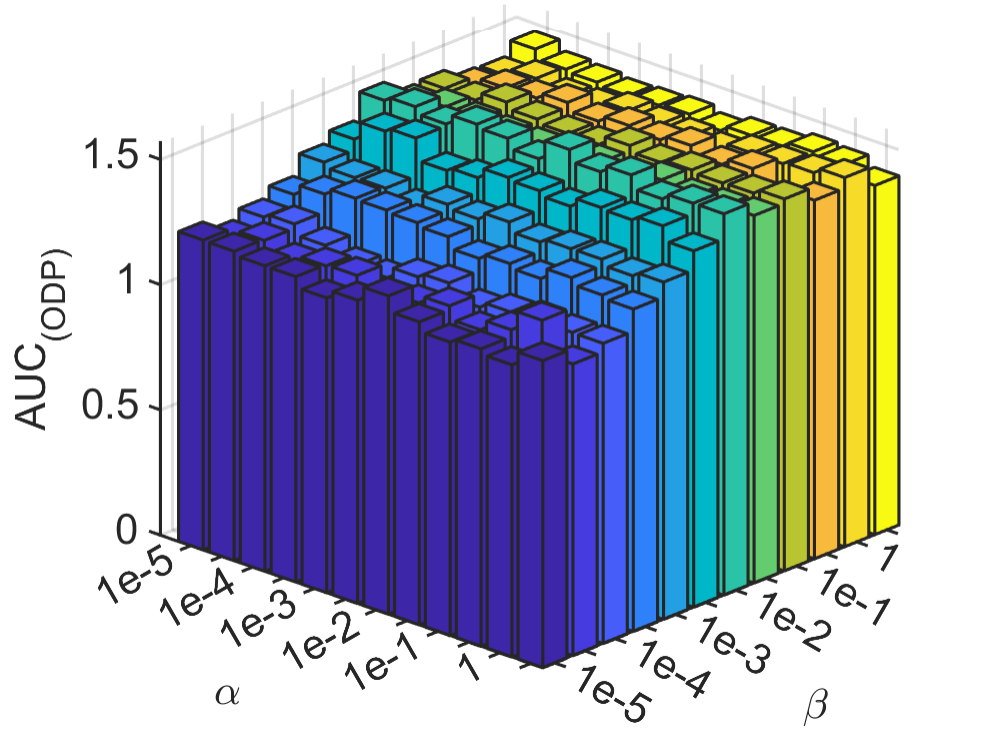}&
\includegraphics[width=1.775in, height=1.39843in]{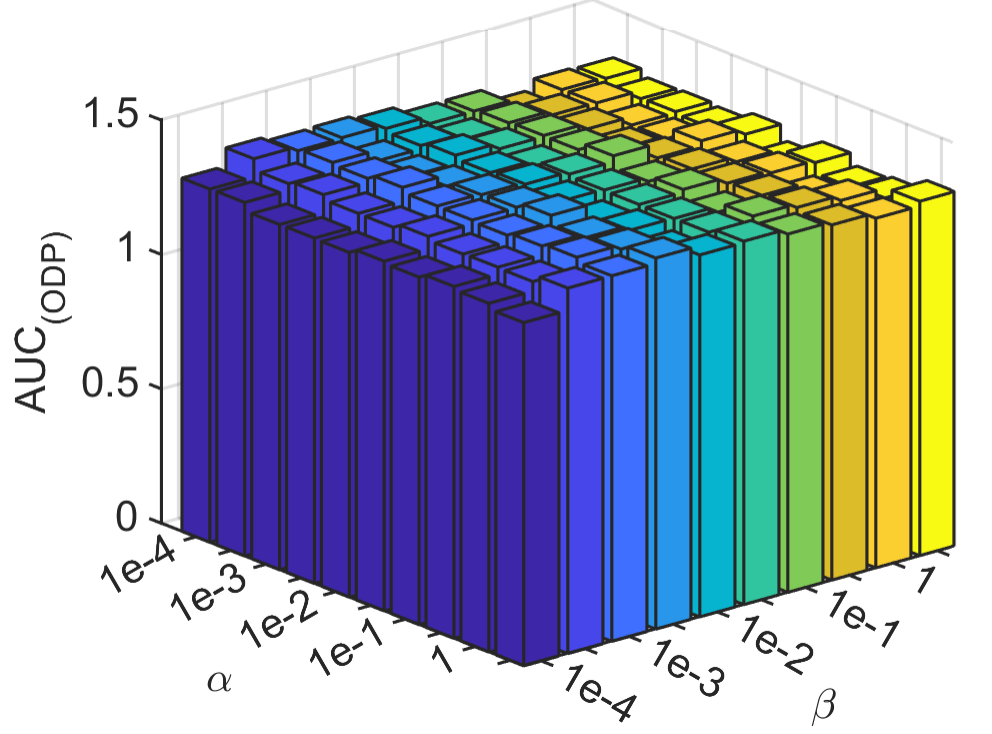}&
\includegraphics[width=1.84in, height=1.39843in]{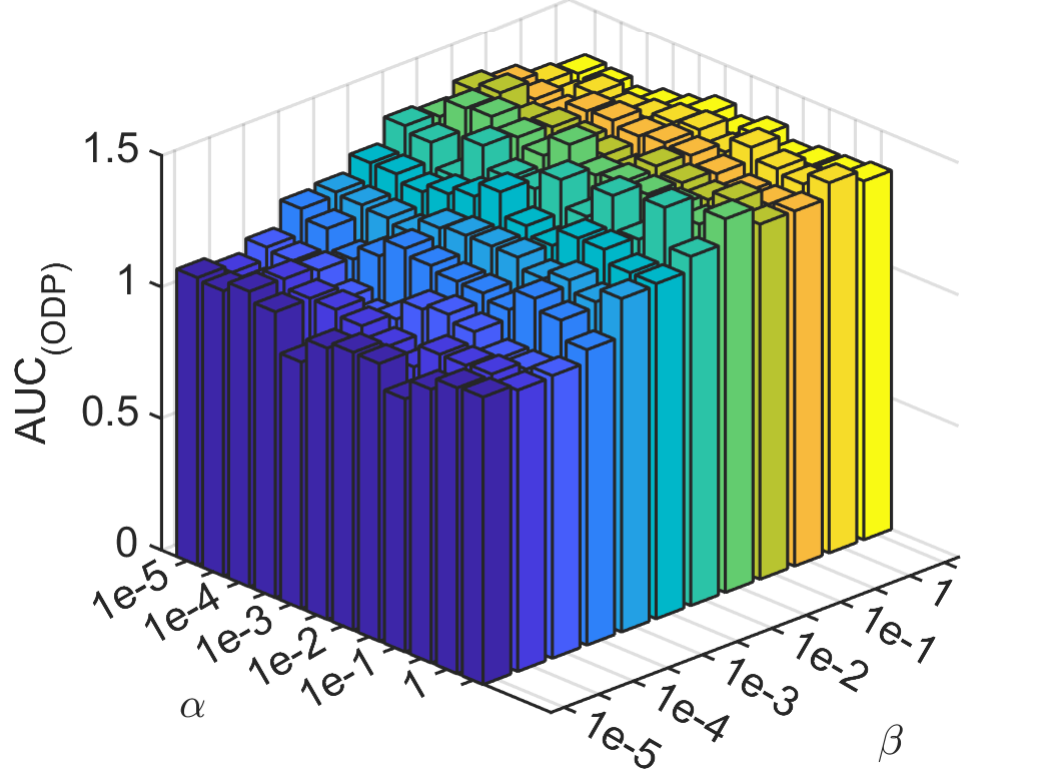}
\\
 \tabincell{c}{
\footnotesize{ {{(a)}} Beach-3}
}
 & 
\tabincell{c}{
\footnotesize{ {{(b)}}   HYDICE}
}
 & \tabincell{c}{
\footnotesize{ {{(c)}} Hyperion}
}
  & \tabincell{c}{
\footnotesize{ {{(d)}} Urban-3}
}
\\
\\
\includegraphics[width=1.84in, height=1.39843in]{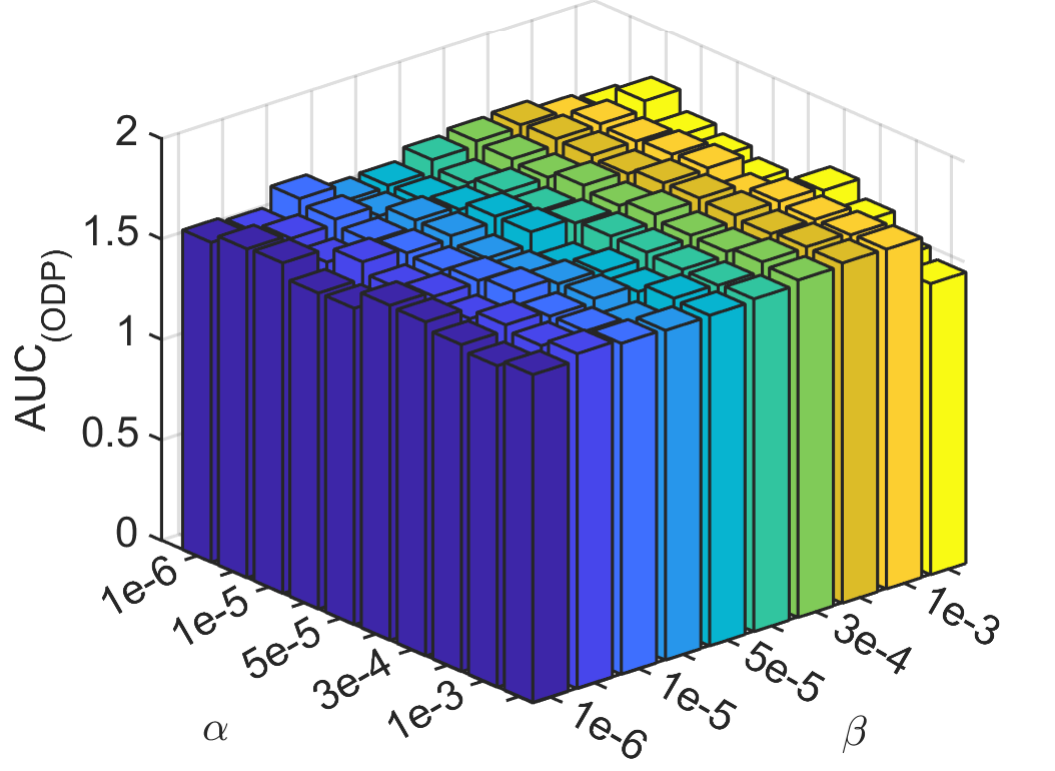}
&
\includegraphics[width=1.775in, height=1.39843in]{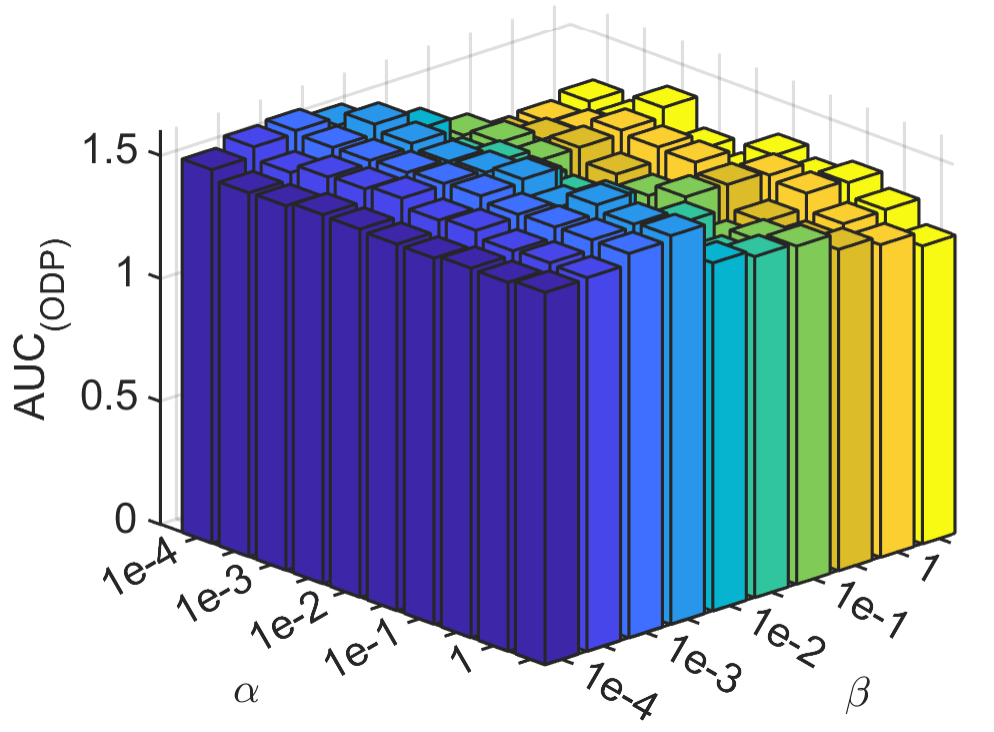}&
\includegraphics[width=1.775in, height=1.39843in]{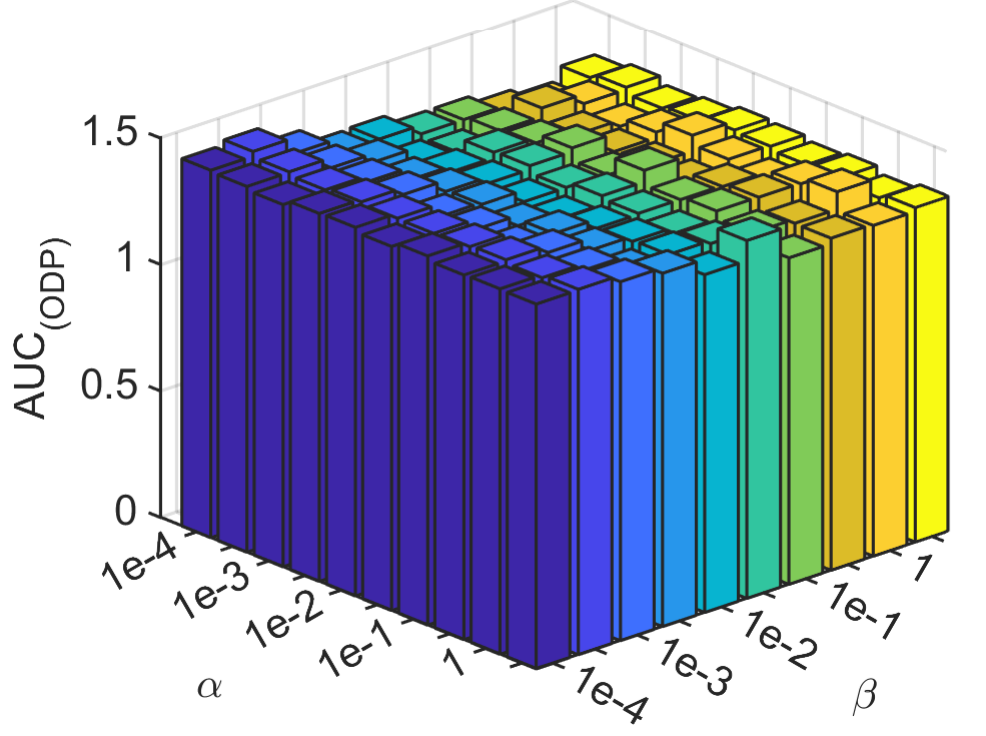}&
\includegraphics[width=1.84in, height=1.39843in]{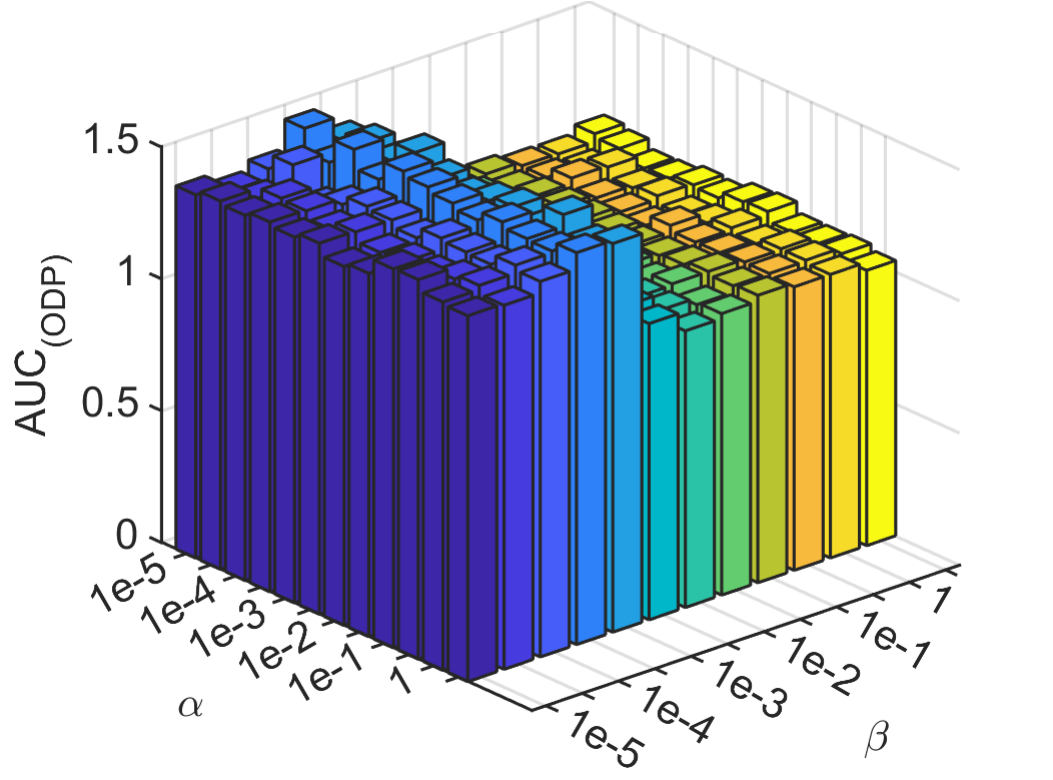}\\


\tabincell{c}{
\footnotesize{ {{(e)}} Salinas}
} &
\tabincell{c}{
\footnotesize{ {{(f)}}   Airport-4}
}  & 
 \tabincell{c}{
\footnotesize{ {{(g)}} Pavia}
}
& 
\tabincell{c}{
\footnotesize{ {{(h)}}  San-Diego}
}

%
%
%
%
%

\end{tabular}
\caption{
 \textcolor[rgb]{0.00,0.00,0.00}{Sensitivity analysis of the 
 trade-off parameters $\alpha$ and $\beta$.
For each HSI data, the TR rank is fixed
as $(r_1,r_2,r_3) = (6, 6, 6)$,
 %
and the nonconvex combination
$ \Phi$+$\psi$ is  set to be the same, i.e., $ \Phi(\cdot) = \psi(\cdot)$=Capped-Log. 
}}
\vspace{-0.35cm}
\label{fig-beta-learning1}
\end{figure*}

This experiment 
focuses on investigating the impact of the 
different trade-off parameters $\alpha$ and   $\beta$ upon
 anomaly detection performance in the context of given TR rank $(r_1,r_2, r_3)$ and nonconvex combinations
 $\Phi(\cdot)$+$\psi(\cdot)$.
 In this part,
 The  TR rank is empirically fixed as
 $(6,6,6)$.
The relevant experimental results are displayed  in Figure \ref{fig-beta-learning1}. 
Due to space constraints, we only present the results achieved 
by 
the Capped-Log function.
We find that
for Salinas, Airport-4, Hyperion, Pavia, and Beach-3, 
 the 
detection performance demonstrates strong robustness to the 
parameters $\alpha$ and   $\beta$, 
 with $\operatorname{AUC} _{(\operatorname{ODP})}$ 
 values experiencing only slight fluctuations as $\alpha$ and   $\beta$ vary.
%
 The robustness of the 
 detection performance in San-Diego varies moderately with
the parameters $\alpha$ and $\beta$  and the $\operatorname{AUC} _{(\operatorname{ODP})}$  value  generally shows
a 
decreasing trend with the increase of $\beta$. The changing trend
of $\operatorname{AUC} _{(\operatorname{ODP})}$ with the rise of $\alpha$ is not apparent.
For  HYDICE and
 Urban-3,
as the parameter $\beta$ increases, 
the $\operatorname{AUC} _{(\operatorname{ODP})}$ value of the proposed HAD algorithm
begins to stabilize.
Overall, the detection performance is insensitive to 
the  trade-off
parameters $\alpha$ and   $\beta$.


\subsubsection{\textbf{Nonconvex Functions $ \Phi(\cdot)$ and   $\psi(\cdot)$}}

In this experiment, the performance of our proposed HAD method  under different nonconvex functions
is verified. 
Based on the 
parameters $\alpha$ and $\beta$ learned in the previous experiment,  Table \ref{odp_L1L21333}
shows the  AUC    values
 obtained by   various nonconvex combinations $\Phi(\cdot)$+$\psi(\cdot)$ for  different
  HSI  datasets.
We pursue a general surrogate for approximating low rank and sparsity, which can be flexibly chosen
according to different scenarios.
 We can observe that the convex function L1
performs worse than all nonconvex functions in terms of AUC values. We can also observe
that the non-Capped-type functions generally perform worse than the corresponding Capped-type
functions in terms of AUC values.
 Among the Capped-type functions, Capped-Log  achieve
relatively good  AUC values in many scenarios. 

\subsubsection{\textbf{TR  Rank $[r_1, r_2, r_3]$}}

The TR rank contains three parameters $r = [r_1,r_2,r_3]$, and the estimation of TR rank is still
an open problem.
In this experiment, we find that the TR rank is usually needed into a pattern that sets the two sides of the rank smaller and the middle larger.
 To simplify the complexity of the parameters analysis,  we choose $r_1 = r_3$ to maintain the consistency of the spatial TR core tensor.
The rank components $r_1$ and $r_3$ undergo adjustments within the range of 2 to 20, while $r_2$ is varied within
the interval of 2 to 42.
Figure \ref{fig_rank-learning} illustrates the trend of the $\operatorname{AUC} _{(\operatorname{ODP})}$  value
 as it varies with 
 TR rank $(r_1, r_2, r_3)$.
From this figure, we find that
under suitable nonconvex functions  and trade-off parameters, a relatively low TR rank can yield satisfactory 
detection performance. On the whole,
the anomaly detection performance of the
proposed algorithm is robust to the combination of $r_1$, $r_2$ and $r_3$.



\begin{table*}[tp]
\renewcommand{\arraystretch}{0.1}
\setlength\tabcolsep{5pt}
  \centering
  \caption{The  
AUC 
  values obtained by   various nonconvex combinations
  $\Phi(\cdot)$+$\psi(\cdot)$
   for  different
  HSI  datasets.}
  \label{odp_L1L21333}
  \scriptsize
  \begin{threeparttable}
    \begin{tabular}{c   ccc ccc ccc ccc }
    \hline
    \tabincell{c}{Nonconvex\\ function}
   &\multicolumn{1}{c}{AUC-Metrics}
   &\multicolumn{1}{c}{Salinas}&\multicolumn{1}{c}{Pavia}&\multicolumn{1}{c}{Hyperion}
    &\multicolumn{1}{c}{HYDICE}& \multicolumn{1}{c}{San-Diego}&\multicolumn{1}{c}{Airport-4}&
    \multicolumn{1}{c}{Beach-3}&\multicolumn{1}{c}{Beach-4} &\multicolumn{1}{c}{Urban-3} &
    \multicolumn{1}{c}{\text{Urban-4}}&\multicolumn{1}{c}{Urban-5}
    \cr

    \hline
    \hline

   \multirow{4}{*}
   {L1}& $\operatorname{AUC} _{({\operatorname{P}}_{\mathnormal{D}}, {\operatorname{P}}_{\mathnormal{F}})}$
   & 0.9986 & 0.9975 &0.9971 &0.9949&0.9895&0.9956&0.9980&0.9809&0.9809&0.9954&0.9638
    \cr

   \qquad	&
  $\operatorname{AUC} _{(\operatorname{ODP})}$
  & 1.6073 &1.3734 &1.3081&1.4140&1.4711&1.5619&1.4455&1.2506&1.2828&1.0893&1.2838
\cr

 \qquad	&
  $\operatorname{AUC} _{(\operatorname{SNPR})}$
  & 17.0345& 24.6140 &13.1800&17.8686& 11.1167&10.4967&15.3768&16.2700&7.3307&20.0167&5.6482
  \cr

  \qquad	&
  $\operatorname{AUC} _{(\operatorname{TDBS})}$
  & 0.6087  &  0.3759&    0.3110 &   0.4191 &   0.4816   & 0.5663 &   0.4475  &  0.2697 &   0.3019 &   0.0939 &   0.3200
  \cr

   \hline

  \multirow{4}{*}{Lp}& $\operatorname{AUC} _{({\operatorname{P}}_{\mathnormal{D}}, {\operatorname{P}}_{\mathnormal{F}})}$
  & 0.9986  & 0.9983 &0.9972  & 0.9968&0.9910&0.9950&0.9973&0.9752&0.9838&0.9953 &0.9793\cr

 \qquad	&
 $\operatorname{AUC} _{(ODP)}$
&1.6204& 1.4010&1.3092 &1.4426&1.4911&1.4771&1.4558&1.2779&1.3470&1.0982&1.2627
   \cr
   \qquad	&
  $\operatorname{AUC} _{(\operatorname{SNPR})}$
& 16.2106&27.4386&14.2062&16.2505&10.2608&12.3109&21.7941&14.2359&8.9213&22.5599&7.5783
\cr

\qquad &
$\operatorname{AUC} _{(\operatorname{TDBS})}$
&0.6218 &   0.4027   & 0.3120  &  0.4458 &   0.5001 &   0.4821  &  0.4585 &   0.3027 &   0.3632   & 0.1029 &   0.2834
\cr
\hline
  \multirow{4}{*}{MCP}& $\operatorname{AUC} _{({\operatorname{P}}_{\mathnormal{D}}, {\operatorname{P}}_{\mathnormal{F}})}$
   & 0.9991&0.9979&0.9972&0.9947&0.9887&0.9971&0.9988&0.9881&0.9916&0.9948&0.9820
    \cr

   \qquad	&
  $\operatorname{AUC} _{(\operatorname{ODP})}$
& 1.7069 &1.4503& 1.3799 &1.4718&1.5072&1.5474&1.5043&1.2705&1.4895&1.1168&1.2850

\cr

 \qquad	&
  $\operatorname{AUC} _{(\operatorname{SNPR})}$
& 15.6435&30.0371&17.0651&23.1564&10.6468&14.5926&29.4970&19.4595&14.7299&26.3305&8.1361

\cr
 \qquad	&
  $\operatorname{AUC} _{(\operatorname{TDBS})}$&
0.7078  &  0.4524  &  0.3827  &  0.4771&    0.5185&    0.5503 &   0.5055   & 0.2824 &   0.4979 &   0.1220&    0.3030
\cr
   \hline
     \multirow{4}{*}{Log}& $\operatorname{AUC} _{({\operatorname{P}}_{\mathnormal{D}}, {\operatorname{P}}_{\mathnormal{F}})}$
    & 0.9996&0.9983&0.9970&0.9943&0.9916&0.9977&0.9991&0.9800&0.9890&0.9935&0.9842
    \cr

   \qquad	&
  $\operatorname{AUC} _{(\operatorname{ODP})}$
&1.6988& 1.4248&1.3402&1.4743&1.5059&1.5572&1.4942&1.2905&1.4353&1.0859&1.3078
\cr

\qquad	&
  $\operatorname{AUC} _{(\operatorname{SNPR})}$
& 22.3044&29.1209 &16.1383 &23.1957&8.0262&18.2023&27.7427&16.4094&13.5974&26.1800&9.1143
\cr

\qquad	&
  $\operatorname{AUC} _{(\operatorname{TDBS})}$ &
 0.6992 &   0.4265 &   0.3432    &0.4800  &  0.5143   & 0.5595 &   0.4951&    0.3105  &  0.4463   & 0.0924   & 0.3236
 \cr
   \hline

   \multirow{4}{*}{Capped-L1}& $\operatorname{AUC} _{({\operatorname{P}}_{\mathnormal{D}}, {\operatorname{P}}_{\mathnormal{F}})}$
   & 0.9997&0.9981&0.9933&0.9942&0.9899&0.9941&0.9985&0.9857&0.9888&0.9950&0.9826
    \cr

   \qquad	&
  $\operatorname{AUC} _{(\operatorname{ODP})}$
&1.7019 &1.4645&1.3350&1.4618&1.4862&1.5695&1.4844&1.2657&1.4293&1.1105&1.2833
\cr

 \qquad	&
  $\operatorname{AUC} _{(\operatorname{SNPR})}$
&30.1980&29.7396 &14.5027 &19.5036&10.1049&13.2079&25.3616&16.9770&12.5657&27.6946&7.8313
\cr

\qquad	&
  $\operatorname{AUC} _{(\operatorname{TDBS})}$&
 0.7022 &   0.4664  &  0.3417  &  0.4676 &   0.4963  &  0.5754 &   0.4859 &   0.2800 &   0.4405  &  0.1155&    0.3007
 \cr
   \hline

  \multirow{4}{*}{Capped-Lp}& $\operatorname{AUC} _{({\operatorname{P}}_{\mathnormal{D}}, {\operatorname{P}}_{\mathnormal{F}})}$
  & 0.9997 &0.9981 &0.9977 &0.9921&0.9916&0.9982&0.9990&0.9883&0.9903&0.9962&0.9811\cr

   \qquad	&
  $\operatorname{AUC} _{(\operatorname{ODP})}$
& 1.6694&1.4442&1.3661&1.5088&1.5213&1.5711&1.4900&1.2877&1.4087&1.1189&1.3418
\cr

   \qquad	&
  $\operatorname{AUC} _{(\operatorname{SNPR})}$
& 26.6731&30.0460&14.7704&20.2671&16.3419&19.3292&27.4201&20.7341&13.3398&28.2700&10.2623
\cr
 \qquad	&
  $\operatorname{AUC} _{(\operatorname{TDBS})}$&
 0.6697  &  0.4461   & 0.3684&    0.5167   & 0.5297  &  0.5729  &  0.4910  &  0.2994&    0.4184 &   0.1227 &   0.3607
 \cr

\hline
  \multirow{4}{*}{Capped-MCP}& $\operatorname{AUC} _{({\operatorname{P}}_{\mathnormal{D}}, {\operatorname{P}}_{\mathnormal{F}})}$
  &0.9986  & 0.9986  &0.9921&0.9908&0.9900&0.9986&0.9993&0.9851&0.9915&0.9956&0.9822\cr

   \qquad	&
  $\operatorname{AUC} _{(\operatorname{ODP})}$
& 1.6512& 1.4464&1.3509&1.5333&1.4486&1.5872&1.4912&1.2880&1.4025&1.1185&1.3505
\cr

  \qquad	&
  $\operatorname{AUC} _{(\operatorname{SNPR})}$
& 24.5450&31.4068&14.8410&19.2955&16.2175&19.9611&26.0997&18.6652&13.5420&23.4183&9.7221
\cr
\qquad	&
  $\operatorname{AUC} _{(\operatorname{TDBS})}$&
 0.6526  &  0.4478  &  0.3588&    0.5425&    0.4586&    0.5886    &0.4919 &   0.3029  &  0.4110 &   0.1229 &   0.3683
 \cr
   \hline

     \multirow{4}{*} 
     {Capped-Log}& $\operatorname{AUC} _{({\operatorname{P}}_{\mathnormal{D}}, {\operatorname{P}}_{\mathnormal{F}})}$
   & 0.9998& 0.9976 & 0.9941&0.9921&0.9877&0.9982&0.9991&0.9863&0.9901&0.9958&0.9847\cr

   \qquad	&
  $\operatorname{AUC} _{(\operatorname{ODP})}$
&1.7360 &1.4716 &1.3691&1.5658&1.4781&1.5965&1.5065&1.2973&1.4375&1.1258&1.3254
\cr
 \qquad	&
  $\operatorname{AUC} _{(\operatorname{SNPR})}$
& 26.1976 &30.4869&14.1624&20.8786&12.6244&19.0670&26.8211&18.3170&13.2782&24.6349&10.3161
\cr
 \qquad	&
  $\operatorname{AUC} _{(\operatorname{TDBS})}$&
0.7362   & 0.4740  &  0.3750&    0.5737  &  0.4904 &   0.5983  &  0.5074   & 0.3110 &   0.4474 &   0.1300&    0.3407
\cr
   \hline
    \end{tabular}
    \end{threeparttable}
    \vspace{-0.52cm}
\end{table*}

\vspace{-0.65cm}
\subsection{\textcolor[rgb]{0.00,0.00,0.00}{\textbf{More 
Discussions}}}

\subsubsection{Part 1}
\textcolor[rgb]{0.00,0.00,0.00}{In this part, 
 we  conduct related ablation experiments on several benchmark HSI datasets. The purpose of these experiments is
 to investigate
the impact of adopting various nonconvex regularization schemes 
  to encode the prior structures 
of  background tensor on the performance of anomaly detection.
Equivalently,
 to verify
the effectiveness of the designed  EUNTRFR regularization term,
we 
 compare the proposed model (\ref{orin_nonconvex}) with the following four models:}

\textbf{Model $1$ (GNTCTV):}  The  regularizer (\ref{GTCTV})  is directly  
imposed  on the background  tensor, i.e.,
\begin{align} \label{mode1}
& \min_{{\boldsymbol{\mathcal{B}}},{\boldsymbol{\mathcal{E}}}}
 \| {\boldsymbol{\mathcal{B}}}\|_{ \operatorname{GNTCTV}}  +
 \lambda_{1} \cdot \|{\boldsymbol{\mathcal{E}}}\|_  {\ell _ {\mathnormal{F},1}^{ \psi}},
\text{s.t.} \; \;
{\boldsymbol{\mathcal{M}}}=
{\boldsymbol{\mathcal{B}}}
+{\boldsymbol{\mathcal{E}}}.
\end{align}

\textbf{Model $2$ (EGNTCTV):} The  regularizer (\ref{Egntctv})  is directly  
imposed  on the background  tensor, i.e.,
\begin{align} \label{mode4}
& \min_{{\boldsymbol{\mathcal{B}}},{\boldsymbol{\mathcal{E}}}}
 \| {\boldsymbol{\mathcal{B}}}\|_{ \operatorname{EGNTCTV}}  +
 \lambda_{2} \cdot \|{\boldsymbol{\mathcal{E}}}\|_  {\ell _ {\mathnormal{F},1}^{ \psi}},
\text{s.t.} \; \;
{\boldsymbol{\mathcal{M}}}=
{\boldsymbol{\mathcal{B}}}
+{\boldsymbol{\mathcal{E}}}.
\end{align}

\textbf{Model $3$ (GNCTV):}  
The  regularizer (\ref{gnctv}) is imposed on  the mode-$2$ unfolding of each TR factor, i.e.,
\begin{align}
& \min_{\textbf{[}{\boldsymbol{\mathcal{G}}}\textbf{]},{\boldsymbol{\mathcal{E}}}}
 \sum_{n=1}^{3}
 \|
 {\bm{{G}}}^{(n)}_{(2)}
 \|_{ \operatorname{GNCTV}}
 +
 \lambda_{3} \cdot \|{\boldsymbol{\mathcal{E}}}\|_  {\ell _ {\mathnormal{F},1}^{ \psi}},
\notag \\ \label{mode2} &
\text{s.t.} \;  {\boldsymbol{\mathcal{M}}}=
\Re
(\textbf{[}{\boldsymbol{\mathcal{G}}}\textbf{]})
+{\boldsymbol{\mathcal{E}}}.
\end{align}



\textbf{Model $4$ (UNTRFR):} The regularizer (\ref{UNTRFR22}) is  imposed on the background tensor.
This is equivalent to imposing the  
GNTCTV regularizer directly
  on gradient 
 TR factors, i.e.,
\begin{align}
& \min_{\textbf{[}{\boldsymbol{\mathcal{G}}}\textbf{]},{\boldsymbol{\mathcal{E}}}}
 \sum_{n=1}^{3}
 \sum_{k=1}^{3}
 \frac{1}{\gamma}
 \|   {\boldsymbol{\mathcal{T}}}^{(n,k)}
 \|_{ \Phi,  {\mathfrak{L}}}
 +
\lambda_{4} \cdot \|{\boldsymbol{\mathcal{E}}}\|_  {\ell _ {\mathnormal{F},1}^{ \psi}},
\notag \\ \label{mode3}
&
\text{s.t.} \;
{\boldsymbol{\mathcal{M}}}=
\Re
(\textbf{[}{\boldsymbol{\mathcal{G}}}\textbf{]})
+{\boldsymbol{\mathcal{E}}},
\nabla_{k} ({\boldsymbol{\mathcal{G}}} ^{(n)})= {\boldsymbol{\mathcal{T}}}^{(n,k)}, 
\; n, k \in [3],
\end{align}
where $\lambda_{1}, \lambda_{2}, \lambda_{3}, \lambda_{4}  $  are the regularization parameters.


%
\begin{figure*}[!htbp]
\renewcommand{\arraystretch}{0.15}
\setlength\tabcolsep{0.65pt}
\centering
\begin{tabular}{ccc c}
\centering


\includegraphics[width=1.77516851in, height=1.4575in]{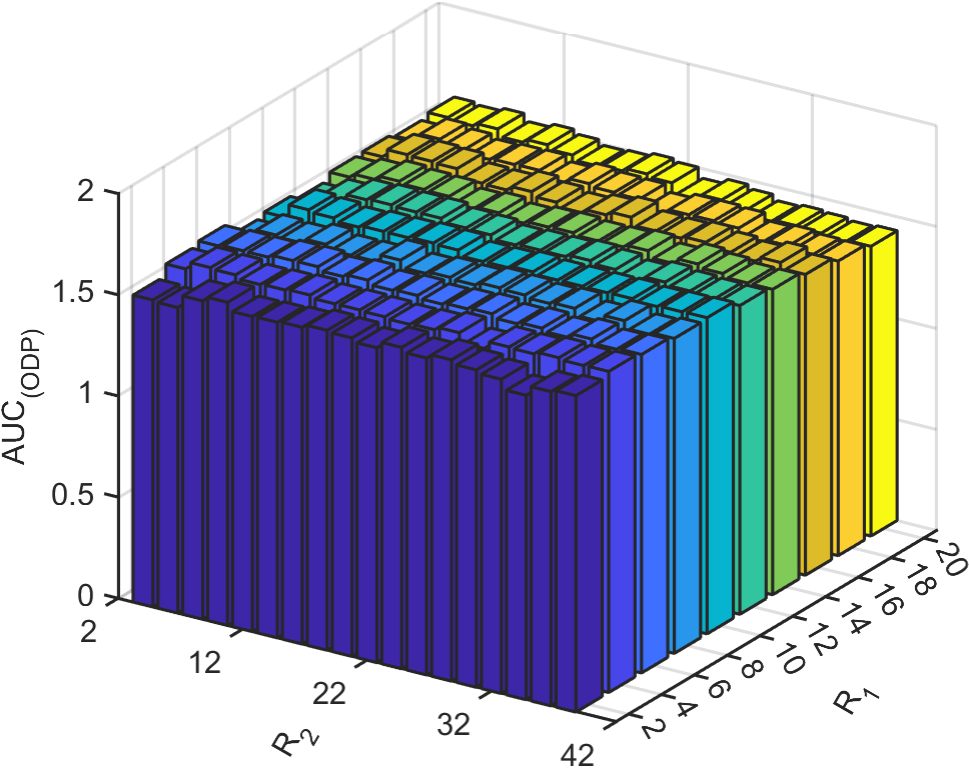}&
\includegraphics[width=1.77516851in, height=1.4575in]{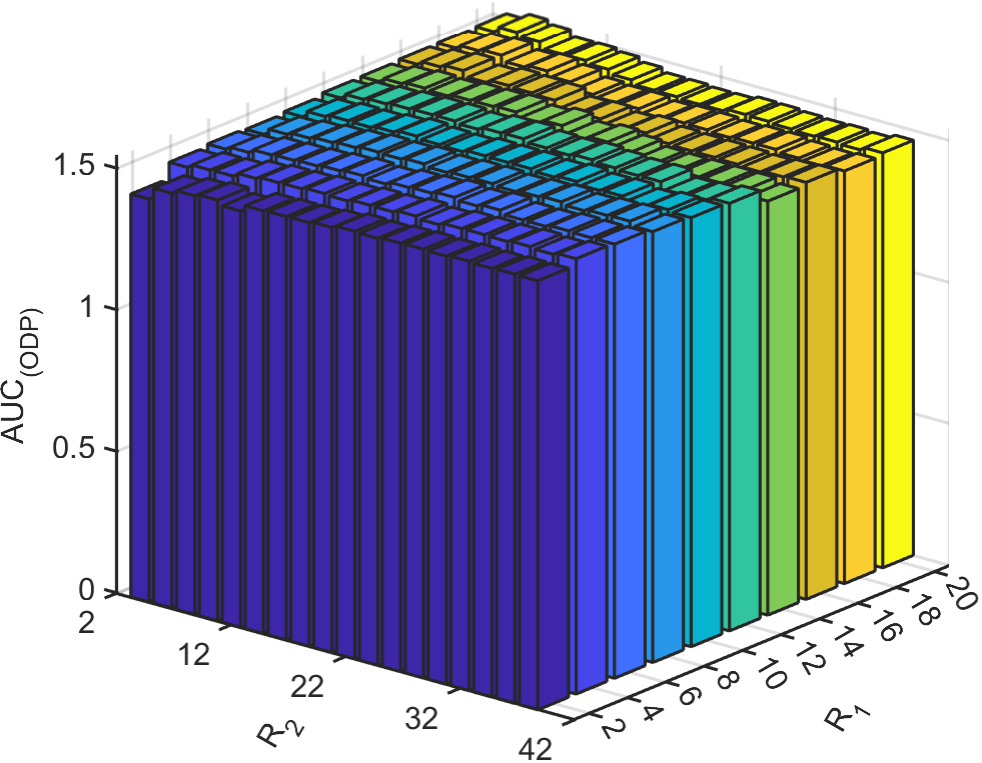}&
\includegraphics[width=1.77516851in, height=1.4575in]{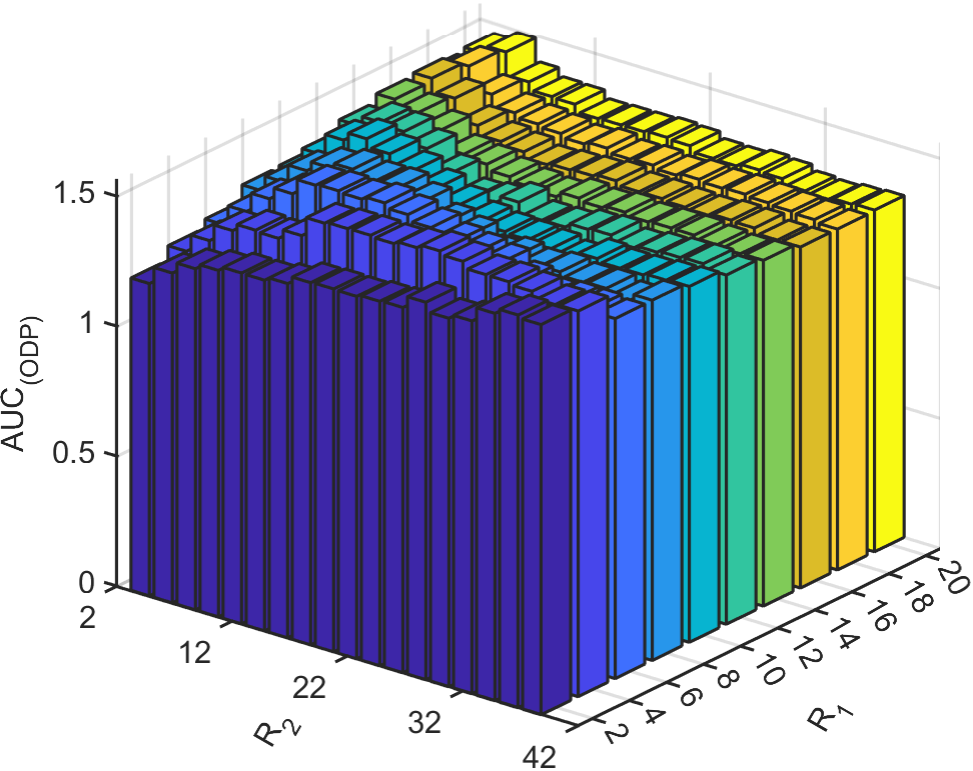}&
\includegraphics[width=1.77516851in, height=1.4575in]{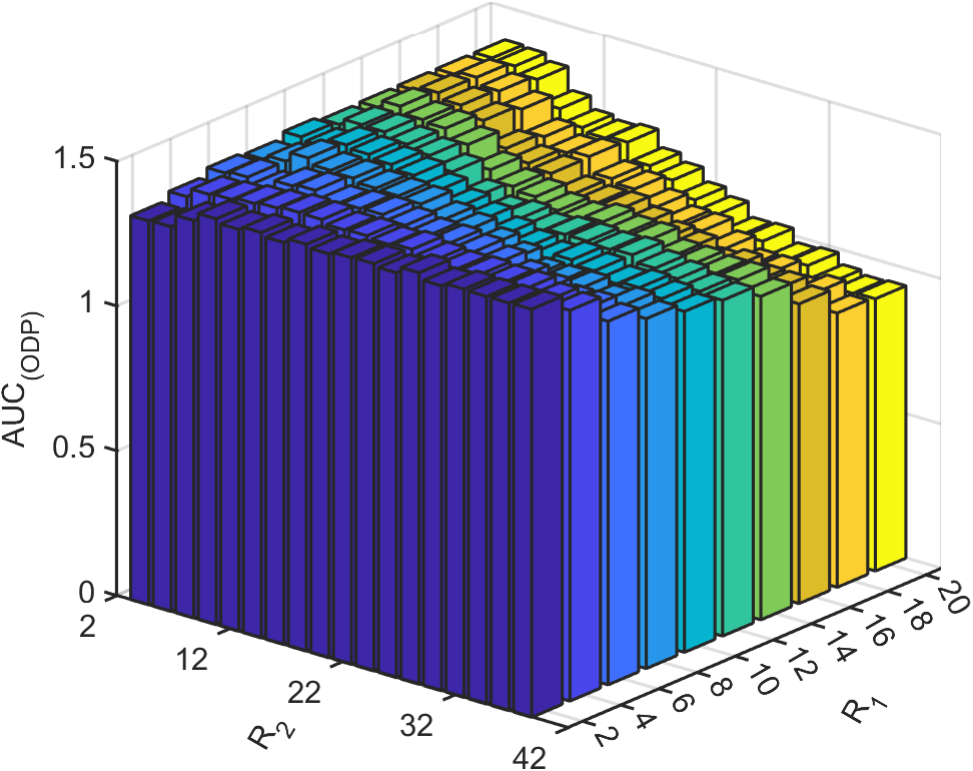}
\\


\tabincell{c}{
\footnotesize{ {{(a)}}} Airport-4: 
\footnotesize{$\beta$=5e-3, $ \alpha $=5e-1}} &
 \tabincell{c}{
\footnotesize{ {{(b)}} } Beach-3: 
 \footnotesize{$\beta$=5e-3, $\alpha$=1e-4}}   & 
 \tabincell{c}{ \footnotesize{ {{(c)}}} HYDICE: 
  \footnotesize{$\beta $=1e-2, $\alpha$=1e-2}}
& 
\tabincell{c}{ \footnotesize{ {{(d)}}} Hyperion: 
 \footnotesize{$\beta $=5e-2, $\alpha$=1}}
\\




\includegraphics[width=1.77516851in, height=1.4575in]{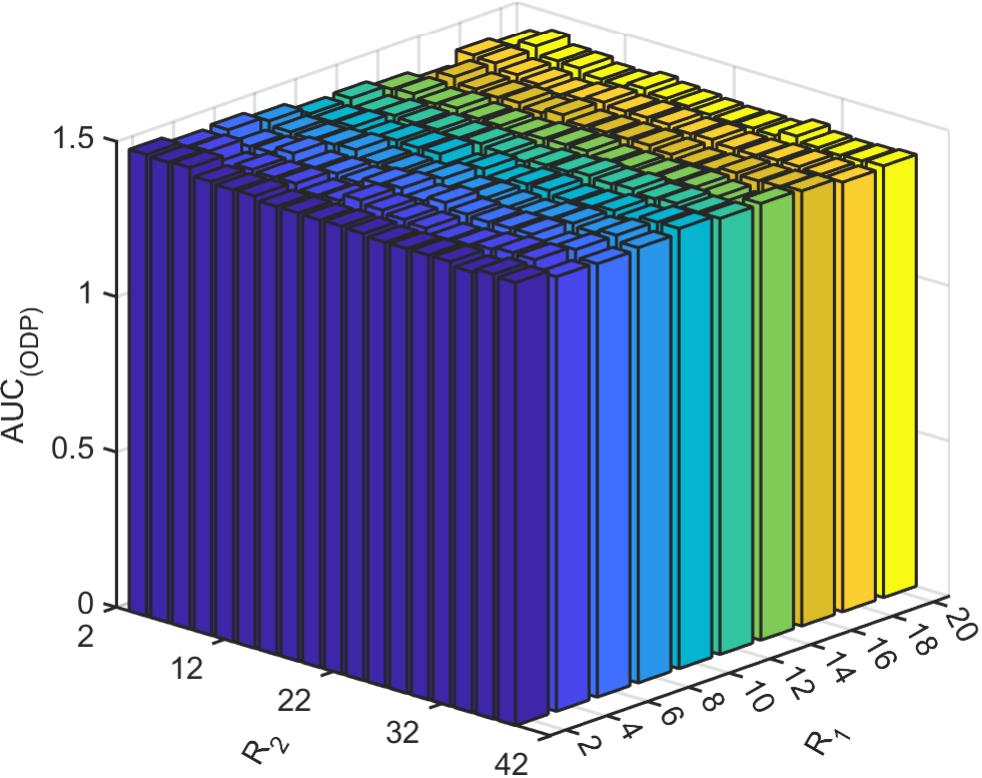}&
\includegraphics[width=1.77516851in, height=1.4575in]{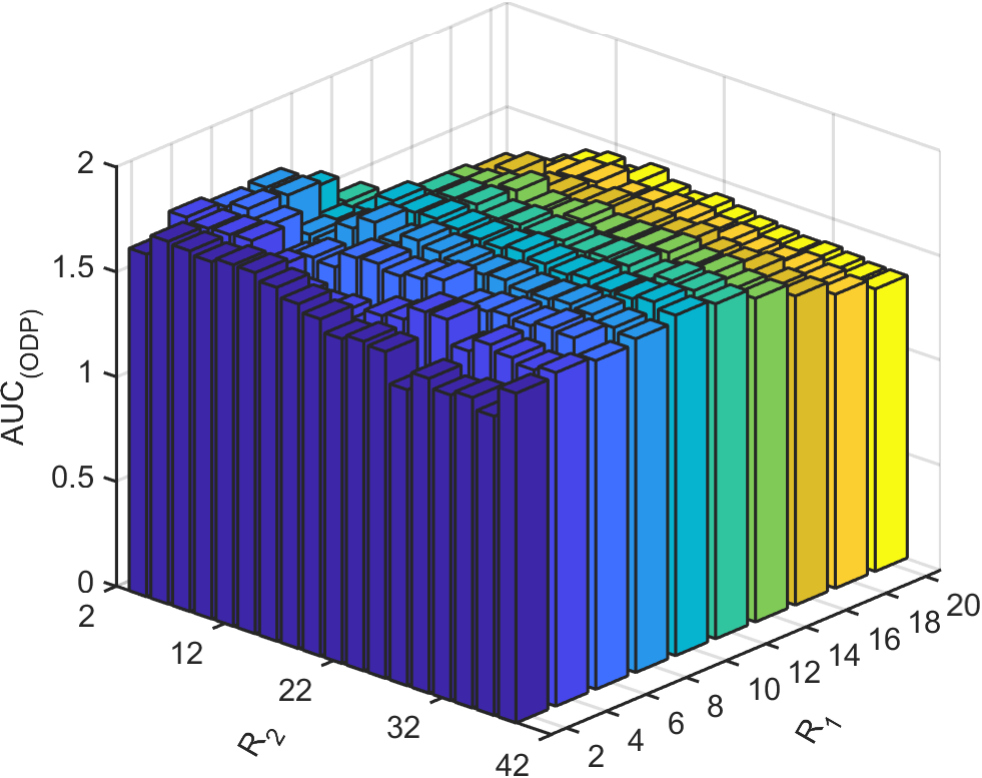}&
\includegraphics[width=1.77516851in, height=1.4575in]{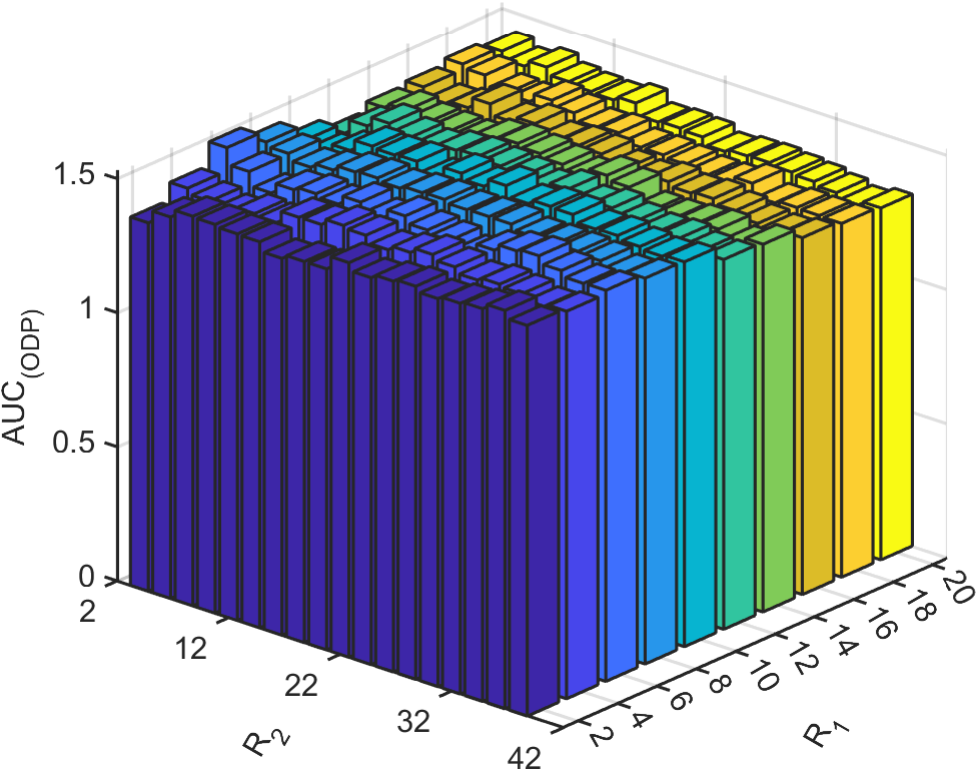}&
\includegraphics[width=1.77516851in, height=1.4575in]{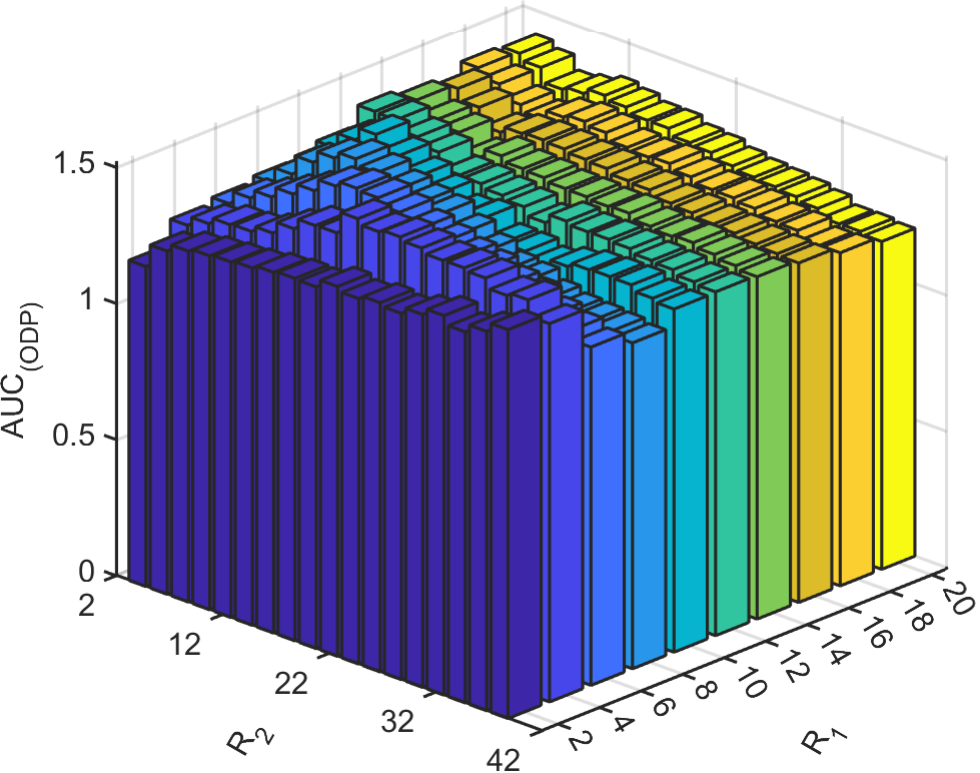}
\\



 \tabincell{c}{ \footnotesize{ {{(e)}} } Pavia: 
 \footnotesize{$ \beta $=5e-4, $\alpha$=1e-4}}
 & 
 \tabincell{c}{ \footnotesize{ {{(f)}} } Salinas: 
  \footnotesize{$\beta $=5e-3, $ \alpha $=5e-6}}
 & \tabincell{c}{ \footnotesize{ {{(g)}}} San-Diego: 
 \footnotesize{$\beta $=1e-3, $\alpha$=5e-3}}
  & \tabincell{c}{ \footnotesize{ {{(h)}} } Urban-3: 
   \footnotesize{$\beta  $=1e-2, $\alpha$=5e-3}}
\end{tabular}
\caption{
 \textcolor[rgb]{0.00,0.00,0.00}{Sensitivity analysis of the TR rank ($r_1,r_2,r_3$).
%
All nonconvex combinations are set to Capped-Log,  
with the exception of Hyperion (MCP), San-Diego (Capped-Lp) and Urban-3 (MCP).
}}
\vspace{-0.305cm}
\label{fig_rank-learning}
\end{figure*}

\textbf{Proposed Model (EUNTRFR):} The regularizer (\ref{UNTRFR}) is  imposed on the background tensor.
This is equivalent to imposing the  
EGNTCTV regularizer directly
  on gradient 
 TR factors.
 Please see the  model (\ref{orin_nonconvex}) for more details.

\textcolor[rgb]{0.00,0.00,0.00}{
It is worth mentioning that 
the models (\ref{mode1}) and  (\ref{mode4}) use  the GNTCTV and EGNTCTV regularizers induced by the T-SVD framework
to encode the $\textbf{L}$+$\textbf{S}$ priors
of HSI's background, respectively.
The  model (\ref{mode2}) utilizes   the GNCTV 
to describe  the $\textbf{L}$ and $\textbf{S}$ priors
of TR factors of HSI's background.
Unlike the  model (\ref{mode2}) that uses the mode-$2$ unfolding scheme, the  model (\ref{mode3}) treats all 
 TR factors  as the low-tubal-rank tensors,
and then employ  
the GNTCTV regularizer to characterize their  prior structures 
in the gradient domain.
Based on the model (\ref{mode3}), the proposed model  (\ref{orin_nonconvex}) additionally  introduces a residual term to enhance its robustness.
All of the above models are optimized by using ADMM framework.}

In our experiments,  the trade-off parameter $\lambda_1,\lambda_2,\lambda_3, \lambda_4$ 
 in (\ref{mode1})-(\ref{mode3})
  are set to be  $\lambda$$=$$
 {\kappa} / {\sqrt{\min(n_1,n_2) \cdot n_3}} $, $\kappa$ is selected from
 $\{0.1, 0.3, 0.5, 0.8, 1, 1.2, 1.5, 1.8,2,2.2\}$.
 The relevant experimental results are shown in Table \ref{ablation-study}.
 Among all the T-SVD-based models, 
 we discovered that Model 1 and Model 2 overcome the 
 TCTV model  in AUC values, with Model 2 exceeding Model 1.
 This demonstrates that the GNTCTV regularizer and its enhanced version
 are  more capable of effectively extracting 
 the prior 
 information
 of the background
 compared to convex TCTV   regularizer.
 Besides, the EGNTCTV regularizer 
 outperforms the GNTCTV regularizer.
 Across all TR-based models, we observed that 
 our model  and Model 4 
 are  superior to Model 3.
 This implies that, compared to the matrix unfolding strategy employed in Model 3, the
  UNTRFR and EUNTRFR schemes  are able to better extract structural information from TR factors,
  thereby enhancing detection performance.

\begin{table*}[tp]
\renewcommand{\arraystretch}{0.36}
\setlength\tabcolsep{5pt}
  \centering
  \caption{The  
AUC 
  values obtained by  the proposed HAD model and its degraded versions 
   for  different
  HSI  datasets.}
  \label{ablation-study}
  \scriptsize
  \begin{threeparttable}
    \begin{tabular}{c   ccc ccc ccc ccc }
    \hline
    \multicolumn{1}{c}{HAD-Model 
    }
   &\multicolumn{1}{c}{AUC-Metrics}
   &\multicolumn{1}{c}{Salinas}&\multicolumn{1}{c}{Pavia}&\multicolumn{1}{c}{Hyperion}
    &\multicolumn{1}{c}{HYDICE}& \multicolumn{1}{c}{San-Diego}&\multicolumn{1}{c}{Airport-4}&
    \multicolumn{1}{c}{Beach-3}&\multicolumn{1}{c}{Beach-4} &\multicolumn{1}{c}{Urban-3} &
    \multicolumn{1}{c}{\text{Urban-4}}&\multicolumn{1}{c}{Urban-5}
    \cr

    \hline
    \hline

   \multirow{4}{*}
   {Model 1}& $\operatorname{AUC} _{({\operatorname{P}}_{\mathnormal{D}}, {\operatorname{P}}_{\mathnormal{F}})}$
   &0.9703&0.9986&0.9935&0.9709&0.9959&0.9897&0.9956&0.9823&0.9730&0.9863&0.9670
    \cr

   \qquad	&
  $\operatorname{AUC} _{(\operatorname{ODP})}$
  & 1.2238&1.4730&1.3674&1.3891&1.3180&1.4215&1.4466&1.2976&1.1945&1.0389&1.1726
\cr

 \qquad	&
  $\operatorname{AUC} _{(\operatorname{SNPR})}$
  & 1.8815&27.2931&13.2716&16.6455&19.0602&3.6785&11.2594&19.2033&8.1959&25.2131&8.1807
  \cr

  \qquad	&
  $\operatorname{AUC} _{(\operatorname{TDBS})}$
  & 0.2535&0.4744&0.3739&0.4182&0.3221&0.4318&04510&0.3153&0.2215&0.0526&0.2056
  \cr

   \hline

  \multirow{4}{*}{Model 2}& $\operatorname{AUC} _{({\operatorname{P}}_{\mathnormal{D}}, {\operatorname{P}}_{\mathnormal{F}})}$
  & 0.9959& 0.9985&0.9940&0.9725&0.9933&0.9966&0.9976&0.9800&0.9742&0.9871&0.9766

  \cr

 \qquad	&
 $\operatorname{AUC} _{(ODP)}$
&1.3840 &1.4733&1.3704&1.3982&1.3357&1.5237&1.5292&1.2822&1.1681&1.0293&1.2596
   \cr
   \qquad	&
  $\operatorname{AUC} _{(\operatorname{SNPR})}$
& 2.9909&27.1495&12.0886&16.7718&17.3890&6.4321&22.9846&17.9461&8.7253&25.7181&4.4842
\cr

\qquad &
$\operatorname{AUC} _{(\operatorname{TDBS})}$
&0.3882&0.4748&0.3764&0.4257&0.3425&0.5272&0.5316&0.3022&0.1939&0.0421&0.2841
\cr
\hline
  \multirow{4}{*}{Model 3}& $\operatorname{AUC} _{({\operatorname{P}}_{\mathnormal{D}}, {\operatorname{P}}_{\mathnormal{F}})}$
   & 0.9955&0.9969&0.9830&0.9514&0.9937&0.9957&0.9962&0.9731&0.9734&0.9940&0.9789
    \cr

   \qquad	&
  $\operatorname{AUC} _{(\operatorname{ODP})}$
& 1.5132&1.3938&1.3328&1.4394&1.3696&1.5837&1.4617&1.2483&1.3891&1.0889&1.2797

\cr

 \qquad	&
  $\operatorname{AUC} _{(\operatorname{SNPR})}$
& 18.9065&21.1303&7.3124&6.3971&8.3112&11.4182&11.5632&12.0305&6.0296&8.7531&6.1895

\cr
 \qquad	&
  $\operatorname{AUC} _{(\operatorname{TDBS})}$&
  0.5138&0.3969&0.3498&0.4880&0.3759&0.5880&0.4655&0.2752&0.4157&0.0949&0.3008
\cr
   \hline
     \multirow{4}{*}{Model 4}& $\operatorname{AUC} _{({\operatorname{P}}_{\mathnormal{D}}, {\operatorname{P}}_{\mathnormal{F}})}$
    & 0.9997 &0.9983&0.9943&0.9902&0.9890&0.9978&0.9986&0.9824&0.9819&0.9824&0.9847
    \cr

   \qquad	&
  $\operatorname{AUC} _{(\operatorname{ODP})}$
&1.5915 &1.3854&1.3652&1.4799&1.4310&1.5976&1.4955&1.2894&1.4370&1.0556&1.3674
\cr

\qquad	&
  $\operatorname{AUC} _{(\operatorname{SNPR})}$
& 23.9790&27.7836&13.7848&17.5180&12.5630&17.9849&21.2692&16.4999&11.4464&19.2204&10.4884
\cr

\qquad	&
  $\operatorname{AUC} _{(\operatorname{TDBS})}$ &
  0.5918 &0.3871&0.3709&0.4897&0.4421&0.5998&0.4969&0.3070&0.4551&0.0732&0.3826
 \cr
   \hline

\multirow{4}{*}[0pt]{Proposed}& $\operatorname{AUC} _{({\operatorname{P}}_{\mathnormal{D}}, {\operatorname{P}}_{\mathnormal{F}})}$
    & 0.9988&0.9975&0.9967&0.9902&0.9927&0.9983&0.9990&0.9846&0.9908&0.9963&0.9821
    \cr

   \qquad	&
  $\operatorname{AUC} _{(\operatorname{ODP})}$
&1.7059&1.4487&1.3705&1.5269&1.5443&1.5961&1.4975&1.2877&1.4974&1.1168&1.3310
\cr

\qquad	&
  $\operatorname{AUC} _{(\operatorname{SNPR})}$
& 26.3255&28.6802&15.6485&19.6848&16.3717&19.0930&25.8585&18.8186&14.9567&31.0597&10.0293
\cr

\qquad	&
  $\operatorname{AUC} _{(\operatorname{TDBS})}$ &
  0.7071&0.4512&0.3738&0.5367&0.5515&0.5978&0.4985&0.3031&0.5066&0.1205&0.3489
 \cr
   \hline

    \end{tabular}
    \end{threeparttable}
    \vspace{-0.325cm}
\end{table*}

%
\begin{figure*}[!htbp]
\renewcommand{\arraystretch}{0.35}
\setlength\tabcolsep{8pt}
\centering
\begin{tabular}{c c}
\includegraphics[width=3.0142345in, height=1.8in]{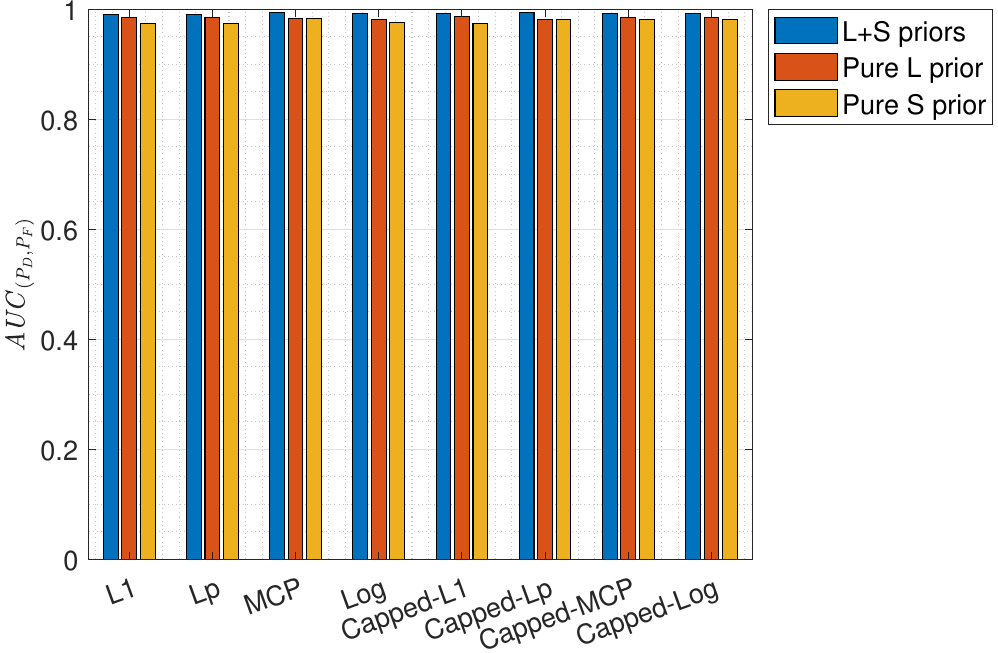} &
\includegraphics[width=3.014234in, height=1.8in]{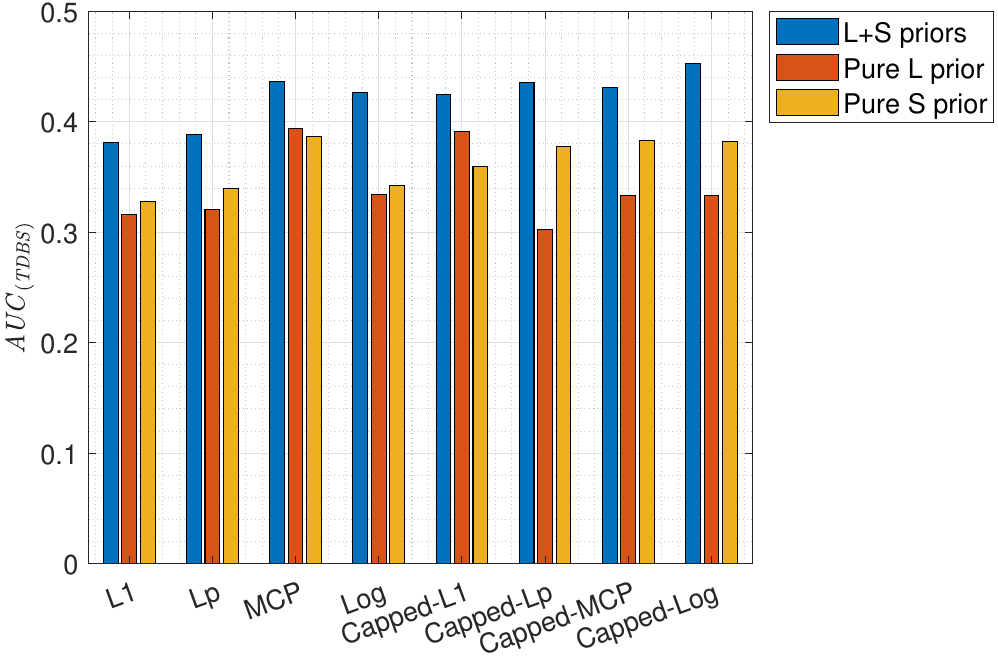} \\
\includegraphics[width=3.014234in, height=1.8in]{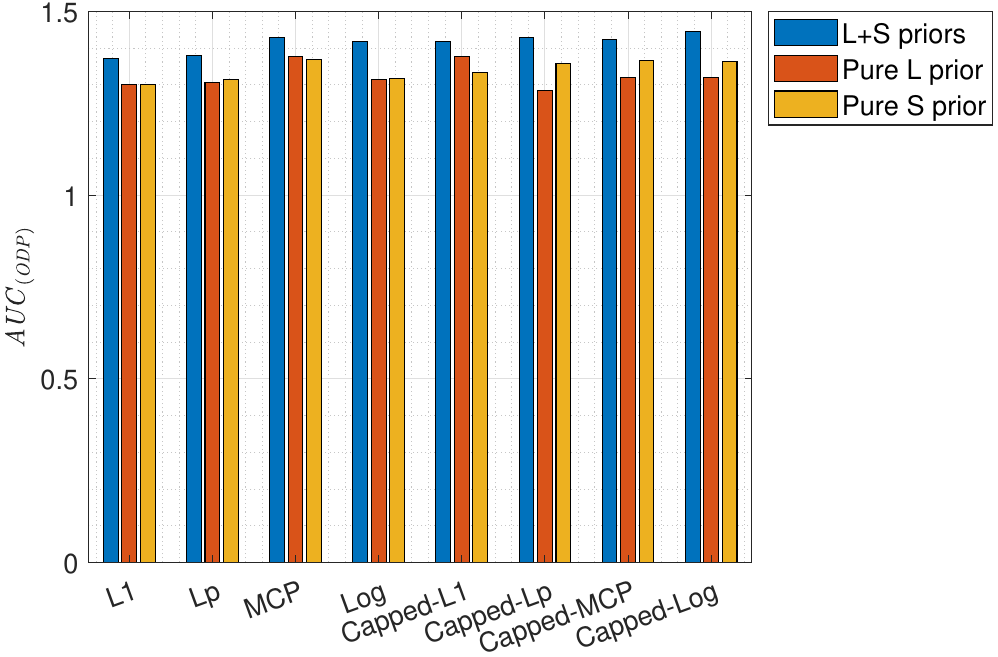}&
\includegraphics[width=3.014234in, height=1.8in]{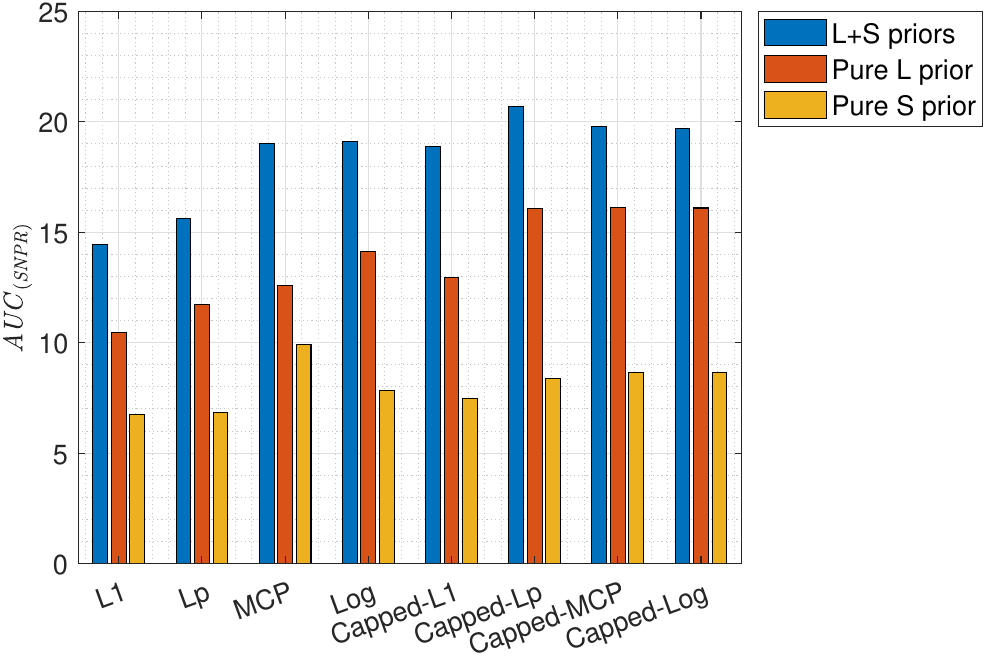}
\end{tabular}
\vspace{-0.15cm}
\caption{
\textcolor[rgb]{0.00,0.00,0.00}{The influence of 
different prior representation methods
upon  anomaly detection 
performance
of the  proposed 
method 
under 
various nonconvex regularization schemes.}
%
%
}
\vspace{-0.38cm}
\label{fig-priors}
\end{figure*}

\subsubsection{Part 2}

\textcolor[rgb]{0.00,0.00,0.00}{To  verify the validity 
of the novel prior representation paradigm 
introduced in our proposed method, we conduct a comparative analysis by juxtaposing our model against the following alternative models:}

\textcolor[rgb]{0.00,0.00,0.00}{
\textbf{Model 1 (Pure $\textbf{L}$ prior):}
Nonconvex regularization scheme is utilized to encode the low-rank property of TR factors,
while disregarding their local smoothness characteristics, i.e.,}
\begin{align}
& \textcolor[rgb]{0.00,0.00,0.00}{\min_{\textbf{[}{\boldsymbol{\mathcal{G}}}\textbf{]},{\boldsymbol{\mathcal{E}}}}
 \sum_{n=1}^{3}
 \|   {\boldsymbol{\mathcal{G}}}^{(n)}
 \|_{ \Phi,  {\mathfrak{L}}}
 +
 \lambda_{5} \cdot \|{\boldsymbol{\mathcal{E}}}\|_  {\ell _ {\mathnormal{F},1}^{ \psi}},}
\notag \\ \label{mode2LLL} &
\textcolor[rgb]{0.00,0.00,0.00}{\text{s.t.} \;  {\boldsymbol{\mathcal{M}}}= {\boldsymbol{\mathcal{B}}} + {\boldsymbol{\mathcal{E}}}=
\Re
(\textbf{[}{\boldsymbol{\mathcal{G}}}\textbf{]})
+{\boldsymbol{\mathcal{E}}},}
\end{align}
\textcolor[rgb]{0.00,0.00,0.00}{where $\lambda_{5}  $  is the regularization parameter,
$\|\cdot\|_  {\ell _ {\mathnormal{F},1}^{ \psi}}
$ and
$\|
 \cdot 
\|_
{
\Phi,
\mathfrak{L} }$
 are defined according to Formulas (\ref{unified-regular22222})  and (\ref{GTCTV}), respectively.}

\textcolor[rgb]{0.00,0.00,0.00}{
\textbf{Model 2 (Pure $\textbf{S}$ prior):}
Nonconvex regularization scheme is utilized  
 to encode the smoothness 
property of TR factors,
while disregarding their low-rankness 
characteristics, i.e.,}
\begin{align}
& \textcolor[rgb]{0.00,0.00,0.00}{\min_{\textbf{[}{\boldsymbol{\mathcal{G}}}\textbf{]},{\boldsymbol{\mathcal{E}}}}
 \sum_{n=1}^{3}
 \|    {\boldsymbol{\mathcal{G}}}^{(n)}  \;{\times}_{2}\;   \bm{D}_{n}
 \|_{ \ell_{1} ^ { \psi} } 
 +
 \lambda_{6} \cdot \|{\boldsymbol{\mathcal{E}}}\|_  {\ell _ {\mathnormal{F},1}^{ \psi}},}
\notag \\ \label{mode2LLL} &
\textcolor[rgb]{0.00,0.00,0.00}{\text{s.t.} \;  {\boldsymbol{\mathcal{M}}}= {\boldsymbol{\mathcal{B}}} + {\boldsymbol{\mathcal{E}}}=
\Re
(\textbf{[}{\boldsymbol{\mathcal{G}}}\textbf{]})
+{\boldsymbol{\mathcal{E}}},}
\end{align}
\textcolor[rgb]{0.00,0.00,0.00}{where $\lambda_{6}  $  is the regularization parameter,
$\|\cdot\|_ { \ell_{1} ^ { \psi} }$
 is defined according to  Formula (\ref{unified-regular22}), 
and $ \bm{D}_{n} $ is the first-order difference matrix.}

\textcolor[rgb]{0.00,0.00,0.00}{The 
experimental results are shown in
Figure \ref{fig-priors}, which  presents 
various AUC values obtained from three prior representation paradigms under different nonconvex regularization strategies.
Wherein, the AUC values refer to the average AUC values obtained on the  $11$ HSI datasets.
We find that compared with the pure low-rankness prior and the pure smoothness prior, the proposed joint 
$\textbf{L}$+$\textbf{S}$ priors
can well improve the detection performance.} 

\begin{table*}[tp]
\renewcommand{\arraystretch}{0.3}
\setlength\tabcolsep{2pt}
  \centering
  \caption{\textcolor[rgb]{0.00,0.00,0.00}{The  
AUC 
  values obtained by  the proposed  HAD model 
  under different nonconvex combinations
  $\Phi(\cdot)$+$\psi(\cdot)$
   for    various
   HSI
datasets with subpixel anomalies.}}
  \label{ablation-study-subpiple}
  \scriptsize
  \begin{threeparttable}
    \begin{tabular}{c   ccc ccc ccc cccc cccc }
    \hline
    \multirow{2}{*}  {HSIs-Name}
   & 
   \multirow{2}{*} {AUC-Metrics}
   &\multicolumn{2}{c}{L1}&\multicolumn{2}{c}{Lp}&\multicolumn{2}{c}{MCP}
    &\multicolumn{2}{c}{Log}& 
     \multicolumn{2}{c}{Capped-L1}
    &
     \multicolumn{2}{c}{Capped-Lp}
    &
    %
   \multicolumn{2}{c}{Capped-MCP}
    & 
     \multicolumn{2}{c}{Capped-Log}
    \cr
    \cmidrule(rl){3-4} \cmidrule(rl){5-6} \cmidrule(rl){7-8} \cmidrule(rl){9-10}
    \cmidrule(rl){11-12} \cmidrule(rl){13-14} \cmidrule(rl){15-16} \cmidrule(rl){17-18}
    \qquad &\qquad & \multicolumn{1}{c}{FFT} &\multicolumn{1}{c}{DCT}
    & \multicolumn{1}{c}{FFT} &\multicolumn{1}{c}{DCT}
    & \multicolumn{1}{c}{FFT} &\multicolumn{1}{c}{DCT}
    & \multicolumn{1}{c}{FFT} &\multicolumn{1}{c}{DCT}
    & \multicolumn{1}{c}{FFT} &\multicolumn{1}{c}{DCT}
    & \multicolumn{1}{c}{FFT} &\multicolumn{1}{c}{DCT}
    & \multicolumn{1}{c}{FFT} &\multicolumn{1}{c}{DCT}
    & \multicolumn{1}{c}{FFT} &\multicolumn{1}{c}{DCT}
\cr

    \hline
    \hline

%

\multirow{4}{*}
   {Hyperion}& $\operatorname{AUC} _{({\operatorname{P}}_{\mathnormal{D}}, {\operatorname{P}}_{\mathnormal{F}})}$

&0.9953& 0.9944  &
0.9967&  0.9956  &0.9906& 0.9926
&
0.9951&0.9950&
0.9948& 0.9934 &
0.9959&0.9924 &
0.9962&0.9935 &
0.9959& 0.9935

\cr
   \qquad	&
  $\operatorname{AUC} _{(\operatorname{ODP})}$

& 1.3101&  1.3354 &1.3027&  1.3706   &1.3371&1.4157 &
1.3017& 1.3611&
1.3221&  1.3954 &
1.3161&1.3653 &
1.3153&1.4149 &
1.3099& 1.4149
\cr
 \qquad	&
  $\operatorname{AUC} _{(\operatorname{SNPR})}$

& 8.2186& 6.4845  &8.1385&6.7609 &10.0104&9.3219 &
10.1093& 8.6432 &
8.1611&  7.6908&
8.4699&9.0822 &
9.1034&7.8352 &
9.0871& 7.8275

  \cr
  \qquad	&
  $\operatorname{AUC} _{(\operatorname{TDBS})}$

& 0.3147&0.3410  &0.3061&0.3750&0.3464&   0.4231&
0.3067&0.3661&
0.3274&   0.4020&
0.3202&0.3729&
0.3191&0.4214 &
0.3141& 0.4214
  \cr
   \hline

   \multirow{4}{*}
   {HYDICE-I}& $\operatorname{AUC} _{({\operatorname{P}}_{\mathnormal{D}}, {\operatorname{P}}_{\mathnormal{F}})}$
  & 0.9964&0.9971  &	0.9964	&  0.9972  &0.9941&  0.9945 	&
  0.9955	&0.9951 &
  0.9918&0.9926 	&
  0.9951&0.9930&
  	0.9953& 0.9961 & 
  	0.9953& 0.9935
\cr
   \qquad	&
  $\operatorname{AUC} _{(\operatorname{ODP})}$
  &1.4463	& 1.5152  &1.3923&1.4314  &	1.4838&1.4800 	&
  1.4521	&1.4682 &
  1.4353&	1.4983 &
  1.4657	& 1.4972 &
   1.4668&  1.4679 
   &
  	1.4662& 1.4736
\cr
 \qquad	&
  $\operatorname{AUC} _{(\operatorname{SNPR})}$
  &15.4641&	16.1052 & 19.9096&20.9871   &	25.0011	& 23.0837 &
  23.0879&24.3862 &
  	19.6751&	20.0651 &
  21.1034&	21.4722 &
  21.7587&  20.5353 
  &
  	21.7211& 21.2262
  \cr
  \qquad	&
  $\operatorname{AUC} _{(\operatorname{TDBS})}$
  & 0.4499	&  0.5181 &0.3959	& 0.4342 & 0.4897&	0.4855&
  0.4566&0.4731 &
  	0.4435&	0.5057 &
  0.4708&	0.5042&
  0.4715&	 0.4720 & 
  0.4709& 0.4801
  \cr
   \hline

   \multirow{4}{*}
   {HYDICE-II}& $\operatorname{AUC} _{({\operatorname{P}}_{\mathnormal{D}}, {\operatorname{P}}_{\mathnormal{F}})}$

 &  0.9914	& 0.9954  &0.9941&0.9910 	&0.9855&	0.9844 &
 0.9947&0.9963 &
 	0.9951&0.9918 	&
 0.9927&0.9943 &
 	0.9901	& 0.9856  &
 0.9903& 0.9914

\cr
   \qquad	&
  $\operatorname{AUC} _{(\operatorname{ODP})}$
 & 1.5017	&1.5932    &1.5591&1.5830  &	1.5395&1.5179 	&
 1.4384	&1.5133 &
 1.5165	&1.5797 &
 1.3904	&1.4317 &
 1.4709	&1.4989 &
 1.4731& 1.4546

\cr
 \qquad	&
  $\operatorname{AUC} _{(\operatorname{SNPR})}$
 & 5.9149&6.9202  &	7.1408& 7.0256 &	9.5867	& 12.1078 &
 11.1171&	10.6459 &
 10.6961	& 9.8834  &
 10.6946&10.3506  &	
 9.8432	&8.2111 &
 9.9273&  11.5243

  \cr
  \qquad	&
  $\operatorname{AUC} _{(\operatorname{TDBS})}$
  &0.5103& 0.5978&	0.5651& 0.5920&	0.5541&	0.5335&
  0.4437&	0.5170 &
  0.5214	&0.55878&
  0.3977&	0.4374 &
  0.4809& 0.5134
  &	0.4828& 0.4632
  \cr
   \hline

   \multirow{4}{*}
   {AVIRIS-II}& $\operatorname{AUC} _{({\operatorname{P}}_{\mathnormal{D}}, {\operatorname{P}}_{\mathnormal{F}})}$

 & 0.9911	& 0.9963  &0.9961&	0.9960 & 0.9942	& 0.9971&
  0.9964& 0.9981 &
 	0.9927&	0.9939&
 0.9942&	0.9972 &
 0.9975	& 0.9955
 &
 0.9974& 0.9941

\cr
   \qquad	&
  $\operatorname{AUC} _{(\operatorname{ODP})}$
& 1.4763&  1.5000  &	1.5067&1.5047 	& 1.5282& 1.5236  &	
1.5232& 1.5287 &
	1.5237	& 1.5348&
1.5241& 1.5271 &
	1.5259& 1.5279
&
	1.5261& 1.5265

\cr
 \qquad	&
  $\operatorname{AUC} _{(\operatorname{SNPR})}$
& 34.1055&  31.8844  &	34.1388	& 34.1388   & 39.3751& 36.4750 &	
34.8498&	37.0606 &
21.5638&	 32.6637  &
33.7411&	34.7466 &
33.4456& 32.9886
&	
33.4628& 33.3617

  \cr
  \qquad	&
  $\operatorname{AUC} _{(\operatorname{TDBS})}$
  &0.4852&0.5037 	&0.5087&0.5087 &	0.5341&	0.5265&
  0.5268&	0.5306 &
  0.5311& 0.5410&
  	0.5299	& 0.5299 &
  0.5285& 0.5324
  &	
  0.5285& 0.5325
  \cr
   \hline

\multirow{4}{*}
   {AVIRIS-I}& $\operatorname{AUC} _{({\operatorname{P}}_{\mathnormal{D}}, {\operatorname{P}}_{\mathnormal{F}})}$

 &0.9732& 0.9800 & 0.9705& 0.9735 & 0.9859& 0.9881 & 
 0.9888&0.9876&
 0.9721& 0.9829 &
 0.9728&0.9846&
 0.9695& 0.9883
 &
 0.9687& 0.9885

\cr
   \qquad	&
  $\operatorname{AUC} _{(\operatorname{ODP})}$
& 1.4352&  1.4456  &1.4072&  1.4172   &1.4584& 1.4528 & 
1.4728& 1.4702 &
1.4132&1.4306 &
1.3659& 1.4668 &
1.3464& 1.4575
&
1.3455& 1.4570

\cr
 \qquad	&
  $\operatorname{AUC} _{(\operatorname{SNPR})}$
& 11.3653& 11.6347  & 12.2208&  11.5077  & 22.0709& 15.9134 & 
18.5783&22.7939&
14.8688& 12.5085 &
14.5268&21.3680 &
21.2485& 16.6486
&
21.2086& 16.4736

  \cr
  \qquad	&
  $\operatorname{AUC} _{(\operatorname{TDBS})}$
  &0.4619&0.4656& 0.4366&0.4437& 0.4725& 0.4647 & 
  0.4841& 0.4825 &
  0.4412& 0.4477&
  0.3932&0.4822 &
  0.3769& 0.4693&
  0.3769& 0.4685
  \cr
   \hline

    \end{tabular}
    \end{threeparttable}
    \vspace{-0.3cm}
\end{table*}

%
\begin{figure}[!htbp]
\renewcommand{\arraystretch}{0.5}
\setlength\tabcolsep{5pt}
\centering
\begin{tabular}{c}
\includegraphics[width=3.42345in, height=1.45in]{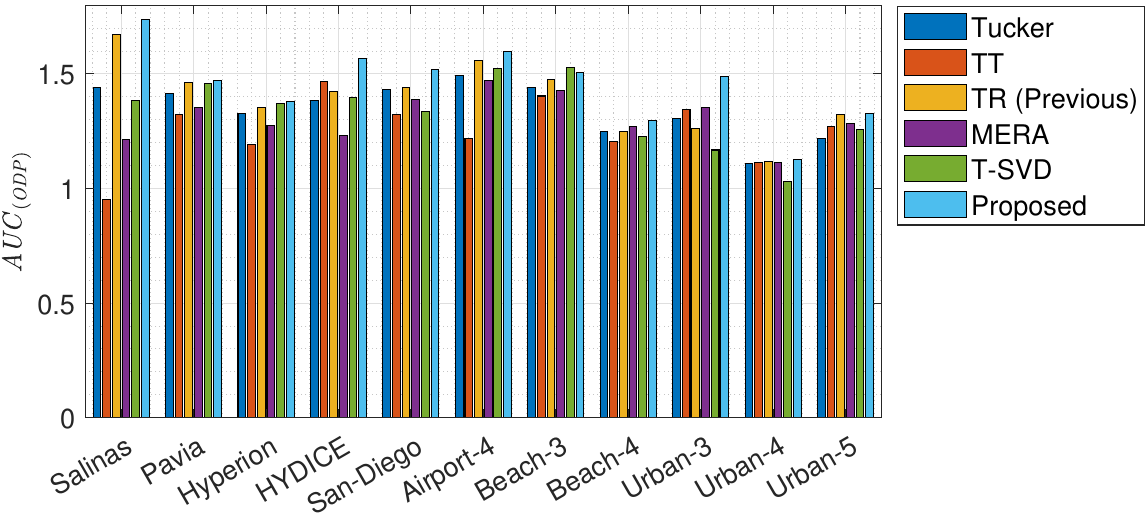} \\
\includegraphics[width=3.4234in, height=1.45in]{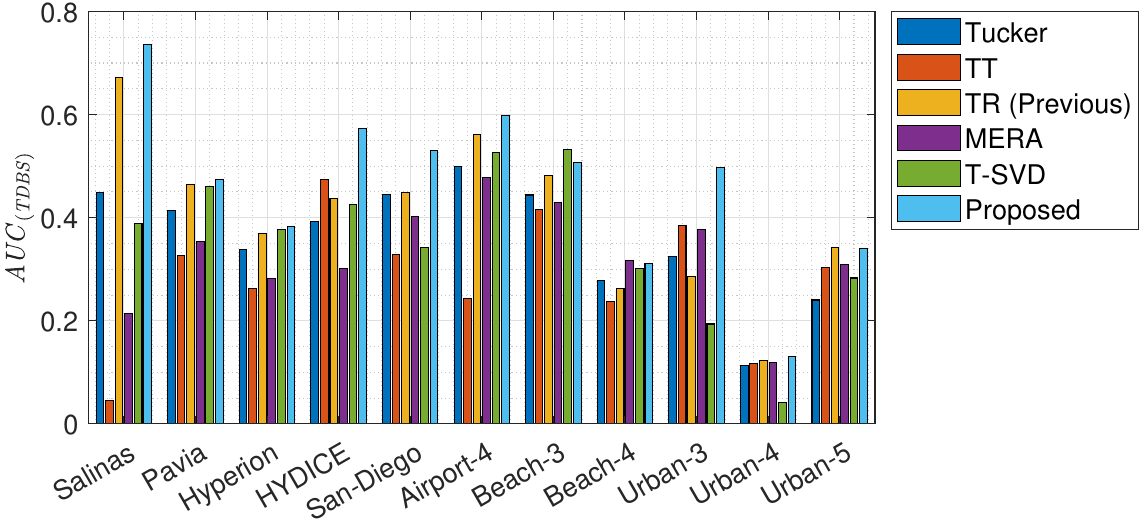} \\
\includegraphics[width=3.4234in, height=1.45in]{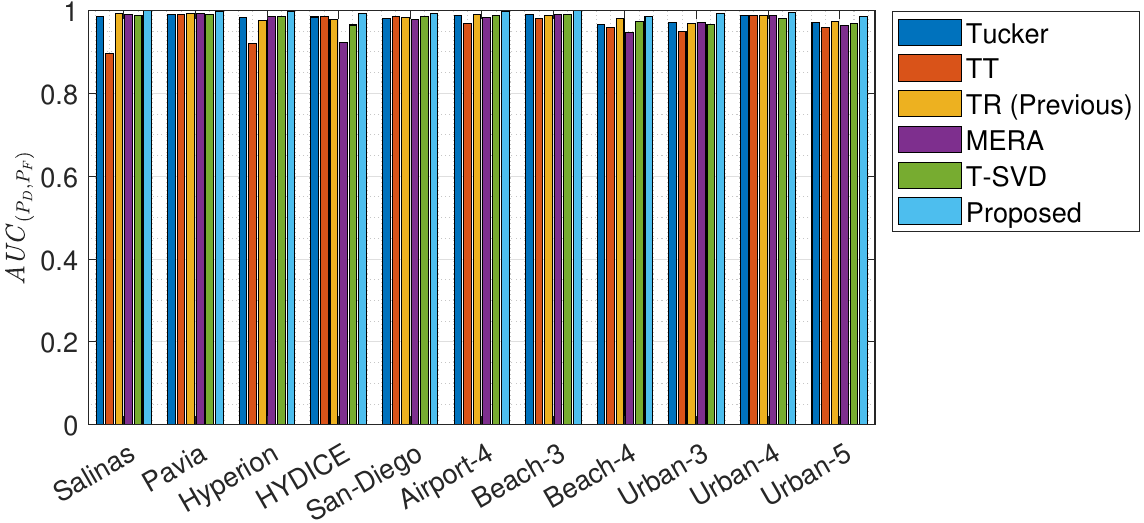}
\end{tabular}
\vspace{-0.15cm}
\caption{
\textcolor[rgb]{0.00,0.00,0.00}{The influence of 
various  tensor decomposition schemes  
upon  anomaly detection 
performance
of the  proposed 
method 
under  different HSIs datasets.}
%
%
}
\vspace{-0.7053cm}
\label{fig-decomposition}
\end{figure}

%
\begin{figure*}[!htbp]
\renewcommand{\arraystretch}{0.5}
\setlength\tabcolsep{1.2pt}
\centering
\begin{tabular}{ccc c}
\centering

\includegraphics[width=1.72in, height=1.43in]{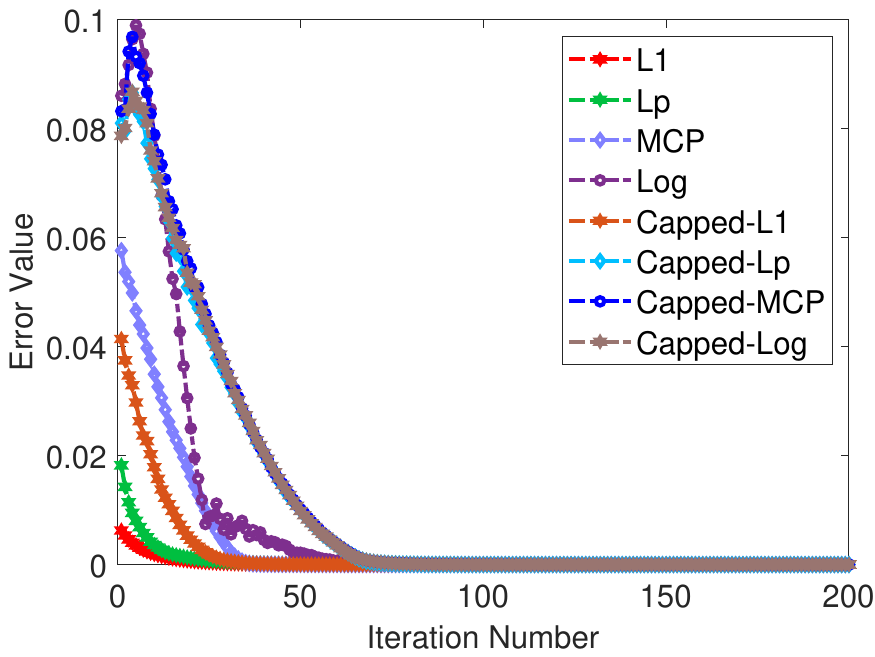}&
\includegraphics[width=1.72in, height=1.43in]{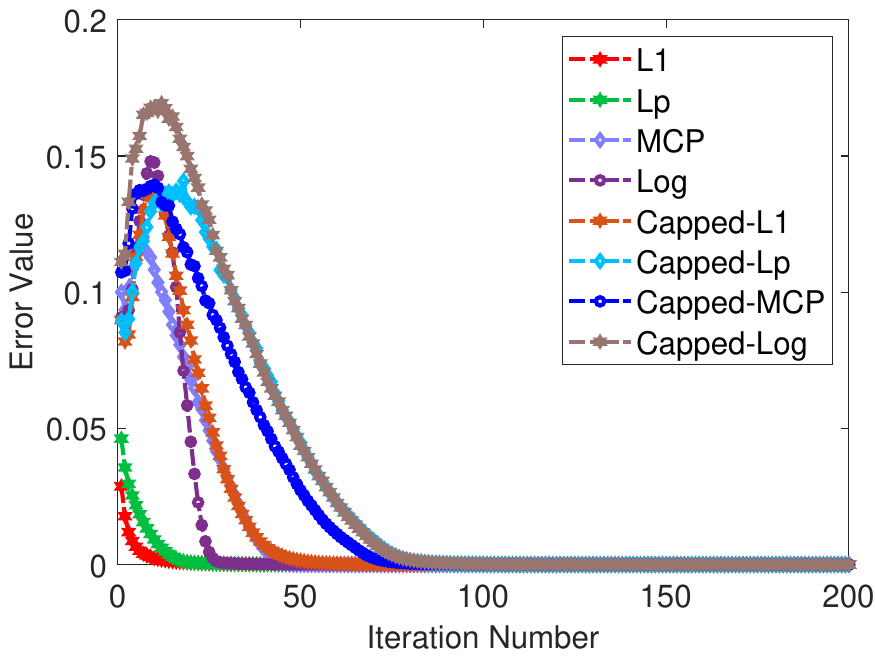}&
\includegraphics[width=1.72in, height=1.43in]{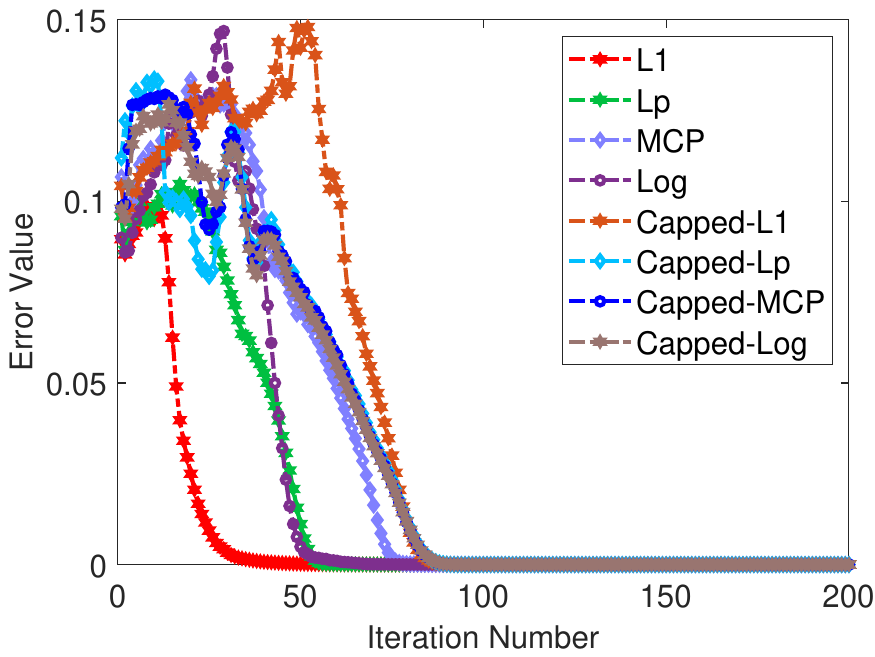}&
\includegraphics[width=1.72in, height=1.43in]{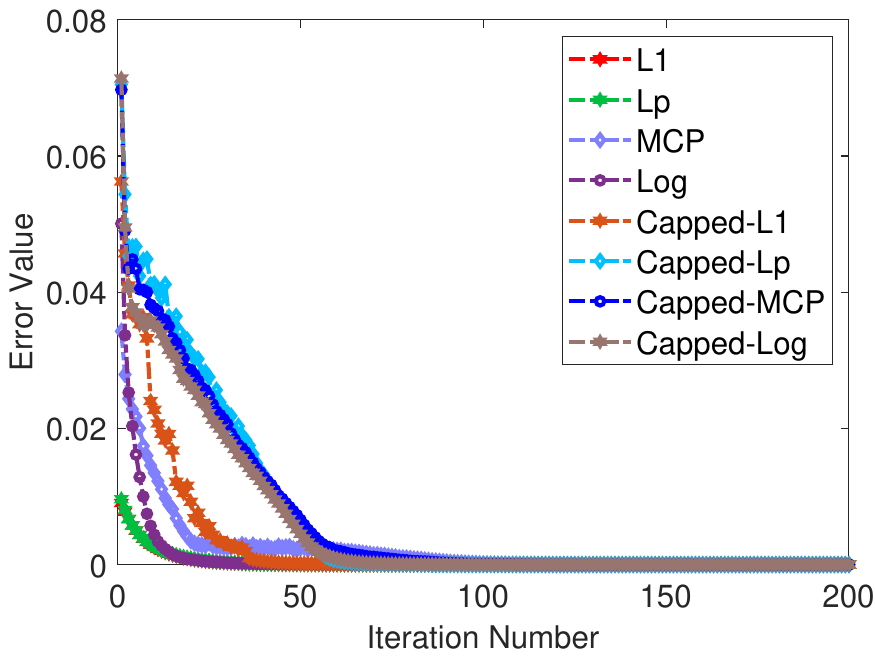}\\
\footnotesize{{{(a)}} Airport-4}  &
  \footnotesize{(b) Beach-3}  & \footnotesize{(c) Hyperion}
&\footnotesize{(d) San-Diego}
\\

\includegraphics[width=1.72in, height=1.43in]{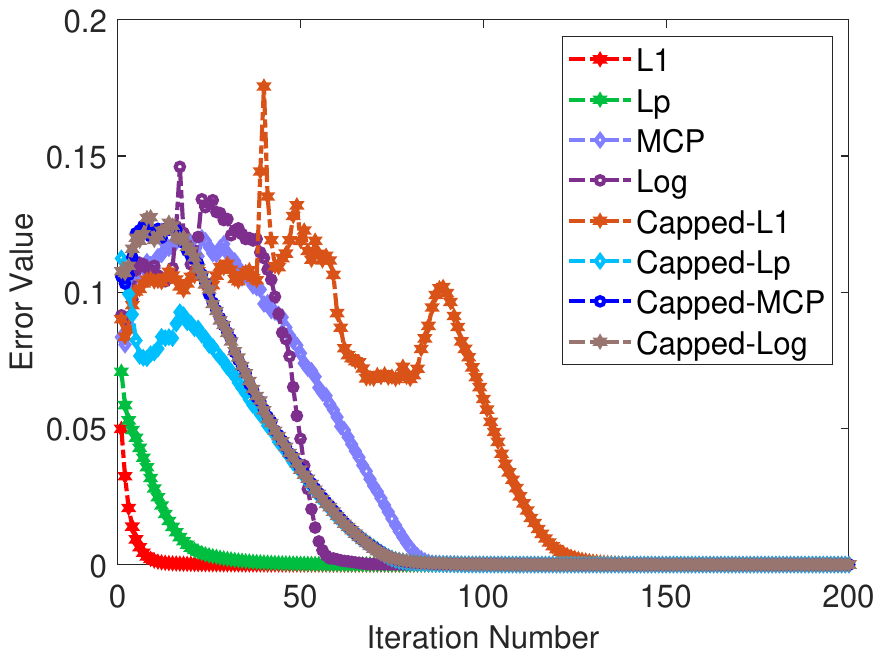}&
\includegraphics[width=1.72in, height=1.43in]{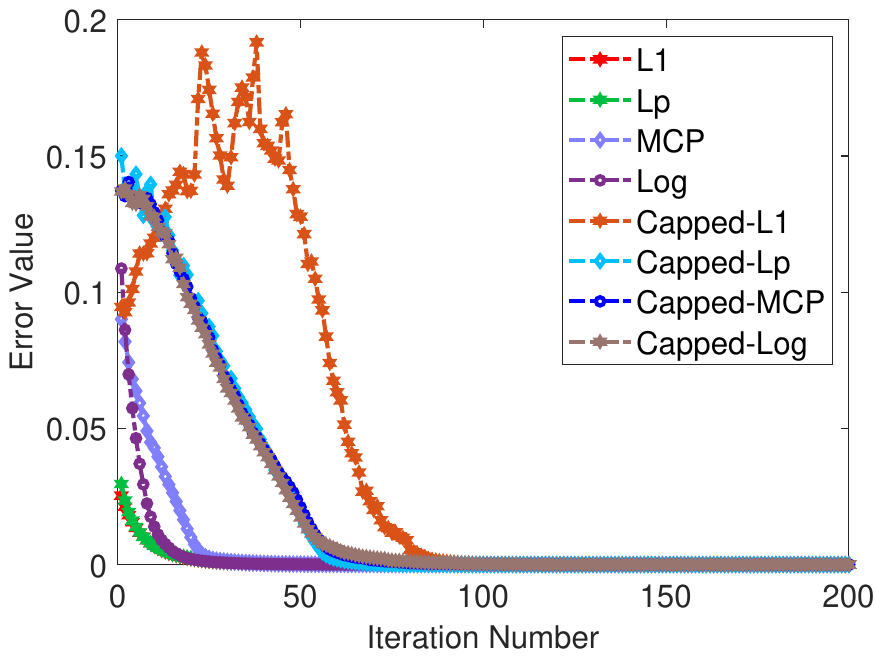}&
\includegraphics[width=1.72in, height=1.43in]{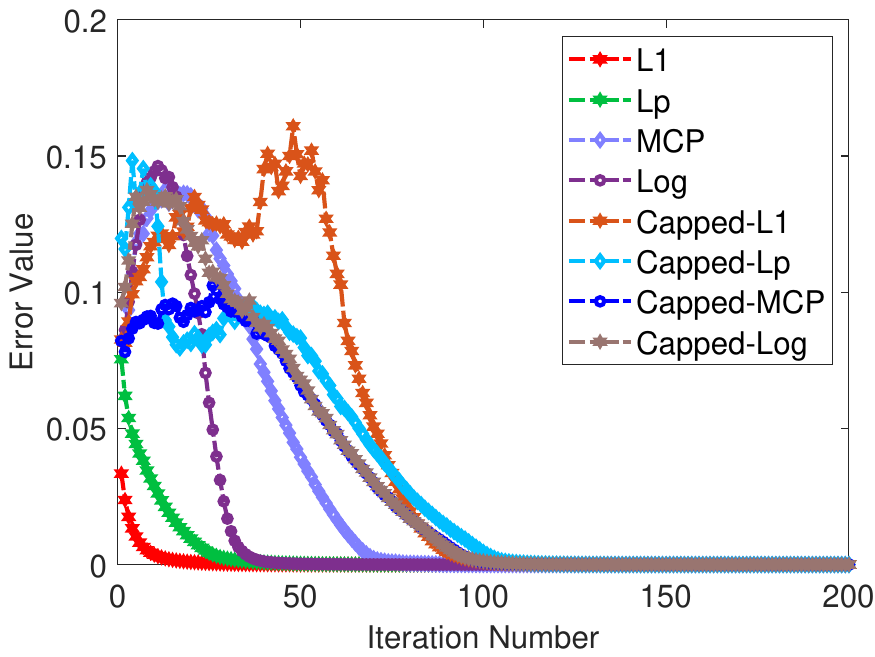}&
\includegraphics[width=1.72in, height=1.43in]{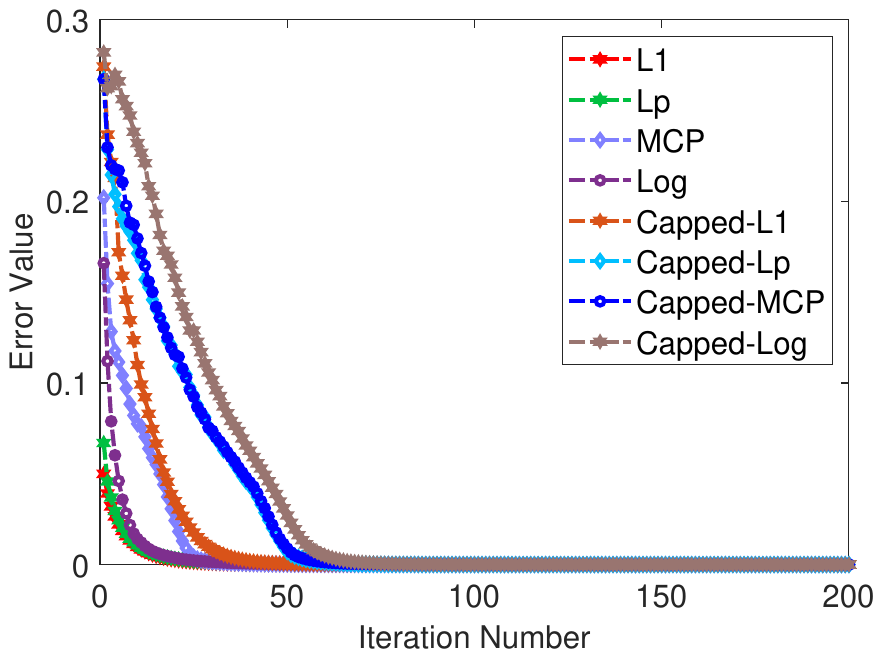}\\

\footnotesize{{{(e)}} HYDICE}  &
  \footnotesize{(f) Beach-4}  & \footnotesize{(g) Urban-3}
&\footnotesize{(h)  Pavia}

\end{tabular}
\caption{\textcolor[rgb]{0.00,0.00,0.00}{The convergence behavior of  the  proposed HAD  algorithm
under  different nonconvex functions.}} 
\vspace{-0.5cm}
\label{fig_convergence}
\end{figure*}

\subsubsection{Part 3}
\textcolor[rgb]{0.00,0.00,0.00}{
To verify
the effectiveness of the  TR decomposition, we 
replace the low-TR-rank  factorization scheme   
in our HAD
model with other tensor decompositions to achieve anomaly detection, such as TT  decomposition,
Tucker  decomposition \cite{shang2023hyperspectral},
 T-SVD decomposition \cite{wang2023guaranteed}, and \textit{multiscale entanglement renormalization ansatz}
 (MERA) \cite{xiao2024hyperspectral33}.} 
 \textcolor[rgb]{0.00,0.00,0.00}{In addition, we also compare  our HAD algorithm 
  with the recently  proposed  HAD method \cite{qin2024tensor}, which is  based on the nonconvex  low-TR-rank  gradient tensor 
approximation. 
This experiment 
keeps on using
the experimental datasets from the previous part.
The experimental results are depicted 
in Figure \ref{fig-decomposition}.
It can be easily noticed from Figure \ref{fig-decomposition} that, 
the proposed 
algorithm achieves optimal AUC values 
in most cases.
This shows that the TR 
decomposition and the nonconvex regularization strategies can
fully exploit the spectral-spatial correlation of the background,
thereby improving detection performance.
}

\subsubsection{Part 4}
\textcolor[rgb]{0.00,0.00,0.00}{We conducted 
experiments on HSI datasets with subpixel anomalies to further verify
the effectiveness and superiority of the  proposed HAD algorithm.
The hyperspectral data utilized  in this part  are consistent with those used in the literature \cite{gao2023bs}, i.e.,
HYDICE-I ($80 \times 100 \times 162$),
HYDICE-II ($60 \times 80 \times 162$),
AVIRIS-I ($100 \times 100 \times 189 $),
AVIRIS-II ($120 \times 120 \times 204 $), and
Hyperion ($150 \times 150 \times 155$).
The parameter settings for this experiment are the same as those in section  \ref{parameter-trtr}.
The 
experimental results can be found in  Table \ref{ablation-study-subpiple}, from which we can see that
our proposed nonconvex HAD framework
exhibits good detection performance in terms of  various output AUC values.
In general, the \textit{Discrete Cosine Transform} (DCT) is slightly better than the \textit{Fast Fourier Transform} (FFT)
in most cases.}

\vspace{-0.353cm}

\subsection{\textcolor[rgb]{0.00,0.00,0.00}{\textbf{Convergence  Analysis}}}

Under different nonconvex combinations:  $\Phi(\cdot)$+$\psi(\cdot)$,
the convergence 
error curves 
of the proposed  HAD algorithm on  all eight tested HSI datasets are presented  in Figure \ref{fig_convergence}.
For brevity,
the nonconvex functions $\Phi(\cdot)$ and $\psi(\cdot)$ are set to be the same.
It can be seen that
compared to the non-Capped-type functions, the Capped-type functions 
 exhibit a slower rate of convergence.
 Although there exist some fluctuations during the first $20$$\sim$$50$ iterations, they finally converge to zero after at most $100$ iterations.
Thereby, the proposed  HAD
algorithm has
good convergence and stability.


\section{\textbf{Conclusions and  Future Work}}\label{conclusion}


In this paper,
we have integrated several effective  technologies, 
including joint tensor decompositions, gradient maps  modeling,  
and  novel  nonconvex  sparsity-inducing 
strategy,
to collaboratively devise  an innovative HAD   method called HAD-EUNTRFR. 
 In our formulated HAD model,
  the powerful TR decomposition is utilized to fully
 mine the essential structural information 
 of background component 
  in the spectral and spatial modes.
Drawing upon the interpretable gradient TR  factor, we further introduce a  unified nonconvex 
regularizer induced by the T-SVD framework.
 %
 This regularizer effectively captures 
 the inherent low-rankness and the transformed sparsity 
  simultaneously, thereby substantially enhancing the model's performance and robustness.
 Meanwhile, another generalized nonconvex constraint is 
  incorporated into our model to promote the structured
 sparsity  of anomalous targets.
Algorithmically,  we deduce  a detailed procedure with an ADMM structure for solving
the 
proposed nonconvex model.
 A series of experiments considering 
 on both synthetic  and
real-world HSIs have verified the 
superiority and effectiveness of our HAD 
method.
\textcolor[rgb]{0.00,0.00,0.00}{In future research, we plan to first develop a  novel data-driven regularization strategy for profound characterization, along with a sketching framework aimed at dimensionality reduction.
 Furthermore, we intend to integrate these key 
  tools
 into tensor-format  deep neural networks
to 
 explore  more accurate and effective 
methods for the HAD task.}
\ifCLASSOPTIONcaptionsoff
  \newpage
\fi

\ifCLASSOPTIONcaptionsoff
  \newpage
\fi


\bibliographystyle{IEEEtran}
\bibliography{rhtc}

\end{document}